\documentclass[10pt,twocolumn,letterpaper]{article}

\usepackage[pagenumbers]{cvpr} 

\usepackage{graphicx}
\usepackage{sidecap}
\usepackage{amsmath}
\usepackage{amssymb}
\usepackage{yfonts}
\usepackage{array}
\usepackage{multirow}
\usepackage{booktabs}
\usepackage{enumitem}
\setlist[itemize]{noitemsep, topsep=0pt}
\setlist[enumerate]{noitemsep, topsep=0pt}

\usepackage[square,sort,comma,numbers]{natbib}

\usepackage[pagebackref,breaklinks,colorlinks]{hyperref}

\usepackage{filecontents}

\usepackage[capitalize]{cleveref}
\crefname{section}{Sec.}{Secs.}
\Crefname{section}{Section}{Sections}
\Crefname{table}{Table}{Tables}
\crefname{table}{Tab.}{Tabs.}

\makeatletter
\@namedef{ver@everyshi.sty}{}
\makeatother
\usepackage{tikz}
\usetikzlibrary{fit,positioning,calc,matrix,shapes.multipart,backgrounds,shapes.geometric, shapes.arrows}

\usepackage{xr}

\makeatletter
\newcommand*{\addFileDependency}[1]{
  \typeout{(#1)}
  \@addtofilelist{#1}
  \IfFileExists{#1}{}{\typeout{No file #1.}}
}
\makeatother

\newcommand*{\myexternaldocument}[1]{%
    \externaldocument{#1}%
    \addFileDependency{#1.tex}%
    \addFileDependency{#1.aux}%
}

\myexternaldocument{./sup}

\def\cvprPaperID{6114} 
\def\confName{CVPR}
\def\confYear{2022}

\newcommand{\ZZ}{\mathcal{Z}}
\newcommand{\WW}{\mathcal{W}}
\newcommand{\Ss}{\mathcal{S}}
\newcommand{\RR}{\mathbb{R}}
\newcommand{\Span}{\mathsf{span}}

\newcommand{\logit}{\mathsf{logit}}

\newcommand{\numel}{\mathsf{numel}}

\begin{document}

\title{Which Style Makes Me Attractive? Interpretable Control Discovery and Counterfactual Explanation on StyleGAN\thanks{Disclaimer: By attractiveness we mean the prediction of attractiveness by a classifier. Please be aware that the classifier might be biased. This paper does not reflect the judgement of the authors on the standard of attractiveness. }}

\author{Bo Li$^{1,2}$ \qquad Qiulin Wang$^{1}$ \qquad Jiquan Pei$^{1}$ \qquad Yu Yang$^{2}$ \qquad Xiangyang Ji$^{2}$\\
$^{1}$JD Technology \\ $^{2}$Tsinghua University\\
{\tt\small prclibo@gmail.com}
}
\maketitle

\begin{abstract}
The semantically disentangled latent subspace in GAN provides rich interpretable controls in image generation. 
This paper includes two contributions on semantic latent subspace analysis in the scenario of face generation using StyleGAN2. First, we propose a novel approach to disentangle latent subspace semantics by exploiting existing face analysis models, e.g., face parsers and face landmark detectors. These models provide the flexibility to construct various criterions with very concrete and interpretable semantic meanings (e.g., change face shape or change skin color) to restrict latent subspace disentanglement. Rich latent space controls unknown previously can be discovered using the constructed criterions. 
Second, we propose a new perspective to explain the behavior of a CNN classifier by generating counterfactuals in the interpretable latent subspaces we discovered. This explanation helps reveal whether the classifier learns semantics as intended.
Experiments on various disentanglement criterions demonstrate the effectiveness of our approach. We believe this approach contributes to both areas of image manipulation and counterfactual explainability of CNNs. The code is available at \url{https://github.com/prclibo/ice}.
\end{abstract}

\begin{figure}
    \centering
     \input{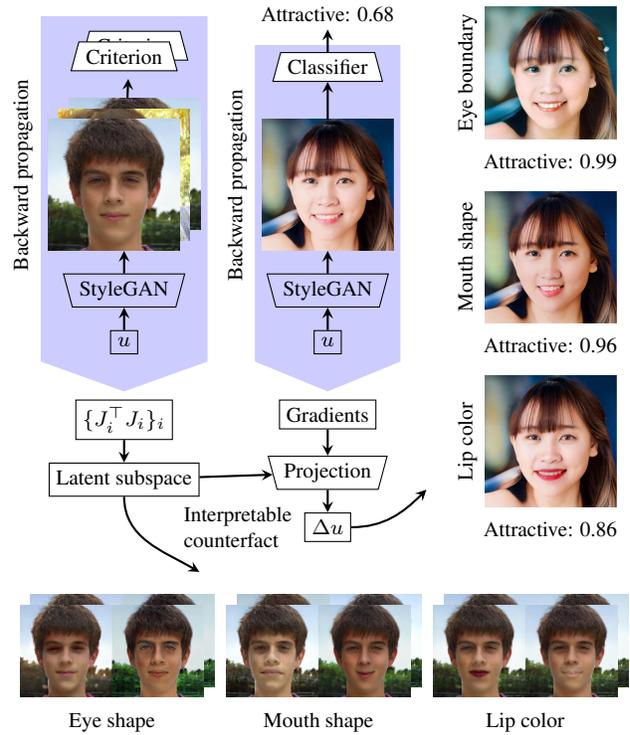}
    \caption{The left part illustrates the procedure of latent subspace discovery. Latent code samples $u$ are fed into StyleGAN and then interpretable criterions sequentially. We execute backward pass to obtain the Jacobian of the cascaded model to solve latent subspaces. The middle part illustrates the procedure of counterfactual generation. We feed a latent code sample $u$ into StyleGAN and then a classifier sequentially. The classification score is then backward passed through the cascaded model. The resulted gradients are projected into the previously solved interpretable subspaces. Thus we obtained interpretable controls on the latent code to produce counterfactuals and increase the classification score, as shown in the right part. See \Cref{sec:subspace-discovery,sec:counterfactual} for details.}
    \label{fig:teaser}
\end{figure}

\begin{figure*}
    \centering
    \footnotesize
    \setlength{\tabcolsep}{1pt}
    \begin{tabular}{cccccccc}
        \includegraphics[width=0.123\textwidth]{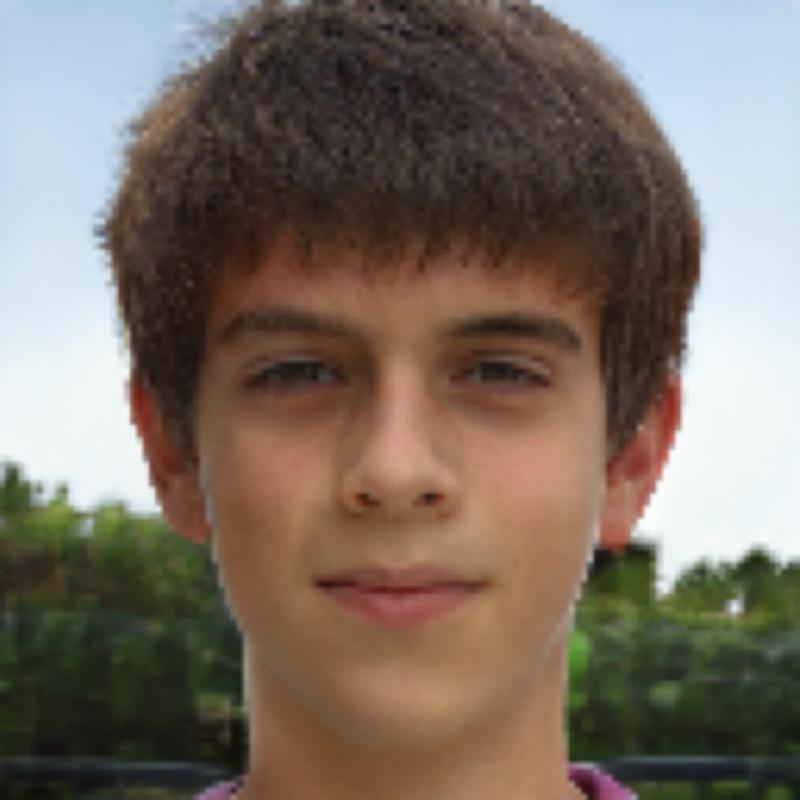}%
        \includegraphics[width=0.123\textwidth]{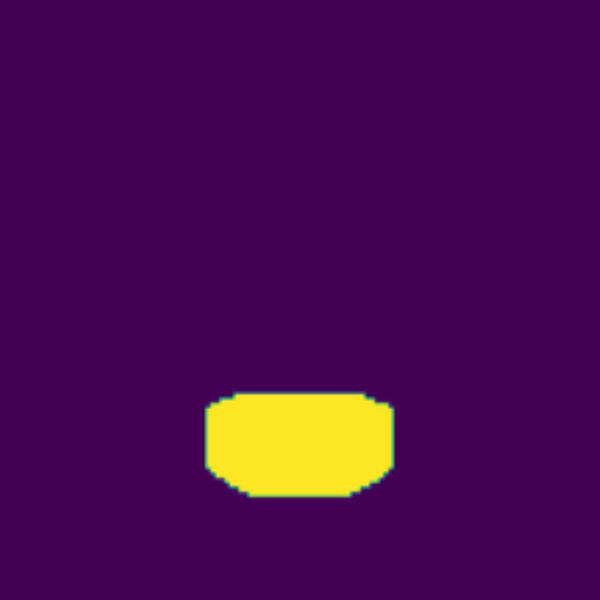} &
        \includegraphics[width=0.246\textwidth]{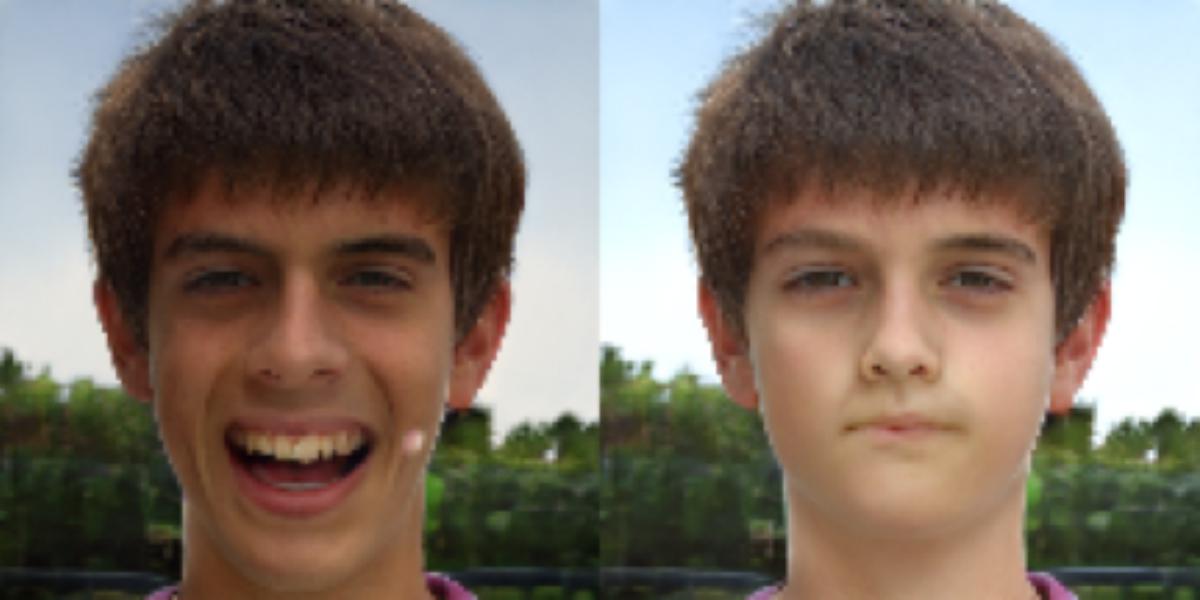} &
        \includegraphics[width=0.246\textwidth]{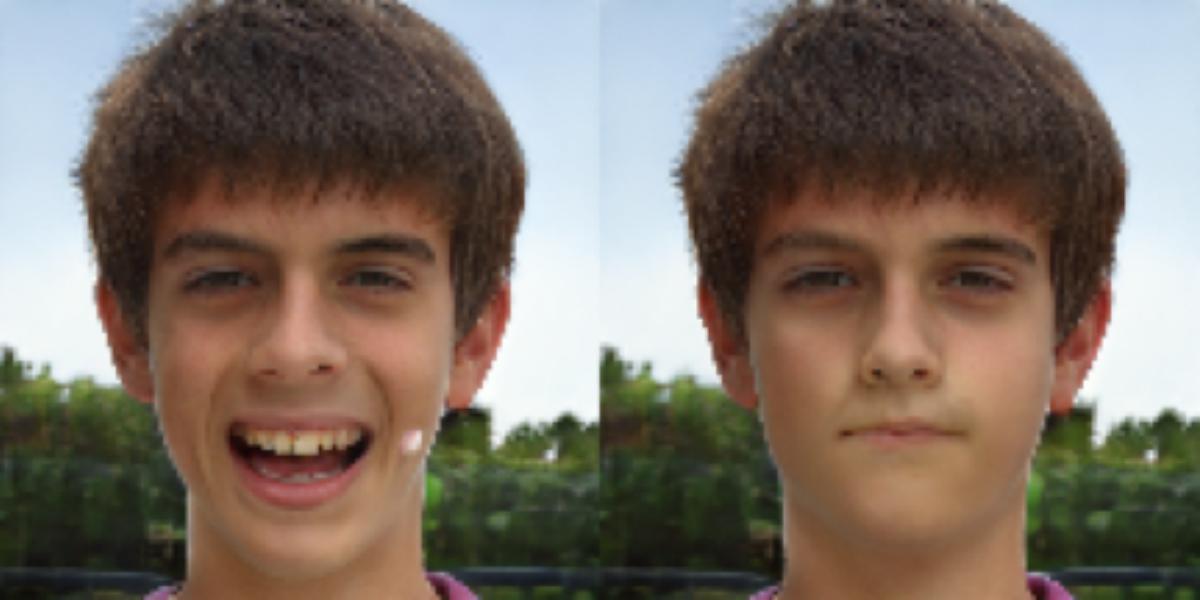}&
        \includegraphics[width=0.123\textwidth]{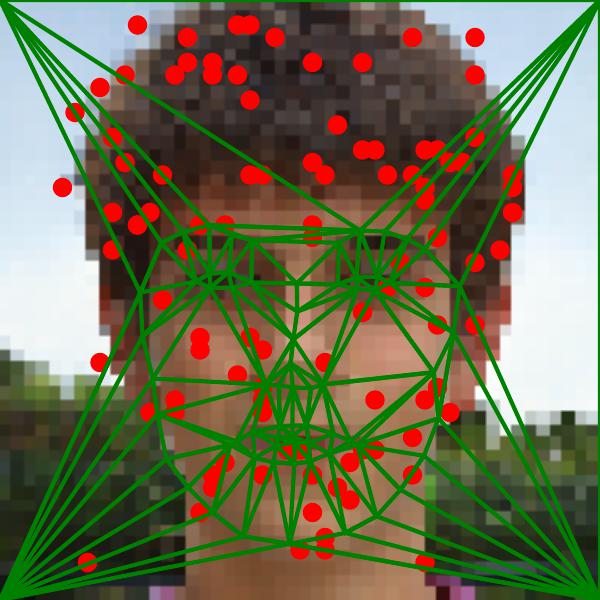}%
        \includegraphics[width=0.123\textwidth]{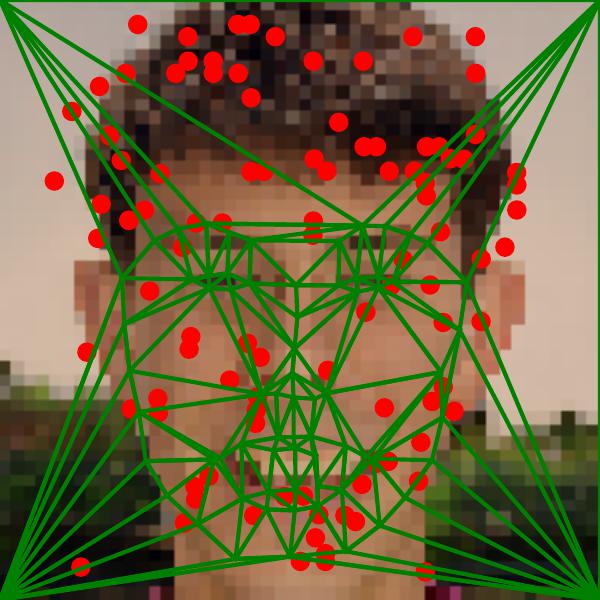}\\
        (a) Original image and mask  & (b) $\Span(V_\text{mp}^\text{mou})$ & (c) $\cdots \cap \Span(W_\text{mp}^{\overline{\text{mou}}})$& (d) Interpolated correspondences\\
        \includegraphics[width=0.246\textwidth]{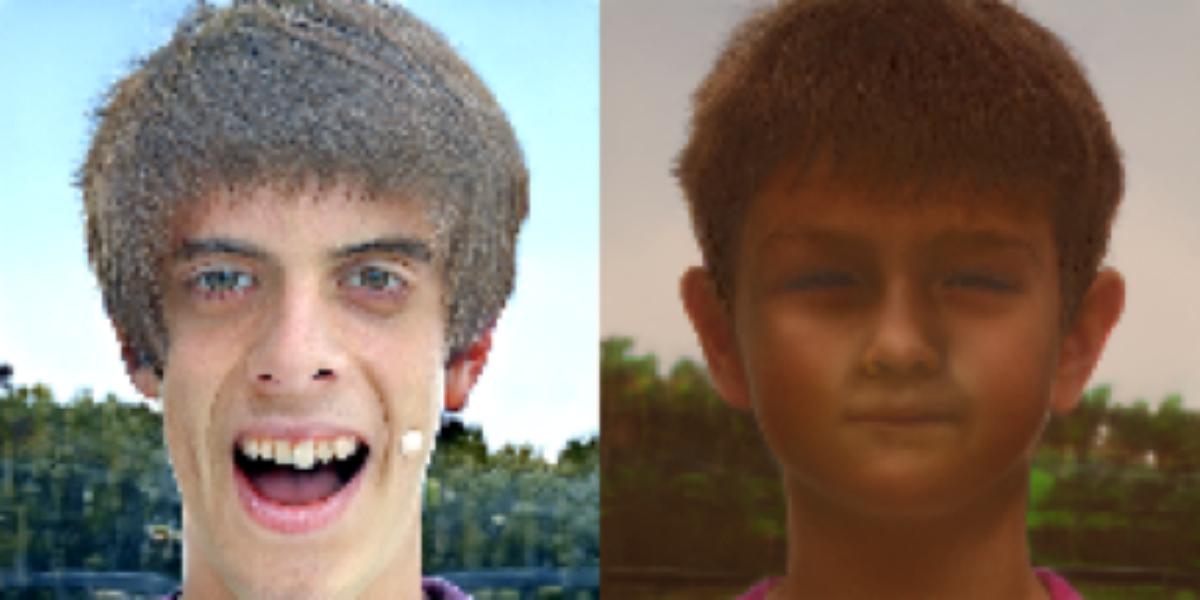} &
        \includegraphics[width=0.246\textwidth]{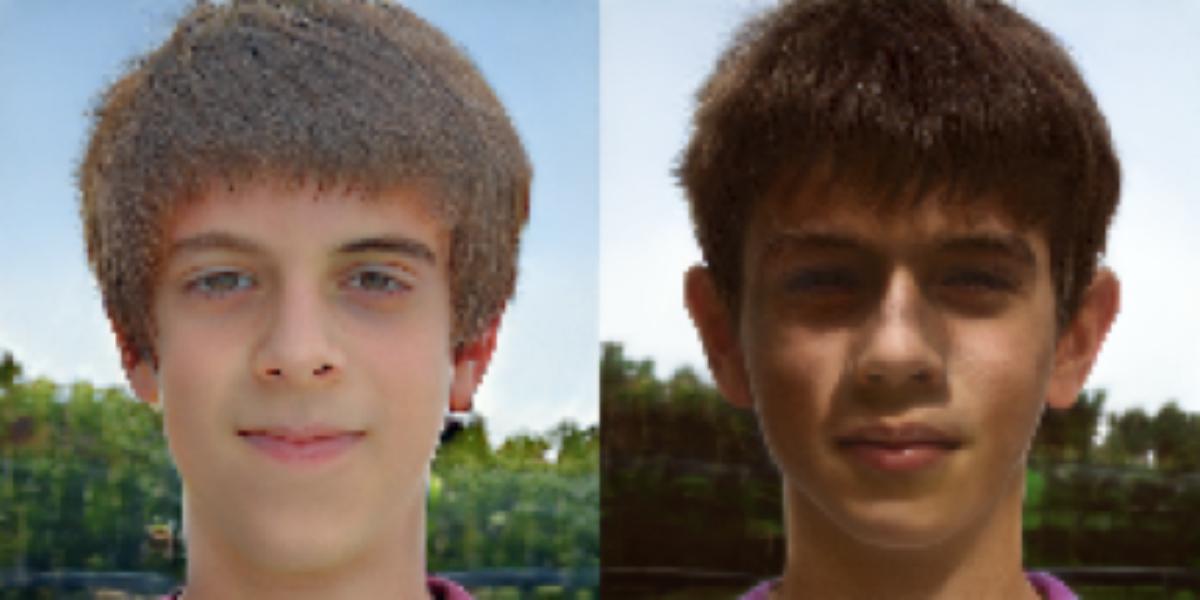} &
        \includegraphics[width=0.246\textwidth]{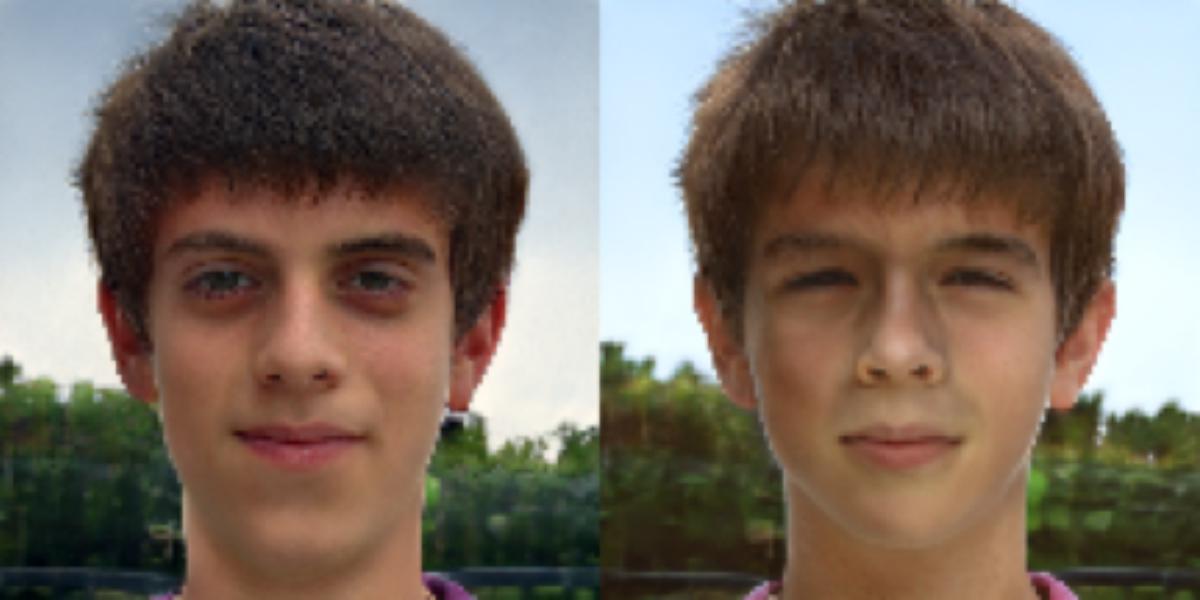} &
        \includegraphics[width=0.246\textwidth]{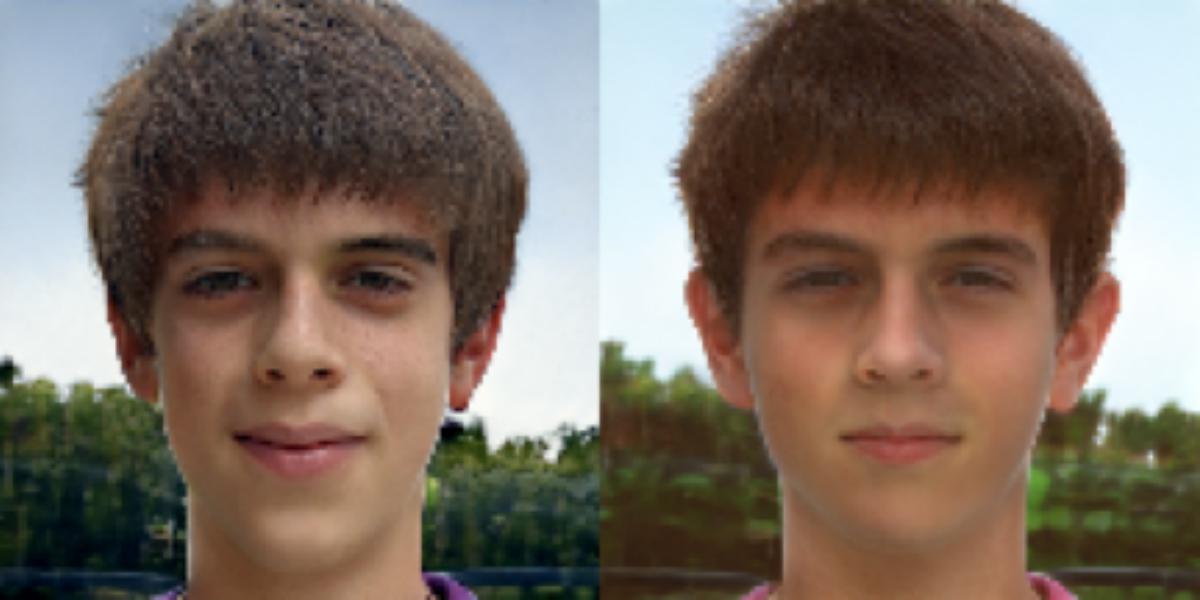} \\
        (e) $\Span(V_\text{fl}^\text{mou})$ & (f) $\cdots\cap\Span(W_\text{fl}^{\overline{\text{mou}}})$ & (g) $\cdots\cap\Span(W_\text{ap})$ & (h) $\cdots\cap\Span(W_\text{id})$
    \end{tabular}
    \caption{Illustration of latent controls generated by different criterions as described in \Cref{sec:criterions}. We use $\cdots$ to denote the subspace denoted in the previous sub-figure. In b-c and e-h, extra subspaces are consecutively intersected with the previous resulted subspaces.}
    \label{fig:criterions}
\end{figure*}

\section{Introduction}
\label{sec:intro}
Modern Generative Adversarial Networks (GANs)~\cite{karras2019style} have shown impressive ability to synthesize highly realistic images. StyleGAN~\cite{karras2019style} and its variants~\cite{karras2020analyzing} are one of the most successful representatives. StyleGAN converts a random latent code to an intermediate style code and injects styles into each layers of the generator network. Recent studies have shown that the latent space of StyleGAN provides well-disentangled representations for visual semantics. This property has derived an active research area on exploring interpretable latent subspaces (also referred as latent controls) to modify visual attributes in synthesized images~\cite{shen2020interfacegan, wu2021stylespace, patashnik2021styleclip, abdal2021styleflow, harkonen2020ganspace, shen2021closed, collins2020editing, zhu2021low}. Recent study has revealed the ability of StyleGAN subspaces to control even very elaborate structures such as eye gaze or earings~\cite{wu2021stylespace}. In this paper, we further show that by exploiting the knowledge from conventional face analysis models, e.g., face parsers and facial landmark detectors, more various interpretable latent subspaces can be discovered for face image generation.

The development of interpretable latent subspaces in image generation reminds a question whether this technique can be exploited to study deep learning interpretability.
In this paper, we answer this question by proposing a new type of counterfactuals which are generated from our discovered interpretable latent subspaces. 

The idea originates from the earlier studies on explaining classifiers of table data~\cite{wachter2017counterfactual, bodria2021benchmarking} and are recently recognized as an important perspective in the ML interpretability study. An ideal counterfactual alters the variables of a sample as little as possible to flip the classifier prediction. These altered variables help explain the factors related to a particular prediction. For example, a counterfactual usually comes in a form of ``If my salary is 10k higher, I will not be rejected for the loan''. Unlike table data, generating valid image counterfactuals is non-trivial. 
On one hand, it is difficult to sample valid counterfactuals from high-dimensional image manifold by simply altering pixels. To address this difficulty, some recent studies consider to exploit image generation techniques to produce counterfactuals~\cite{sixt2020interpretability, dombrowski2021diffeomorphic, lang2021explaining}. On the other hand, even though image counterfactuals can be sampled using generators, it is sometimes still difficult to explain what attributes or concepts are altered. Some works~\cite{sixt2020interpretability, dombrowski2021diffeomorphic} make use of heatmaps to highlight the difference between the original sample and its counterfactuals. However, when the difference is non-local and noisy, such explanation is still not satisfactory. 

In order to produce counterfactuals which can be explained concretely, we realize that interpretable latent subspaces serve as an possible solution. If counterfactuals are produced by altering with an interpretable latent subspace, their difference with the original sample can be better interpreted. Based on this assumption, we propose the following two contributions in this paper:
\begin{itemize}
    \item We propose a novel approach to discover interpretable latent subspaces. By exploiting face analysis models (e.g., face parsers and facial landmark detectors) as supervision, we are able to discover various new latent subspaces with good interpretability and concrete semantics. The flexibility of our approach could significantly enriches the latent subspaces for face generation.
    \item By restricting the latent code within the discovered latent subspace, we can produce counterfactuals with concrete semantics to explain the behavior of a face attribution classifier. We believe this approach provides a new perspective to the interpretability study for CNNs. 
\end{itemize}




\section{Related Works}
\label{sec:related}
\subsection{Interpretable Subspaces in StyleGAN}
The study of interpretable latent subspace disentanglement has become an active area recently~\cite{zhang2021survey}. Current approaches can be roughly categorized into two categories based on the requirement of supervision.

Supervised methods~\cite{shen2020interfacegan, yang2021semantic, bau2018gan, wu2021stylespace, patashnik2021styleclip, abdal2021styleflow, nitzan2020face} utilize annotated data or pre-trained networks to obtain domain specific knowledge. InterFaceGAN~\cite{shen2020interfacegan} uses pre-trained face attribute classifiers trained on attribute annotations to find hyper-planes in the latent space that can linearly separate positive and negative samples. \citet{yang2021semantic} and \citet{bau2018gan} explore and visualize the semantic hierarchy embedded in the intermediate layers of GANs through the correspondences with annotated parts or objects in the images. \citet{wu2021stylespace}~explore the more disentangled StyleSpace and discover the locally-active or attribute-specific style channels through the pre-trained segmentation and classification networks. StyleCLIP~\cite{patashnik2021styleclip} adopts the Contrastive Language-Image Pre-training (CLIP) model~\cite{radford2021learning} as supervision to perform the text-guided manipulation by optimization in the latent space. StyleFlow~\cite{abdal2021styleflow} utilizes a pre-trained normalizing flow, which defines an attribute-conditioned mapping between the latent space and the style space.

Unsupervised methods discover meaningful latent codes or directions by analyzing the space distribution under certain constraints~\cite{harkonen2020ganspace, shen2021closed, collins2020editing, zhu2021low, jahanian2019steerability, tewari2020stylerig}.
\citet{jahanian2019steerability} and \citet{spingarn2021gan} explore the "steerability" of pre-trained GANs and find "steering" directions correspond to image transformations by optimization or a closed form solution from the generator's weights.
StyleRig~\cite{tewari2020stylerig} proposes to train a new rigging network in a self-supervised pattern, which controls the StyleGAN via a 3D morphable face model (3DMM). GANSpace~\cite{harkonen2020ganspace} applies Principal Component Analysis (PCA) either in the latent space or feature space to identify useful control directions. \citet{plumerault2020controlling} find interpretable latent directions that correspond to continuous factors of variations (e.g. the position or scale of an object). SeFa~\cite{shen2021closed} delves into the very first fully-connected layer of the generator. It proposes an unsupervised and closed-form semantic factorization algorithm that discovers interpretable directions. \citet{collins2020editing} demonstrate that the StyleGAN's latent space can be spatially disentangled and decomposed into semantic objects and parts, which enables simple and effective local image editing. \citet{voynov2020unsupervised} propose an unsupervised and model-agnostic approach to discover interpretable directions in the GAN latent space by optimizing a joint classification and regression objective. \citet{wang2021geometry} compute the Hessian matrix of the squared distance function and perform eigen-decomposition to discover interpretable axes in the latent space. LowRankGAN~\cite{zhu2021low} proposes a low-rank factorization algorithm on the Jacobian matrix that corresponds to a specific local region to perform precise local control via null space projection. 

In this paper, our subspace discovery approach combines both supervised and unsupervised strategies. We formulate various novel supervisions by exploiting face analysis models to narrow the range of discovery, and then use the unsupervised strategy similar to~\cite{zhu2021low} to discover rich latent subspaces which is unknown previously.

\subsection{Generative Counterfactual Images}
Counterfactuals have been regarded as an important means to explain the behavior of black-box ML models. Comprehensive general reviews of this topic could be found in recent literature such as~\cite{bodria2021benchmarking}. Besides ML models on table data, a number of works have studied to generate counterfactuals of image data. Due to the difficulty to synthesize realistic images, earlier literature~\cite{dhurandhar2018explanations, van2019interpretable} mostly works with simple datasets like MNIST. Recently, the development of generative models (e.g., GANs~\cite{goodfellow2014generative}, VAEs~\cite{kingma2013auto}, Normalizing Flows~\cite{kingma2018glow}, etc) provides the possibility to generate very realistic counterfactual images since they are able to learn mappings from the latent space to the image manifold. \citet{singla2019explanation} adopt a conditional GAN to produce realistic perturbations on a given image that gradually change the predictions of a classifier. GANalyze~\cite{goetschalckx2019ganalyze} takes advantages of the GAN-based model to visualize what a CNN model learns about high-level cognitive properties. StylEx~\cite{lang2021explaining} proposes to incorporate the classifier into the training process of StyleGAN and learn a classifier-specific StyleSpace. 
\citet{sauer2021counterfactual} propose to disentangle object shape, object texture and background in the image generation process and generate structured conterfacturals which help improve the robustness and interpretability of classifiers.
\citet{goyal2019explaining} adopt a VAE as the generative model to discover the Causal Concept Effects (CaCE) that explain a specific classifier. \citet{sixt2020interpretability} combine an invertible network with a linear classifier to generate faithful counterfactual examples and find the isosurface of the classifier through PCA. \citet{dombrowski2021diffeomorphic} perform gradient ascent in the latent space of a well-trained normalizing flow to maximize the probability of the target class, then invert it back to the image space to generate counterfactuals.

As mentioned in \Cref{sec:intro}, after the counterfactuals are produced by generators, it is still unclear what semantic features are altered. We propose in this paper to alleviate this problem by restricting the alternation in a subspace with concrete semantics. 


\subsection{Interpretable Face Analysis}
To explain the motivation of this paper in a wider perspective, we further briefly review the study of interpretability on deep learning face recognition task. 
Interpretability has been an active topic in ML and also an important component to establish trustworthy AI~\cite{li2021trustworthy}. In the area of computer vision, recent works on interpretable mostly focus on image classification tasks on dataset like MNIST or ImageNet~\cite{bodria2021benchmarking}.
In terms of face recognition or attribution tasks which distinguish fine-grained structures, these approaches are usually not applicable. We notice there have been some recent literature working on explaining either face recognition or face attribution~\cite{jiang2021explainable, williford2020explainable, yin2019towards, sixt2020interpretability, dombrowski2021diffeomorphic}.
We believe the proposed approach also makes contribution to this underdeveloped area.

\section{Supervised Subspace Discovery}
\label{sec:subspace-discovery}

\subsection{Notations}

StyleGAN is known to have three different internal feature spaces, namely the input space $\ZZ=\RR^{512}$, the latent space $\WW=\RR^{512}$, the augmented latent space $\WW+=\RR^{9216}$, and the style space $\Ss=\RR^{9088}$. In this paper, we use $u$ to denote a feature code in any of the above spaces. A generator (e.g., a StyleGAN) $g$ transforms $u$ to a generated image $x = g(u)$. We use $p$ to denote a pixel. $x(p)$ can be obtained by bilinear interpolation.

We use $f$ to denote a differentiable function on $x$. For a face image in this paper, $f$ can be a classifier, a facial landmark regression model, a face pose estimator or any customized functions. In \Cref{sec:criterions}, we will describe a series of examples of $f$. We can cascade $f$ with $g$, and obtain a predictor on $u$ as $h(u) = f(g(u))$.

We use $\odot$ to denote element-wise multiplication. We use $\overline{m}$ to denote the opposite of a mask $m$. We use a capital letter to denote a matrix and sometimes abuse it as a set made up of its column vectors. We use $\Span(X)$ to denote the linear subspace spanned by column vectors in $X$. We use $\numel(x)$ to denote the size of a vector, an image, or a feature map. We use $\begin{bmatrix} x \end{bmatrix}_\text{condition(x)}$ to denote the matrix stacked by all the vectors which satisfy given conditions.

In this section, we mainly study the eigenvector matrix of $\frac{\partial h_*^*}{\partial u}^\top\frac{\partial h_*^*}{\partial u}$. We split the eigenvector matrix into two blocks $\begin{bmatrix} V_*^* & W_*^* \end{bmatrix}$, where $V_*^* = V(h_*^*)$ contains eigenvectors with non-zero eigenvalues while $W_*^* = W(h_*^*)$ corresponds to eigenvalues near zero. $*$ refers to any superscripts or subscripts. 
\subsection{Jacobian Analysis}
\label{sec:jacobian-analysis}

\subsubsection{Basic Formulation}
We start by a setup similar to~\cite{zhu2021low}. Suppose we would like to find a direction vector $n$ in the StyleGAN feature space which can change the appearance of the mouth. We denote a predictor as
\begin{equation}
    h_0(u) = g(u) \odot m,
    \label{eq:masked-photo}
\end{equation}
where $m$ is a binary mask indicating the mouth region. We use the following formulation to denote the change of appearance:
\begin{equation}
    \ell(u, n, \alpha) = \| h_0(u + \alpha n) - h_0(u) \|^2 \approx \alpha^2 n^\top J^\top J n,
    \label{eq:l2-diff}
\end{equation}
where $J_0 = \frac{\partial h_0}{\partial u}$ is the Jacobian matrix of $h$, $n$ is a unit vector, and $\alpha$ is a small step on the direction of $n$. Obviously \eqref{eq:l2-diff} is maximized by the eigenvector with the largest eigenvalue of $J_0^\top J_0$ and minimized by the eigenvector the smallest eigenvalue. We hereby denote the eigenvector matrix of $J_0^\top J_0$ as $\begin{bmatrix} V_0 & W_0 \end{bmatrix}$, where $W_0$ corresponds to eigenvalues near zero. Perturbing $u$ in subspace $\Span(V_0)$ changes (activate) the value of $h_0(u)$, while perturbation in subspace $\Span(W_0)$ does not change (suppress) the value of $h_0(u)$.

We illustrate this procedure by an example in \Cref{fig:criterions}a. As can be noticed in \Cref{fig:criterions}b, perturbations in $V_0$ might also cause $h(u)$ to change in the region of $\overline{m}$ (i.e. the face color). To suppress this unnecessary change, we repeat the above procedure for $h_1(u) = g(u) \odot \overline{m}$ and obtain $\begin{bmatrix} V_1 & W_1 \end{bmatrix}$. Perturbations in the linear span intersection
\begin{equation}
    \Span(V_0) \cap \Span(W_1),
    \label{eq:span-intersection}
\end{equation}
will change the value of $x$ in the region of $m$ but preserve the value of $x$ in $\overline{m}$ (\Cref{fig:criterions}c). 

\subsubsection{Implementation Tricks}

$J = \frac{\partial h}{\partial u}$ can be obtained by executing $\numel(h(u))$ times of backward pass for each element of $h(u)$. When $\numel(h(u))$ is large, e.g., $h(u)$ can be an image-like map, it is computationally expensive to directly compute $J$. Not to mention that the computational cost of $J^\top J$ is also high for large $J$. To alleviate this cost, we can approximate $J^\top J$ directly by considering $\frac{\partial\ell}{\partial n} \approx 2 \alpha^2 n^\top J^\top J$.
$\alpha$ is a selected constant parameter. If we select $n$ as a one-hot vector and execute backward pass to compute $\frac{\partial\ell}{\partial n}$, we obtain one row of $J^\top J$. By enumerating all the one-hot $n$, we can compute $J^\top J$ by $\numel(n)$ times of backward pass. In practice, the above two implementations of $J^\top J$ can be selected according to the dimension of $h(u)$.

We next consider to compute the basis vectors $V$ of $\Span(V_0) \cap \Span(W_1)$. In~\cite{zhu2021low}, the basis vectors are obtained as $V = W_1 {W_1}^\top V_0$. In practice, we would like the columns of $V$ to be sorted by their impact on $h(u)$. $W_1 {W_1}^\top V_0$ does not guarantee this, even though $V_0$ is sorted by the eigenvalue of $J_0^\top J_0$. A trick to obtain sorted basis vectors is to compute $V$ as the eigenvectors of
\begin{equation}
    {W_1}^\top J_0^\top J_0 W_1.
    \label{eq:wjjw}
\end{equation}
\subsection{Criterion for Interpretable Control}
\label{sec:criterions}

\subsubsection{Masked Photometry}

\eqref{eq:masked-photo} provides a straight-forward criterion to measure the photometric change of the generated images. The mask $m$ can be generated by a face parsing network or by a manually labeled region as in~\cite{zhu2021low}. We have illustrated the example in \Cref{fig:criterions}a-c previously. To easy reference, we denote this criterion as $h_\text{mp}^\text{mou}$ by setting $m$ as a mouth mask $m^\text{mou}$ for \eqref{eq:masked-photo}. The corresponding subspace for \eqref{eq:span-intersection} is thus written as 
\begin{equation}
 \Span(V_\text{mp}^\text{mou}) \cap \Span(V_\text{mp}^{\overline{\text{mou}}}).
 \label{eq:mouth-mp}
\end{equation}

\subsubsection{Facial Landmarks}
\label{sec:facial-landmarks}

Using masked photometry to compute \eqref{eq:span-intersection} enables to locally manipulate the generated images. 
However, such image change might be entangled with both appearance and geometry. We believe that better disentanglement of appearance and geometry could contribute to better interpretability and control in the generation procedure. In \Cref{sec:facial-landmarks} and~\ref{sec:aligned-photometry}, we use facial landmarks to construct criterions to achieve elaborate control over the face geometry.

Consider a facial landmark detector $f_\text{fl}$ which is differentiable to images. We define
\begin{align}
    h^m_\text{fl}(u) &= \begin{bmatrix} q \end{bmatrix}_{q \in Q \wedge m(q)=1},\\
    Q &= f_\text{fl}(g(u)),
\end{align}
where $m$ is a mask to select targeted facial landmarks. $h^m_\text{fl}$ provides a flexible approach to manipulate face geometry. We show an example in \Cref{fig:criterions}e-f to manipulate the shape of face boundary. Denote $m$ to select face boundary landmarks. We obtain $V^m_\text{fl}$ as in \Cref{sec:jacobian-analysis}. \Cref{fig:criterions}e shows the manipulation results on a direction in $\Span(V^m_\text{fl})$. On one hand, we notice that supervision by landmarks enables us to flexibly discover geometric controls which are unknown previously. On the other hand, we notice that the manipulation on face boundary will also deform face geometry at other areas, e.g., open the mouth. To alleviate this entanglement, we can compute $W^{\overline{m}}_\text{fl}$ and project the manipulation vector into $\Span(V^{m}_\text{fl}) \cap \Span(W^{\overline{m}}_\text{fl})$ as shown in \Cref{fig:criterions}f.

\subsubsection{Aligned Photometry}
\label{sec:aligned-photometry}

We have disentangled the geometric shape change in face generation using facial landmarks. We then use facial landmarks to disentangle the facial appearance change. This is useful in the cases of changing the face expression or face shape while preserving the face appearance. We first decompose the whole image by Delaunay mesh $\mathsf{Delaunay}(Q, i, c)$ triangulated on a facial landmark set $Q$:
\begin{align}
    \mathsf{Delaunay}(Q, i, c) &= \begin{bmatrix} q_{v_0(i)} & q_{v_1(i)} & q_{v_2(i)} \end{bmatrix} c,\\
    Q &= f_\text{fl}(g(u)),
\end{align}
where $i$ denotes the triangle facet index. $q_*$ are landmarks in $Q$ and $v_*(i)$ denotes the landmark index of each triangle vertex in $Q$. $c$ denotes the 3D bary-centric coordinates of a point which sum up to $1$. Face shape change can be approximately modeled by the deformation of the Delaunay mesh. Consider a set of points uniformly distributed on the Delaunay mesh $P = \{ (i, c) \}$. No matter how the face deformed, $P$ persists consistent semantics and relative positions on a face. 
We show an example of $P$ in \Cref{fig:criterions}d, where the red points from $P$ keep their relative positions with respect to the face under face motion.
We can then use Delaunay mesh to denote a aligned photometry on a generated image:
\begin{align}
    h^m_\text{ap}(u) &= \begin{bmatrix} x(p) \end{bmatrix}_{p \in P' \wedge m(p) = 1},\label{eq:align-interpolate}\\
    P' &= \{ \mathsf{Delaunay}(Q, i, c) | (i, c) \in P \},
\end{align}
where \eqref{eq:align-interpolate} can be realized by bilinear interpolation. $m$ defines a mask to filter the control region.

We illustrate the effectiveness of aligned photometry in \Cref{fig:criterions}g. We notice that the face color is changed in \Cref{fig:criterions}f. To suppress this appearance change, we further intersect the latent subspace with $W_\text{ap}$. $W_\text{ap}$ is obtained from~\eqref{eq:align-interpolate} by setting $m$ as true on the foreground region. $\Span(W_\text{ap})$ allows controls to deform the image and preserves photometry in corresponding areas. We can see in \Cref{fig:criterions}g that the face color is preserved when modifying the face boundary. In addition, we observe that the magnitude of face boundary change is reduced in more restricted subspaces along \Cref{fig:criterions}d-f, though the manipulation magnitudes are the same.


\subsubsection{Face Identity Feature}

In some scenarios, we might require the face identity not to be deviated too much. To preserve the face identity, we can exploit a differentiable face feature extractor $f_\text{id}$ to create a criterion to restrict the subspace,
\begin{equation}
    h_\text{id}(u) = f_\text{id}(g(u)).
\end{equation}
\Cref{fig:criterions}h illustrates the effect of $h_\text{id}$. We notice that the eye shape is more similar to the original image and the face identity is better preserved.

\subsubsection{More Customized Criterions}

As is shown in this section, the proposed formulation is able to make use of any flexibly designed criterion to restrict latent subspace discovery. We provide the above examples which possess good interpretability. As will be shown in our experiments, we are able to disentangle a variety of facial features, e.g., shape, color, appearance, and id, by combining the above criterions. 
Beyond the above criterions, we would like to further note that readers can customize more flexible criterions to discover unknown latent subspaces. We provide two more examples below.

We can control the face color of using the formulation by computing a masked average color. We denote
\begin{align}
    h_\text{mac}^m(u) &= \underset{m(p) = 1}{\mathsf{mean}}(x(p)),\\
    h_\text{res}^m(u) &= \begin{bmatrix} x(p) - h_\text{mac}^m(u) \end{bmatrix}_{m(p) = 1},
\end{align}
where $m$ represents the face skin area. The masked average color $h_\text{mac}^m(u)$ captures the rough color of the face and $h_\text{res}^m(u)$ captures the residual face appearance. By constructing the subspace $\Span(V_\text{mac}^m) \cap \Span(W_\text{res}^m)$, we obtain control directions which manipulate the face color but preserves the face appearance.

\begin{figure*}
    \centering
    \scriptsize
    \setlength{\tabcolsep}{1pt}
    \begin{tabular}{ccccccccccc}
         \rotatebox{90}{Skin color}& 
        \includegraphics[width=0.19\textwidth]{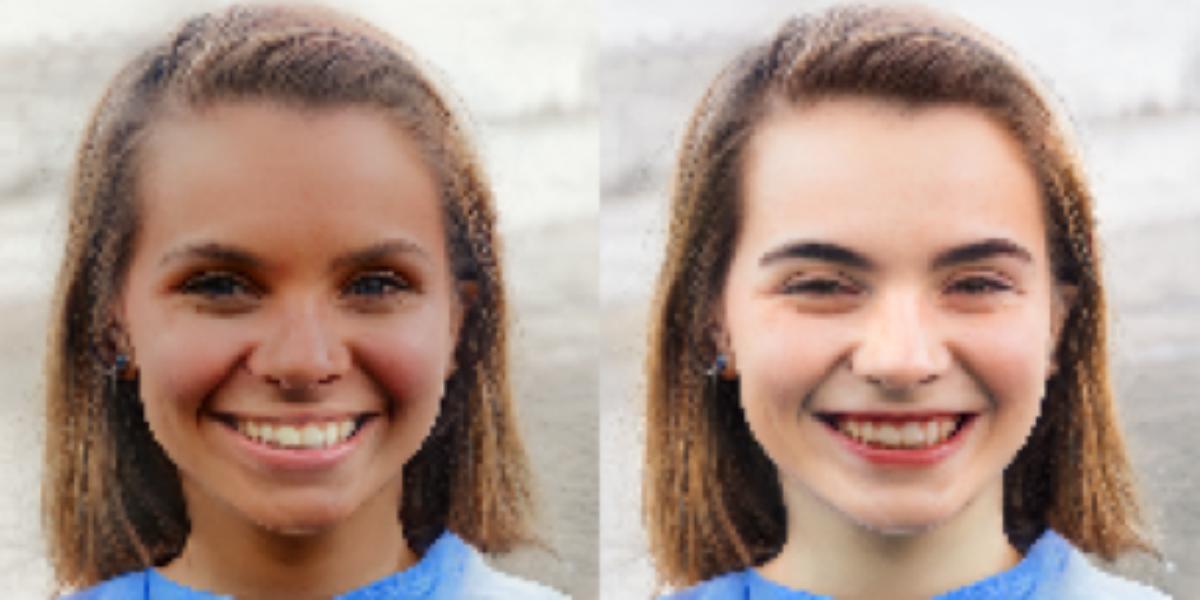} &
        \includegraphics[width=0.19\textwidth]{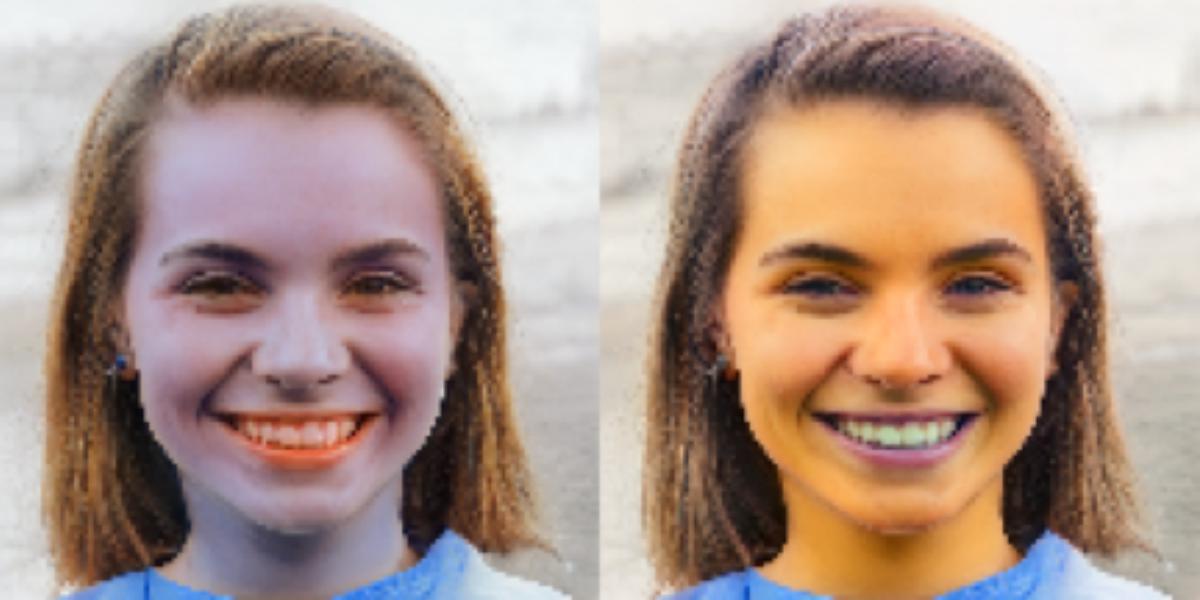} &
        \includegraphics[width=0.19\textwidth]{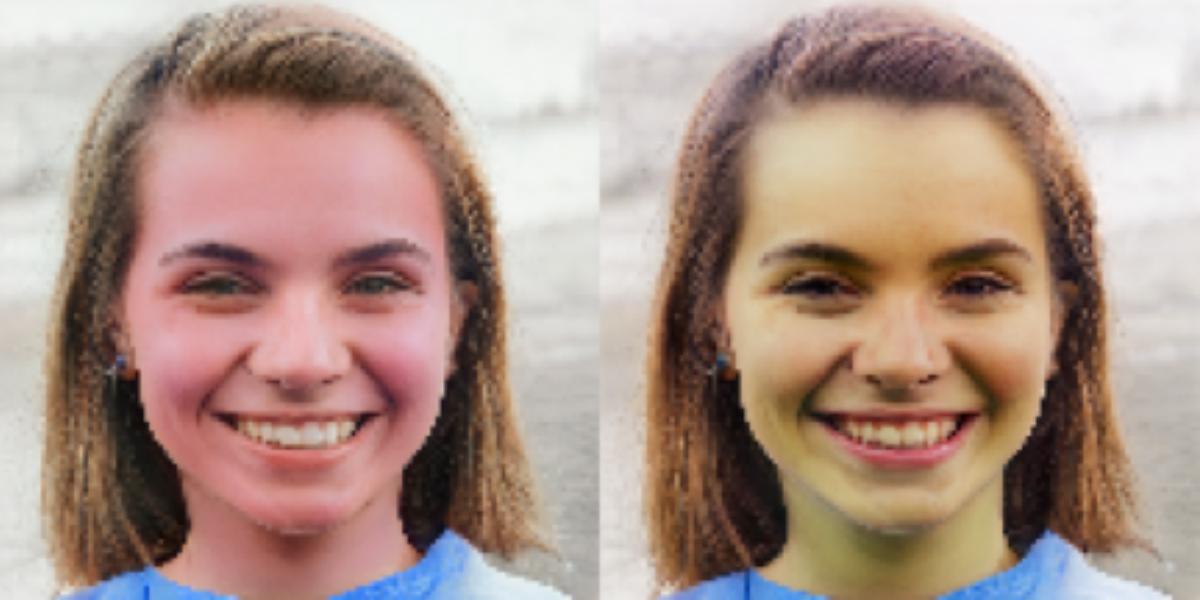} &
        \includegraphics[width=0.19\textwidth]{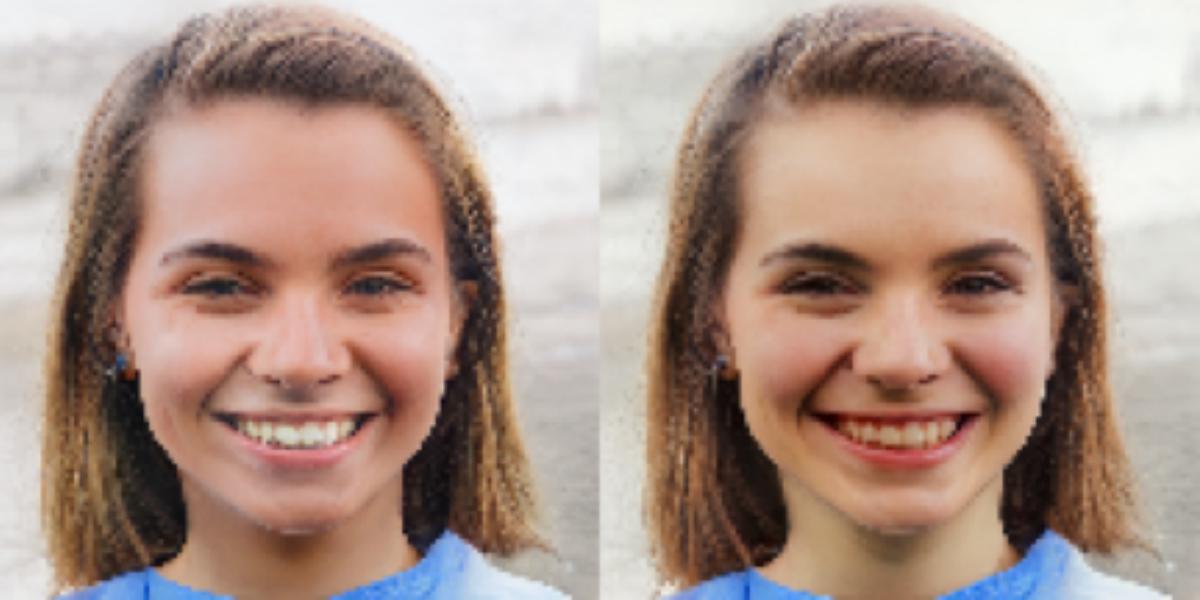} &
        \includegraphics[width=0.19\textwidth]{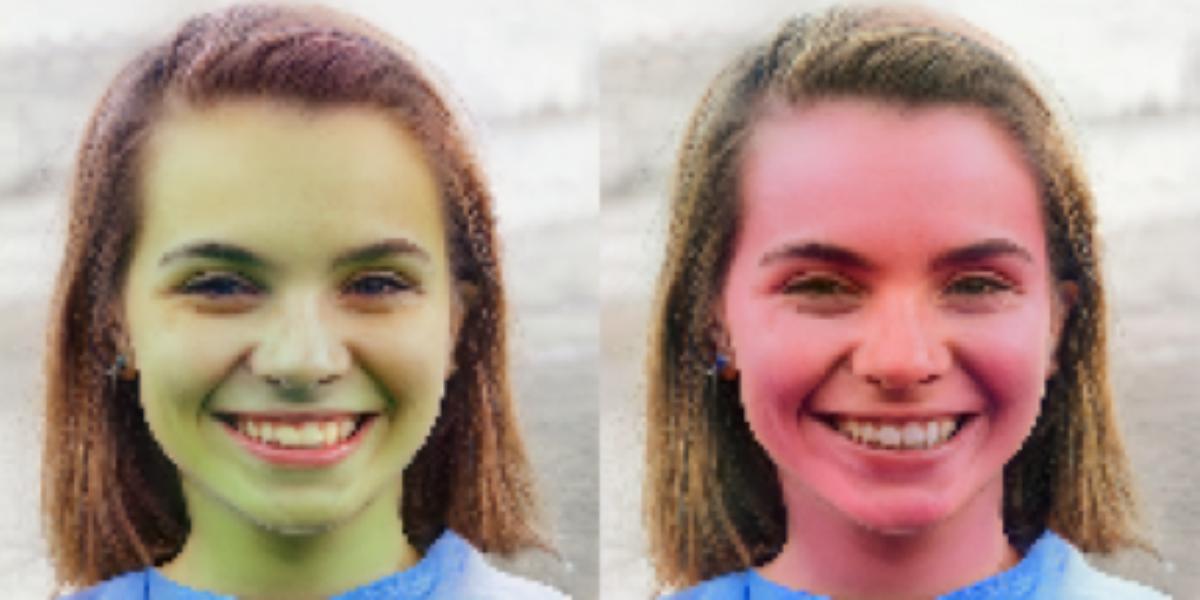} \\
        & Black/white & Blue/yellow & Red/green & Pink/white & Green/red\\
         \rotatebox{90}{Mouth shape}& 
        \includegraphics[width=0.19\textwidth]{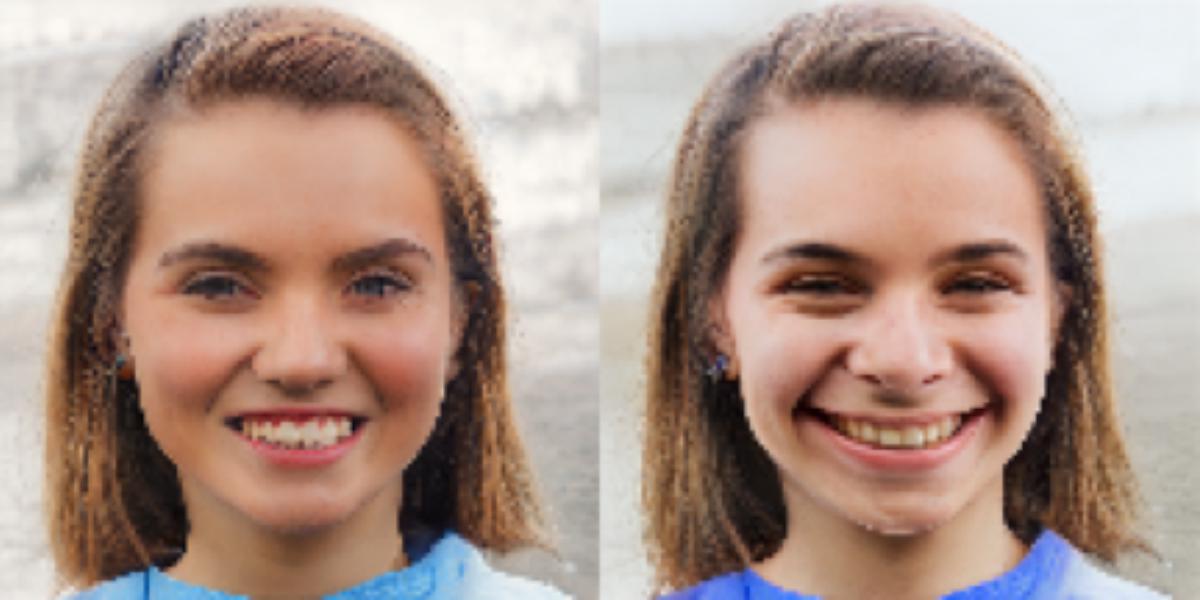} &
        \includegraphics[width=0.19\textwidth]{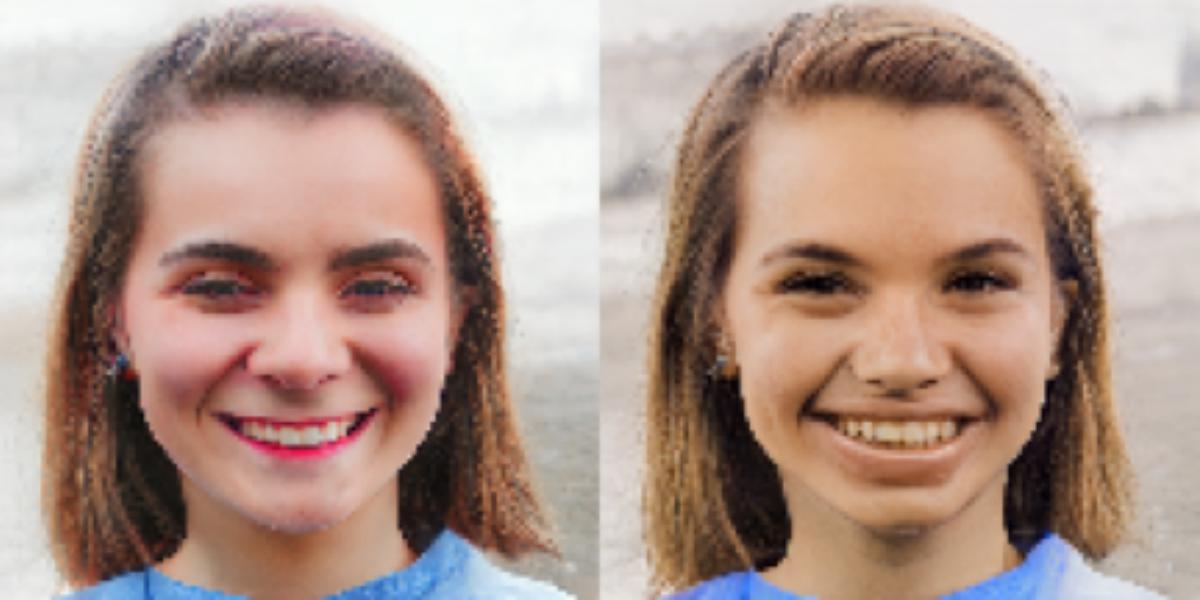} &
        \includegraphics[width=0.19\textwidth]{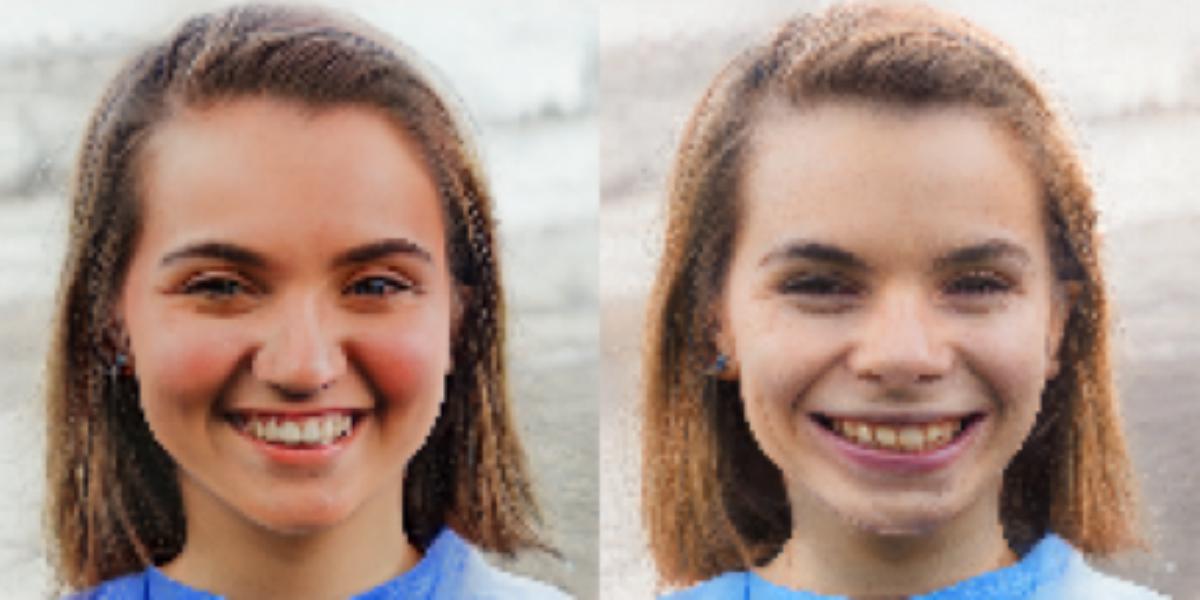} &
        \includegraphics[width=0.19\textwidth]{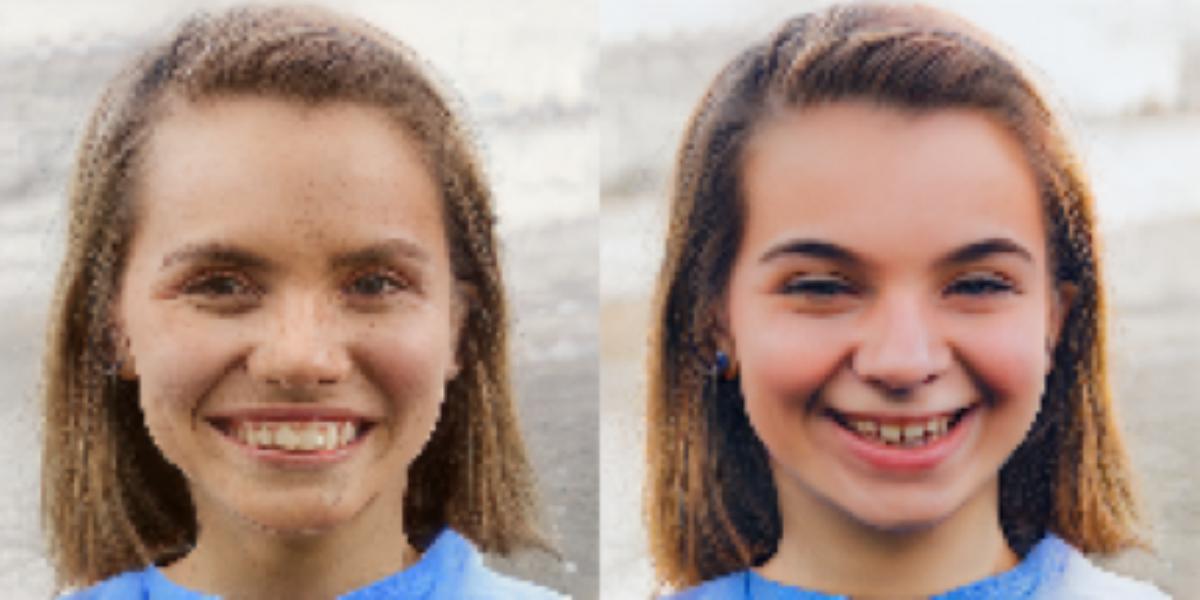} &
        \includegraphics[width=0.19\textwidth]{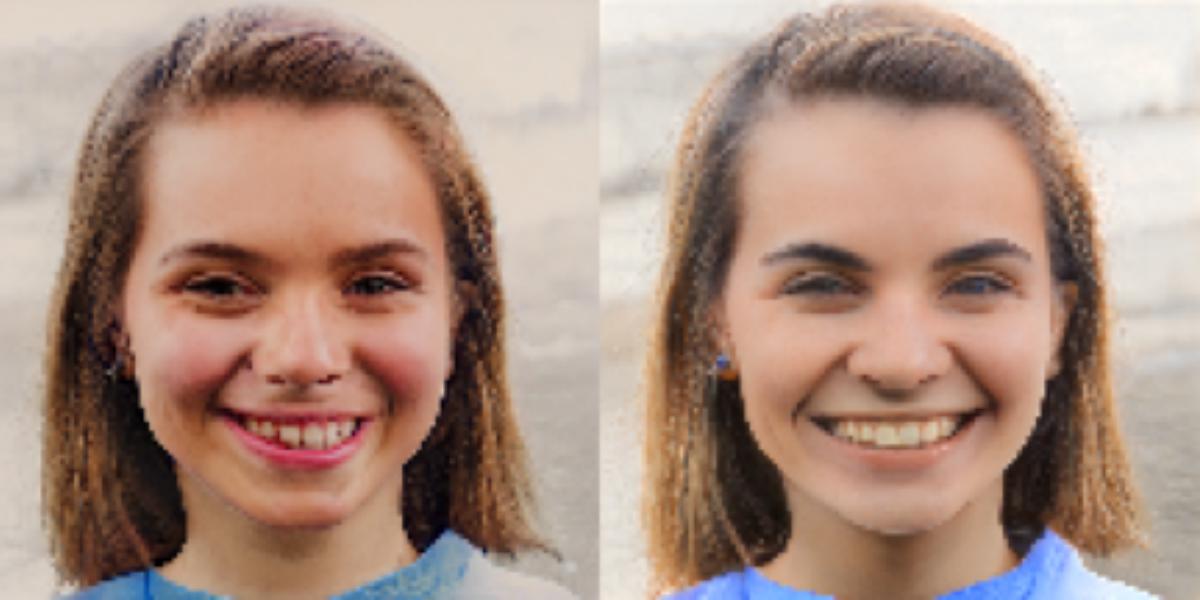} \\
        & Curvy and wide smile & Curvey and narrow smile & Smile width & Curvy and wide smile & Asymmetric smile\\
         \rotatebox{90}{Mouth shape (id)}& 
        \includegraphics[width=0.19\textwidth]{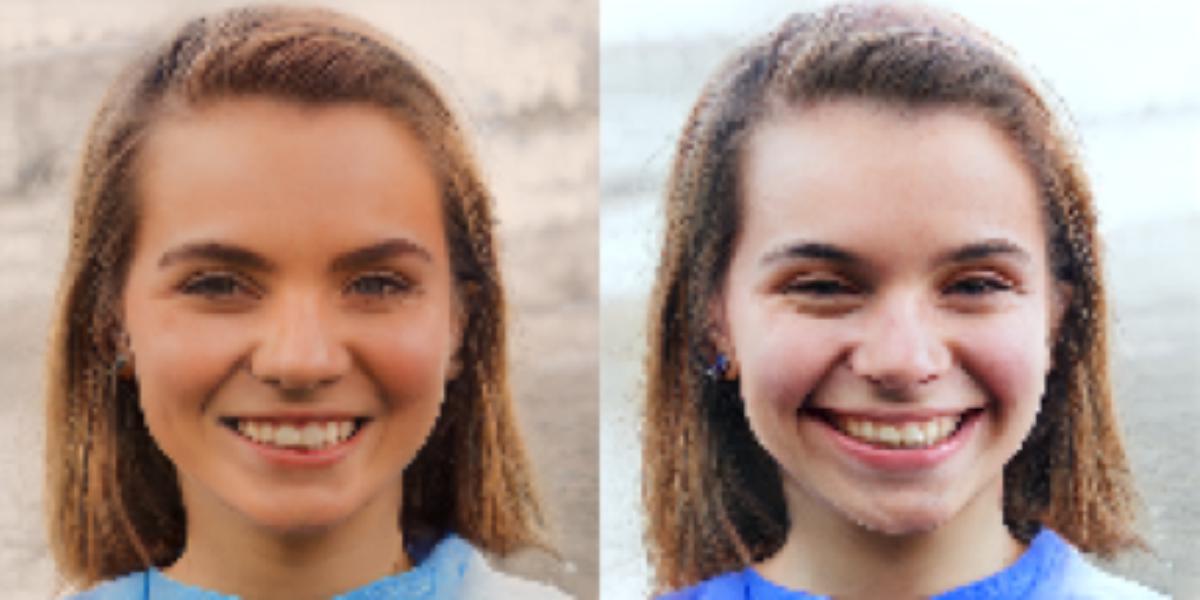} &
        \includegraphics[width=0.19\textwidth]{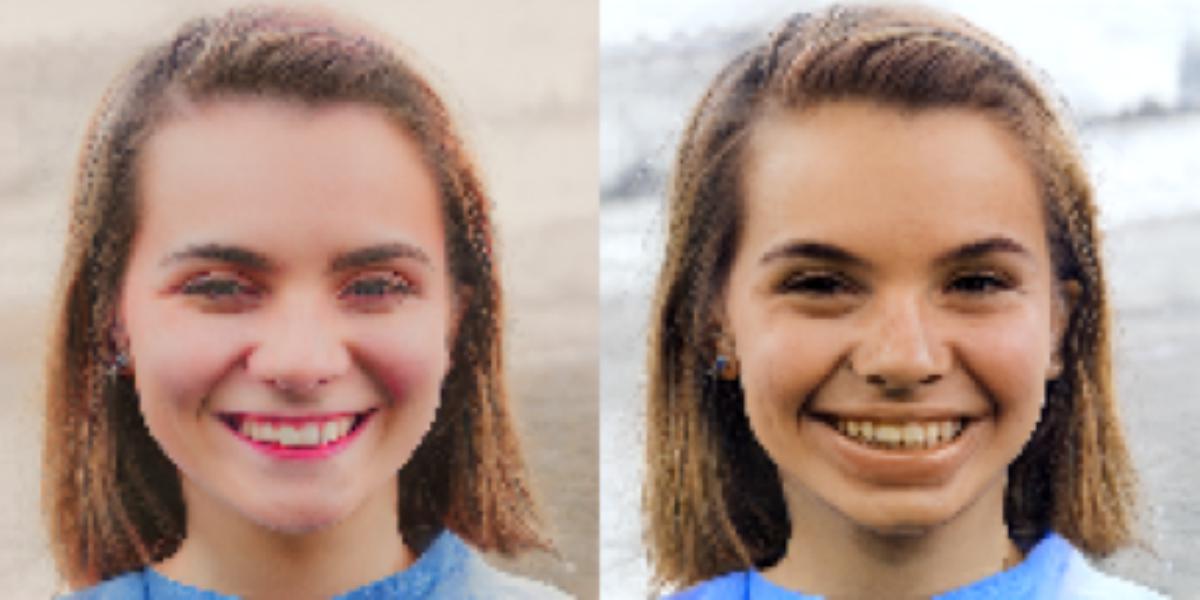} &
        \includegraphics[width=0.19\textwidth]{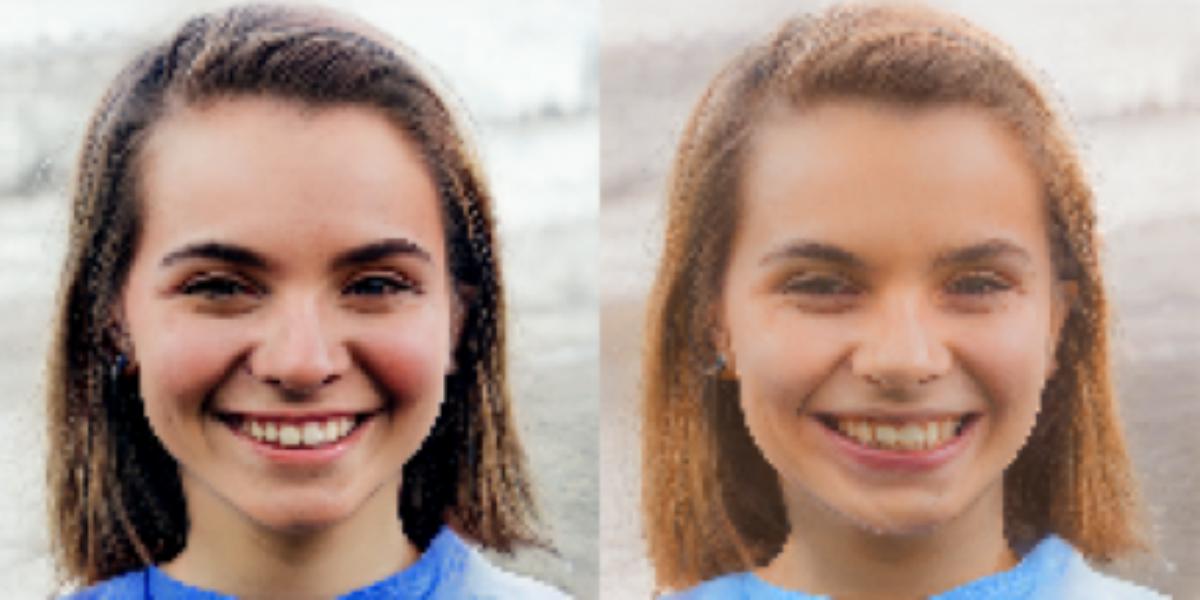} &
        \includegraphics[width=0.19\textwidth]{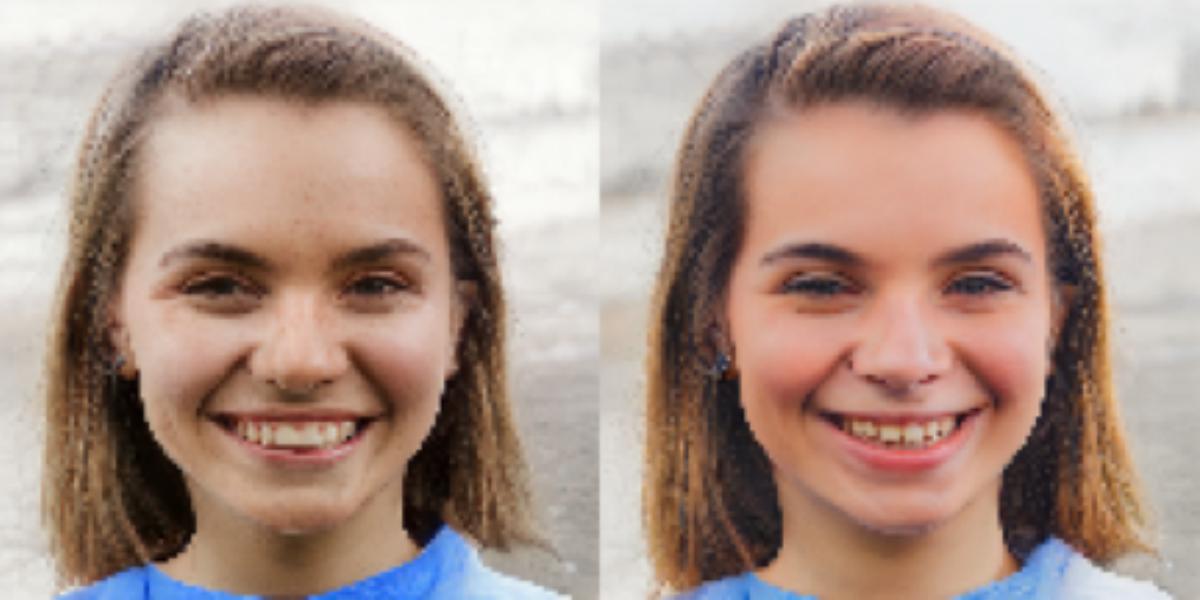} &
        \includegraphics[width=0.19\textwidth]{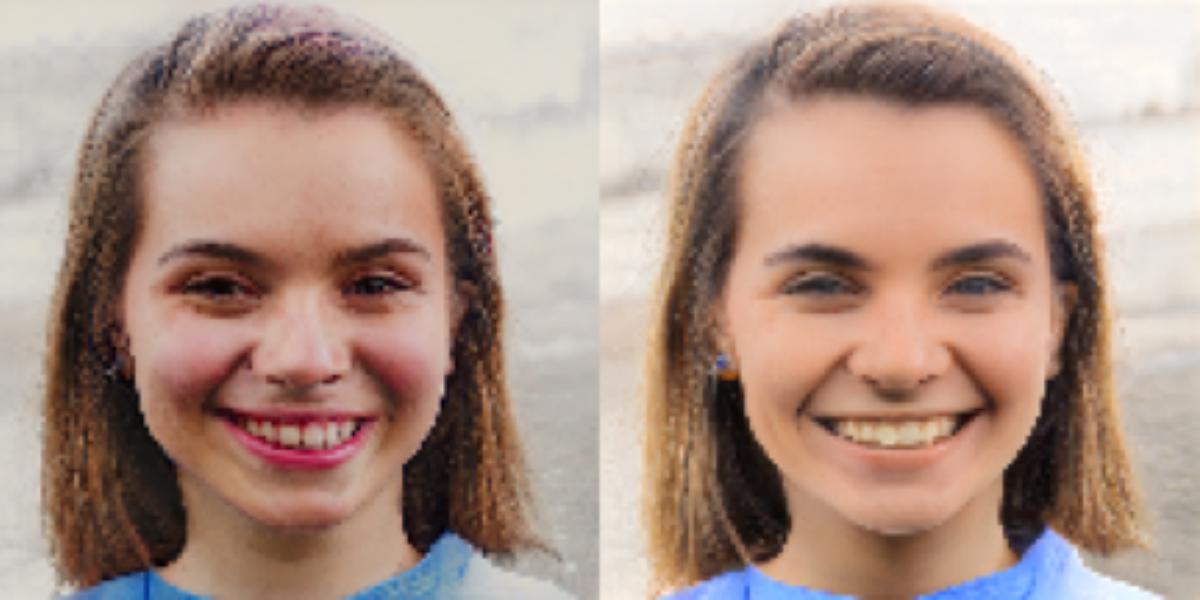} \\
        & Curvy and wide smile & Curvey and narrow smile & Smile width & Curvy and wide smile & Asymmetric smile\\
         \rotatebox{90}{Eye photomet}& 
        \includegraphics[width=0.19\textwidth]{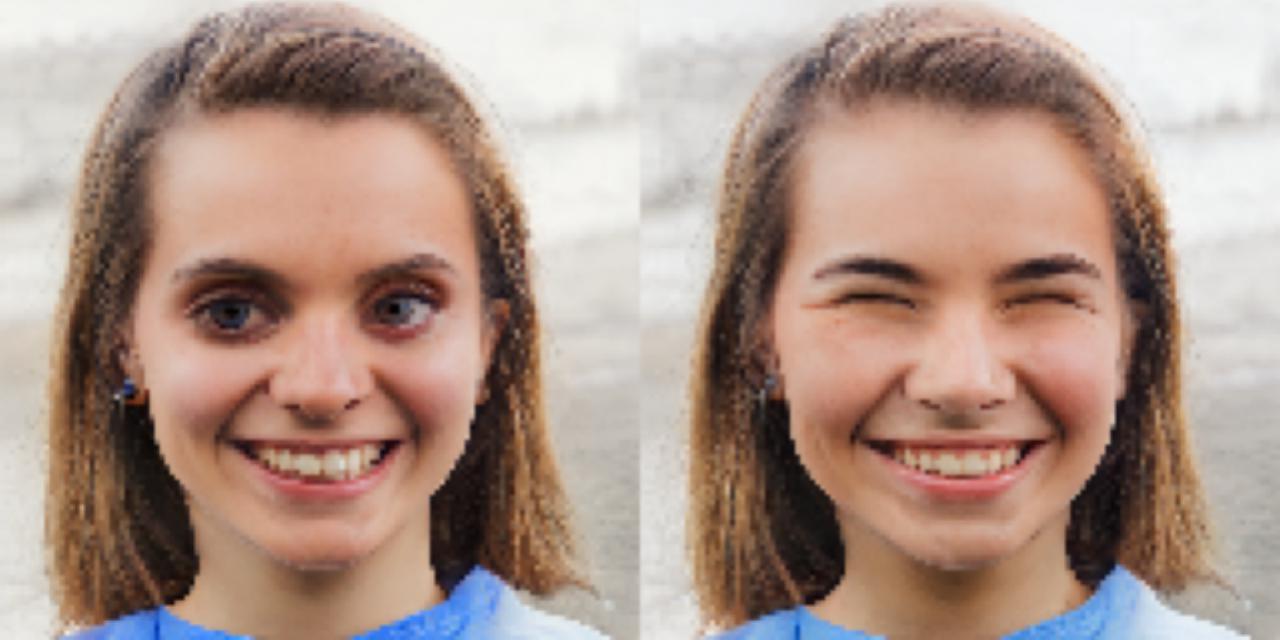} &
        \includegraphics[width=0.19\textwidth]{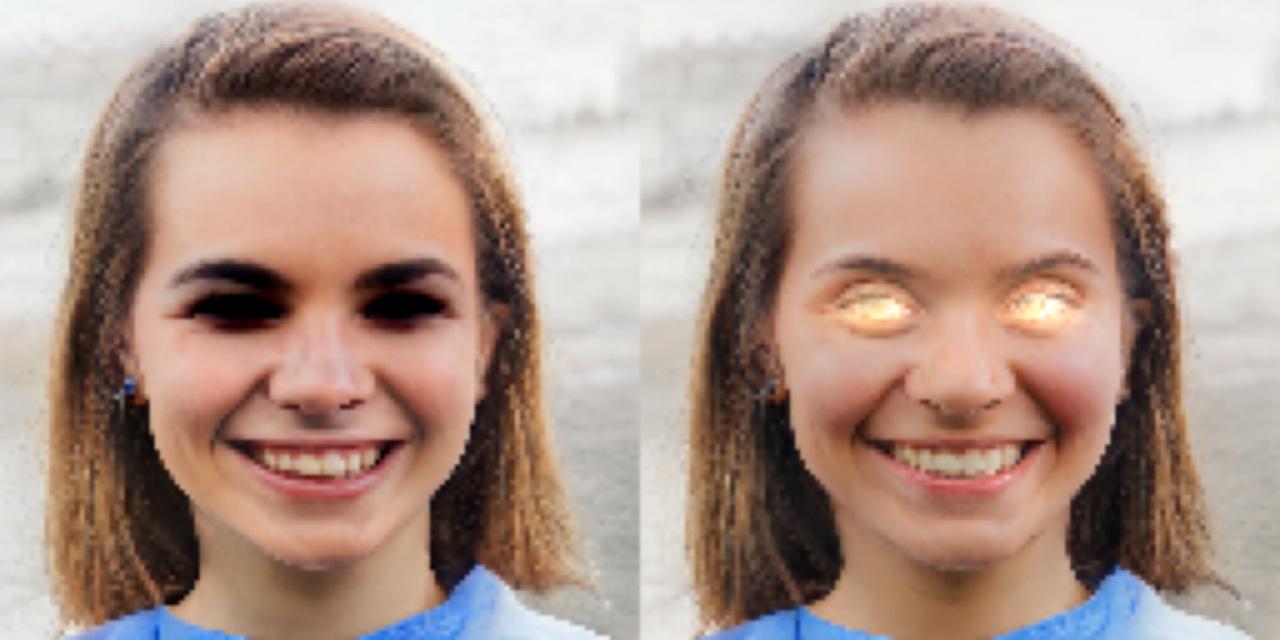} &
        \includegraphics[width=0.19\textwidth]{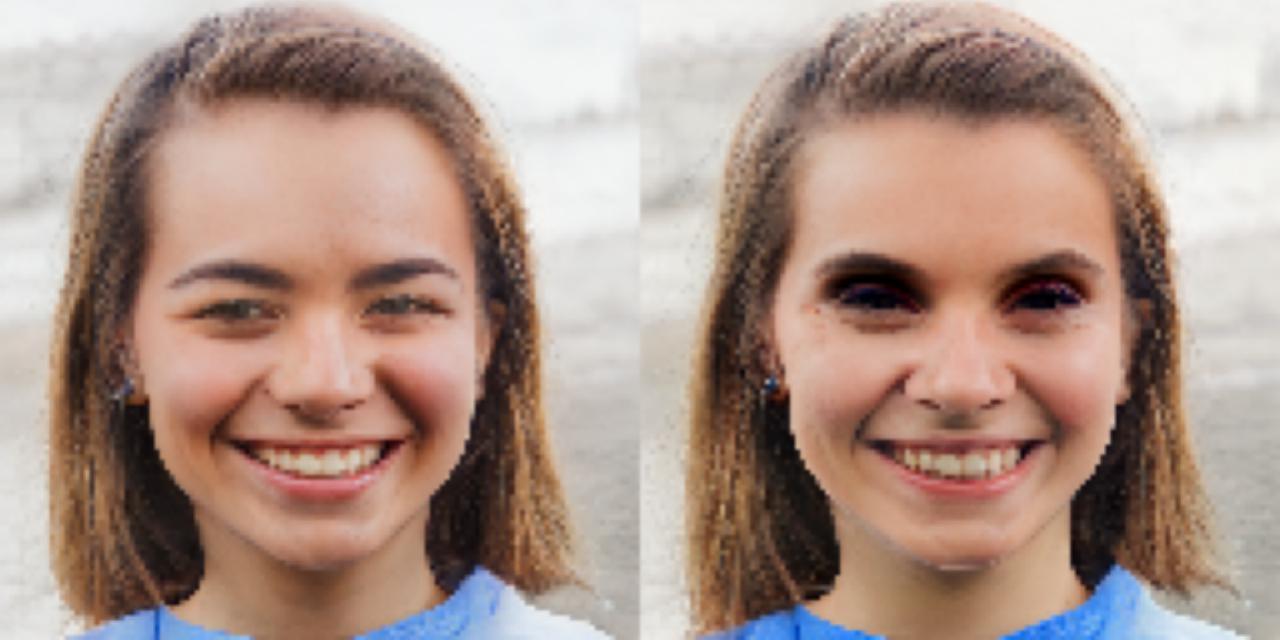} &
        \includegraphics[width=0.19\textwidth]{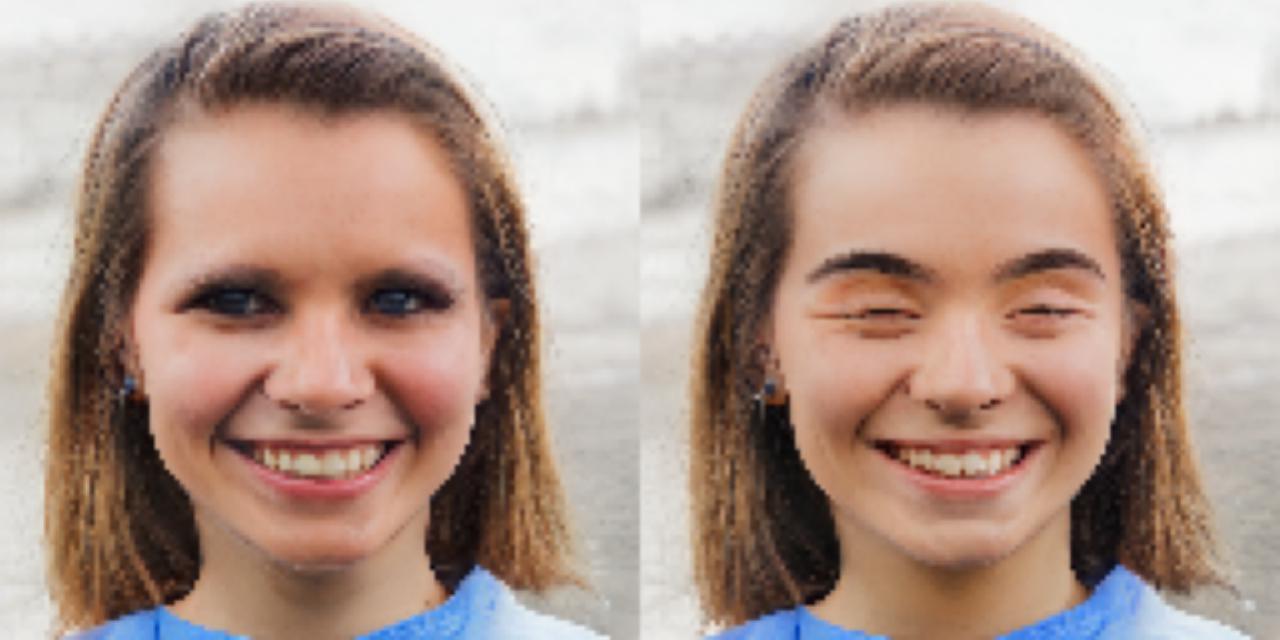} &
        \includegraphics[width=0.19\textwidth]{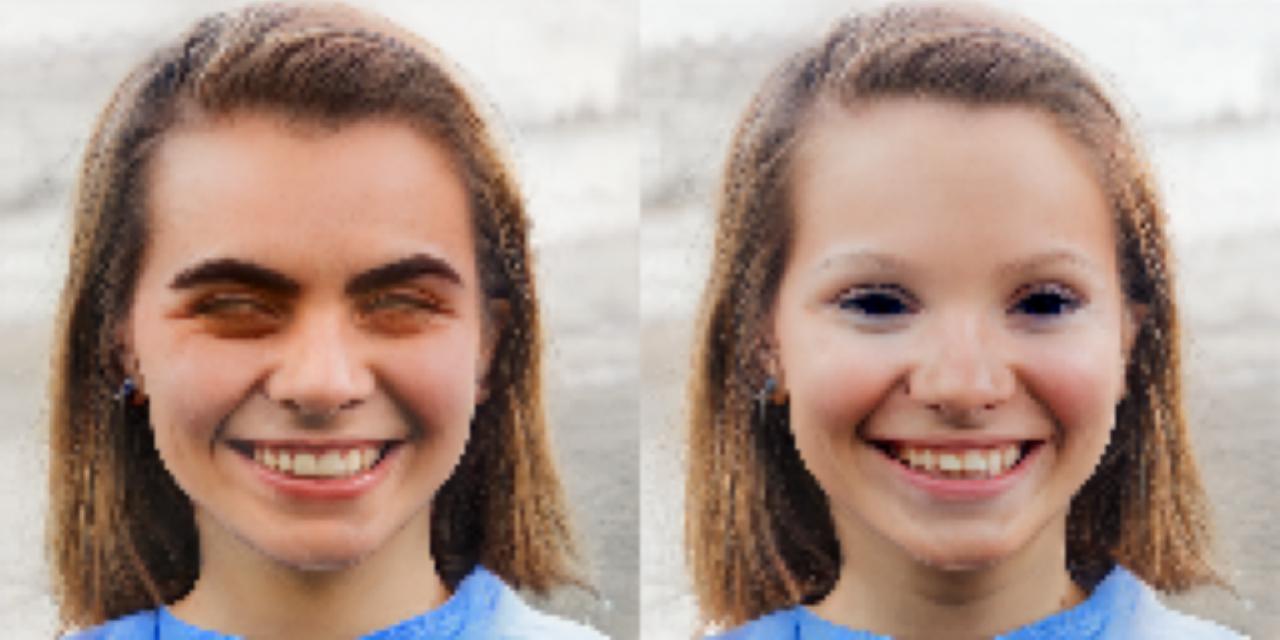} \\
        & Open/close & Dark/bright & Brow ridge & Eye height and brow & Eye width and brow\\
         \rotatebox{90}{Nose photomet}& 
        \includegraphics[width=0.19\textwidth]{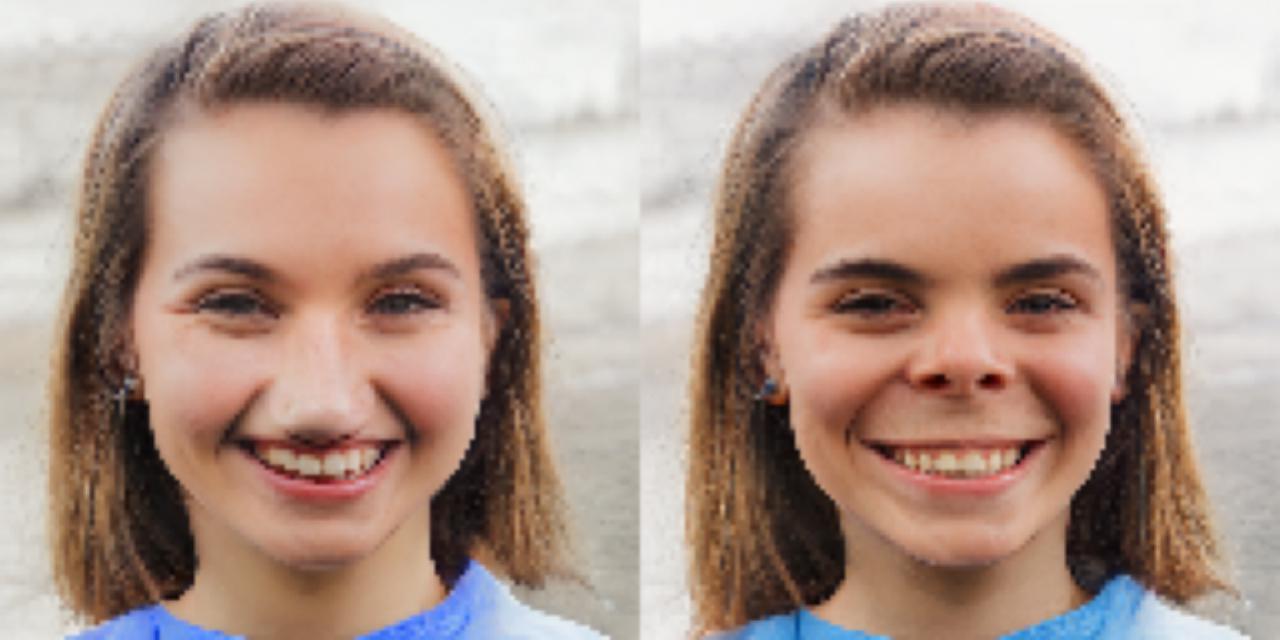} &
        \includegraphics[width=0.19\textwidth]{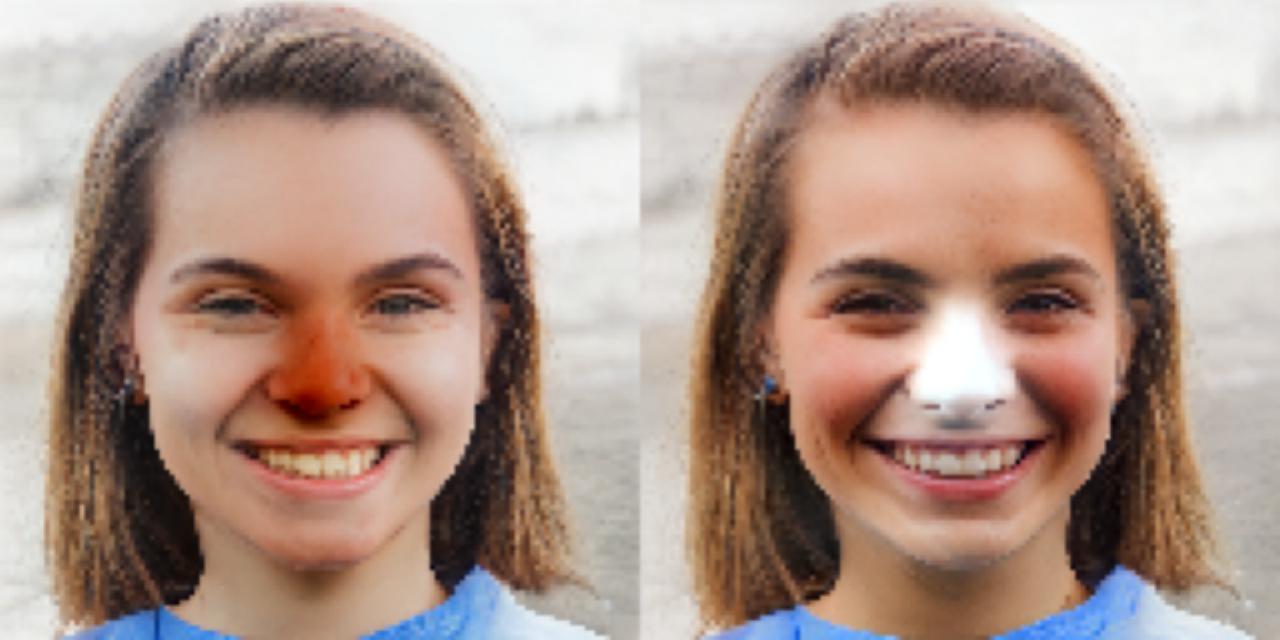} &
        \includegraphics[width=0.19\textwidth]{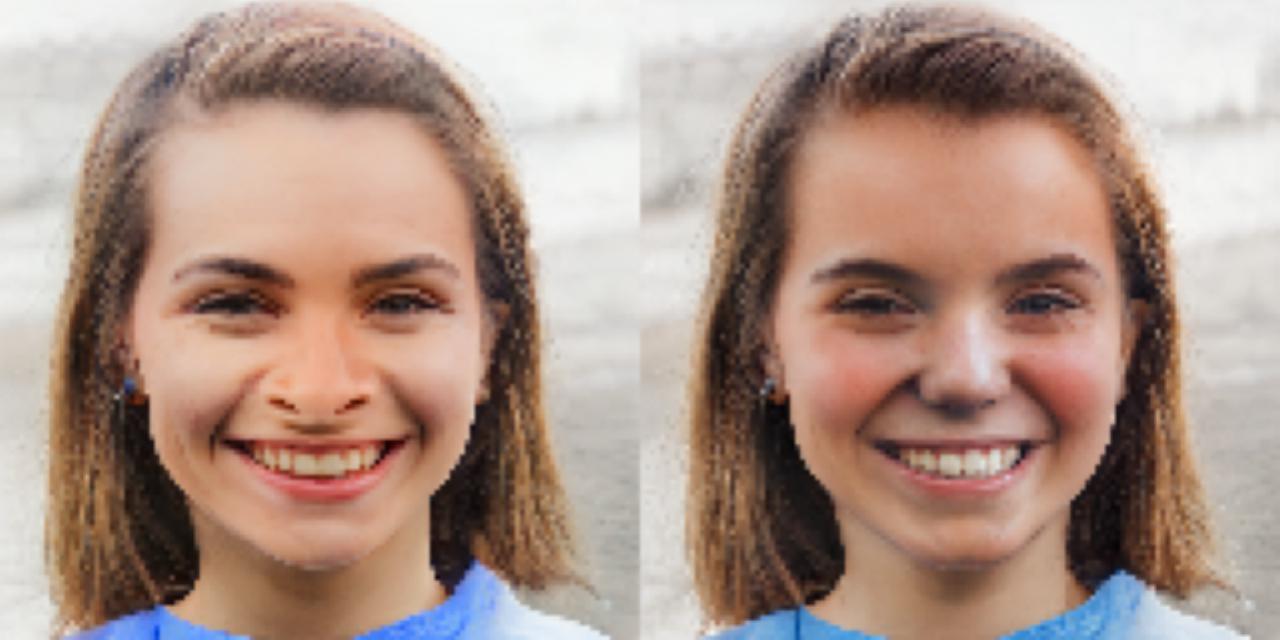} &
        \includegraphics[width=0.19\textwidth]{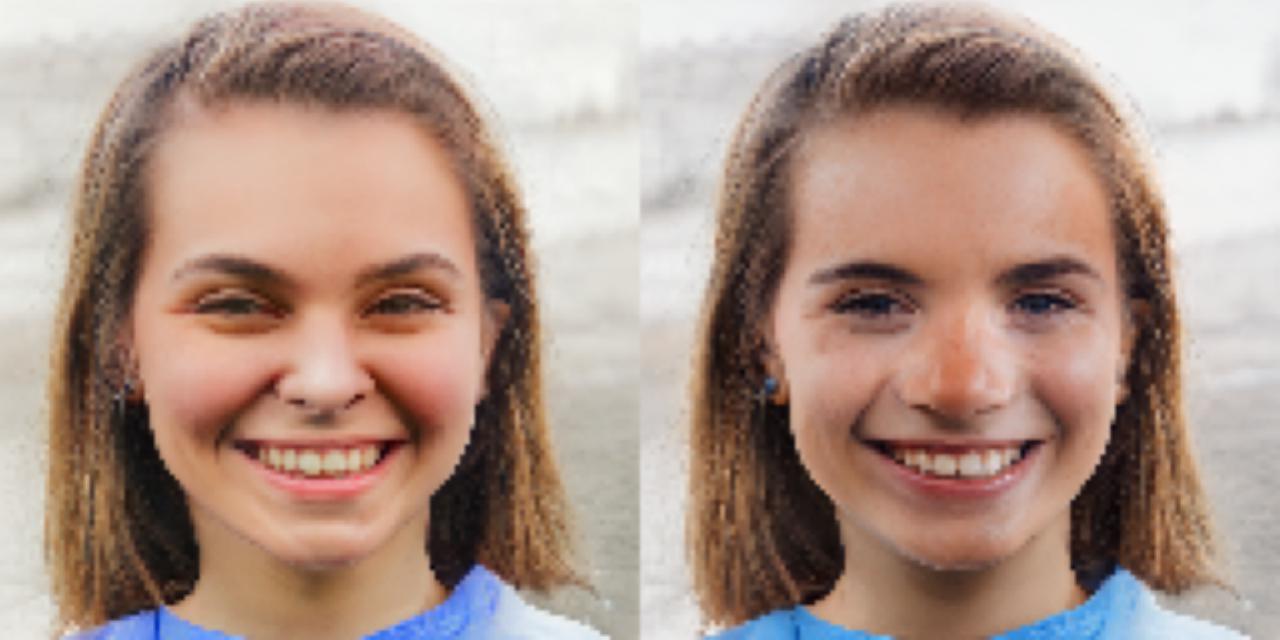} &
        \includegraphics[width=0.19\textwidth]{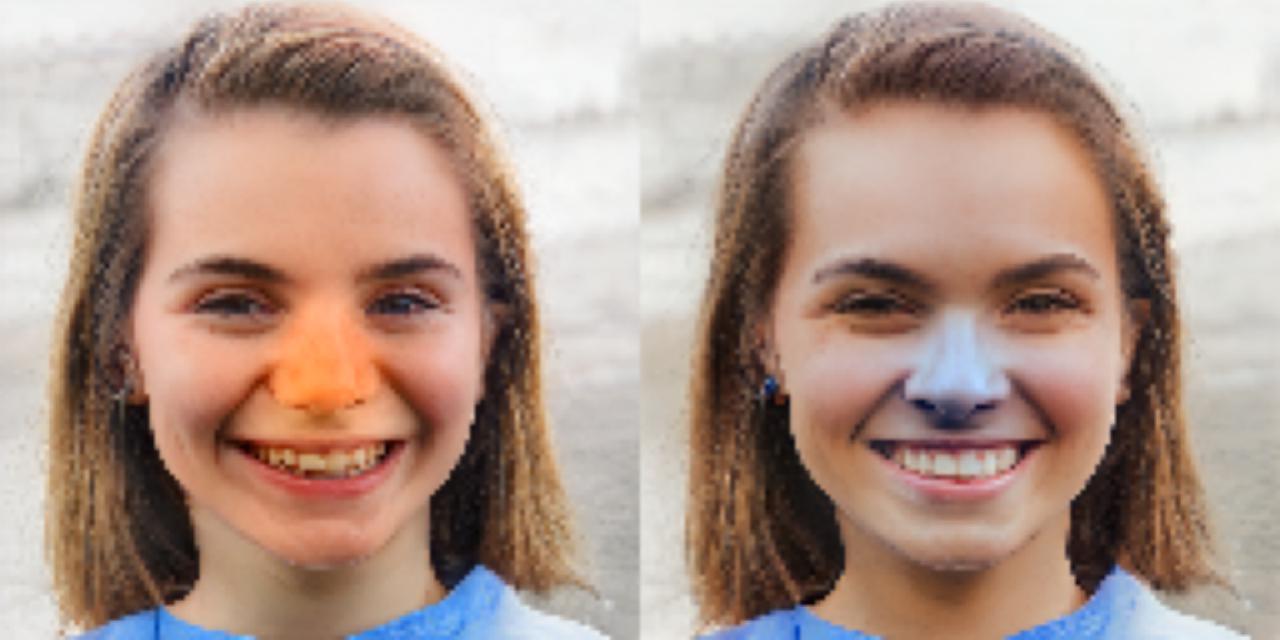} \\
        & Vertical length & Red/white & Nose size and nostril & Nose width & Orange/blue \\
         \rotatebox{90}{High frequency}& 
        \includegraphics[width=0.19\textwidth]{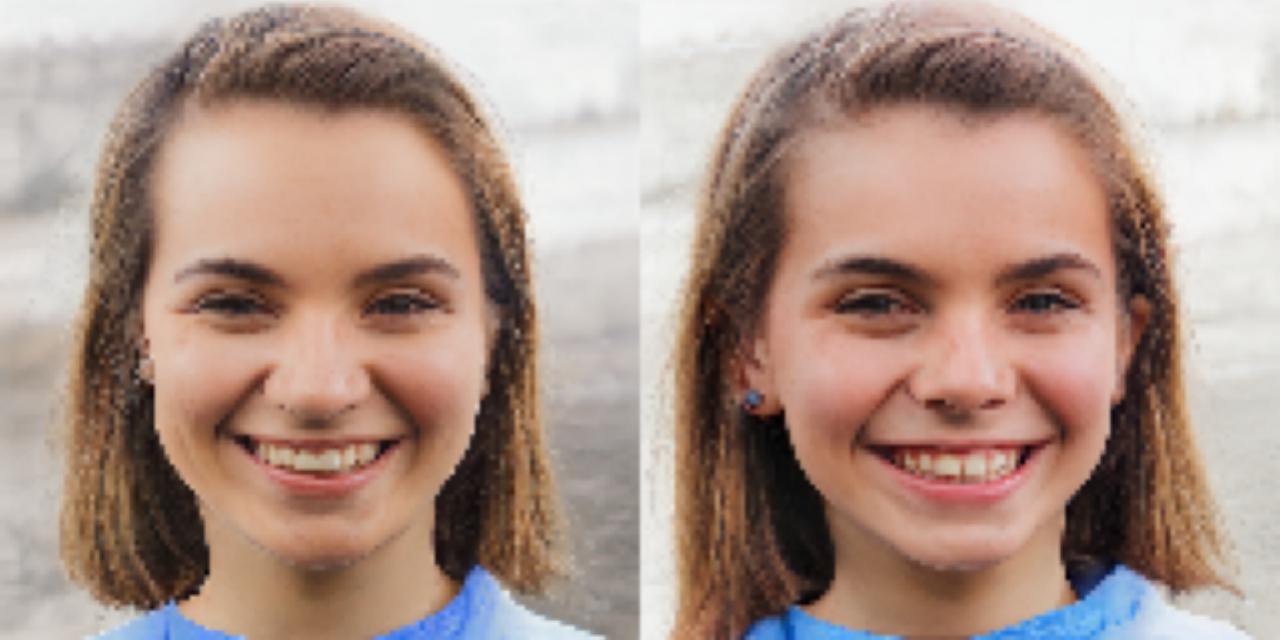} &
        \includegraphics[width=0.19\textwidth]{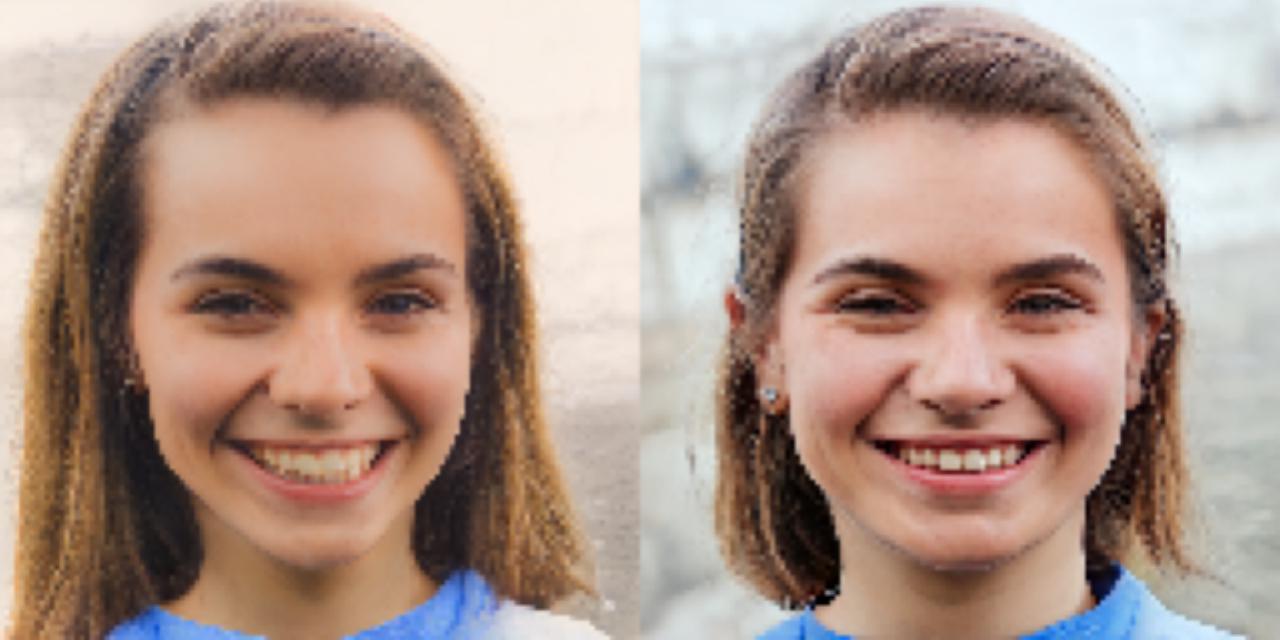} &
        \includegraphics[width=0.19\textwidth]{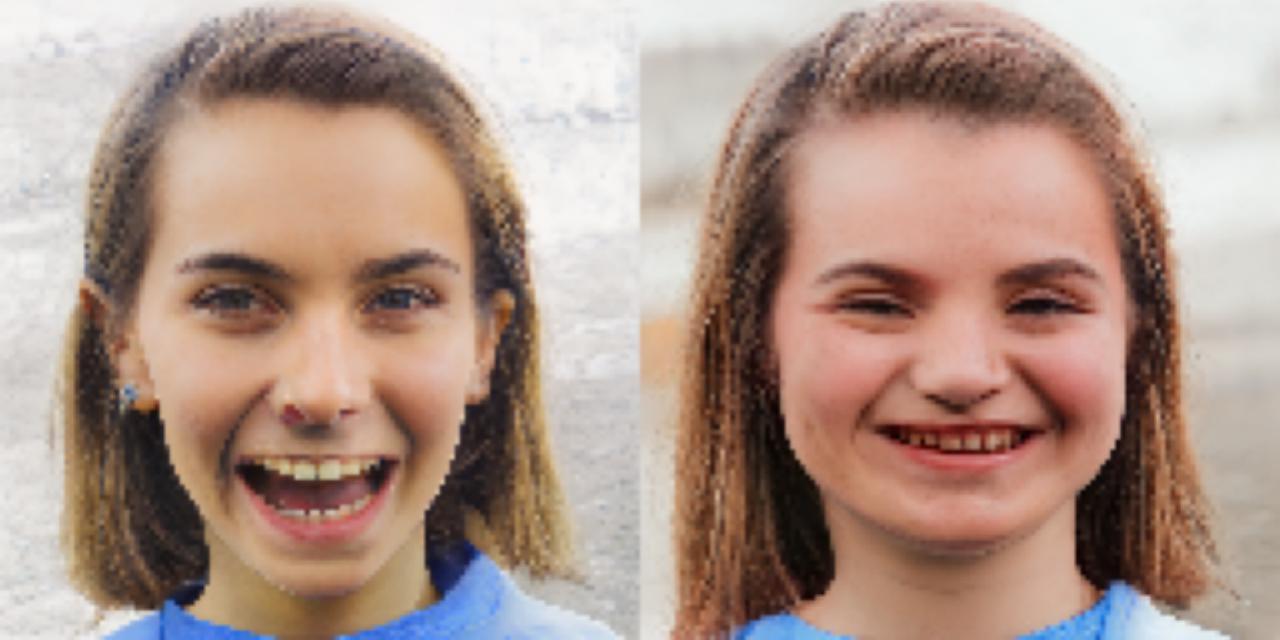} &
        \includegraphics[width=0.19\textwidth]{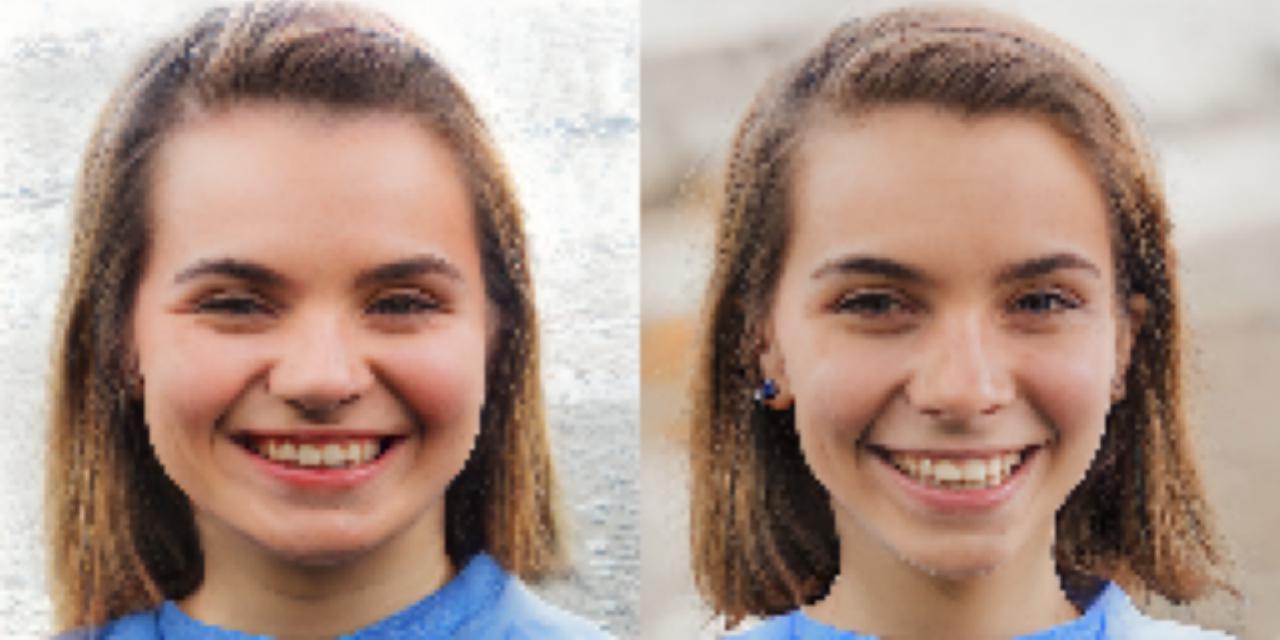} &
        \includegraphics[width=0.19\textwidth]{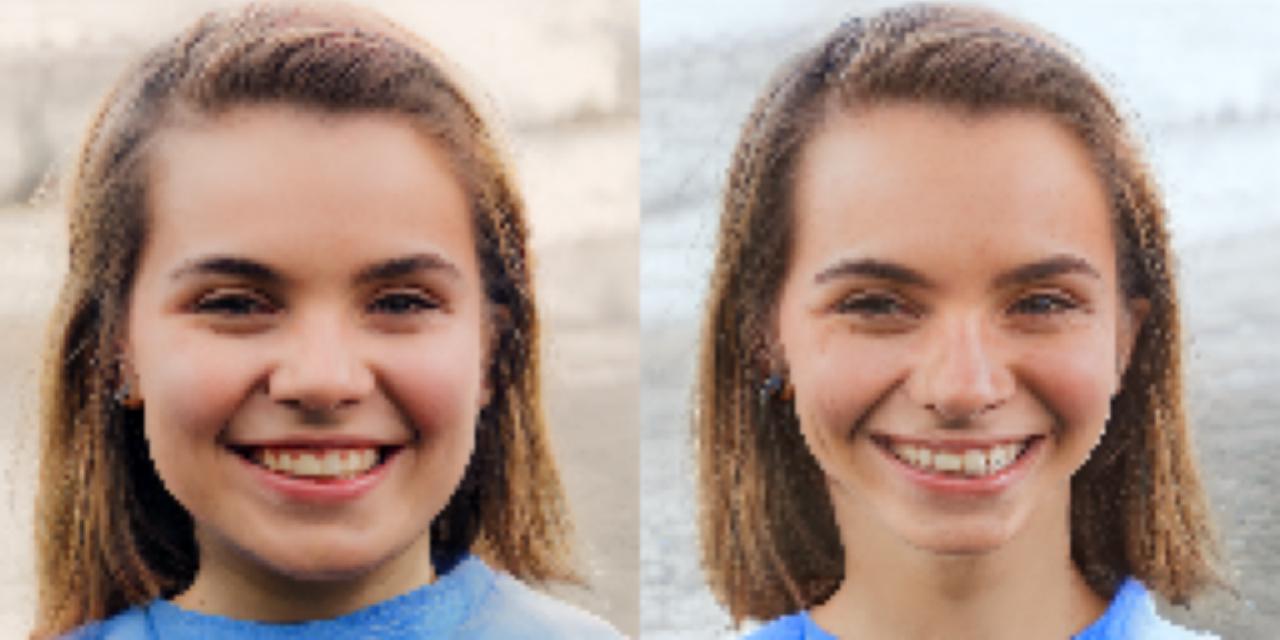} \\
    \end{tabular}
    \caption{Visualization of the discovered interpretable latent subspaces in $\Ss$. We select and visualize the manipulation effects of 5 representative principle components in each subspace. The figure is trimmed to save space and please check \Cref{fig:exp-subspace-full} in the appendix for the full version. See \Cref{sec:exp-subspace} for details.}
    \label{fig:exp-subspace}
\end{figure*}

We can also control to change the dominant edges by constructing a criterion reflecting the high frequency component of an image. Define $f_\text{low}(x)$ as the low-frequency component of an image, which can be obtained by Discrete Wavelet Transform (DWT) or just up/down-sampling. We can then denote the low and high frequency components as
\begin{align}
    h_\text{low}^m &= f_\text{low}(g(u)),\\
    h_\text{high}^m &= g(u) - h_\text{low}(g(u)),
\end{align}
As will be shown in \Cref{sec:experiments} and \Cref{fig:exp-subspace}, this criterion discovers interesting face deformation subspaces.
   
\section{Counterfactual Analysis on Face Attribution}
\label{sec:counterfactual}

By constructing interpretable criterions, we can solve various interpretable latent subspaces to control image generation.
With the discovered subspaces, we can now answer the question in our title, i.e. we can execute counterfactual analysis to explain how the attractiveness perceived by a classifier is affected by the modification in interpretable latent subspaces.


Consider a latent space sample $u$ and a face attribution classifier $\logit(x)$. A counterfactual can be obtained by gradient descent to optimize a perturbation $\Delta u$ on the classifier output $\logit(g(u + \Delta u))$. To guarantee that the generated counterfactuals have interpretable semantics, we restrict $\Delta u \in \Span(V)$, where $V$ denotes the bases of any subspaces or their intersection as derived in \Cref{sec:criterions}.

\Cref{fig:teaser} illustrates an example of generated counterfactuals on a given sampled image. The counterfactuals are generated in three latent subspaces: eye shape, mouth shape, and lip color. See \Cref{tab:exp-formulations} in the appendix for their formulations.
By optimizing $\logit(g(u + \Delta u))$ in each derived subspace, we can obtain counterfactuals with concrete semantics as restricted by the corresponding latent subspaces. The attractiveness score perceived by the classifier is increased in each counterfactual and the modifications are restricted to preserve intepretability and locality. We notice that the resulted bigger eyes, smaller mouth, and red lips are also consistent to common sense of attractiveness. We will discuss more examples in the experiment section.


\section{Experiments}
\label{sec:experiments}

\subsection{Experiment Setup}
\label{sec:exp-setup}
We take the official StyleGAN2 PyTorch implementation~\cite{karras2020training} and the $1024\times1024$ model pre-trained on FFHQ as our generator. We use a face parsing pre-trained model~\cite{zllrunning2019face} based on BiSeNet and a pre-trained 68-facial-landmark detector from~\cite{bulat2017far}. The face attribute classifier to explain is trained on CelebA dataset~\cite{liu2015faceattributes} using~\cite{d2019face}.    

We split an eigenvector matrix into $V$ and $W$ according to relative eigenvalue magnitudes. Note that the modification magnitude is coupled by the eigenvalue and the perturbation magnitude. To uniformly reflect the modification magnitude in different criterions, we define a relative eigenvalue as follows. Denote the largest eigenvalue as $\lambda_0$. Eigenvectors whose eigenvalues are smaller than $\epsilon\lambda_0$ are put in $W$ and the rest are kept in $V$. $\epsilon$ is selected between $10^{-3}\text{-}10^{-2}$ to trade-off the change of modified areas and the stability of the suppressed areas. 

\subsection{Latent Subspace Discovery}
\label{sec:exp-subspace}

\subsubsection{Comparison of Latent Space}
\label{sec:comparison-latent-space}
In \Cref{sec:subspace-discovery}, our formulation is capable to solve latent subspaces in different latent spaces of $\Ss, \WW+, \WW$ and $\ZZ$. In our experiments, we mainly explore the latent subspaces in $\Ss$, since it possesses better disentanglement and richer semantics. Readers can find qualitative and quantitative comparison between the 4 latent spaces in \Cref{sec:space-comparison-sup} in the appendix.

\subsubsection{Discovered Subspaces}
\label{sec:exp-subspace}

In this section, we construct 12 different formulations to restrict latent subspaces to show the effectiveness of our approach in discovering subspaces. The corresponding subspace formulations are summarized in \Cref{tab:exp-formulations}. For each criterion, we solve the subspace on 1-2 training images, select $5$ representative principle components, and visualize the modified images in \Cref{fig:exp-subspace}. We exaggerate the control magnitude to make the modification more recognizable for readers. As pointed in~\cite{wu2021stylespace}, such exaggeration might introduce artifacts and harm image quality. Several interesting properties are observed from the visualization.

\begin{itemize}
    \item The visualized principle components represent the modification directions with top magnitudes, but do not necessarily correspond to realistic face attributes. For example, we see various unrealistic colors are decomposed for the face skin and the nose. It is easy to understand that we can obtain realistic colors in the subspace by linearly combining the components as bases.
    \item We observe that modification overflows to unintended regions in some cases, indicating that the disentanglement is not perfect. We notice that such overflow happens mainly in cases where landmark-related criterions are used, e.g. face boundary and eye boundary. The fact that we exaggerate the modification magnitudes also increases the overflow.
    \item We illustrate the effect of face identity features by comparing subspaces ``mouth shape'' and ``mouth shape (id)''. The restriction on face identity effectively reduces the unintended modification and generates more similar faces. We observe that the reduced modification is mainly on eyes and mouth.
    
\end{itemize}

\begin{figure*}
    \centering
    \footnotesize
    \setlength{\tabcolsep}{1pt}
    \begin{tabular}{ccccccccc}
        \includegraphics[width=0.12\textwidth]{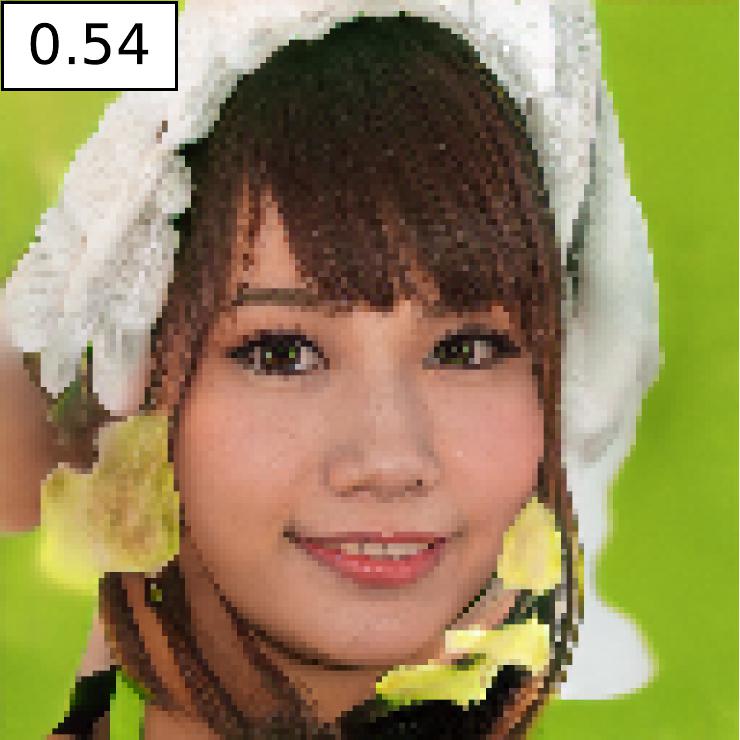} &
        \includegraphics[width=0.12\textwidth]{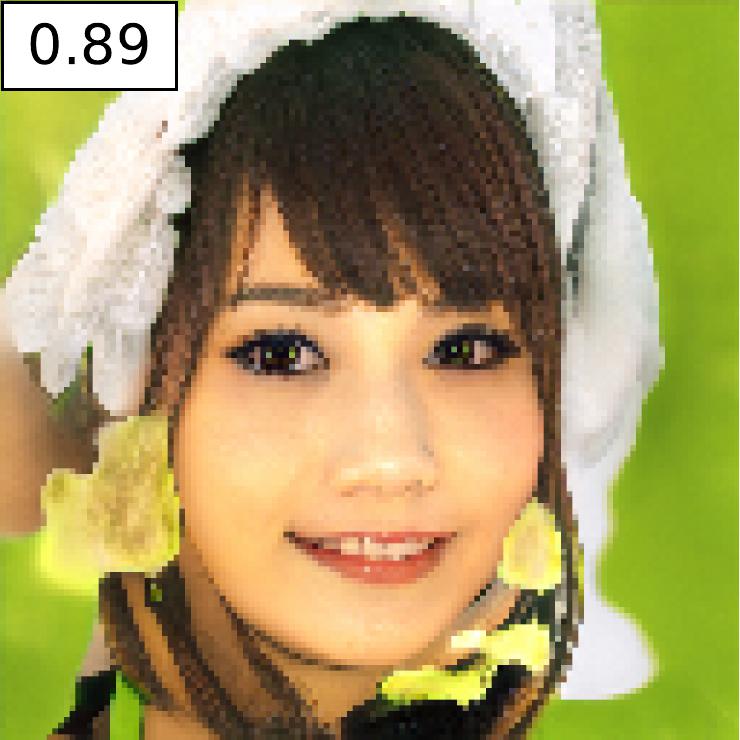} &
        \includegraphics[width=0.12\textwidth]{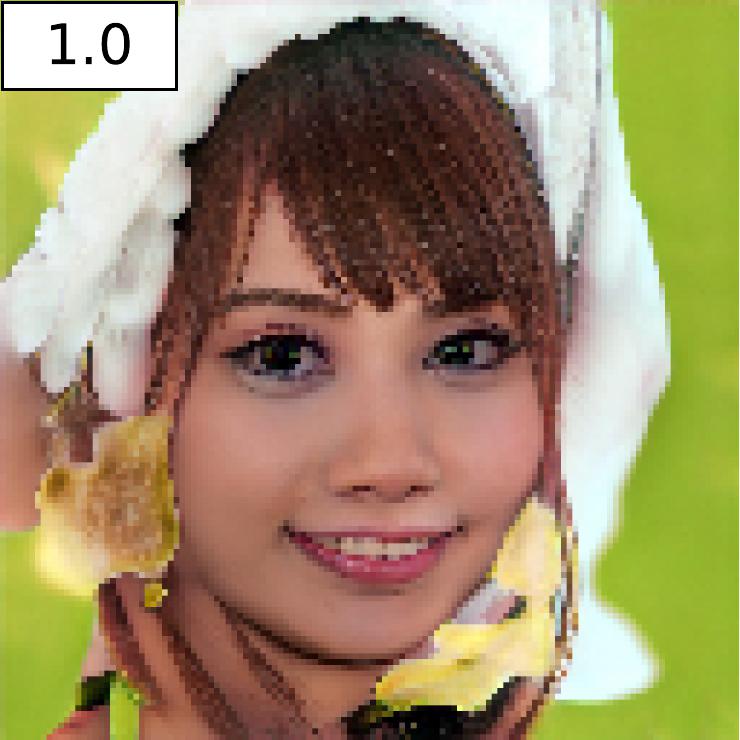} &
        \includegraphics[width=0.12\textwidth]{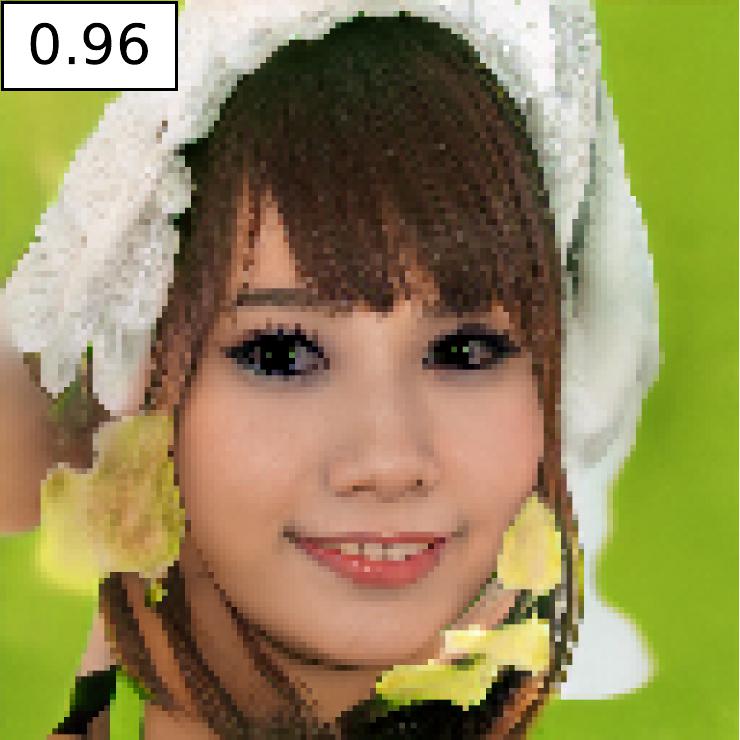} &
        \includegraphics[width=0.12\textwidth]{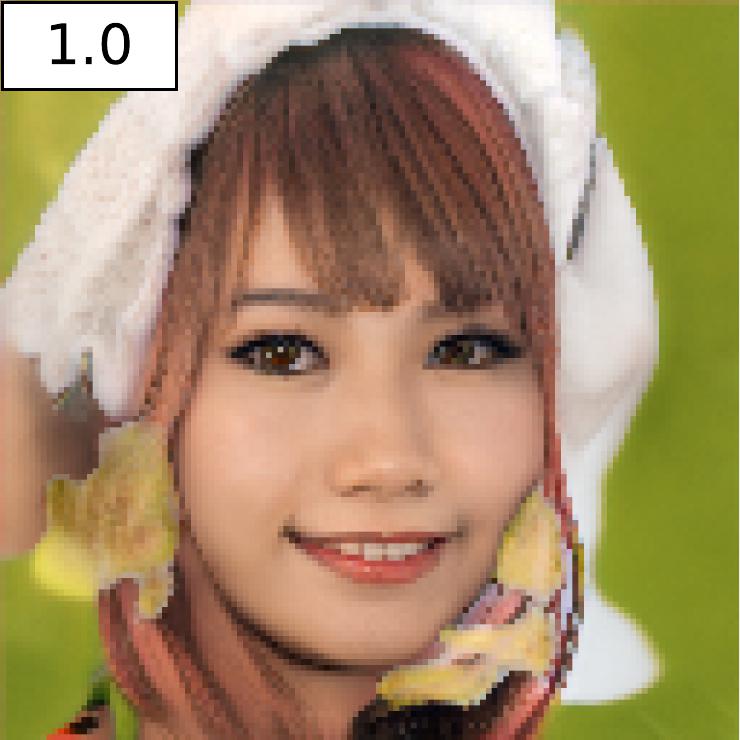} &
        \includegraphics[width=0.12\textwidth]{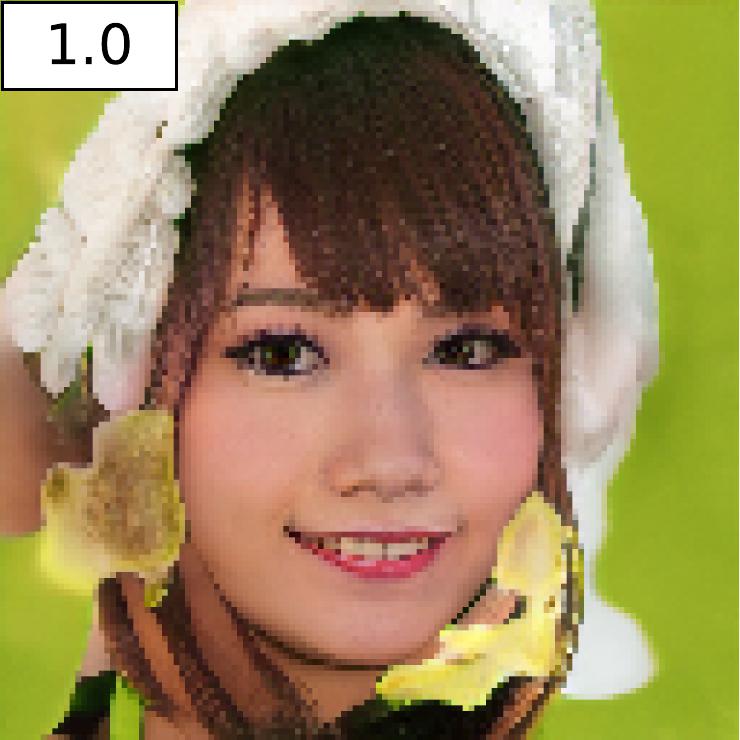} &
        \includegraphics[width=0.12\textwidth]{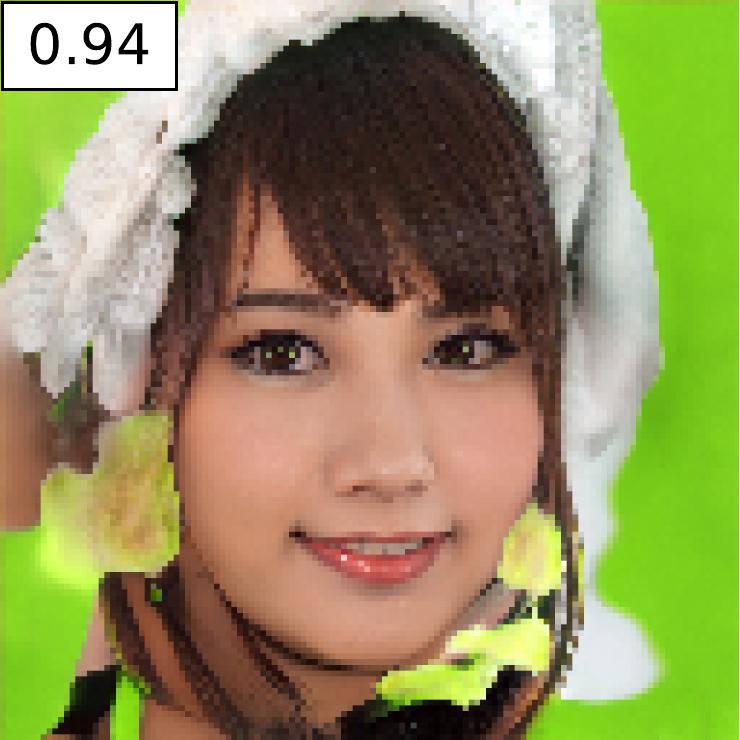} &
        \includegraphics[width=0.12\textwidth]{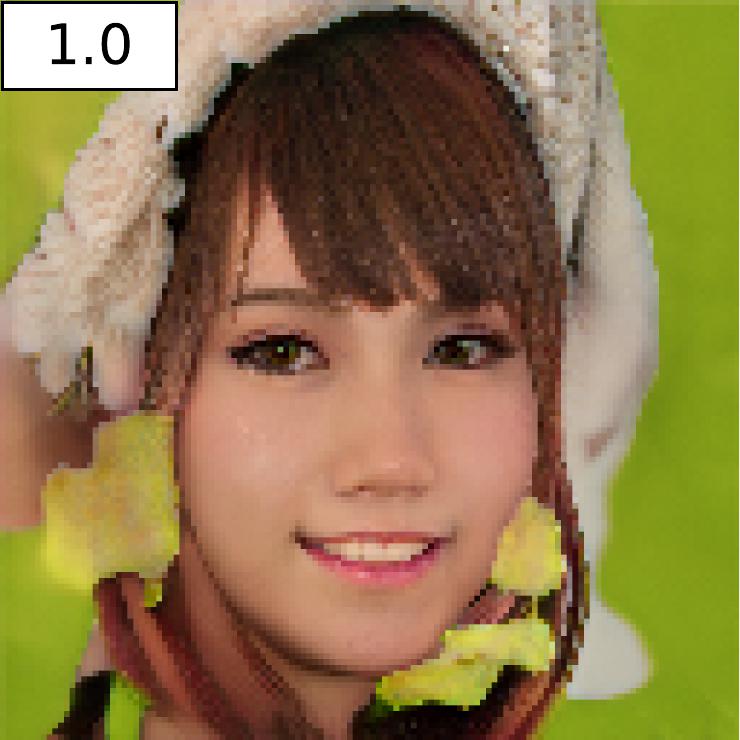} &
        \\
                \includegraphics[width=0.12\textwidth]{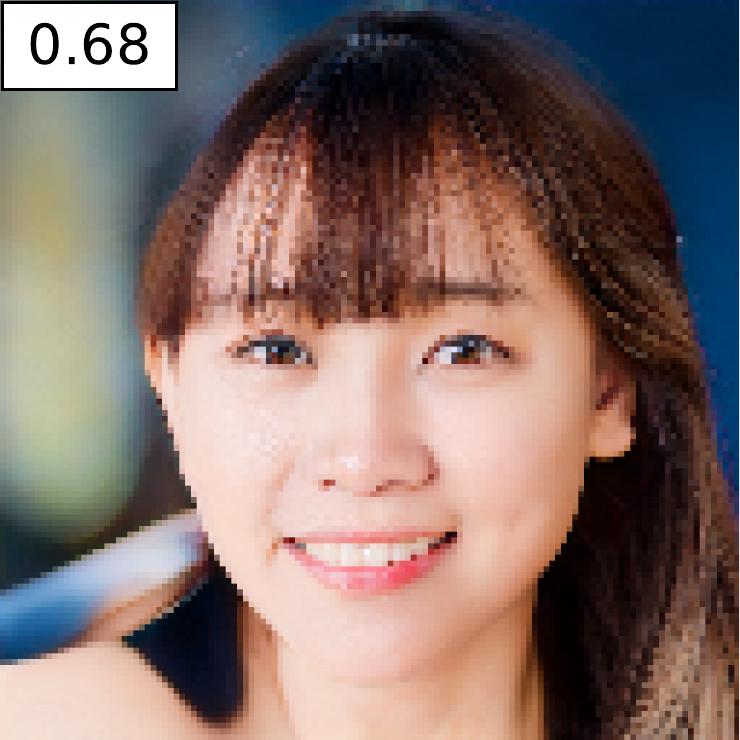} &
        \includegraphics[width=0.12\textwidth]{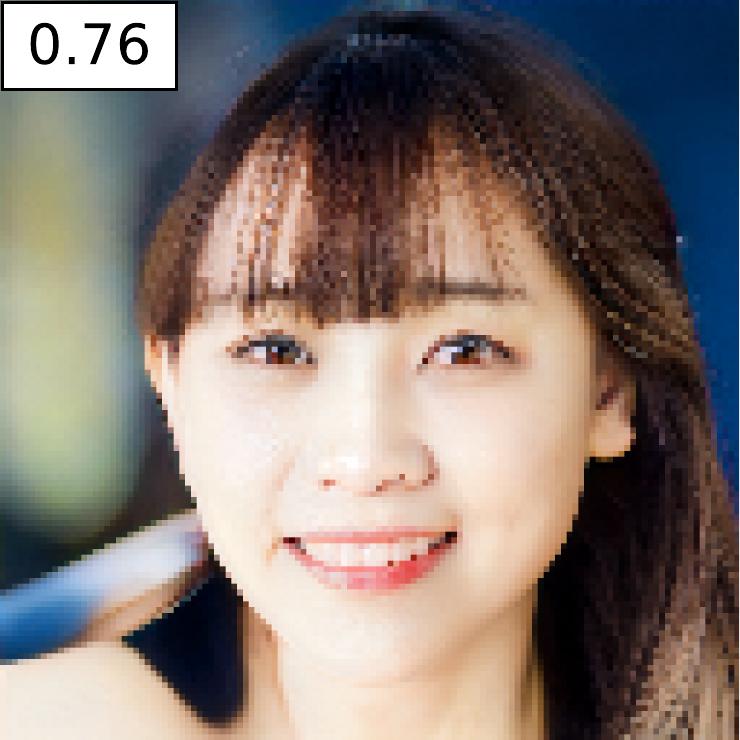} &
        \includegraphics[width=0.12\textwidth]{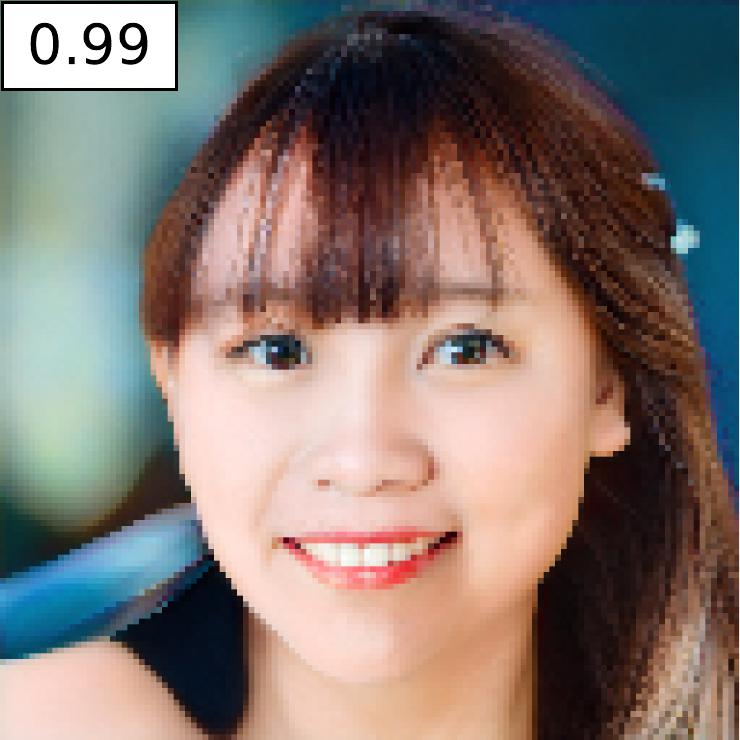} &
        \includegraphics[width=0.12\textwidth]{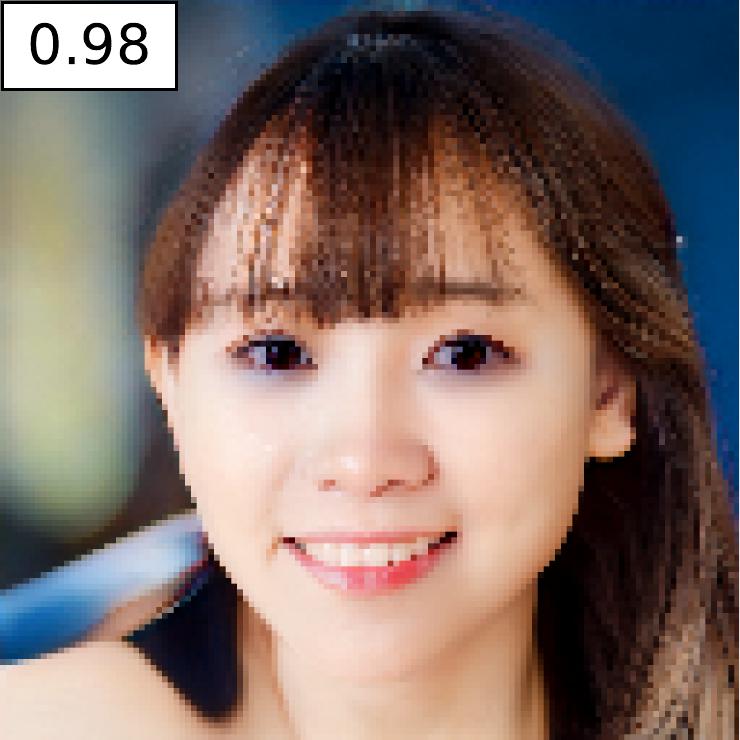} &
        \includegraphics[width=0.12\textwidth]{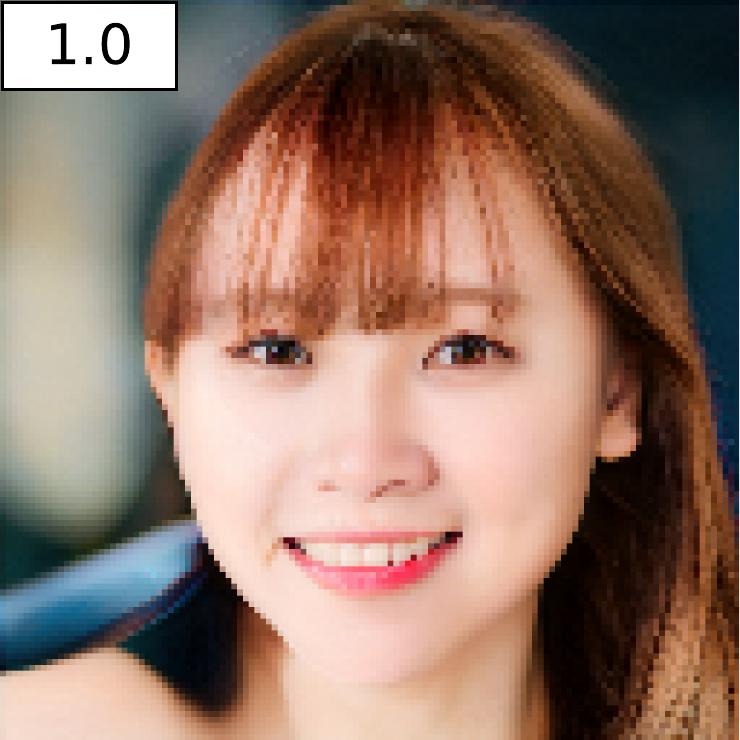} &
        \includegraphics[width=0.12\textwidth]{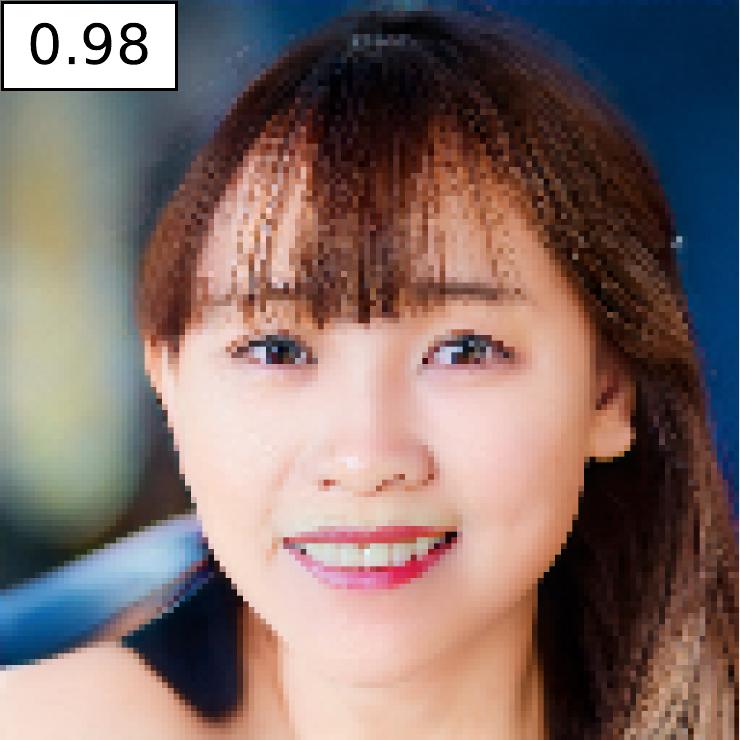} &
        \includegraphics[width=0.12\textwidth]{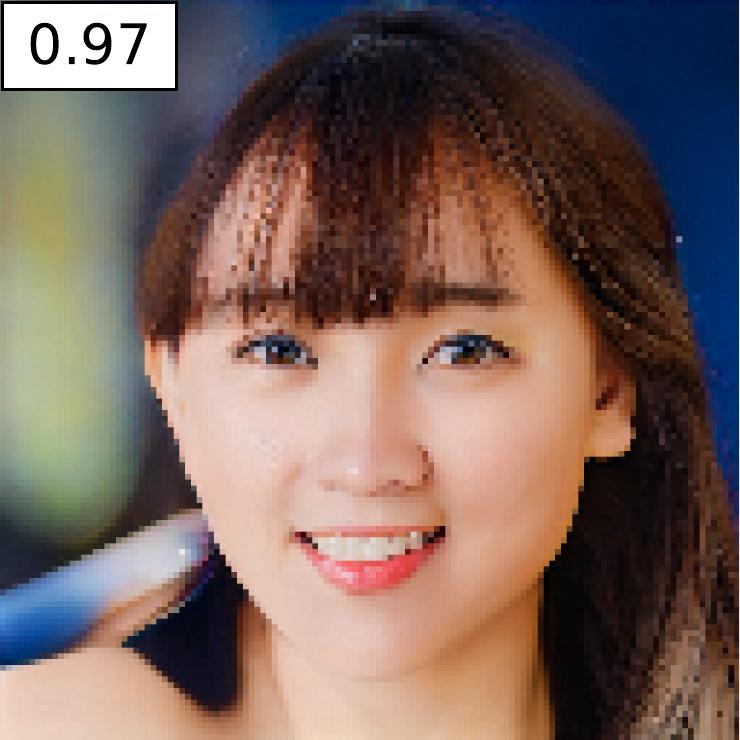} &
        \includegraphics[width=0.12\textwidth]{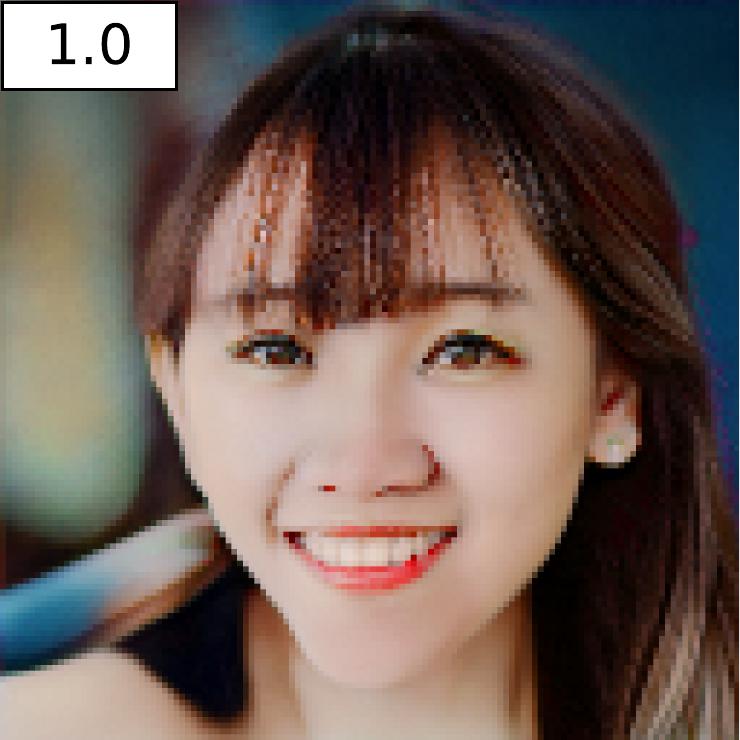} &
        \\
                \includegraphics[width=0.12\textwidth]{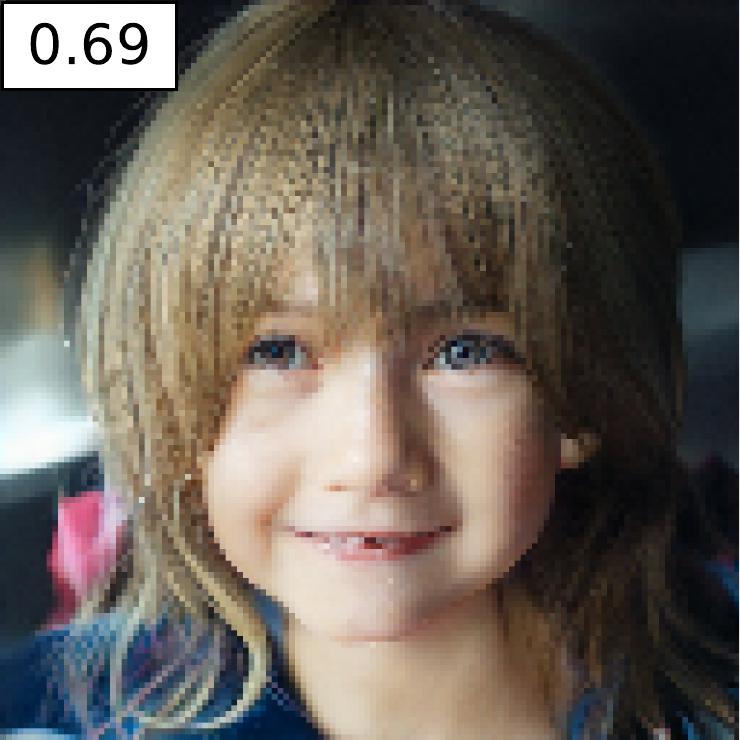} &
        \includegraphics[width=0.12\textwidth]{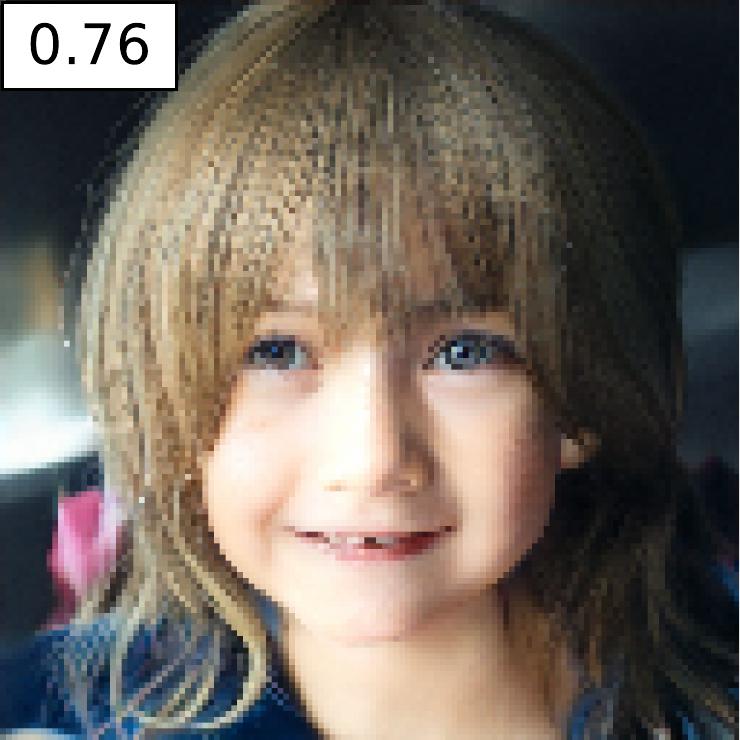} &
        \includegraphics[width=0.12\textwidth]{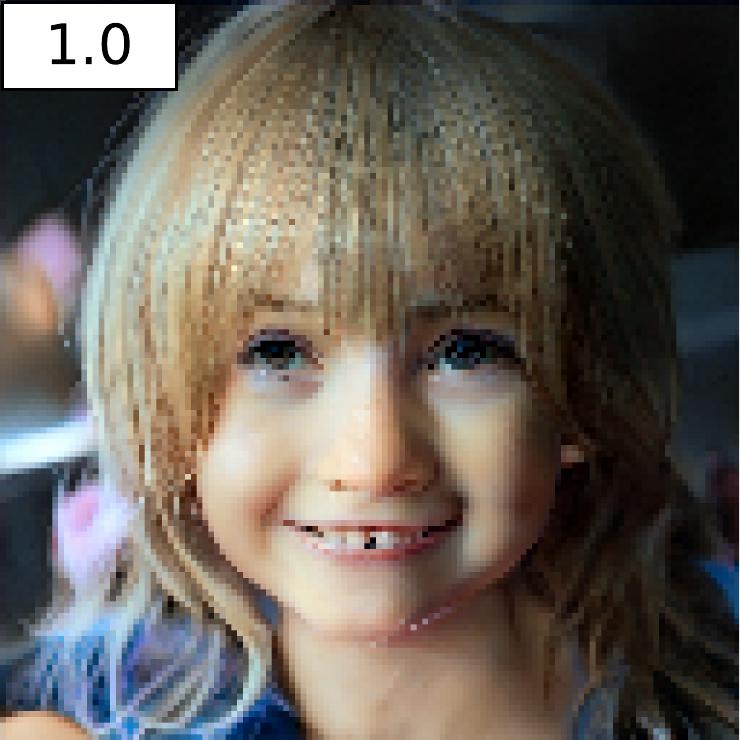} &
        \includegraphics[width=0.12\textwidth]{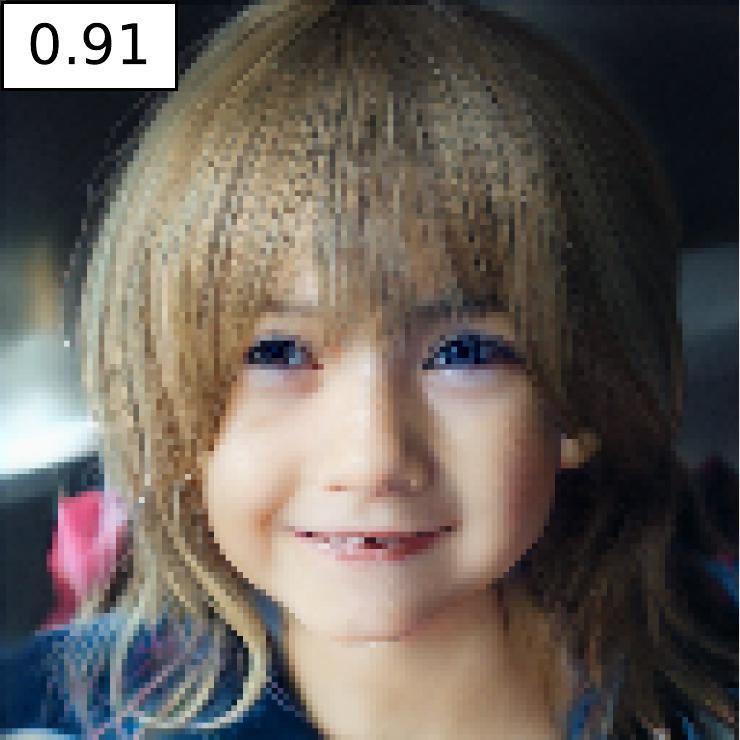} &
        \includegraphics[width=0.12\textwidth]{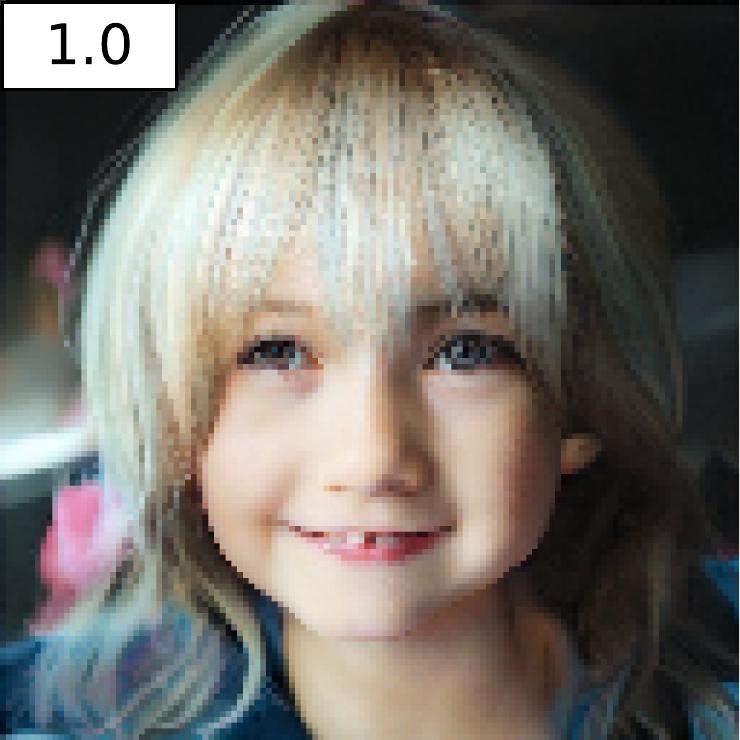} &
        \includegraphics[width=0.12\textwidth]{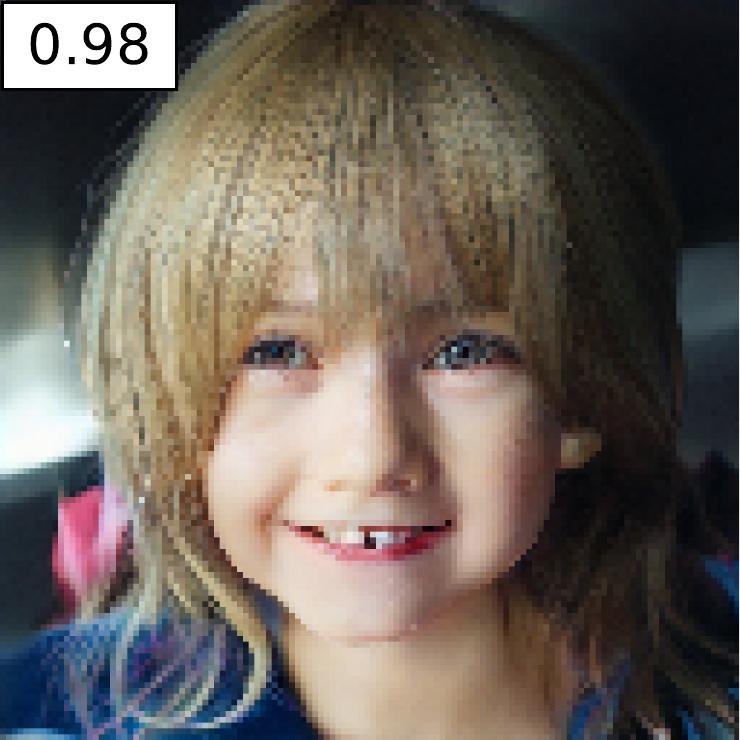} &
        \includegraphics[width=0.12\textwidth]{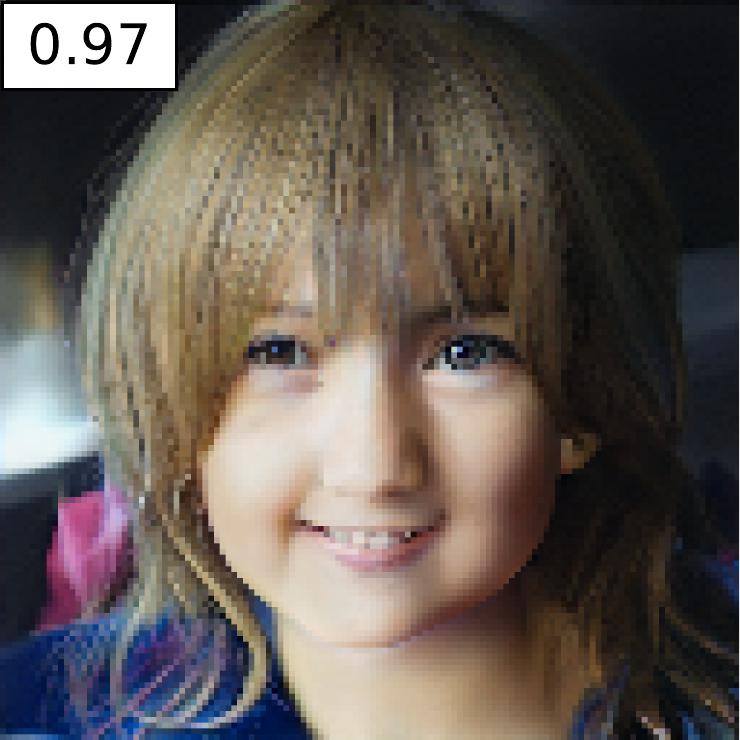} &
        \includegraphics[width=0.12\textwidth]{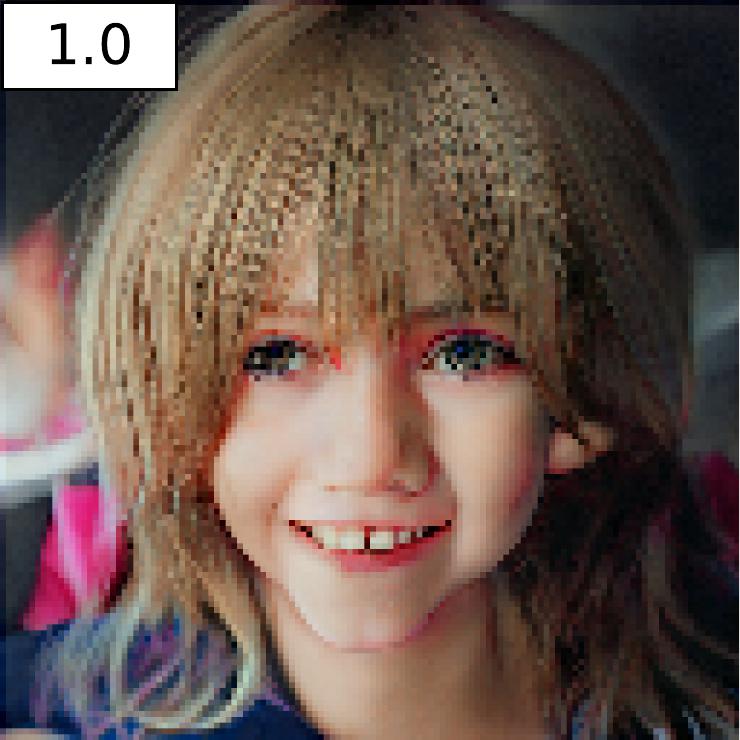} &
        \\
                \includegraphics[width=0.12\textwidth]{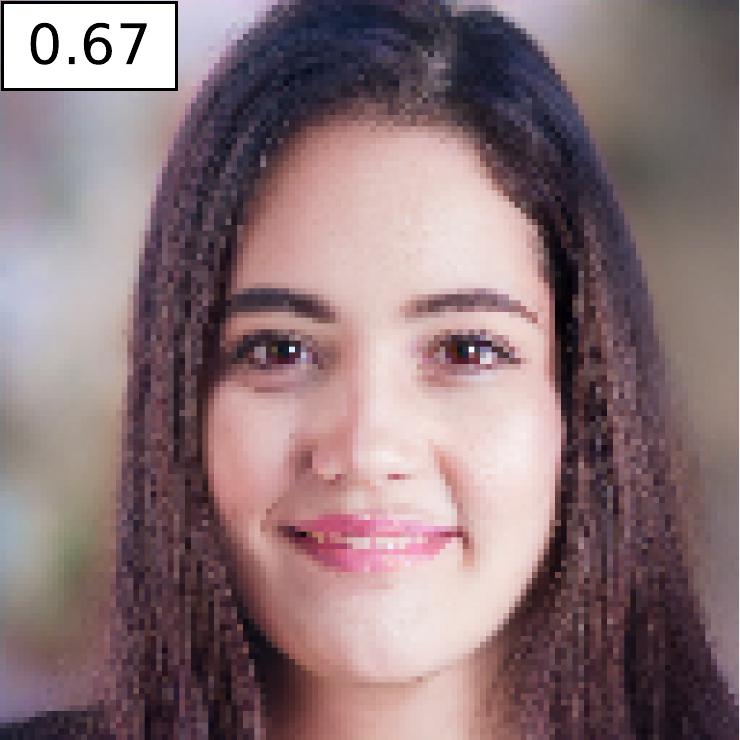} &
        \includegraphics[width=0.12\textwidth]{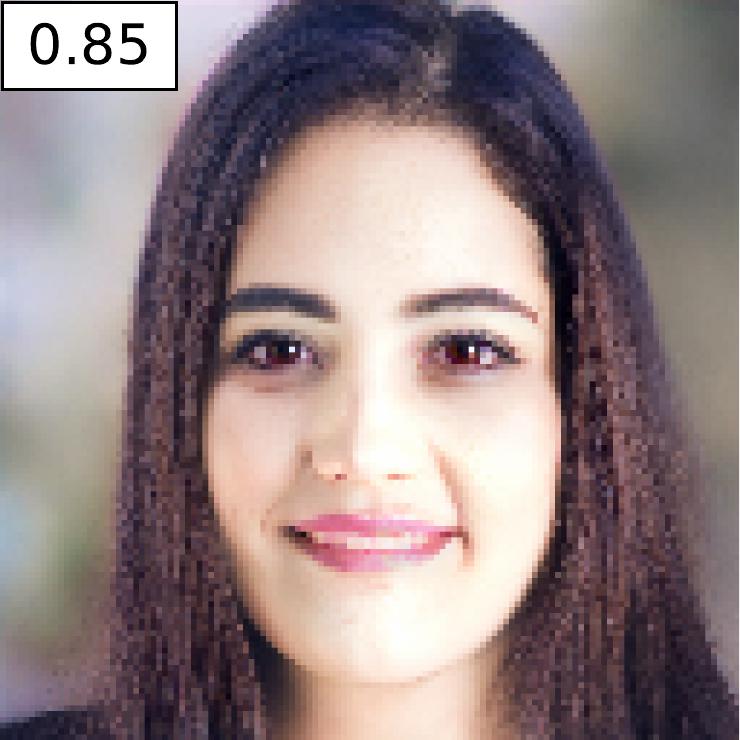} &
        \includegraphics[width=0.12\textwidth]{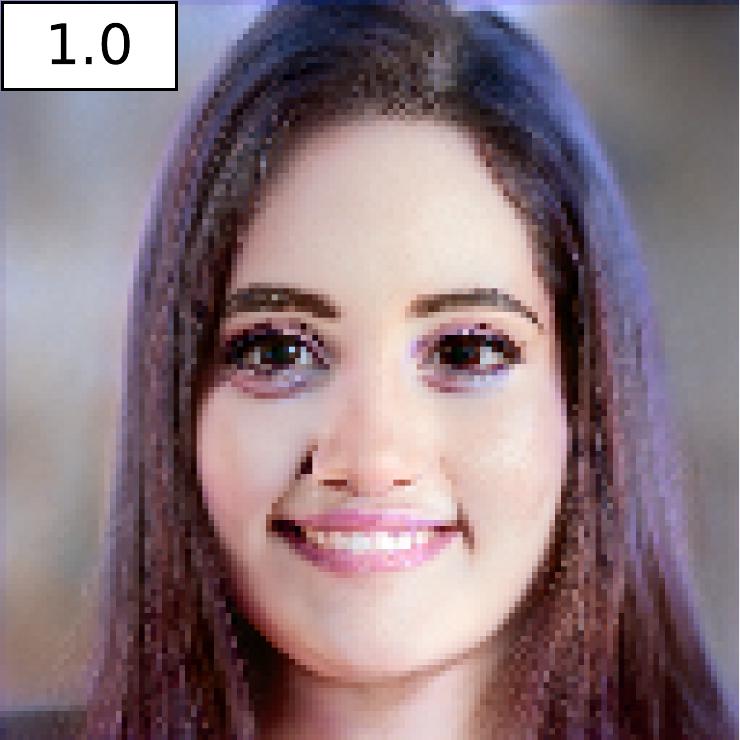} &
        \includegraphics[width=0.12\textwidth]{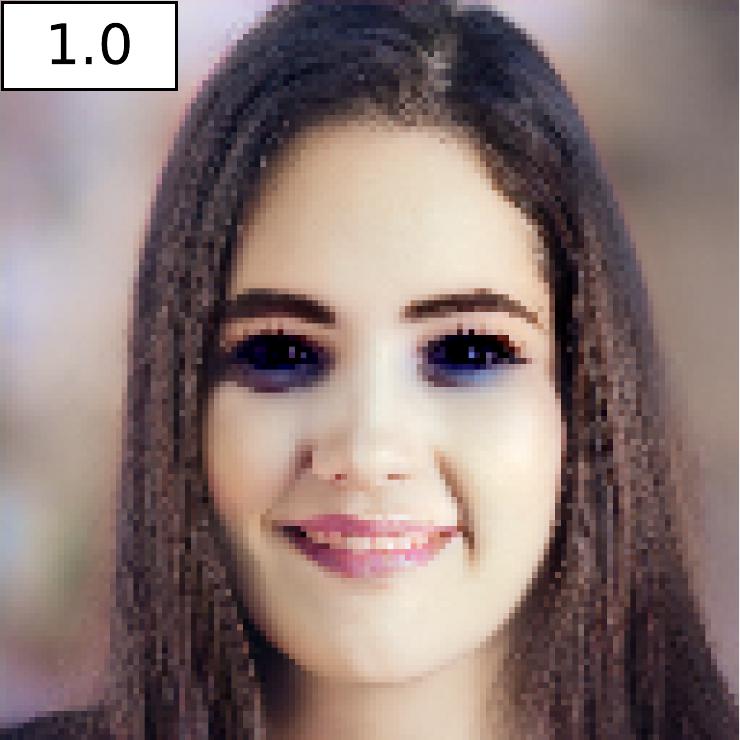} &
        \includegraphics[width=0.12\textwidth]{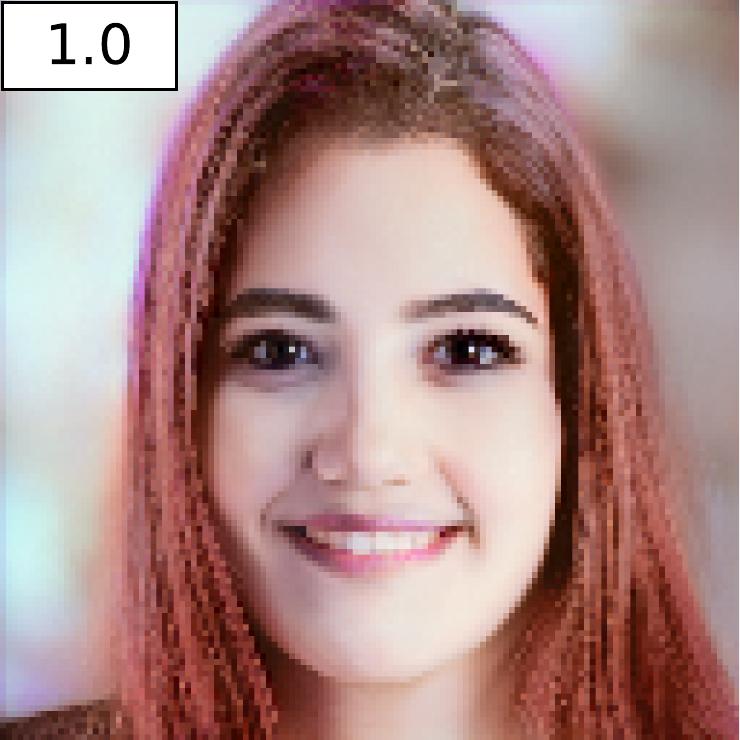} &
        \includegraphics[width=0.12\textwidth]{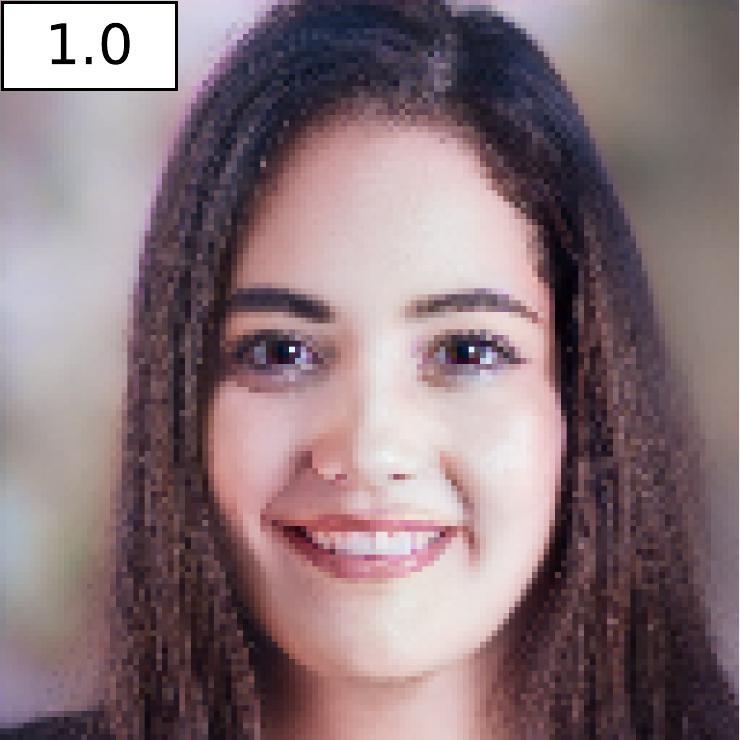} &
        \includegraphics[width=0.12\textwidth]{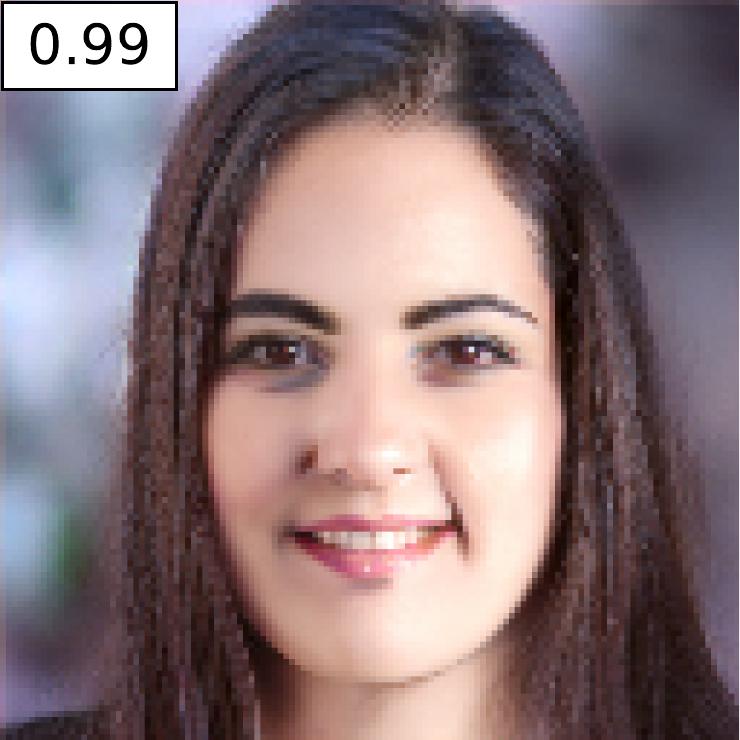} &
        \includegraphics[width=0.12\textwidth]{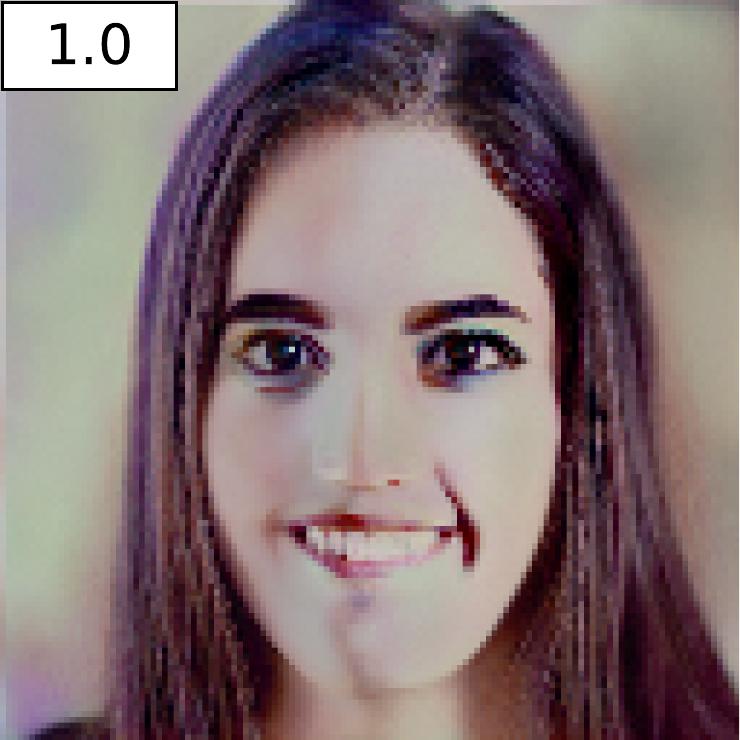} &
        \\
        Original image & Face color & Eye shape & Eye photometry & Hair photometry & Lip color & Mouth shape & Face boundary
    \end{tabular}
    \caption{Counterfactuals generated in interpretable subspaces to increase the predicted scores of the attractiveness classifier. See \Cref{sec:exp-counterfactual} for details. Corresponding difference maps can be found in \Cref{fig:attractive-diff} in the appendix.}
    \label{fig:attractive}
\end{figure*}

\subsubsection{Comparison with Previous Approaches}
We further conduct experiments to compare and analyze the performance with StyleSpace~\cite{wu2021stylespace}. Please find qualitative and quantitative results in \Cref{sec:comparison-sup} in the appendix.

\subsubsection{Transferability, Generality and Individuality}
Similar to~\cite{zhu2021low}, our subspaces discovery approach can be either directly applied on images for modification, or applied on some images for training to re-use the solved controls on other test images. In our experiments, most solved subspaces show good transferability. Unless specifically explained, we use subspaces solved from 1-2 training images, and generate figures in this paper from test set.

On the other hand, we notice that the discovered subspaces are sometimes also related to the individuality of training images. We include some further experiments in \Cref{sec:transferability-sup} in our appendix.

\subsection{Counterfactual Analysis}
\label{sec:exp-counterfactual}

To examine the effectiveness of our proposed approach of counterfactual analysis,
We selected 4 representative face attributes from CelebA including ``attractive'', ``heavy make-up'', ``smiling'', and ``pale skin'' to experiment on counterfactual explanations.

\Cref{fig:attractive} shows example counterfactuals generated to increase the score of the attractiveness classifier. As can be noticed, counterfactuals generated in the subspace of eye shape and eye photometry tend to have bigger and darker eyes. Counterfactuals with lip color modification tend to be redder lipsticked. This goes consistent with the common standard of beauty. Analysis in other subspaces further discovers that the classifier favors light skin, pointed face, small mouth and colorful hair, which is similar to the modern Asian standard of beauty. We suspect this might be related to the preference of dataset annotators, who are probably Asian since CelebA is created by an Asian institution. To highlight the difference of the counterfactuals, we visualize the corresponding difference map in \Cref{fig:attractive-diff} in our appendix.

As is pointed in~\cite{wu2021stylespace}, the attribute of attractiveness is highly subjective. Our counterfactual analysis does not aim to study the public standard of beauty, but rather to analyze the behavior of the specific classifier. This analysis does not reflect the opinion of the authors or the dataset creators on the judgement of beauty. There exist multiple directions to enhance the classifier score and our proposed approach only produces one possible solution.

The average accuracy of our classifier is $90\%$. It should also be noticed that explanation effectiveness is affected by both the quality of subspace disentanglement in the generator and the prediction accuracy of the classifier.

More results on ``heavy make-up'', ``smiling'', and ``pale skin'' are shown in \Cref{sec:exp-counterfactual-materials} in our appendix.

\section{Discussions}

We proposed an approach to flexibly explore interpretable latent subspaces for generators like StyleGAN. This approach not only helps discover unknown controls to manipulate images, but also can be used to generate counterfactual explanations for image recognition. We believe this approach provides a preliminary but new perspective to the explainability study of CNN. In this section, we discuss the limitation and future plan to finalize the paper.

\textbf{Limitation.}
The fundamental assumption for all the recent literature on latent subspace control lies in the semantic disentanglement of the latent space of StyleGAN. The richness of our discovered subspace further proves this assumption. However, we find that the disentanglement is still limited. The poor transferability of the face boundary control and the overflowed modification in unintended regions indicate that the disentanglement could be further improved.

As is shown in \Cref{fig:heavy_makeup} in the appendix, the counterfactuals does not always provide satisfactory explanation. For example, the subspace modification might overflow to unintended region. The unintended modification might compensate or conflict with the desired manipulation. The overfitting effect might also drive the classifier to favor certain controls but ignore others.

\textbf{Outlook.}
As mention in \Cref{sec:related}, though explainability has been an active ML research area recently, studies to explain image recognition on fixed structures like faces is still limited. We believe our proposed approach is valuable to promote the study in this area. We find the approach might inspire several future studies. First, the flexibility of our formulation could inspire the discovery of more unknown latent subspaces for GAN. Second, the interpretable subspaces are only a very small portion of the latent space. It remains an important problem on how the rest subspaces are affecting the classification results. Last but not the least, besides explaining face attribution, the effectiveness of the face identity feature criterion further implies that the approach might also be potential to explain face verification.

In terms of the long-term impact, we believe our new perspective of interpretable counterfactuals will serve as a valuable promotion to practical applications of interpretable AI such as vision-based medical diagnosis.

\section*{Acknowledgement}
We thank Zongze Wu for his explanation of the StyleSpace source code.

{\small
\bibliographystyle{plainnat}
\bibliography{egbib}

\begin{thebibliography}{44}
\providecommand{\natexlab}[1]{#1}
\providecommand{\url}[1]{\texttt{#1}}
\expandafter\ifx\csname urlstyle\endcsname\relax
  \providecommand{\doi}[1]{doi: #1}\else
  \providecommand{\doi}{doi: \begingroup \urlstyle{rm}\Url}\fi

\bibitem[Abdal et~al.(2021)Abdal, Zhu, Mitra, and Wonka]{abdal2021styleflow}
Rameen Abdal, Peihao Zhu, Niloy~J Mitra, and Peter Wonka.
\newblock Styleflow: Attribute-conditioned exploration of stylegan-generated
  images using conditional continuous normalizing flows.
\newblock \emph{ACM Transactions on Graphics (TOG)}, 40\penalty0 (3):\penalty0
  1--21, 2021.

\bibitem[Bau et~al.(2018)Bau, Zhu, Strobelt, Zhou, Tenenbaum, Freeman, and
  Torralba]{bau2018gan}
David Bau, Jun-Yan Zhu, Hendrik Strobelt, Bolei Zhou, Joshua~B Tenenbaum,
  William~T Freeman, and Antonio Torralba.
\newblock Gan dissection: Visualizing and understanding generative adversarial
  networks.
\newblock \emph{arXiv preprint arXiv:1811.10597}, 2018.

\bibitem[Bodria et~al.(2021)Bodria, Giannotti, Guidotti, Naretto, Pedreschi,
  and Rinzivillo]{bodria2021benchmarking}
Francesco Bodria, Fosca Giannotti, Riccardo Guidotti, Francesca Naretto, Dino
  Pedreschi, and Salvatore Rinzivillo.
\newblock Benchmarking and survey of explanation methods for black box models.
\newblock \emph{arXiv preprint arXiv:2102.13076}, 2021.

\bibitem[Bulat and Tzimiropoulos(2017)]{bulat2017far}
Adrian Bulat and Georgios Tzimiropoulos.
\newblock How far are we from solving the 2d \& 3d face alignment problem? (and
  a dataset of 230,000 3d facial landmarks).
\newblock In \emph{International Conference on Computer Vision}, 2017.

\bibitem[Collins et~al.(2020)Collins, Bala, Price, and
  Susstrunk]{collins2020editing}
Edo Collins, Raja Bala, Bob Price, and Sabine Susstrunk.
\newblock Editing in style: Uncovering the local semantics of gans.
\newblock In \emph{Proceedings of the IEEE/CVF Conference on Computer Vision
  and Pattern Recognition}, pages 5771--5780, 2020.

\bibitem[Dhurandhar et~al.(2018)Dhurandhar, Chen, Luss, Tu, Ting, Shanmugam,
  and Das]{dhurandhar2018explanations}
Amit Dhurandhar, Pin-Yu Chen, Ronny Luss, Chun-Chen Tu, Paishun Ting,
  Karthikeyan Shanmugam, and Payel Das.
\newblock Explanations based on the missing: Towards contrastive explanations
  with pertinent negatives.
\newblock \emph{arXiv preprint arXiv:1802.07623}, 2018.

\bibitem[Dombrowski et~al.(2021)Dombrowski, Gerken, and
  Kessel]{dombrowski2021diffeomorphic}
Ann-Kathrin Dombrowski, Jan~E Gerken, and Pan Kessel.
\newblock Diffeomorphic explanations with normalizing flows.
\newblock In \emph{ICML Workshop on Invertible Neural Networks, Normalizing
  Flows, and Explicit Likelihood Models}, 2021.

\bibitem[Goetschalckx et~al.(2019)Goetschalckx, Andonian, Oliva, and
  Isola]{goetschalckx2019ganalyze}
Lore Goetschalckx, Alex Andonian, Aude Oliva, and Phillip Isola.
\newblock Ganalyze: Toward visual definitions of cognitive image properties.
\newblock In \emph{Proceedings of the IEEE/CVF International Conference on
  Computer Vision}, pages 5744--5753, 2019.

\bibitem[Goodfellow et~al.(2014)Goodfellow, Pouget-Abadie, Mirza, Xu,
  Warde-Farley, Ozair, Courville, and Bengio]{goodfellow2014generative}
Ian Goodfellow, Jean Pouget-Abadie, Mehdi Mirza, Bing Xu, David Warde-Farley,
  Sherjil Ozair, Aaron Courville, and Yoshua Bengio.
\newblock Generative adversarial nets.
\newblock \emph{Advances in neural information processing systems}, 27, 2014.

\bibitem[Goyal et~al.(2019)Goyal, Feder, Shalit, and Kim]{goyal2019explaining}
Yash Goyal, Amir Feder, Uri Shalit, and Been Kim.
\newblock Explaining classifiers with causal concept effect (cace).
\newblock \emph{arXiv preprint arXiv:1907.07165}, 2019.

\bibitem[H{\"a}rk{\"o}nen et~al.(2020)H{\"a}rk{\"o}nen, Hertzmann, Lehtinen,
  and Paris]{harkonen2020ganspace}
Erik H{\"a}rk{\"o}nen, Aaron Hertzmann, Jaakko Lehtinen, and Sylvain Paris.
\newblock Ganspace: Discovering interpretable gan controls.
\newblock \emph{arXiv preprint arXiv:2004.02546}, 2020.

\bibitem[Jahanian et~al.(2019)Jahanian, Chai, and
  Isola]{jahanian2019steerability}
Ali Jahanian, Lucy Chai, and Phillip Isola.
\newblock On the" steerability" of generative adversarial networks.
\newblock \emph{arXiv preprint arXiv:1907.07171}, 2019.

\bibitem[Jiang and Zeng(2021)]{jiang2021explainable}
Haoran Jiang and Dan Zeng.
\newblock Explainable face recognition based on accurate facial compositions.
\newblock In \emph{Proceedings of the IEEE/CVF International Conference on
  Computer Vision}, pages 1503--1512, 2021.

\bibitem[Karras et~al.(2019)Karras, Laine, and Aila]{karras2019style}
Tero Karras, Samuli Laine, and Timo Aila.
\newblock A style-based generator architecture for generative adversarial
  networks.
\newblock In \emph{Proceedings of the IEEE/CVF Conference on Computer Vision
  and Pattern Recognition}, pages 4401--4410, 2019.

\bibitem[Karras et~al.(2020{\natexlab{a}})Karras, Aittala, Hellsten, Laine,
  Lehtinen, and Aila]{karras2020training}
Tero Karras, Miika Aittala, Janne Hellsten, Samuli Laine, Jaakko Lehtinen, and
  Timo Aila.
\newblock Training generative adversarial networks with limited data.
\newblock \emph{arXiv preprint arXiv:2006.06676}, 2020{\natexlab{a}}.

\bibitem[Karras et~al.(2020{\natexlab{b}})Karras, Laine, Aittala, Hellsten,
  Lehtinen, and Aila]{karras2020analyzing}
Tero Karras, Samuli Laine, Miika Aittala, Janne Hellsten, Jaakko Lehtinen, and
  Timo Aila.
\newblock Analyzing and improving the image quality of stylegan.
\newblock In \emph{Proceedings of the IEEE/CVF Conference on Computer Vision
  and Pattern Recognition}, pages 8110--8119, 2020{\natexlab{b}}.

\bibitem[Kingma and Dhariwal(2018)]{kingma2018glow}
Diederik~P Kingma and Prafulla Dhariwal.
\newblock Glow: Generative flow with invertible 1x1 convolutions.
\newblock \emph{arXiv preprint arXiv:1807.03039}, 2018.

\bibitem[Kingma and Welling(2013)]{kingma2013auto}
Diederik~P Kingma and Max Welling.
\newblock Auto-encoding variational bayes.
\newblock \emph{arXiv preprint arXiv:1312.6114}, 2013.

\bibitem[Lang et~al.(2021)Lang, Gandelsman, Yarom, Wald, Elidan, Hassidim,
  Freeman, Isola, Globerson, Irani, et~al.]{lang2021explaining}
Oran Lang, Yossi Gandelsman, Michal Yarom, Yoav Wald, Gal Elidan, Avinatan
  Hassidim, William~T Freeman, Phillip Isola, Amir Globerson, Michal Irani,
  et~al.
\newblock Explaining in style: Training a gan to explain a classifier in
  stylespace.
\newblock \emph{arXiv preprint arXiv:2104.13369}, 2021.

\bibitem[Li et~al.(2021)Li, Qi, Liu, Di, Liu, Pei, Yi, and
  Zhou]{li2021trustworthy}
Bo~Li, Peng Qi, Bo~Liu, Shuai Di, Jingen Liu, Jiquan Pei, Jinfeng Yi, and Bowen
  Zhou.
\newblock Trustworthy ai: From principles to practices.
\newblock \emph{arXiv preprint arXiv:2110.01167}, 2021.

\bibitem[Liu et~al.(2015)Liu, Luo, Wang, and Tang]{liu2015faceattributes}
Ziwei Liu, Ping Luo, Xiaogang Wang, and Xiaoou Tang.
\newblock Deep learning face attributes in the wild.
\newblock In \emph{Proceedings of International Conference on Computer Vision
  (ICCV)}, December 2015.

\bibitem[Nitzan et~al.(2020)Nitzan, Bermano, Li, and Cohen-Or]{nitzan2020face}
Yotam Nitzan, Amit Bermano, Yangyan Li, and Daniel Cohen-Or.
\newblock Face identity disentanglement via latent space mapping.
\newblock \emph{ACM Transactions on Graphics (TOG)}, 39\penalty0 (6):\penalty0
  1--14, 2020.

\bibitem[Patashnik et~al.(2021)Patashnik, Wu, Shechtman, Cohen-Or, and
  Lischinski]{patashnik2021styleclip}
Or~Patashnik, Zongze Wu, Eli Shechtman, Daniel Cohen-Or, and Dani Lischinski.
\newblock Styleclip: Text-driven manipulation of stylegan imagery.
\newblock In \emph{Proceedings of the IEEE/CVF International Conference on
  Computer Vision}, pages 2085--2094, 2021.

\bibitem[Plumerault et~al.(2020)Plumerault, Borgne, and
  Hudelot]{plumerault2020controlling}
Antoine Plumerault, Herv{\'e}~Le Borgne, and C{\'e}line Hudelot.
\newblock Controlling generative models with continuous factors of variations.
\newblock \emph{arXiv preprint arXiv:2001.10238}, 2020.

\bibitem[Radford et~al.(2021)Radford, Kim, Hallacy, Ramesh, Goh, Agarwal,
  Sastry, Askell, Mishkin, Clark, et~al.]{radford2021learning}
Alec Radford, Jong~Wook Kim, Chris Hallacy, Aditya Ramesh, Gabriel Goh,
  Sandhini Agarwal, Girish Sastry, Amanda Askell, Pamela Mishkin, Jack Clark,
  et~al.
\newblock Learning transferable visual models from natural language
  supervision.
\newblock \emph{arXiv preprint arXiv:2103.00020}, 2021.

\bibitem[Sauer and Geiger(2021)]{sauer2021counterfactual}
Axel Sauer and Andreas Geiger.
\newblock Counterfactual generative networks.
\newblock In \emph{International Conference on Learning Representations}, 2021.
\newblock URL \url{https://openreview.net/forum?id=BXewfAYMmJw}.

\bibitem[Shen and Zhou(2021)]{shen2021closed}
Yujun Shen and Bolei Zhou.
\newblock Closed-form factorization of latent semantics in gans.
\newblock In \emph{Proceedings of the IEEE/CVF Conference on Computer Vision
  and Pattern Recognition}, pages 1532--1540, 2021.

\bibitem[Shen et~al.(2020)Shen, Yang, Tang, and Zhou]{shen2020interfacegan}
Yujun Shen, Ceyuan Yang, Xiaoou Tang, and Bolei Zhou.
\newblock Interfacegan: Interpreting the disentangled face representation
  learned by gans.
\newblock \emph{IEEE transactions on pattern analysis and machine
  intelligence}, 2020.

\bibitem[Singla et~al.(2019)Singla, Pollack, Chen, and
  Batmanghelich]{singla2019explanation}
Sumedha Singla, Brian Pollack, Junxiang Chen, and Kayhan Batmanghelich.
\newblock Explanation by progressive exaggeration.
\newblock \emph{arXiv preprint arXiv:1911.00483}, 2019.

\bibitem[Sixt et~al.(2020)Sixt, Schuessler, Wei{\ss}, and
  Landgraf]{sixt2020interpretability}
Leon Sixt, Martin Schuessler, Philipp Wei{\ss}, and Tim Landgraf.
\newblock Interpretability through invertibility: A deep convolutional network
  with ideal counterfactuals and isosurfaces.
\newblock 2020.

\bibitem[Spingarn et~al.(2021)Spingarn, Banner, and Michaeli]{spingarn2021gan}
Nurit Spingarn, Ron Banner, and Tomer Michaeli.
\newblock {\{}GAN{\}} ''steerability'' without optimization.
\newblock In \emph{International Conference on Learning Representations}, 2021.
\newblock URL \url{https://openreview.net/forum?id=zDy_nQCXiIj}.

\bibitem[Tewari et~al.(2020)Tewari, Elgharib, Bharaj, Bernard, Seidel,
  P{\'e}rez, Zollhofer, and Theobalt]{tewari2020stylerig}
Ayush Tewari, Mohamed Elgharib, Gaurav Bharaj, Florian Bernard, Hans-Peter
  Seidel, Patrick P{\'e}rez, Michael Zollhofer, and Christian Theobalt.
\newblock Stylerig: Rigging stylegan for 3d control over portrait images.
\newblock In \emph{Proceedings of the IEEE/CVF Conference on Computer Vision
  and Pattern Recognition}, pages 6142--6151, 2020.

\bibitem[\texttt{d-li14}(2019)]{d2019face}
\texttt{d-li14}.
\newblock \texttt{face-attribute-prediction}.
\newblock \url{https://github.com/d-li14/face-attribute-prediction}, 2019.

\bibitem[\texttt{zllrunning}(2019)]{zllrunning2019face}
\texttt{zllrunning}.
\newblock \texttt{face-parsing.PyTorch}.
\newblock \url{https://github.com/zllrunning/face-parsing.PyTorch}, 2019.

\bibitem[Van~Looveren and Klaise(2019)]{van2019interpretable}
Arnaud Van~Looveren and Janis Klaise.
\newblock Interpretable counterfactual explanations guided by prototypes.
\newblock \emph{arXiv preprint arXiv:1907.02584}, 2019.

\bibitem[Voynov and Babenko(2020)]{voynov2020unsupervised}
Andrey Voynov and Artem Babenko.
\newblock Unsupervised discovery of interpretable directions in the gan latent
  space.
\newblock In \emph{International Conference on Machine Learning}, pages
  9786--9796. PMLR, 2020.

\bibitem[Wachter et~al.(2017)Wachter, Mittelstadt, and
  Russell]{wachter2017counterfactual}
Sandra Wachter, Brent Mittelstadt, and Chris Russell.
\newblock Counterfactual explanations without opening the black box: Automated
  decisions and the gdpr.
\newblock \emph{Harv. JL \& Tech.}, 31:\penalty0 841, 2017.

\bibitem[Wang and Ponce(2021)]{wang2021geometry}
Binxu Wang and Carlos~R Ponce.
\newblock The geometry of deep generative image models and its applications.
\newblock \emph{arXiv preprint arXiv:2101.06006}, 2021.

\bibitem[Williford et~al.(2020)Williford, May, and
  Byrne]{williford2020explainable}
Jonathan~R Williford, Brandon~B May, and Jeffrey Byrne.
\newblock Explainable face recognition.
\newblock In \emph{European Conference on Computer Vision}, pages 248--263.
  Springer, 2020.

\bibitem[Wu et~al.(2021)Wu, Lischinski, and Shechtman]{wu2021stylespace}
Zongze Wu, Dani Lischinski, and Eli Shechtman.
\newblock Stylespace analysis: Disentangled controls for stylegan image
  generation.
\newblock In \emph{Proceedings of the IEEE/CVF Conference on Computer Vision
  and Pattern Recognition}, pages 12863--12872, 2021.

\bibitem[Yang et~al.(2021)Yang, Shen, and Zhou]{yang2021semantic}
Ceyuan Yang, Yujun Shen, and Bolei Zhou.
\newblock Semantic hierarchy emerges in deep generative representations for
  scene synthesis.
\newblock \emph{International Journal of Computer Vision}, 129\penalty0
  (5):\penalty0 1451--1466, 2021.

\bibitem[Yin et~al.(2019)Yin, Tran, Li, Shen, and Liu]{yin2019towards}
Bangjie Yin, Luan Tran, Haoxiang Li, Xiaohui Shen, and Xiaoming Liu.
\newblock Towards interpretable face recognition.
\newblock In \emph{Proceedings of the IEEE/CVF International Conference on
  Computer Vision}, pages 9348--9357, 2019.

\bibitem[Zhang et~al.(2021)Zhang, Ti{\v{n}}o, Leonardis, and
  Tang]{zhang2021survey}
Yu~Zhang, Peter Ti{\v{n}}o, Ale{\v{s}} Leonardis, and Ke~Tang.
\newblock A survey on neural network interpretability.
\newblock \emph{IEEE Transactions on Emerging Topics in Computational
  Intelligence}, 2021.

\bibitem[Zhu et~al.(2021)Zhu, Feng, Shen, Zhao, Zha, Zhou, and
  Chen]{zhu2021low}
Jiapeng Zhu, Ruili Feng, Yujun Shen, Deli Zhao, Zhengjun Zha, Jingren Zhou, and
  Qifeng Chen.
\newblock Low-rank subspaces in gans.
\newblock \emph{arXiv preprint arXiv:2106.04488}, 2021.

\end{thebibliography}
}
\makeatletter\@input{PaperForReviewaux.tex}\makeatother
\makeatletter\@input{supaux.tex}\makeatother

\end{document}


\title{Which Style Makes Me Attractive? Interpretable Control Exploration and Counterfactual Explanation by StyleGAN\\\textit{Appendix}}

\maketitle


\section{Latent Space Comparison Experiments}
\label{sec:space-comparison-sup}

\begin{figure}
    \centering
    \includegraphics[width=0.8\linewidth]{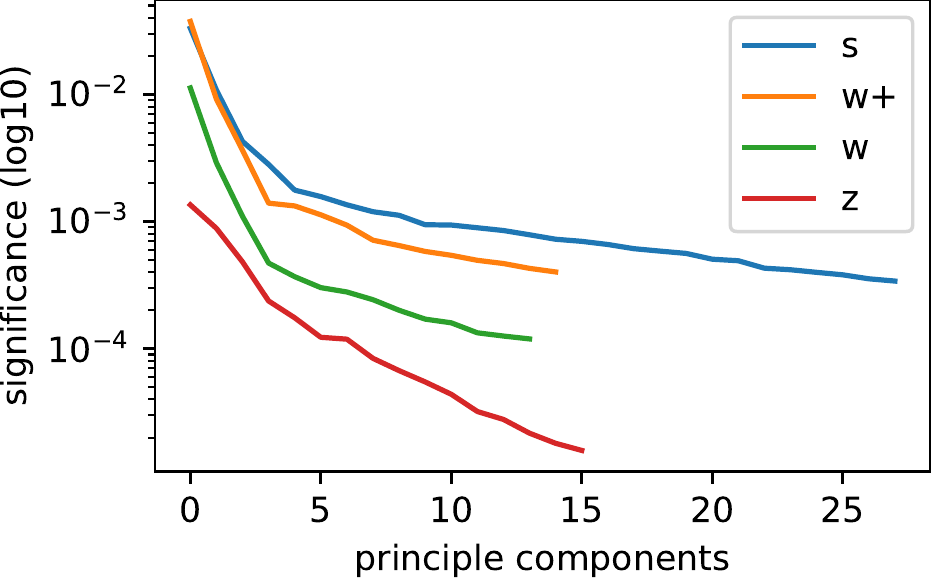}
    \caption{Plots of the control significance in the spaces of $\Ss, \WW+, \WW$ and $\ZZ$. See \Cref{sec:exp-subspace} for details.}
    \label{fig:attenuation}
\end{figure}

We consider a case same as \Cref{fig:criterions}bc, where we aim to modify the masked photometry in the mouth area. We select $\epsilon=10^{-2}$ and compute the principle components of \eqref{eq:wjjw}. For each principle component, we compute the ratio $\lambda / \lambda_0^{\overline{\text{mou}}}$, where $\lambda$ is the corresponding eigenvalue. $\lambda_0^{\overline{\text{mou}}}$ is the maximum eigenvalue of ${\frac{\partial h_\text{mp}^{\overline{\text{mou}}}}{\partial u}}^\top\frac{\partial h_\text{mp}^{\overline{\text{mou}}}}{\partial u}$, i.e. $\lambda_0^{\overline{\text{mou}}}$ is the reference value $\lambda_0$ as mentioned in \Cref{sec:exp-setup} to split $V_\text{mp}^{\overline{\text{mou}}}$ and $W_\text{mp}^{\overline{\text{mou}}}$. This ratio roughly evaluates how much the mouth area is modification given same change at the non-mouth area (which should be kept unchanged).
\Cref{fig:attenuation} plots the logarithm value of $\lambda / \lambda_0^{\overline{\text{mou}}}$ by principle component index. As can be noticed, $\Ss$ possesses the highest modification magnitude. The modification magnitude of $\WW$ and $\ZZ$ is noticeable at the first several components, but attenuates very fast. This plot indicates that $\Ss$ and $\WW+$ possess much richer semantic controls on images. This conclusion is also consistent with~\cite{wu2021stylespace}. 

\begin{figure*}
    \footnotesize
    \setlength{\tabcolsep}{1pt}
    \centering
    \begin{tabular}{ccccccccccc}
        \multirow{2}{*}{ \rotatebox{90}{$\Ss, 0\text{-}9$} } &
        \includegraphics[width=0.19\textwidth]{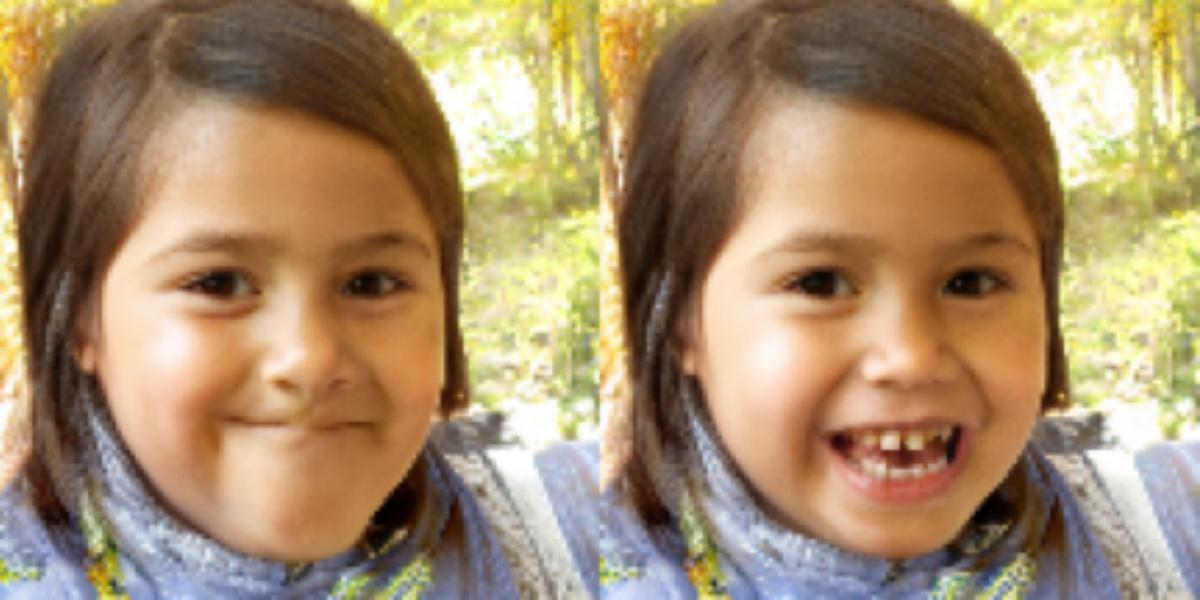} &
        \includegraphics[width=0.19\textwidth]{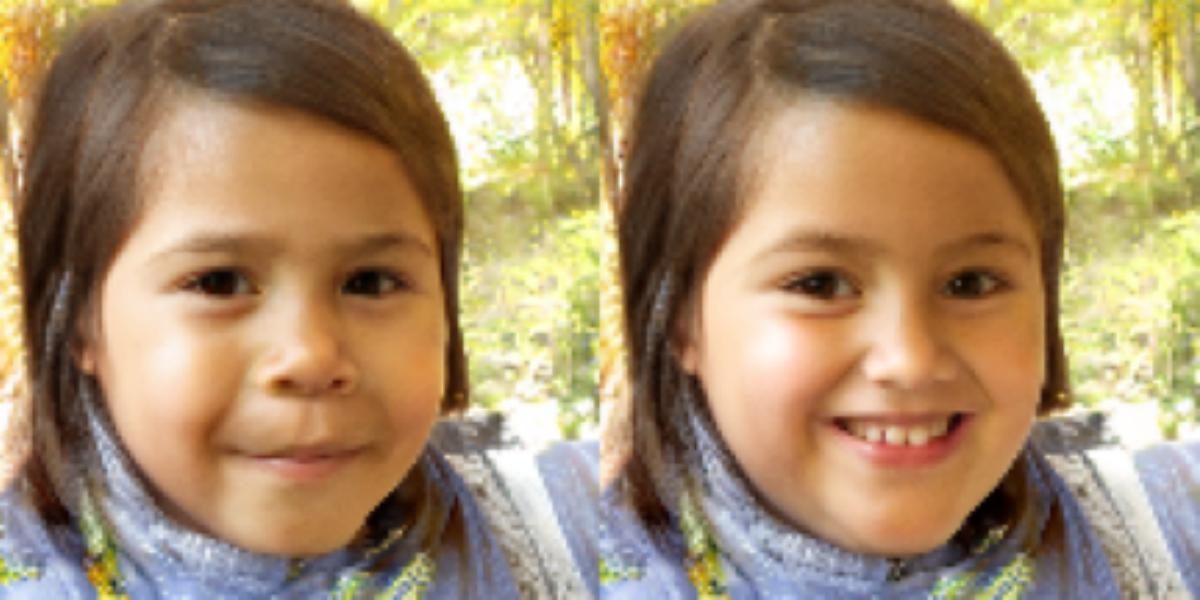} &
        \includegraphics[width=0.19\textwidth]{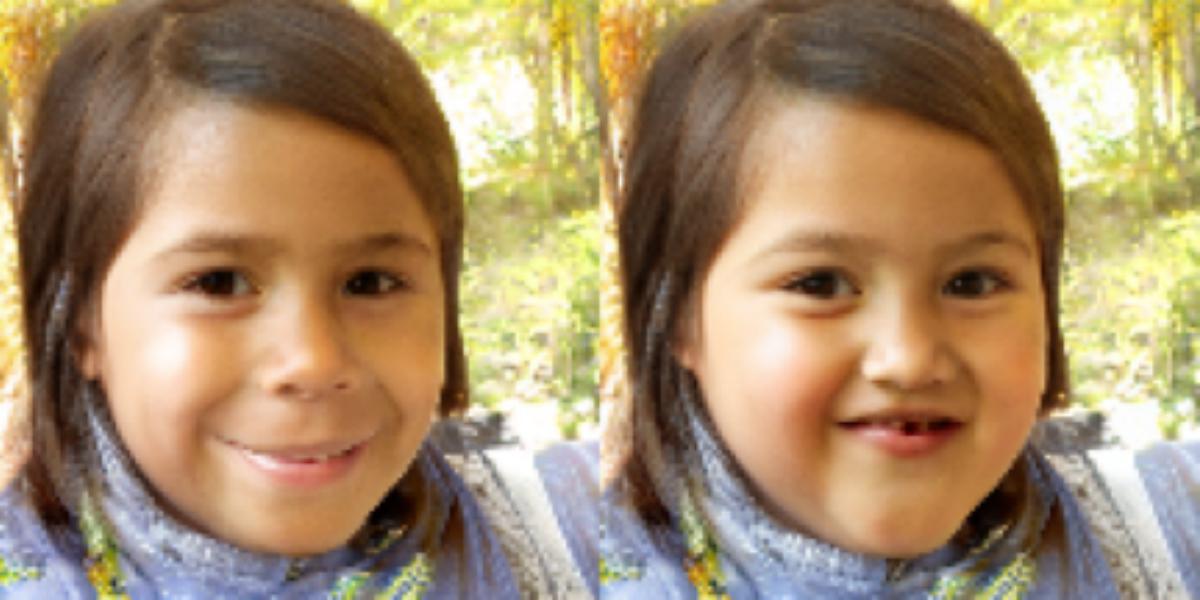}  &
        \includegraphics[width=0.19\textwidth]{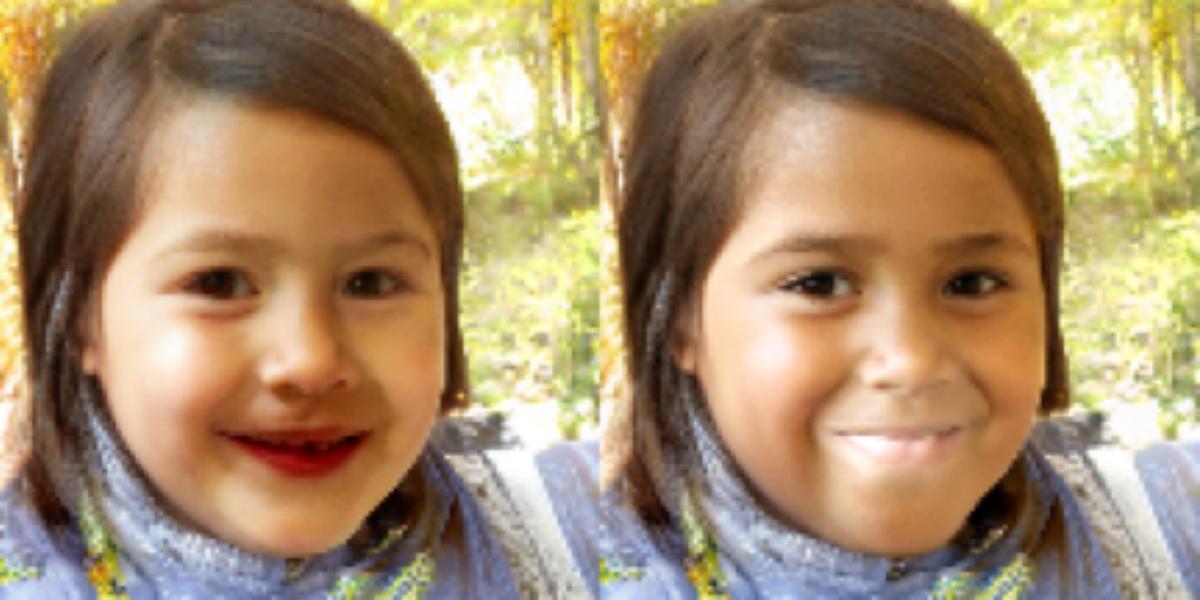}  &
        \includegraphics[width=0.19\textwidth]{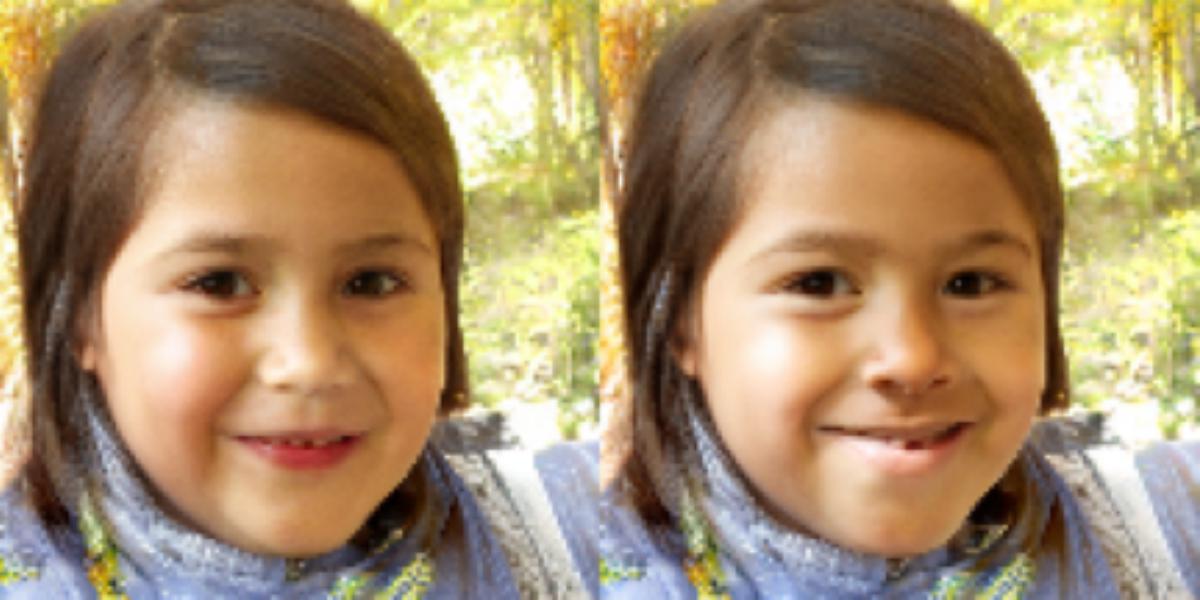}\\
        &\includegraphics[width=0.19\textwidth]{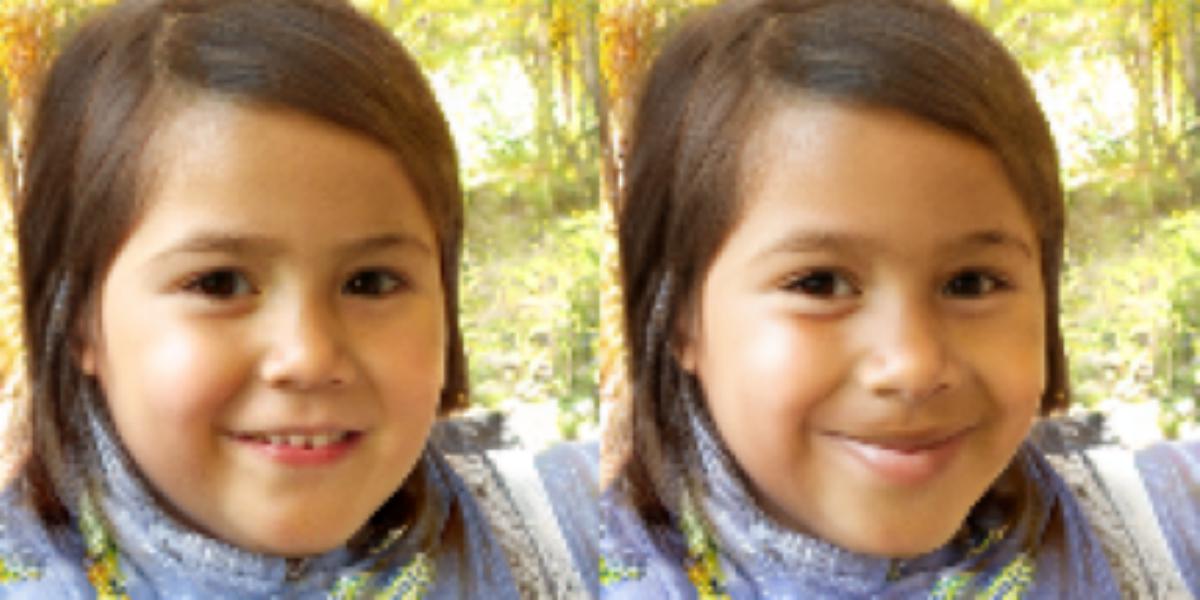} &
        \includegraphics[width=0.19\textwidth]{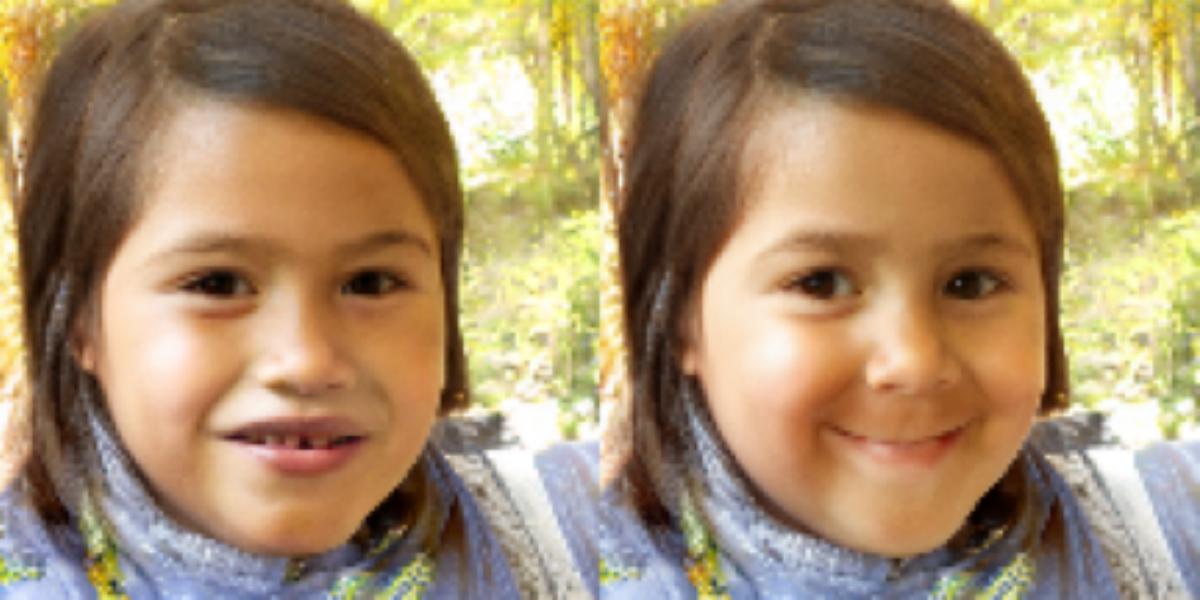} &
        \includegraphics[width=0.19\textwidth]{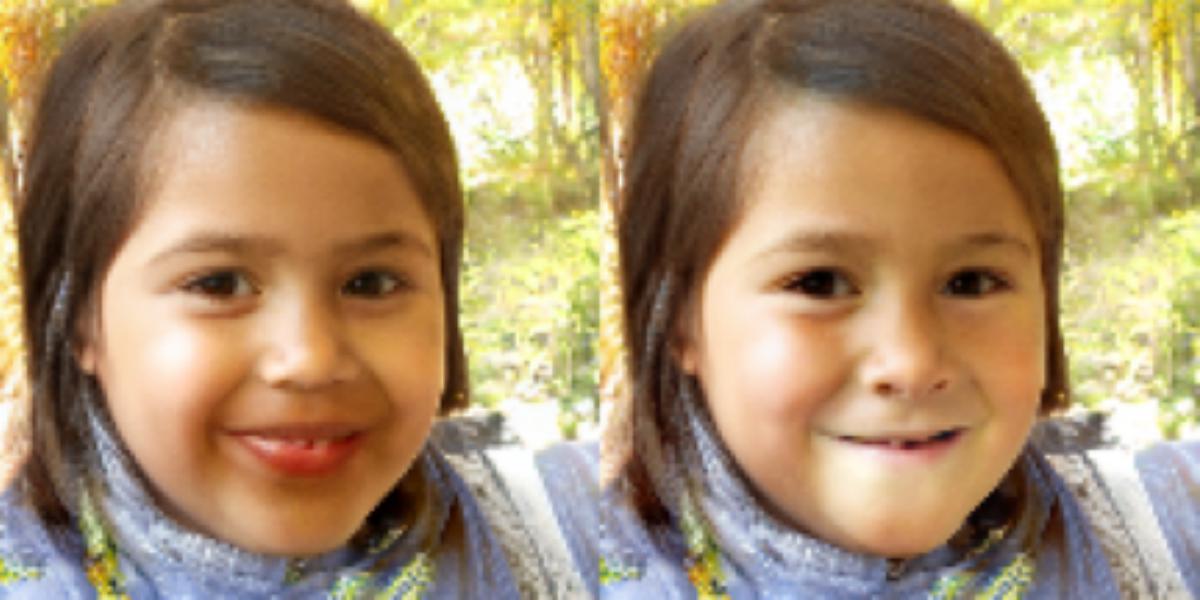}  &
        \includegraphics[width=0.19\textwidth]{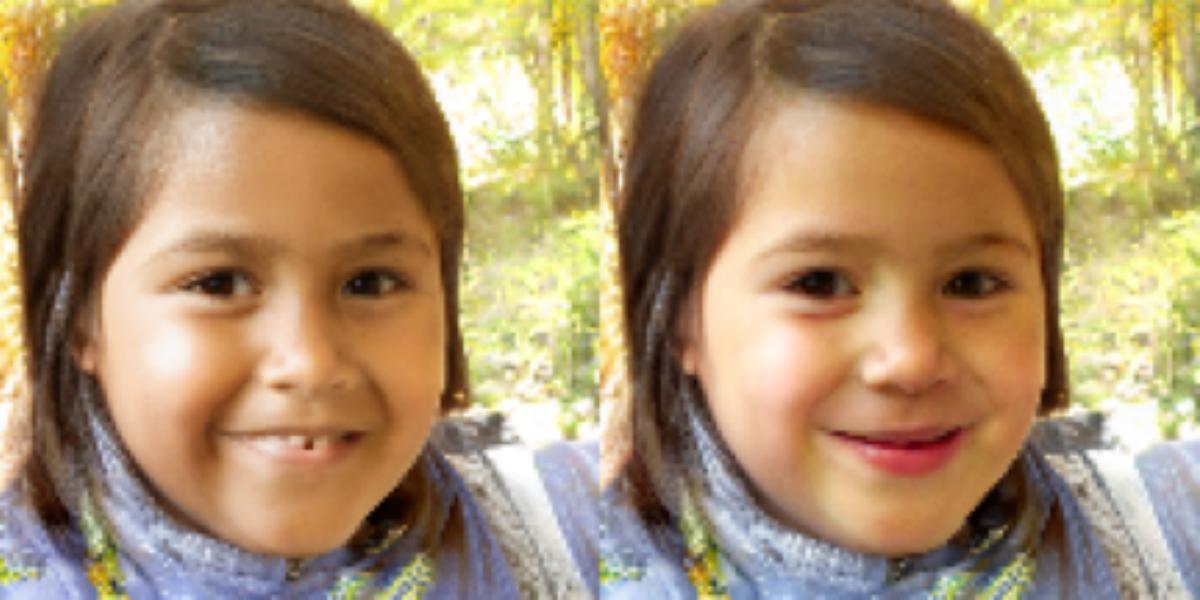}  &
        \includegraphics[width=0.19\textwidth]{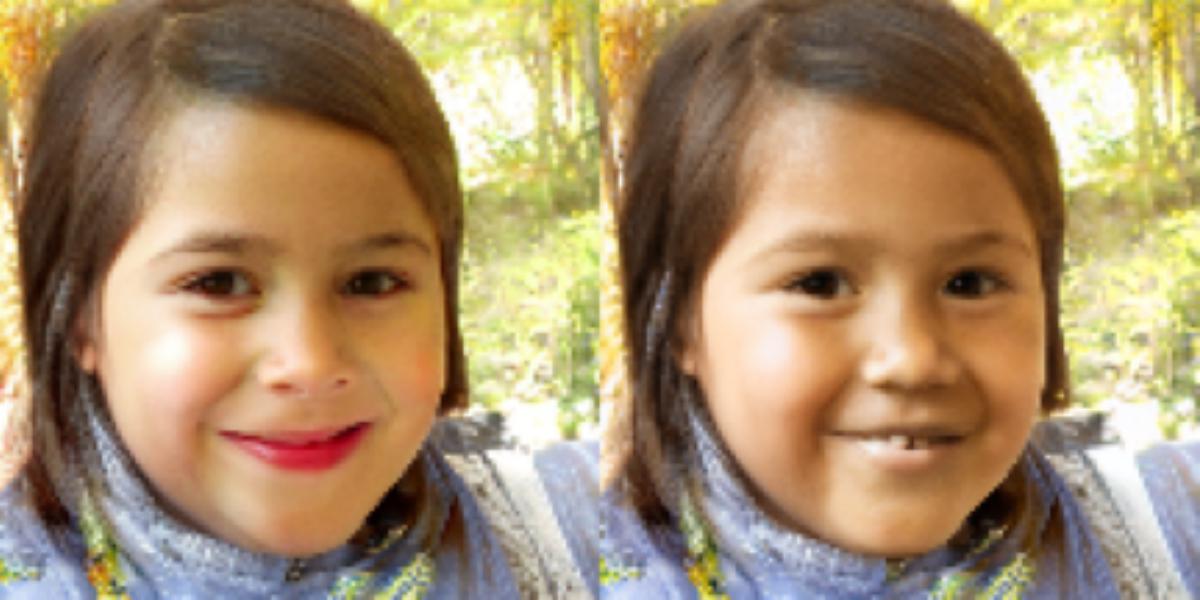}\\\hline
        \rule{0pt}{0.1\textwidth} \multirow{2}{*}{ \rotatebox{90}{$\WW+, 0\text{-}9$} } &
        \includegraphics[width=0.19\textwidth]{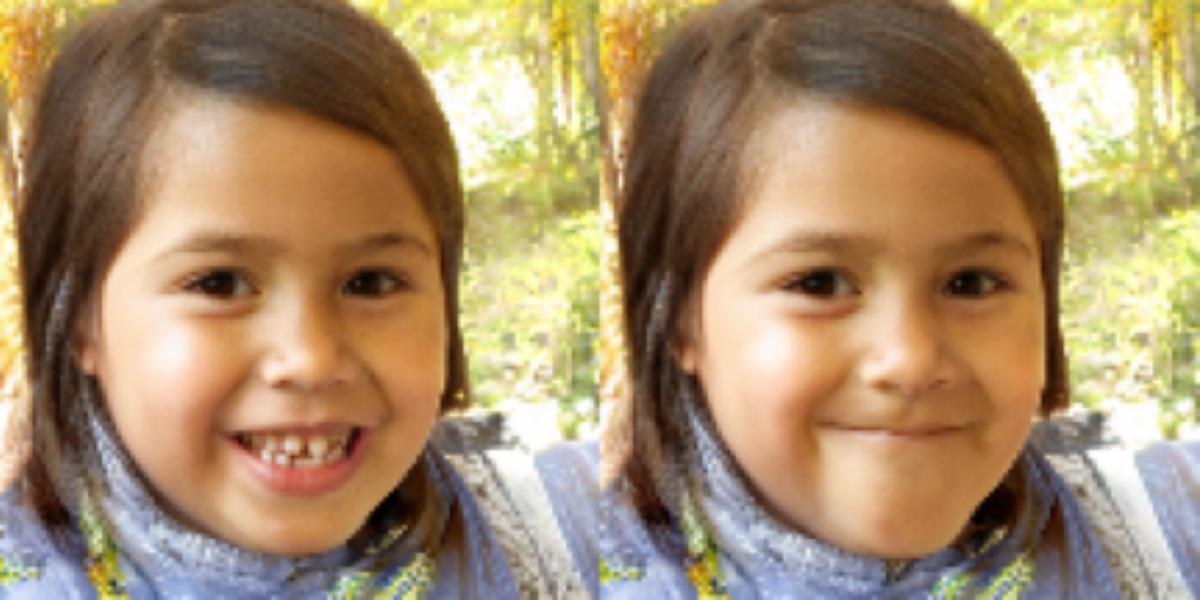} &
        \includegraphics[width=0.19\textwidth]{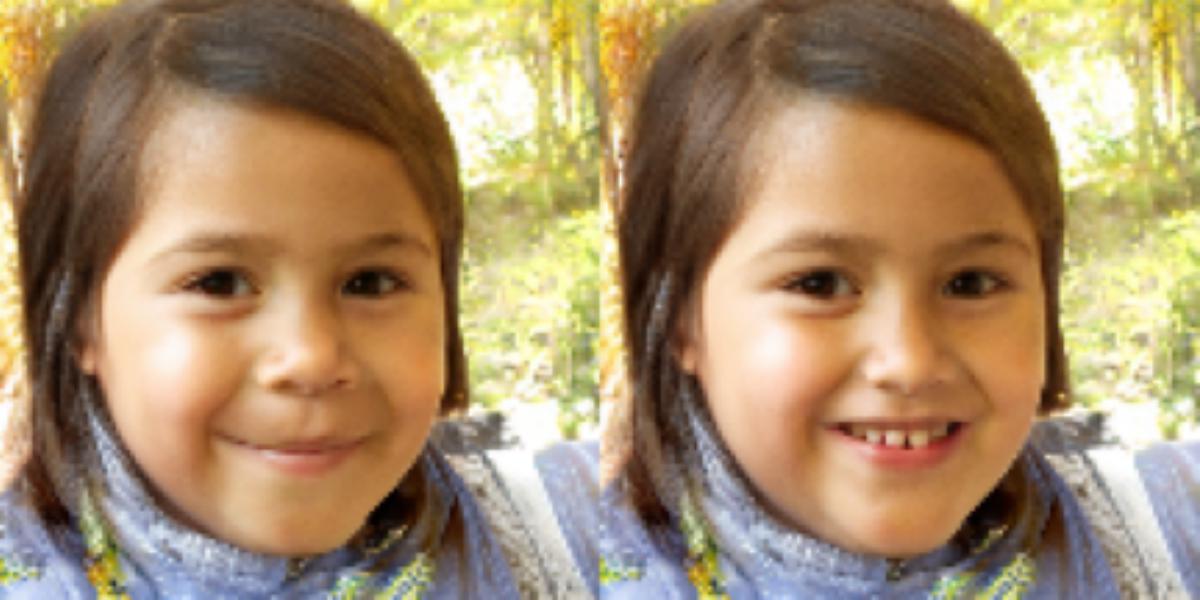} &
        \includegraphics[width=0.19\textwidth]{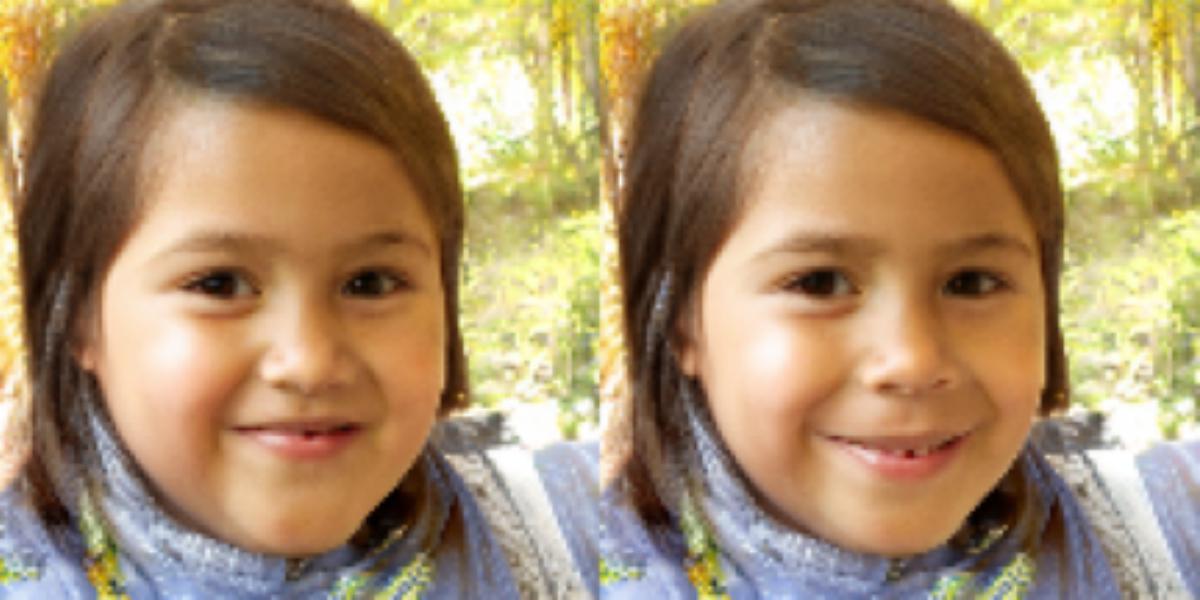}  &
        \includegraphics[width=0.19\textwidth]{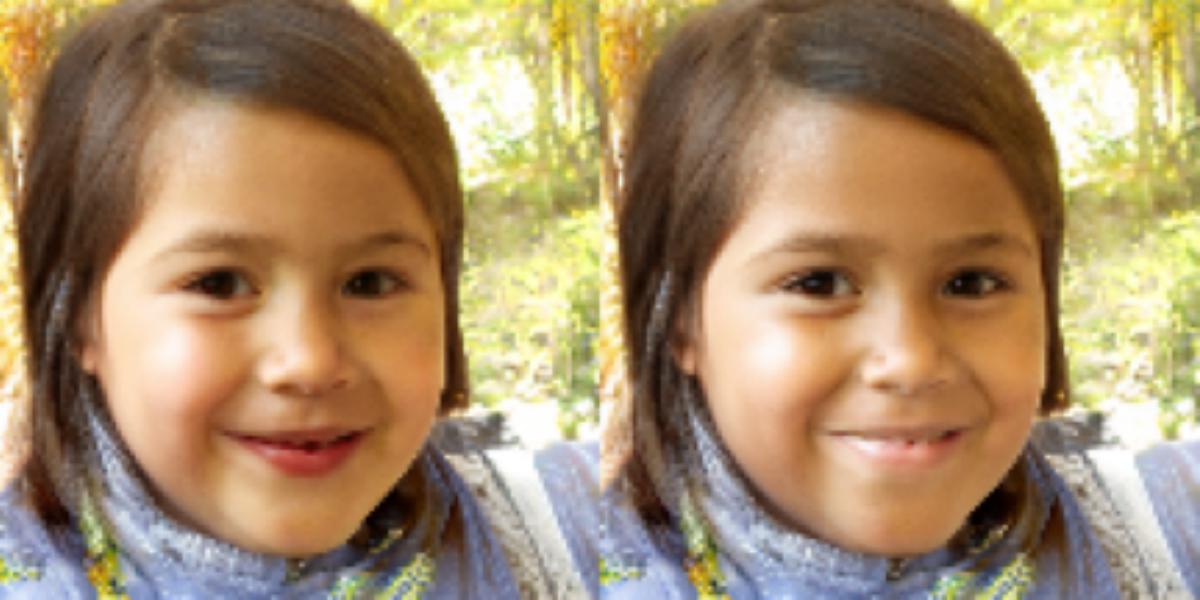}  &
        \includegraphics[width=0.19\textwidth]{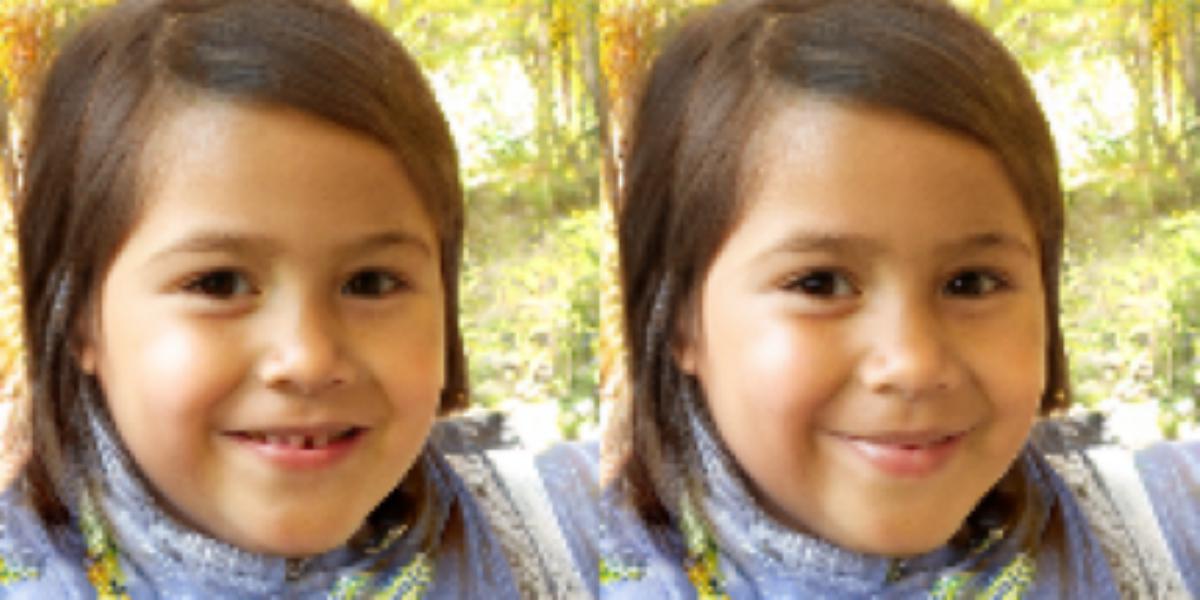}\\
        &\includegraphics[width=0.19\textwidth]{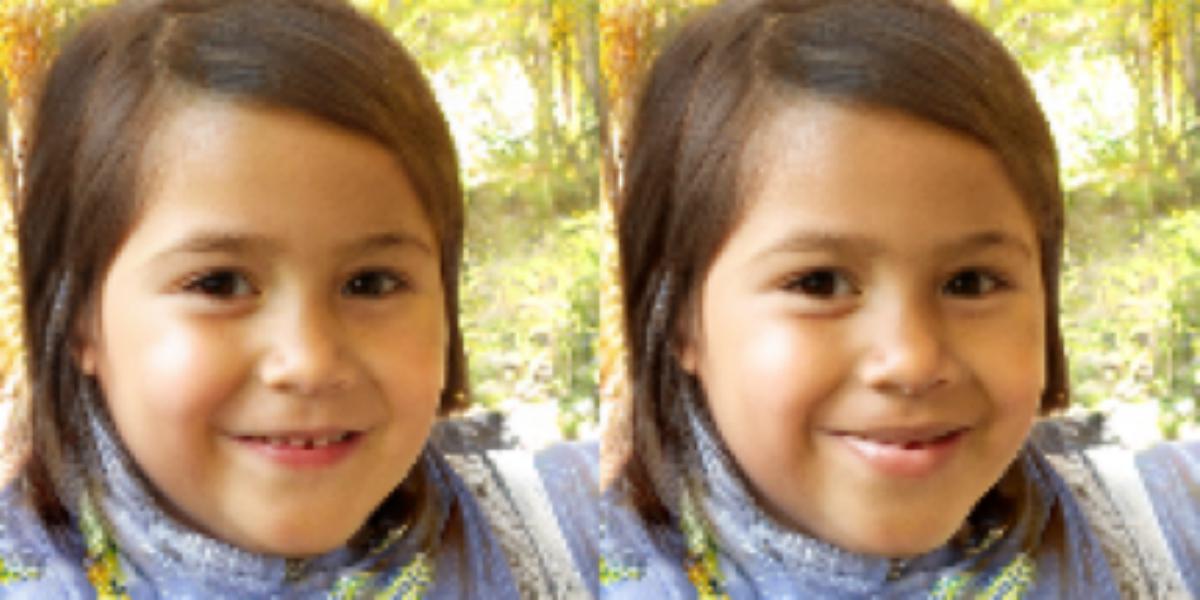} &
        \includegraphics[width=0.19\textwidth]{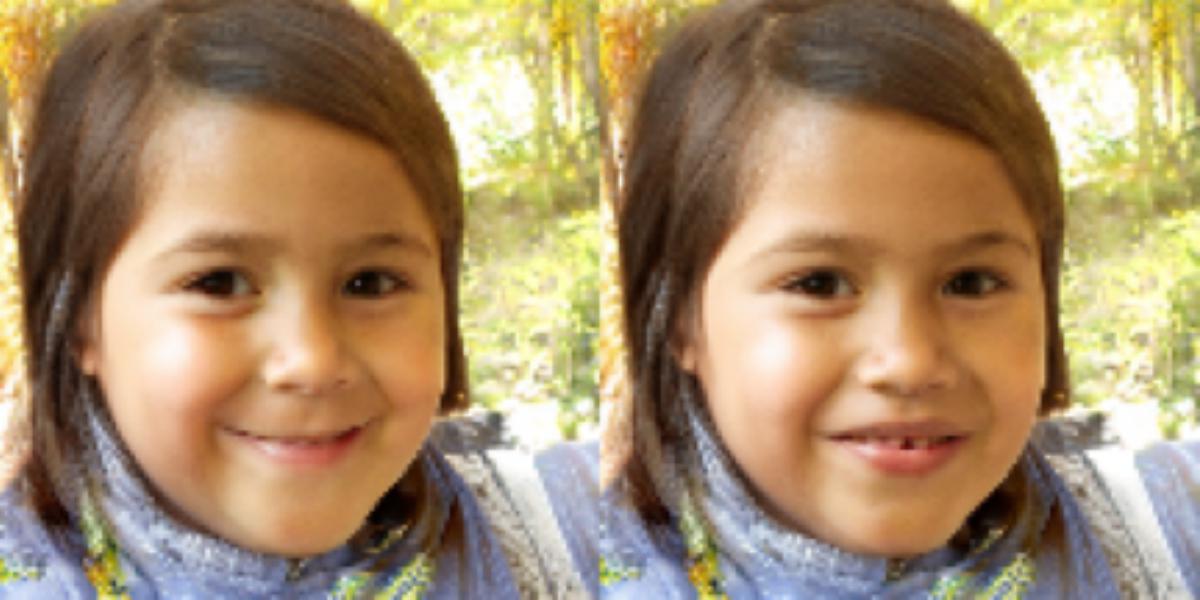} &
        \includegraphics[width=0.19\textwidth]{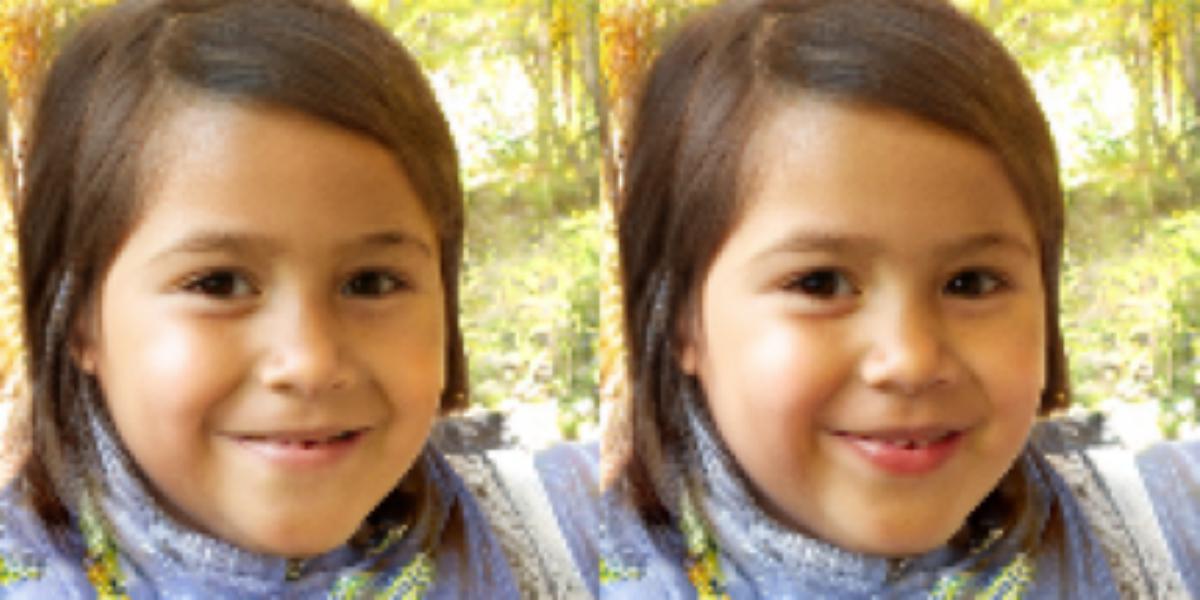}  &
        \includegraphics[width=0.19\textwidth]{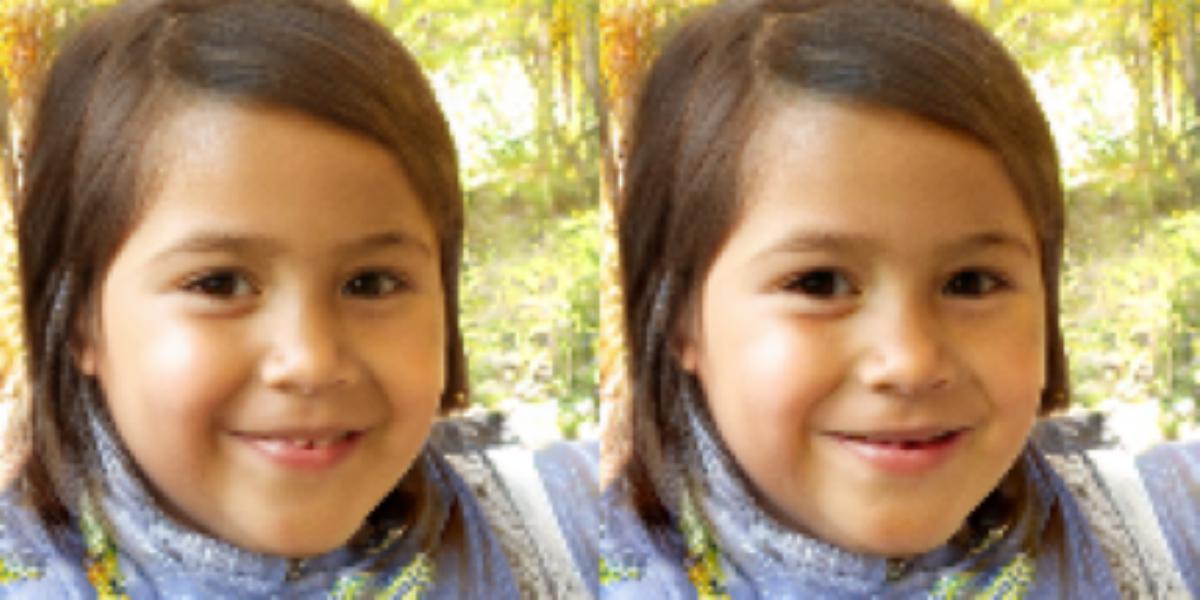}  &
        \includegraphics[width=0.19\textwidth]{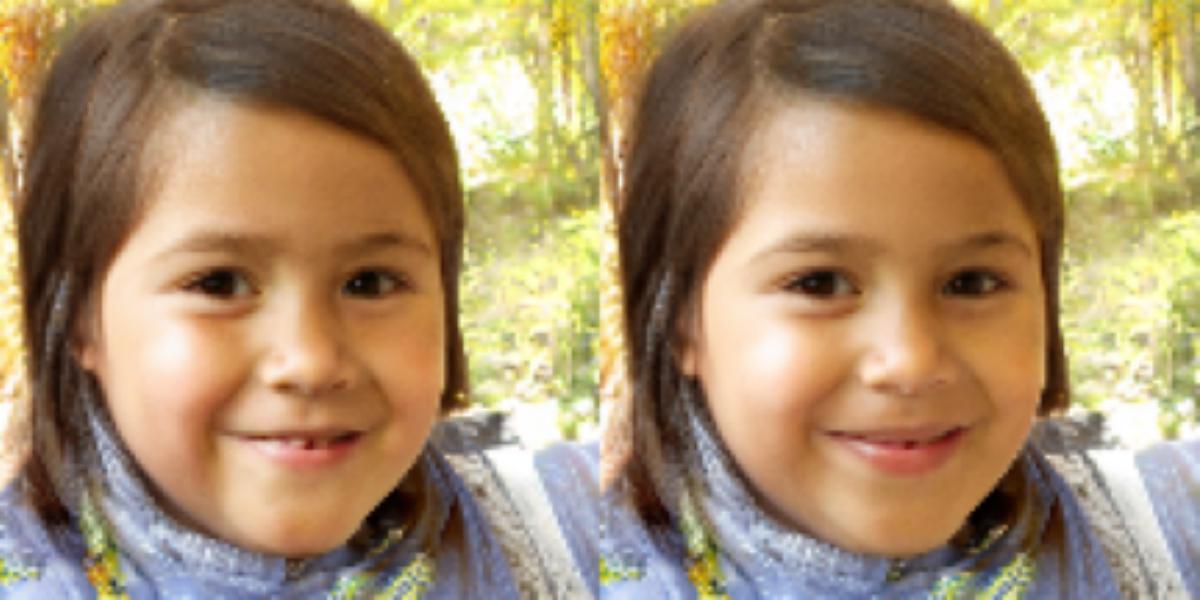}\\\hline
        \rule{0pt}{0.1\textwidth} \multirow{2}{*}{ \rotatebox{90}{$\WW, 0\text{-}9$} } &
        \includegraphics[width=0.19\textwidth]{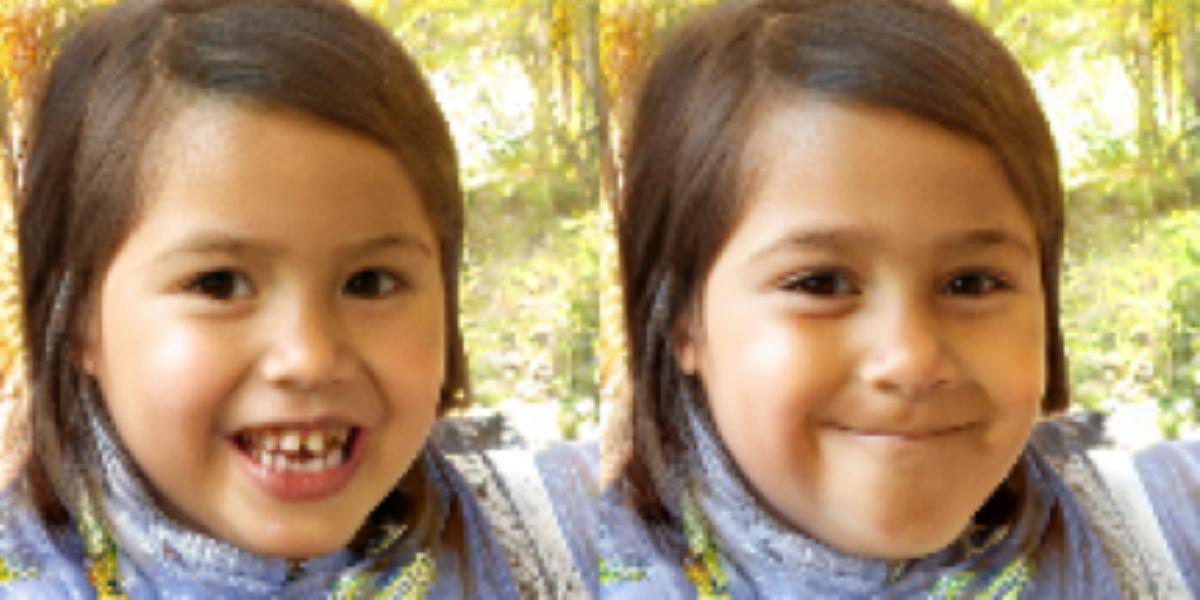} &
        \includegraphics[width=0.19\textwidth]{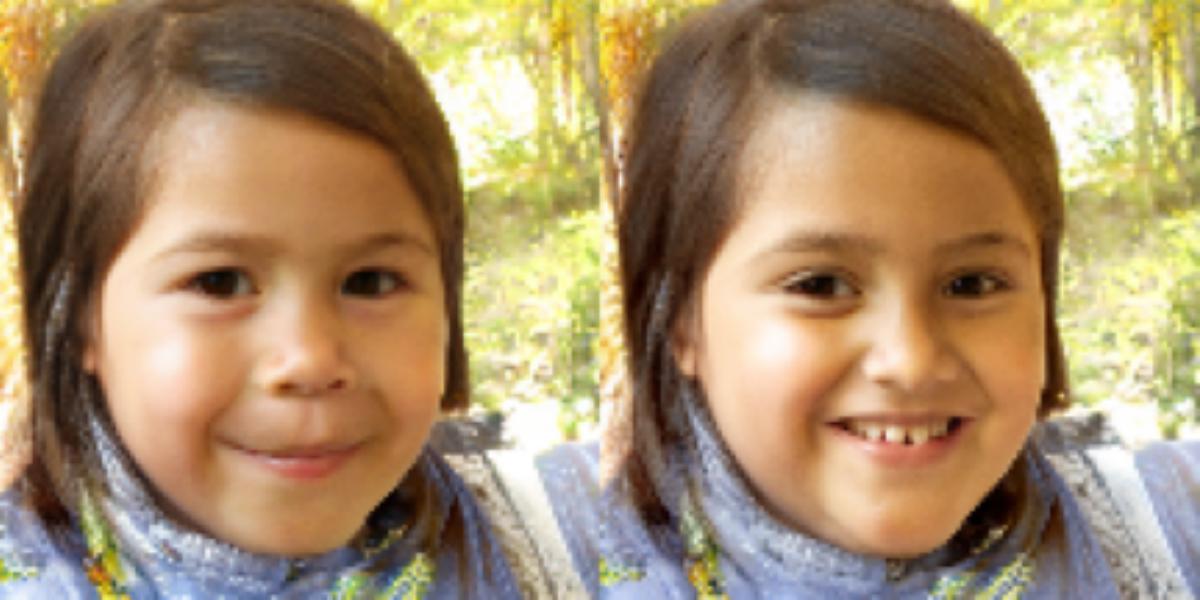} &
        \includegraphics[width=0.19\textwidth]{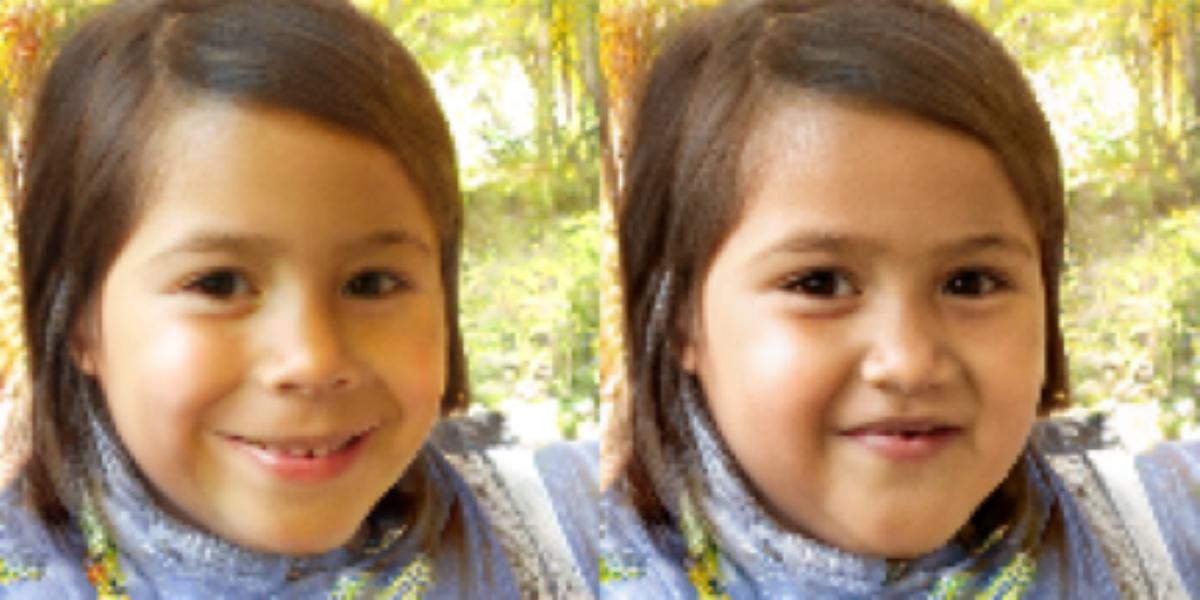}  &
        \includegraphics[width=0.19\textwidth]{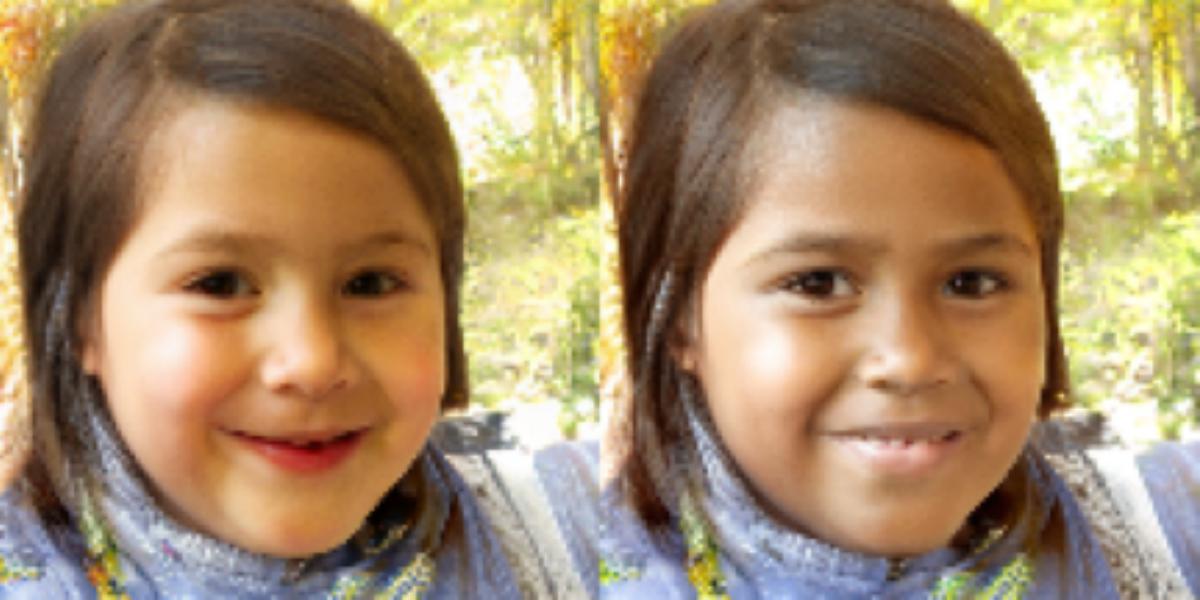}  &
        \includegraphics[width=0.19\textwidth]{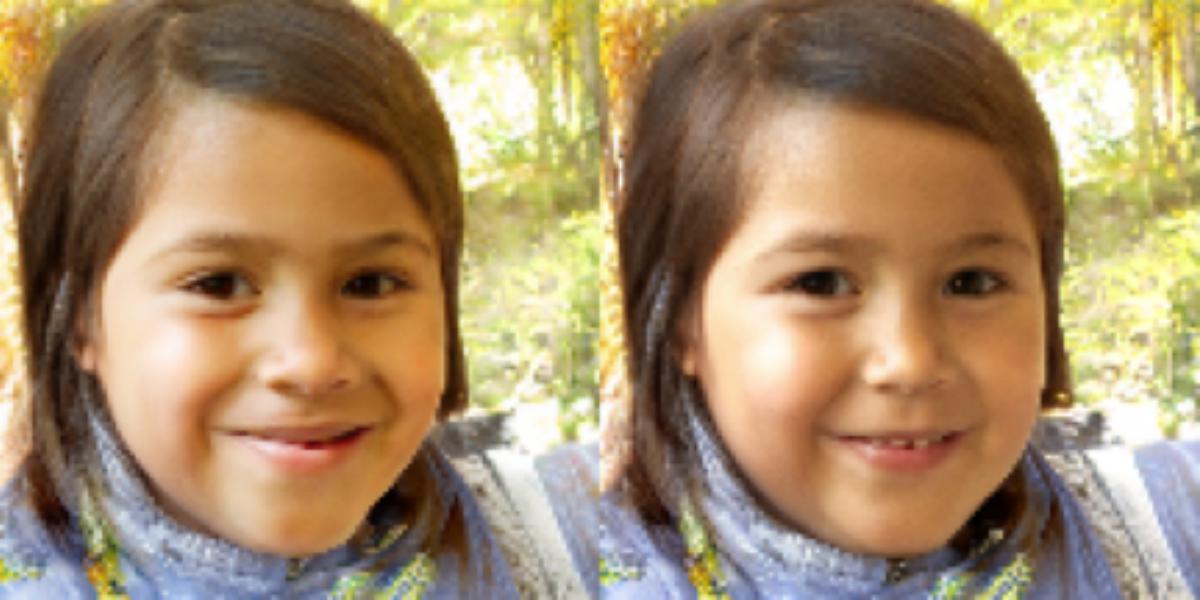}\\
        &\includegraphics[width=0.19\textwidth]{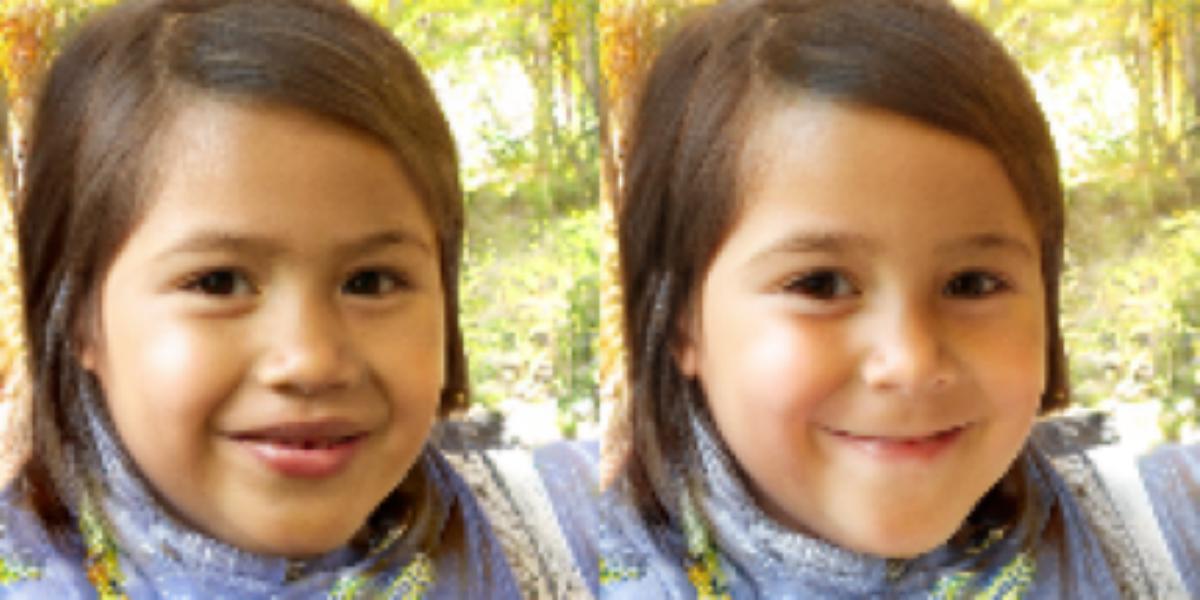} &
        \includegraphics[width=0.19\textwidth]{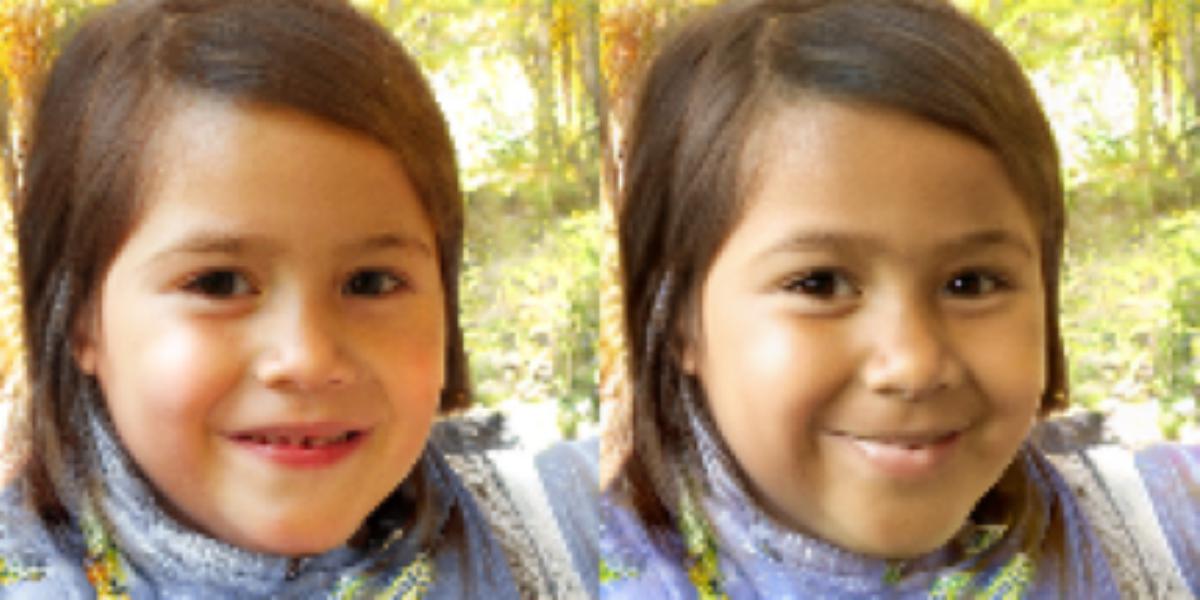} &
        \includegraphics[width=0.19\textwidth]{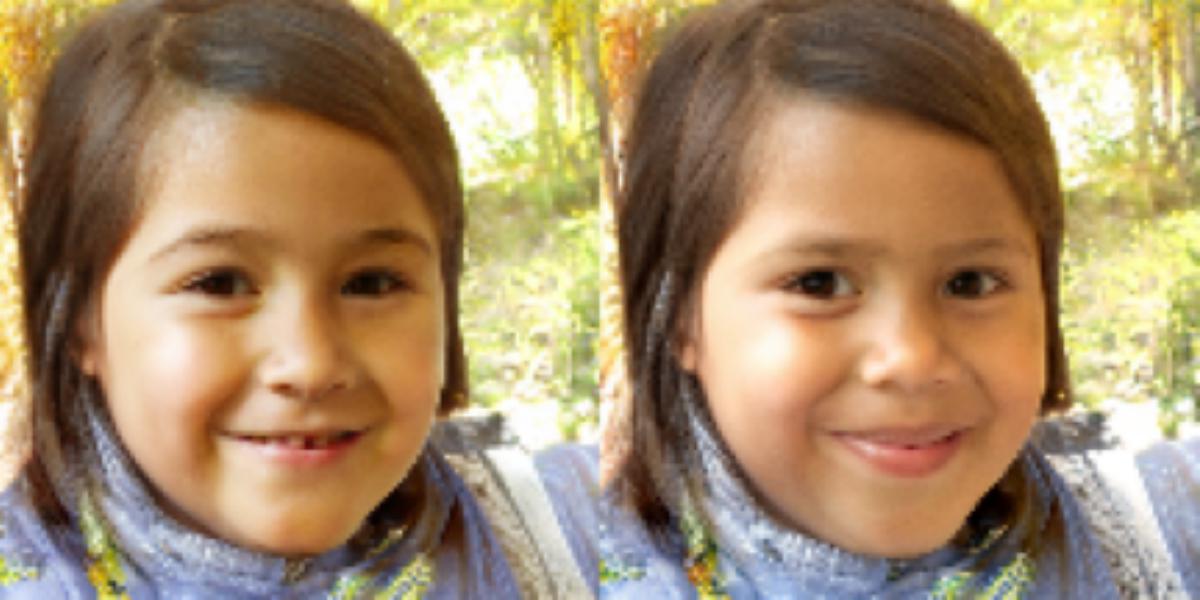}  &
        \includegraphics[width=0.19\textwidth]{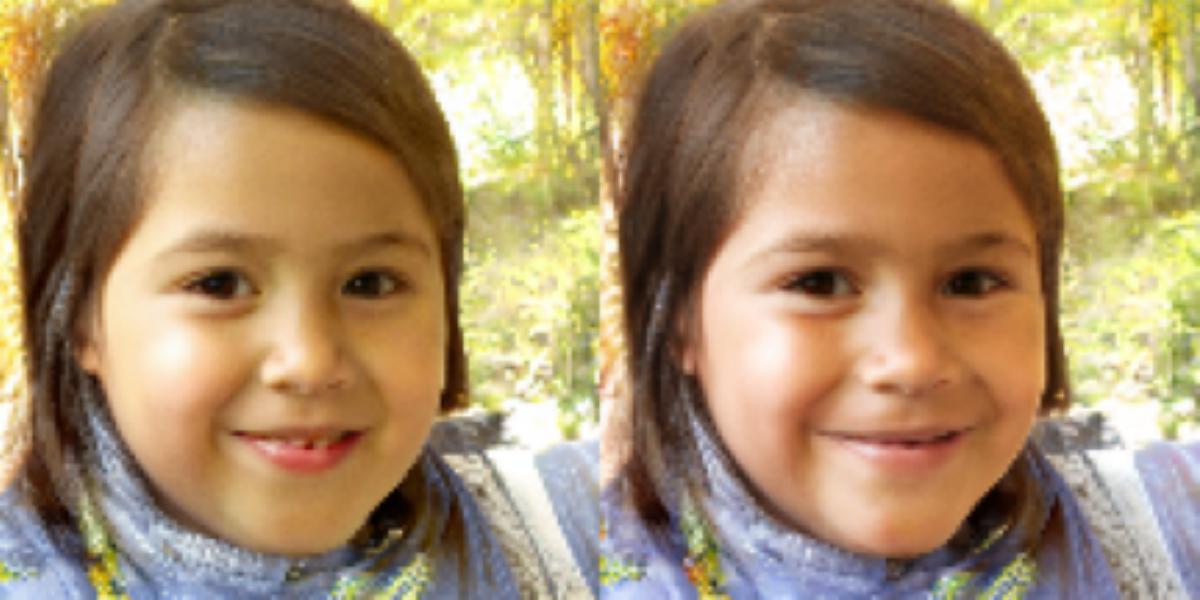}  &
        \includegraphics[width=0.19\textwidth]{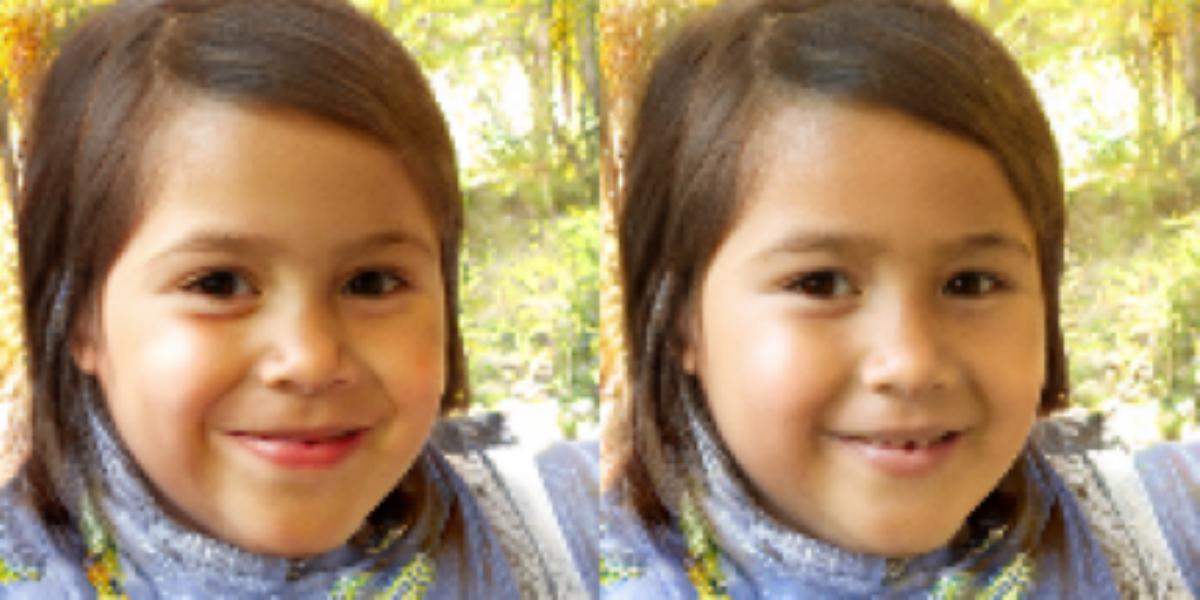}\\\hline
        \rule{0pt}{0.1\textwidth} \multirow{2}{*}{ \rotatebox{90}{$\ZZ, 0\text{-}9$} } &
        \includegraphics[width=0.19\textwidth]{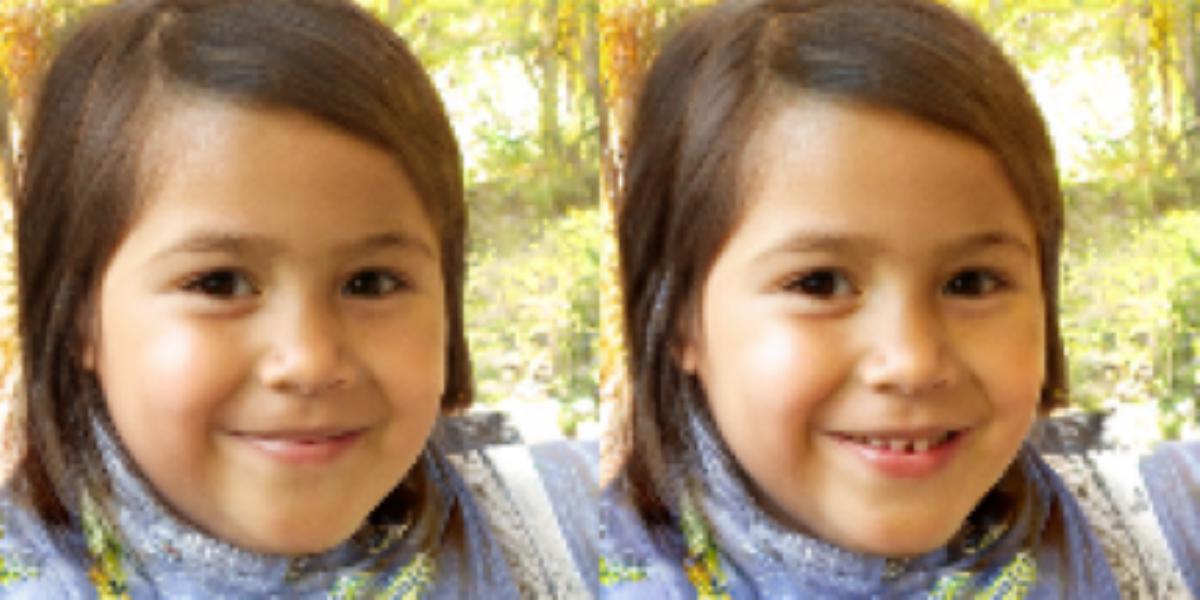} &
        \includegraphics[width=0.19\textwidth]{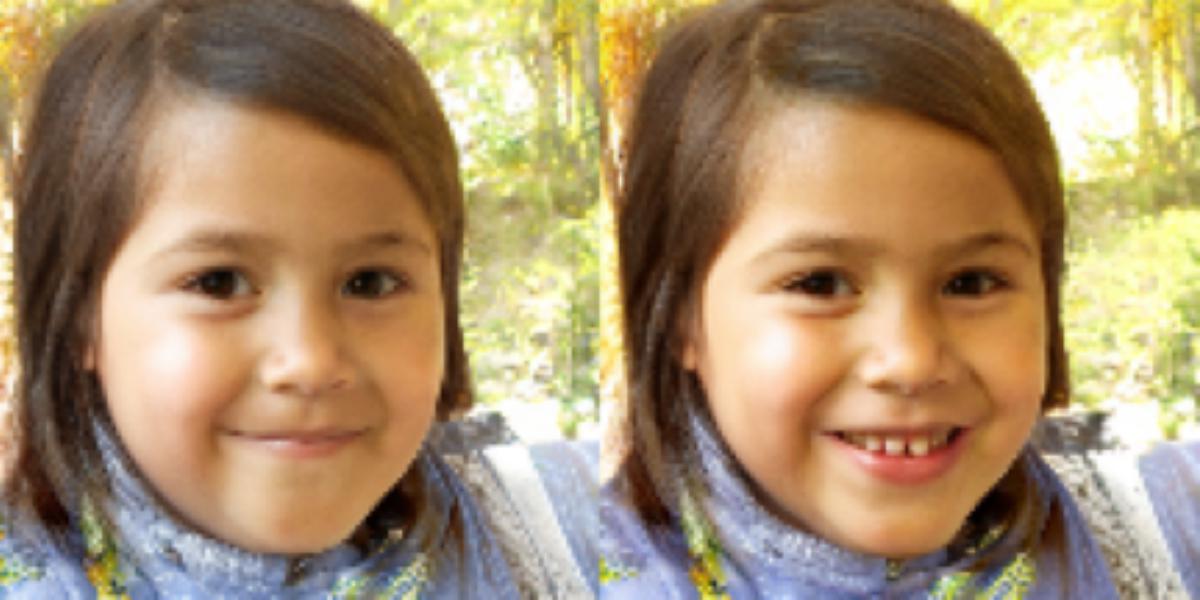} &
        \includegraphics[width=0.19\textwidth]{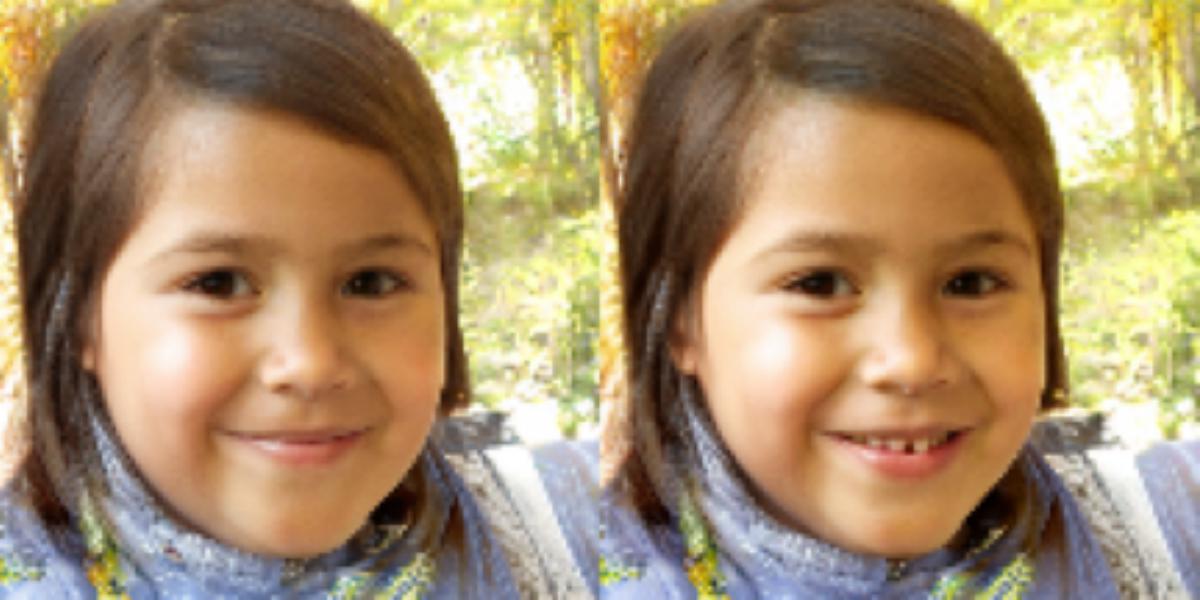}  &
        \includegraphics[width=0.19\textwidth]{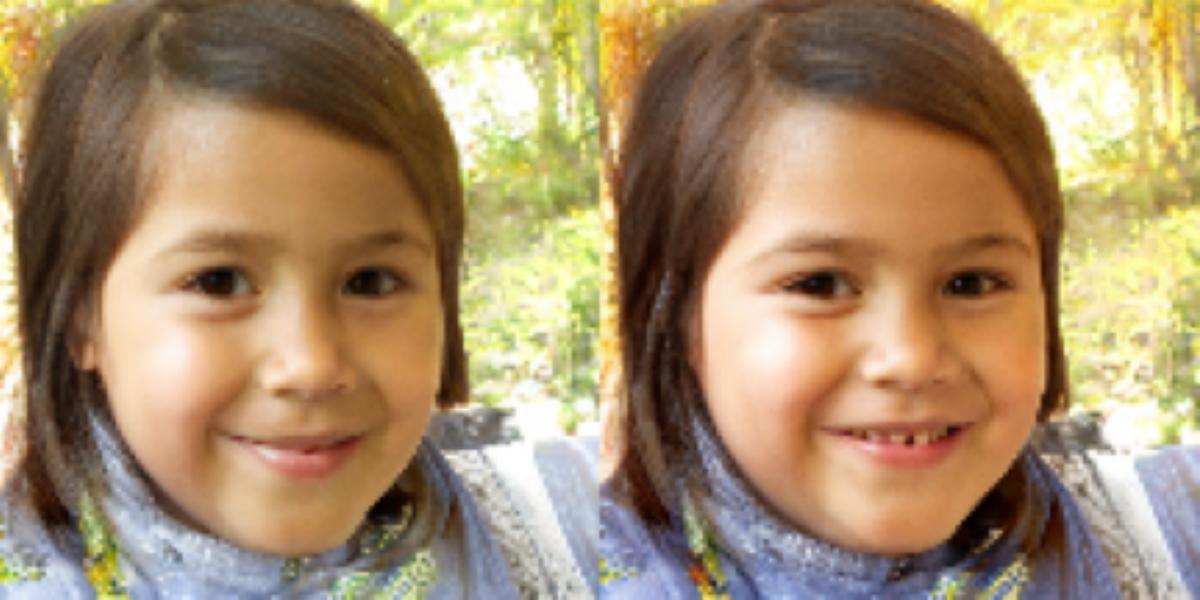}  &
        \includegraphics[width=0.19\textwidth]{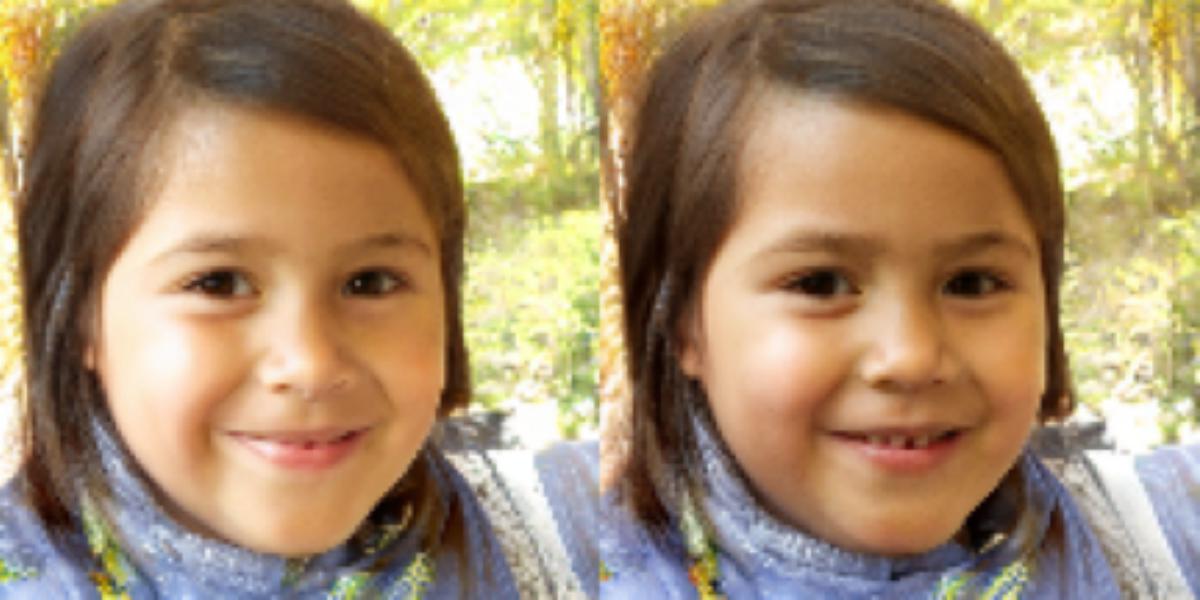}\\
        &\includegraphics[width=0.19\textwidth]{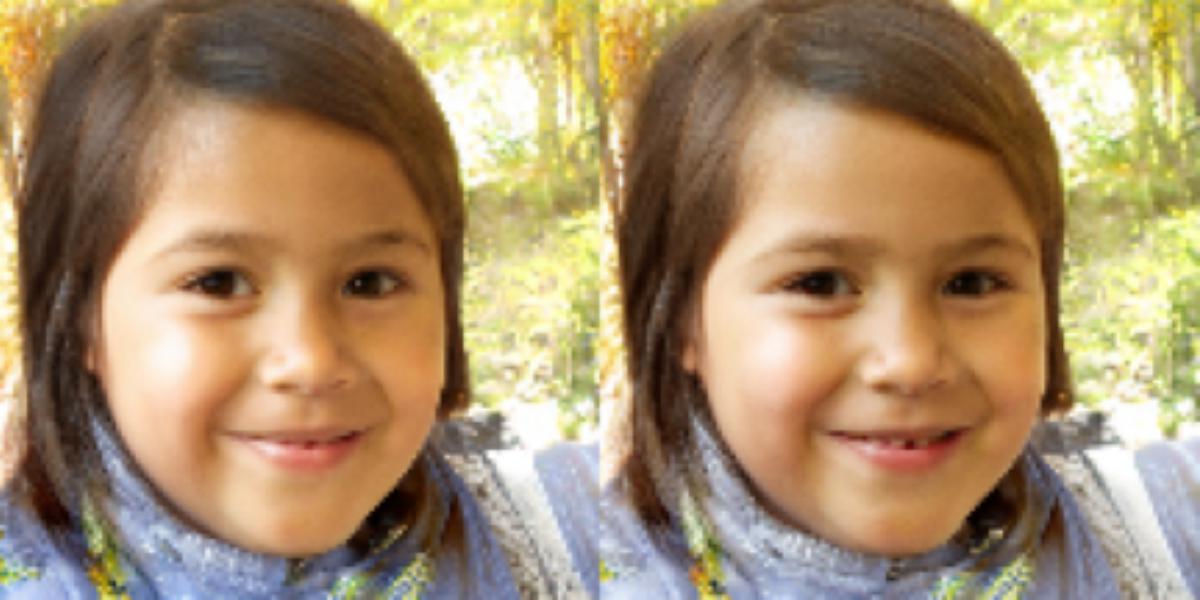} &
        \includegraphics[width=0.19\textwidth]{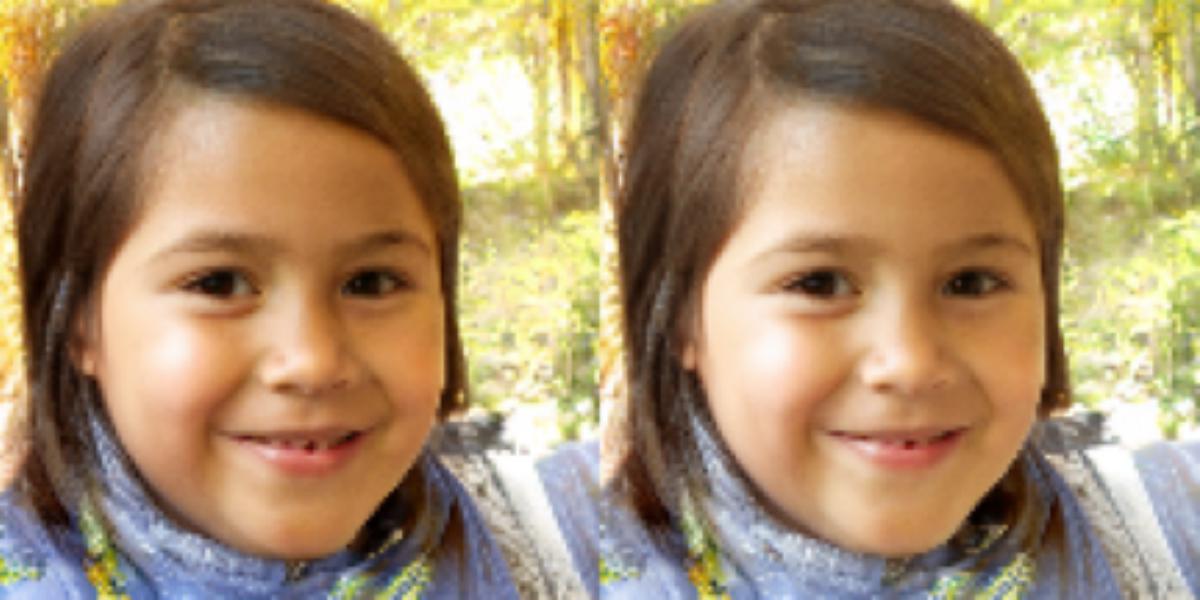} &
        \includegraphics[width=0.19\textwidth]{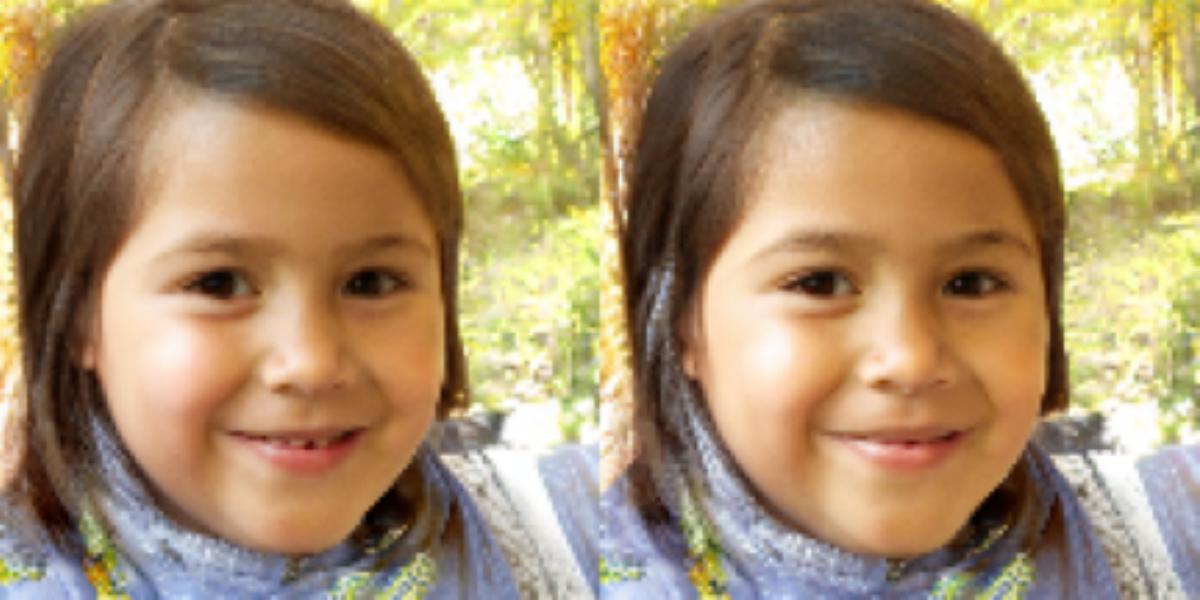}  &
        \includegraphics[width=0.19\textwidth]{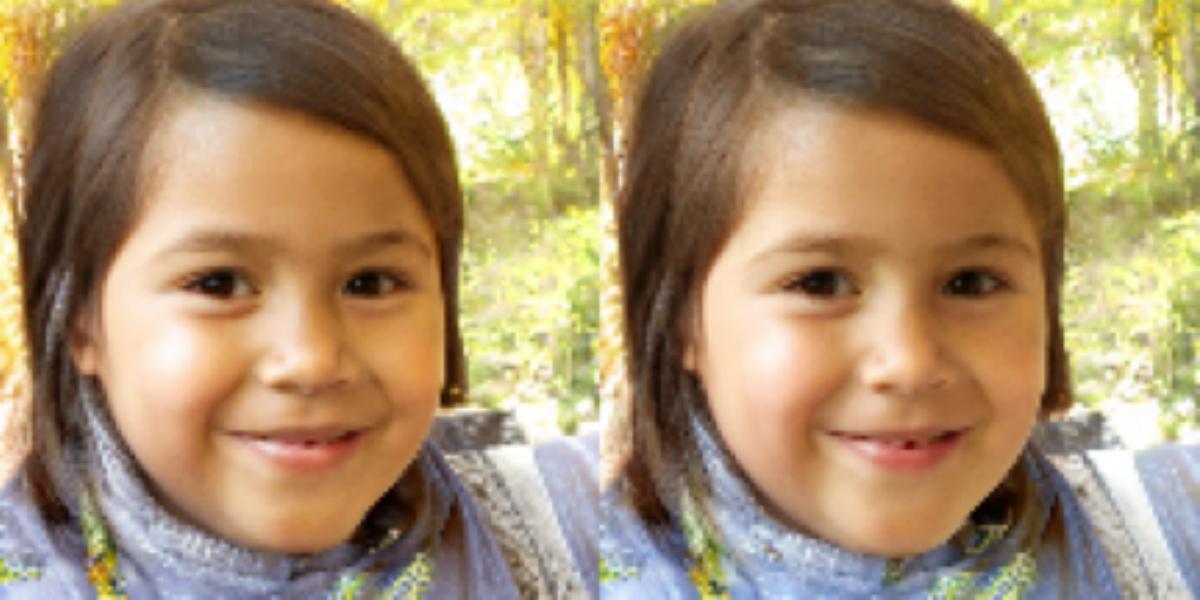}  &
        \includegraphics[width=0.19\textwidth]{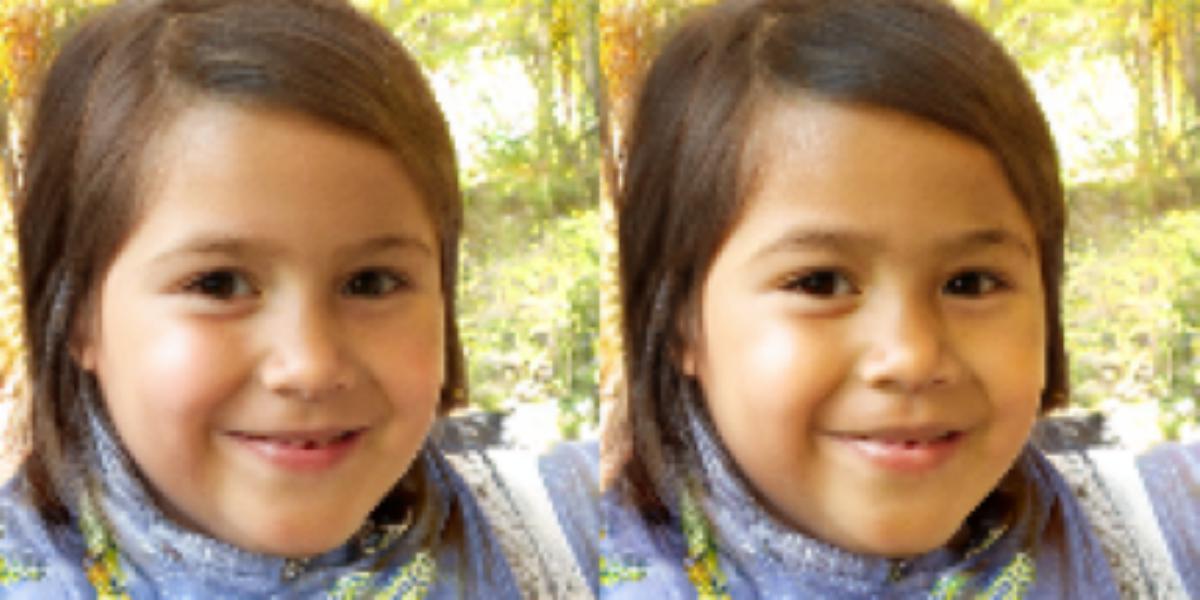}\\
    \end{tabular}
    \caption{Comparison of the control significance in the latent space of $\Ss, \WW+, \WW$, and $\ZZ$. See \Cref{sec:space-comparison-sup,fig:attenuation}. }
    \label{fig:space-comparison-image}
\end{figure*}

In \Cref{fig:space-comparison-image}, we show the modified face images by the first $10$ components in the above four latent spaces. A generated image is manipulated in corresponding latent space positively and negatively along the subspace principle component vectors. The principle components are the same with those in Figure~\ref{fig:attenuation}. It can be noticed that the manipulation significance attenuates fast in $\ZZ$ and $\WW$, while $\WW+$ and $\Ss$ possesses much richer manipulation directions.

Note that the basic formulation of \eqref{eq:l2-diff} follows~\cite{zhu2021low}. If we only consider the masked photometry criterion, the major difference is that we explore $\Ss$ while \cite{zhu2021low} works with $\ZZ$. Thus our latent space comparison experiments can be also considered as a comparison with~\cite{zhu2021low}.

\section{Subspaces Discovery Experiments}
\label{sec:exp-subspace-sup}

\begin{figure*}
    \centering
    \scriptsize
    \setlength{\tabcolsep}{1pt}
    \begin{tabular}{ccccccccccc}
         \rotatebox{90}{Skin color}& 
        \includegraphics[width=0.19\textwidth]{figs/subspace_viz/face_color/r0_b0_v0.jpg} &
        \includegraphics[width=0.19\textwidth]{figs/subspace_viz/face_color/r1_b0_v0.jpg} &
        \includegraphics[width=0.19\textwidth]{figs/subspace_viz/face_color/r2_b0_v0.jpg} &
        \includegraphics[width=0.19\textwidth]{figs/subspace_viz/face_color/r3_b0_v0.jpg} &
        \includegraphics[width=0.19\textwidth]{figs/subspace_viz/face_color/r4_b0_v0.jpg} \\
        & Black/white & Blue/yellow & Red/green & Pink/white & Green/red\\
        \rotatebox{90}{Face boundary}& 
        \includegraphics[width=0.19\textwidth]{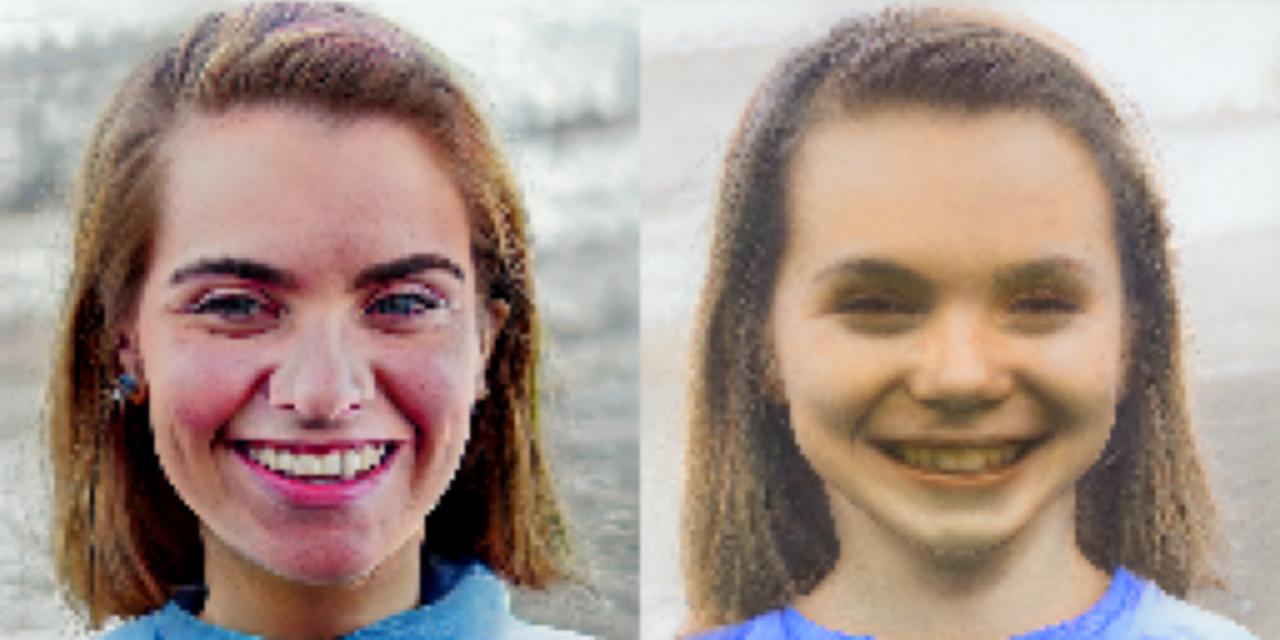} &
        \includegraphics[width=0.19\textwidth]{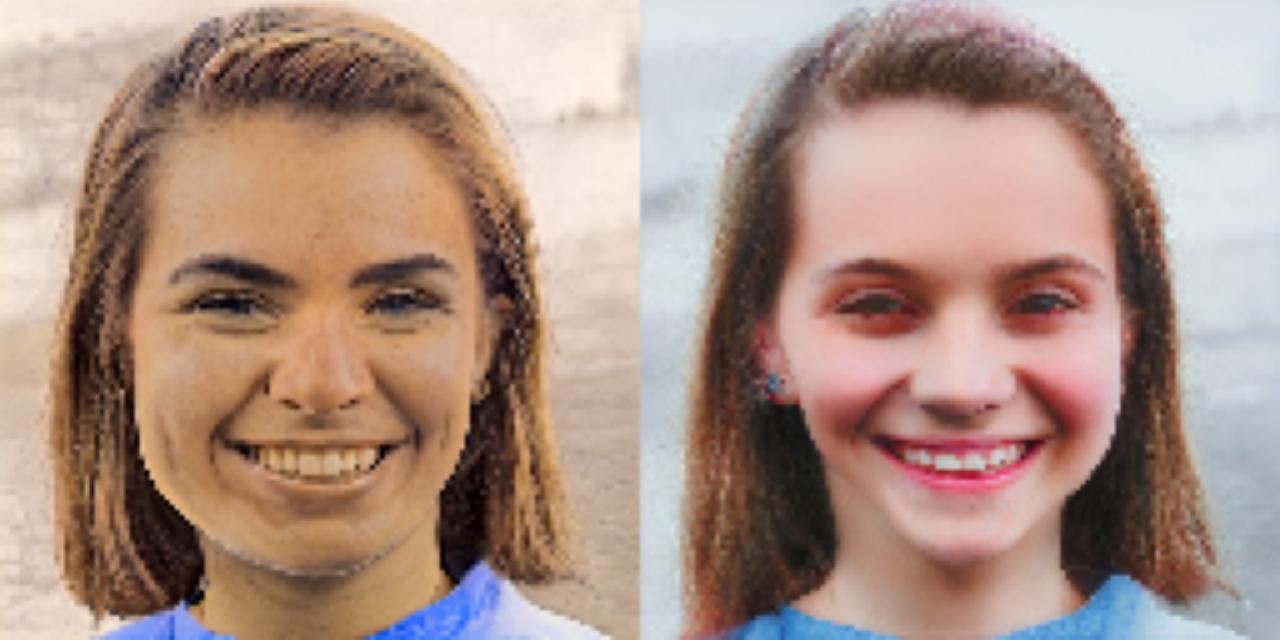} &
        \includegraphics[width=0.19\textwidth]{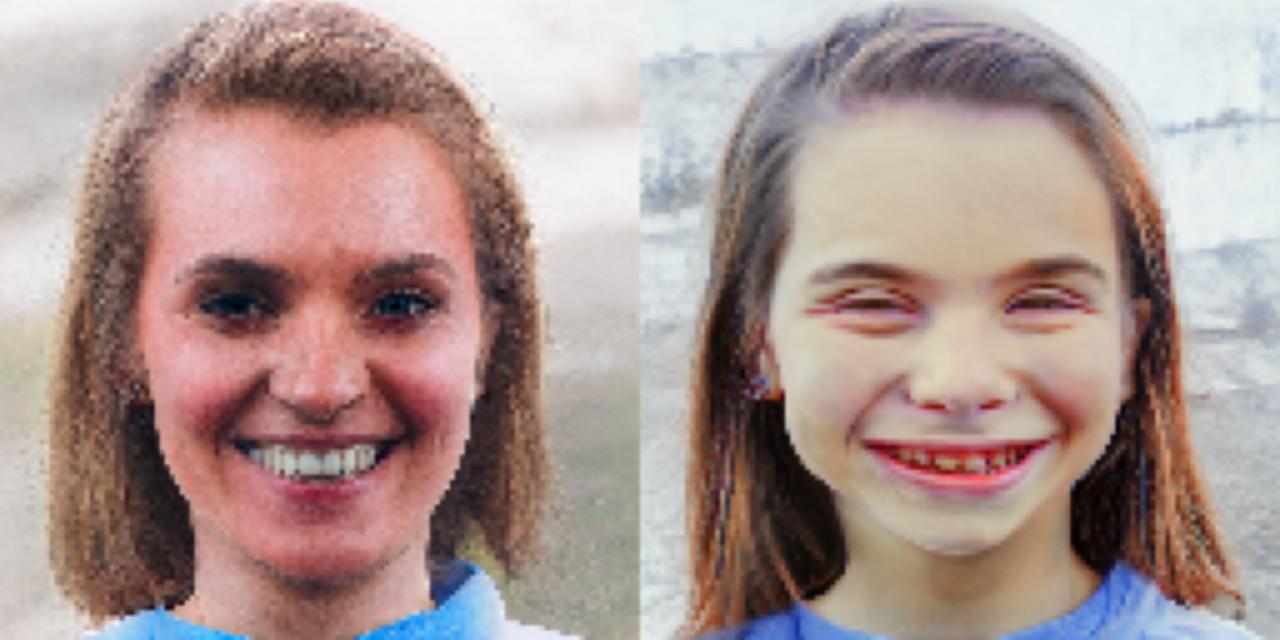} &
        \includegraphics[width=0.19\textwidth]{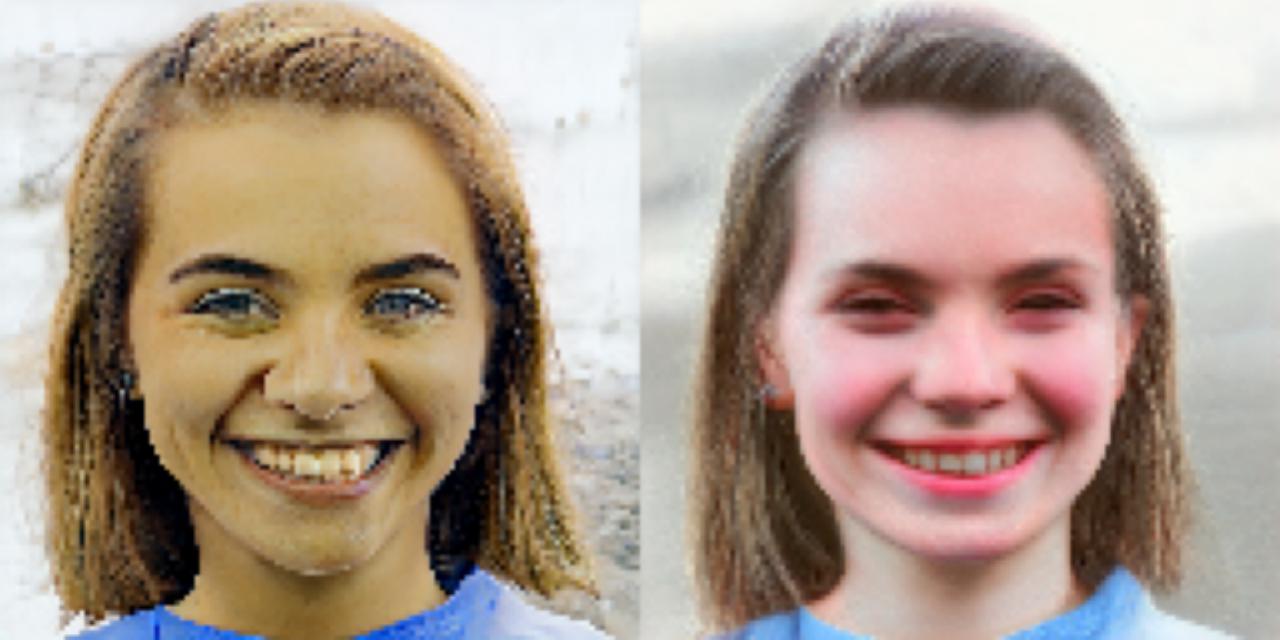} &
        \includegraphics[width=0.19\textwidth]{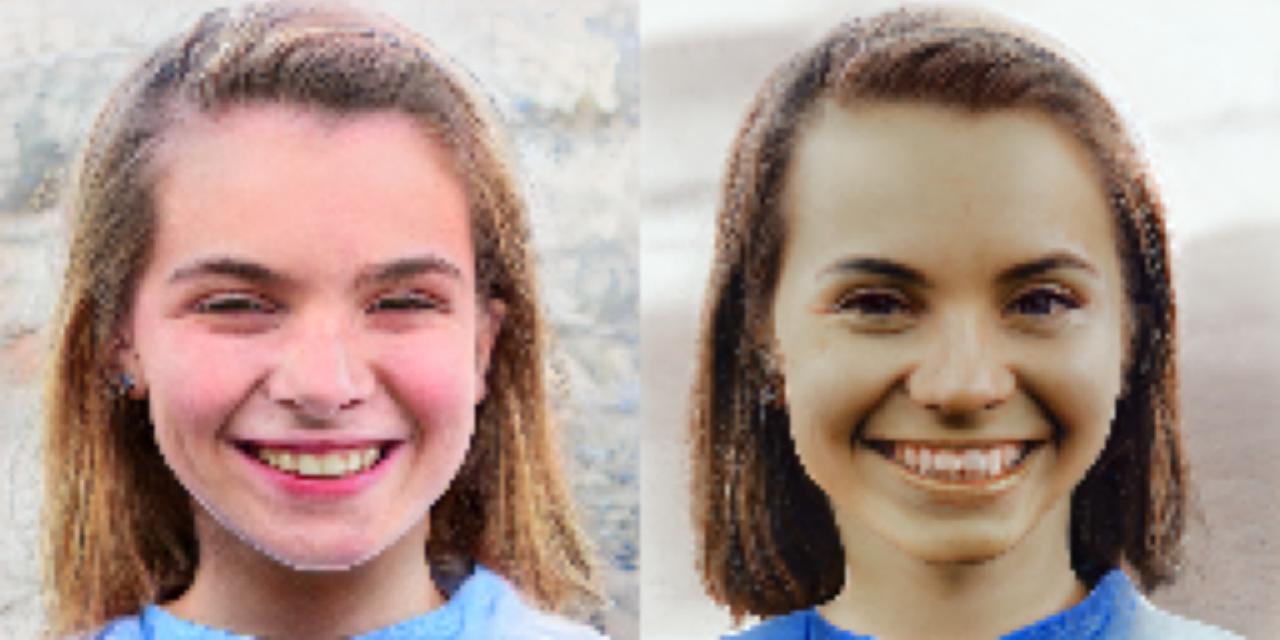} \\
        & Chin length & Asymmetric cheek & Fat chin & Pointed chin & Chin length \\
        \rotatebox{90}{Face boundary (id)}& 
        \includegraphics[width=0.19\textwidth]{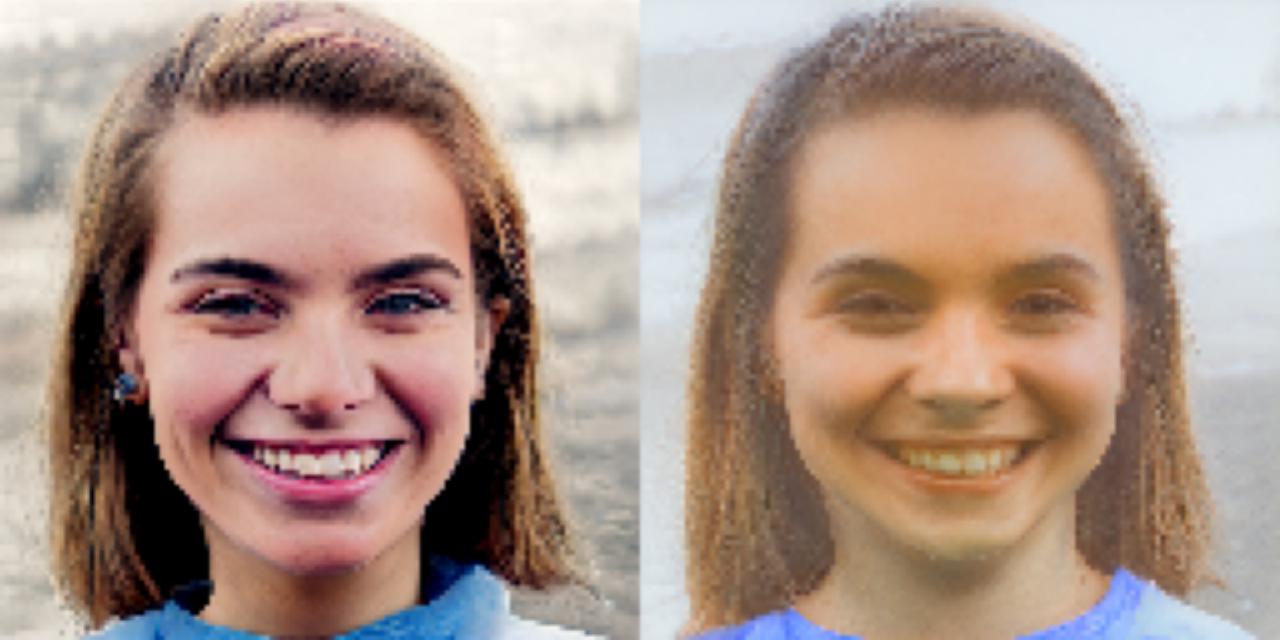} &
        \includegraphics[width=0.19\textwidth]{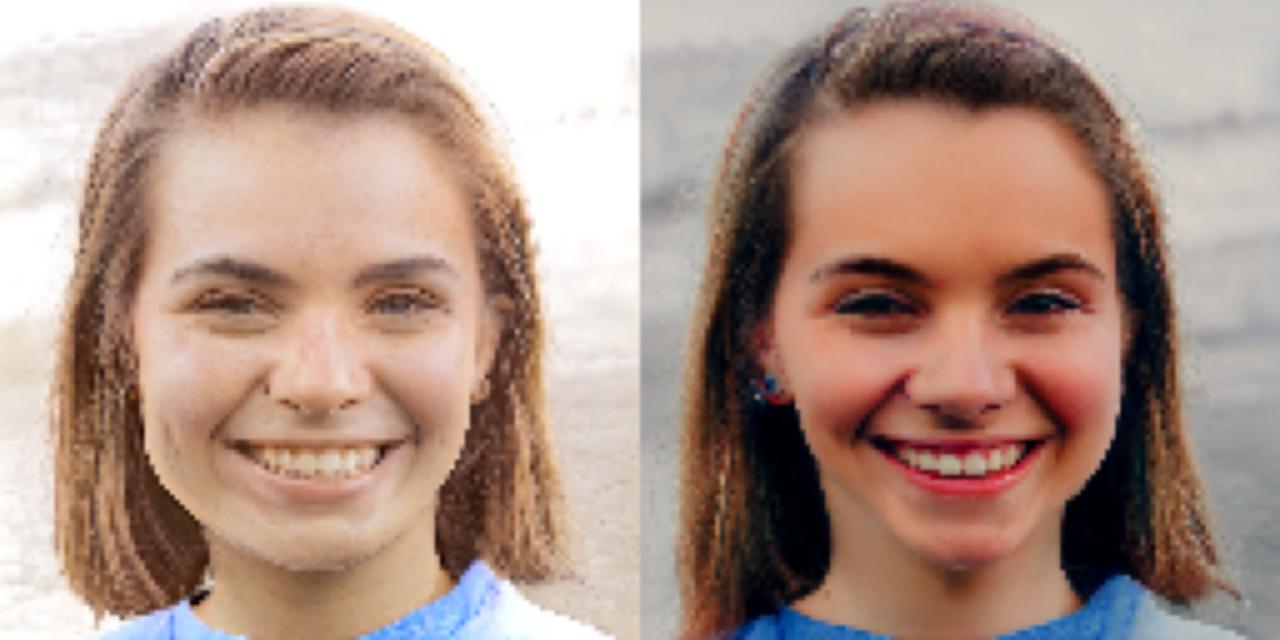} &
        \includegraphics[width=0.19\textwidth]{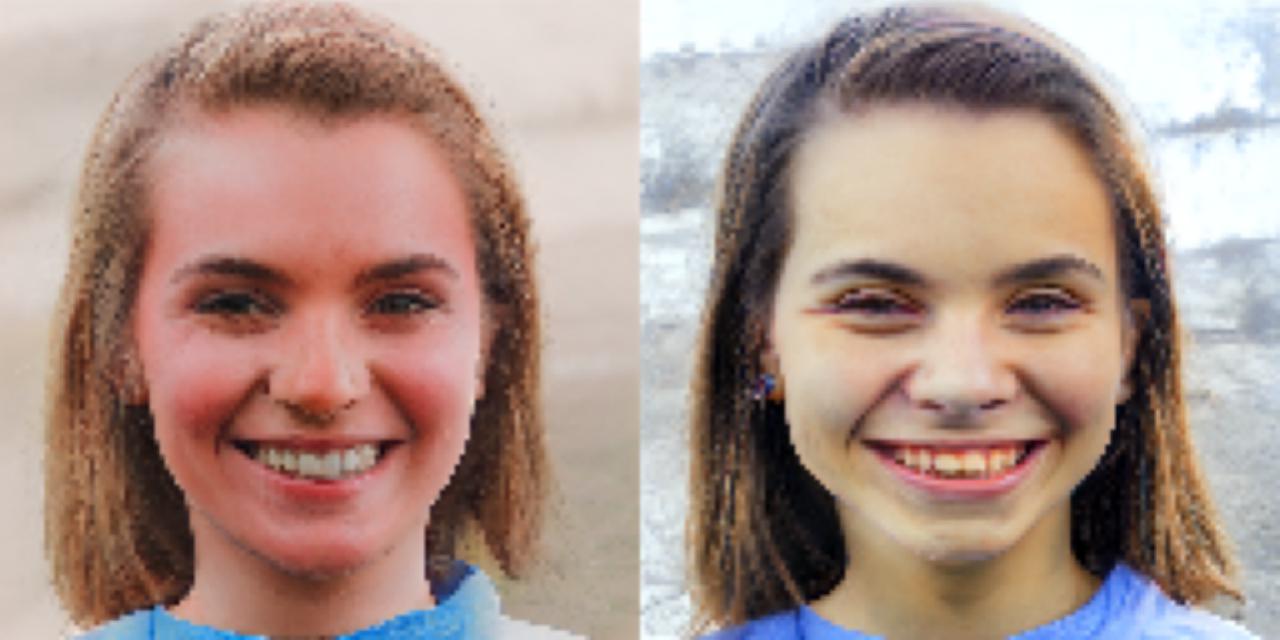} &
        \includegraphics[width=0.19\textwidth]{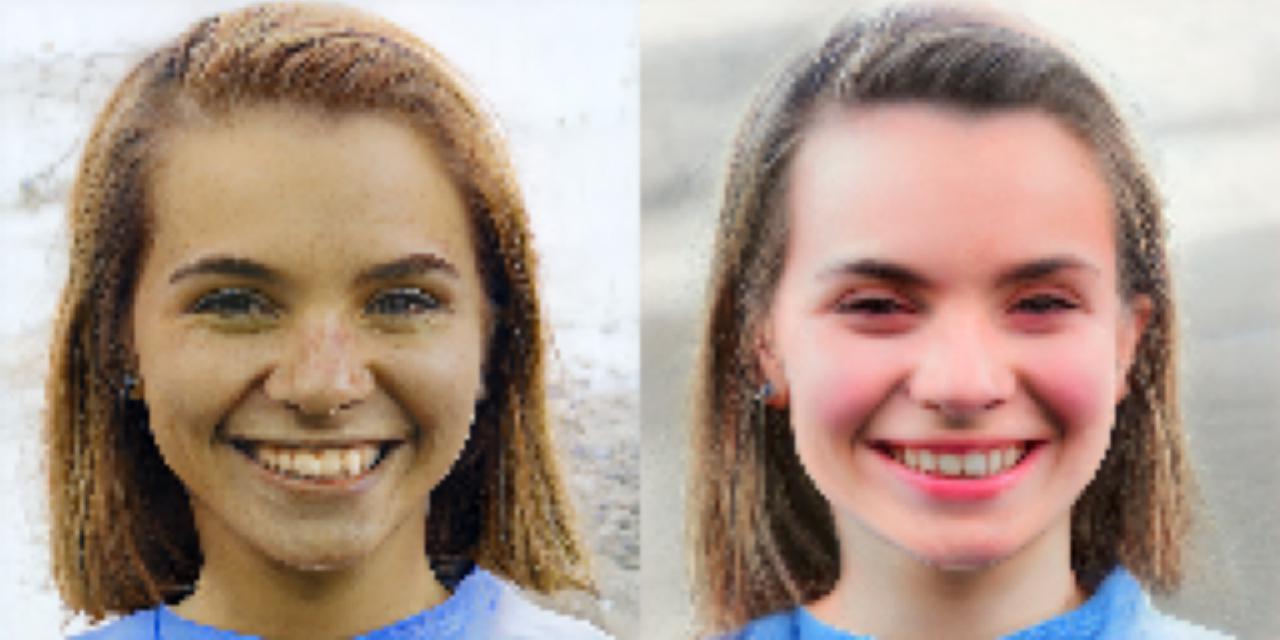} &
        \includegraphics[width=0.19\textwidth]{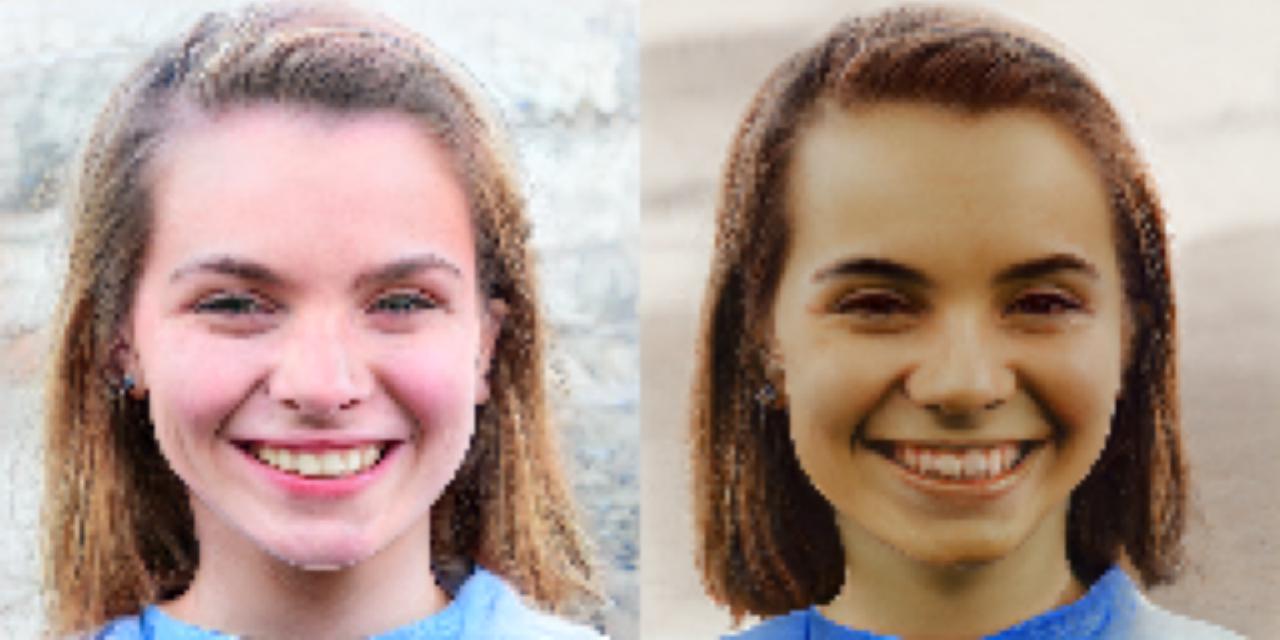} \\
        & Chin length & Asymmetric cheek & Fat chin & Pointed chin & Chin length \\
         \rotatebox{90}{Mouth shape}& 
        \includegraphics[width=0.19\textwidth]{figs/subspace_viz/mouth_shape_id/r0_b0_v0.jpg} &
        \includegraphics[width=0.19\textwidth]{figs/subspace_viz/mouth_shape_id/r1_b0_v0.jpg} &
        \includegraphics[width=0.19\textwidth]{figs/subspace_viz/mouth_shape_id/r3_b0_v0.jpg} &
        \includegraphics[width=0.19\textwidth]{figs/subspace_viz/mouth_shape_id/r4_b0_v0.jpg} &
        \includegraphics[width=0.19\textwidth]{figs/subspace_viz/mouth_shape_id/r5_b0_v0.jpg} \\
        & Curvy and wide smile & Curvey and narrow smile & Smile width & Curvy and wide smile & Asymmetric smile\\
         \rotatebox{90}{Mouth shape (id)}& 
        \includegraphics[width=0.19\textwidth]{figs/subspace_viz/mouth_shape_id/r0_b0_v1.jpg} &
        \includegraphics[width=0.19\textwidth]{figs/subspace_viz/mouth_shape_id/r1_b0_v1.jpg} &
        \includegraphics[width=0.19\textwidth]{figs/subspace_viz/mouth_shape_id/r3_b0_v1.jpg} &
        \includegraphics[width=0.19\textwidth]{figs/subspace_viz/mouth_shape_id/r4_b0_v1.jpg} &
        \includegraphics[width=0.19\textwidth]{figs/subspace_viz/mouth_shape_id/r5_b0_v1.jpg} \\
        & Curvy and wide smile & Curvey and narrow smile & Smile width & Curvy and wide smile & Asymmetric smile\\
         \rotatebox{90}{Lip color}& 
        \includegraphics[width=0.19\textwidth]{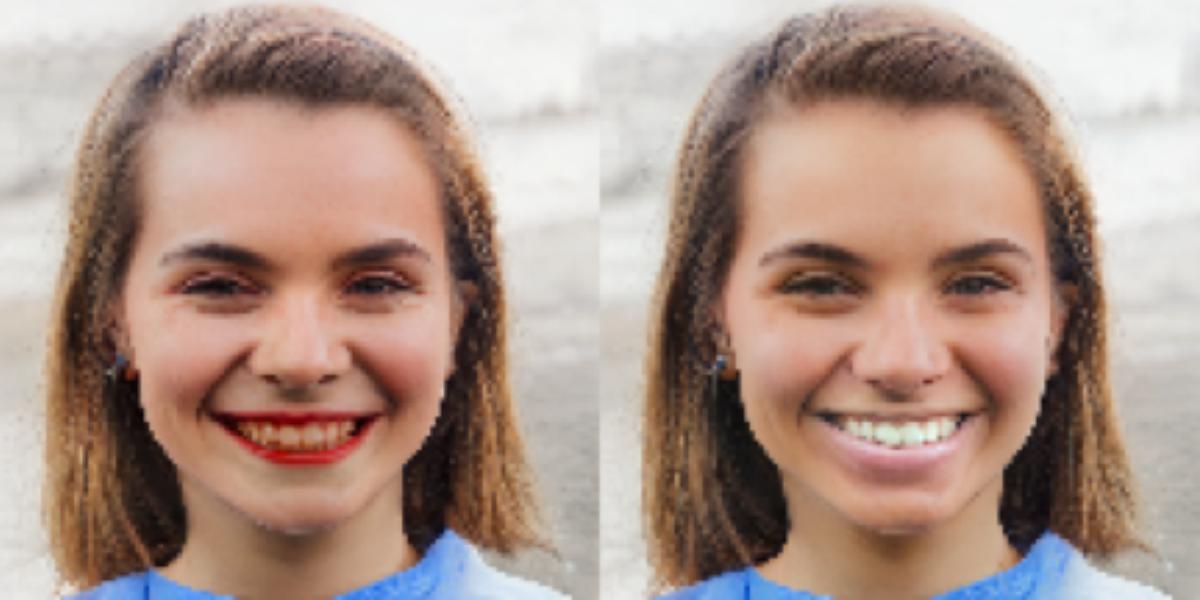} &
        \includegraphics[width=0.19\textwidth]{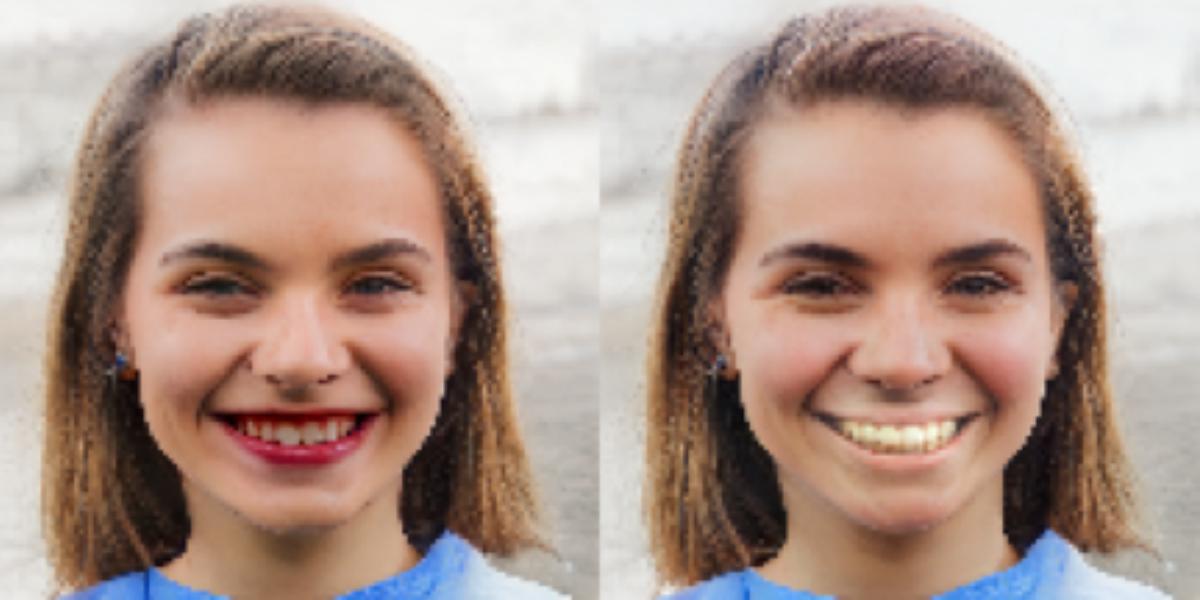} &
        \includegraphics[width=0.19\textwidth]{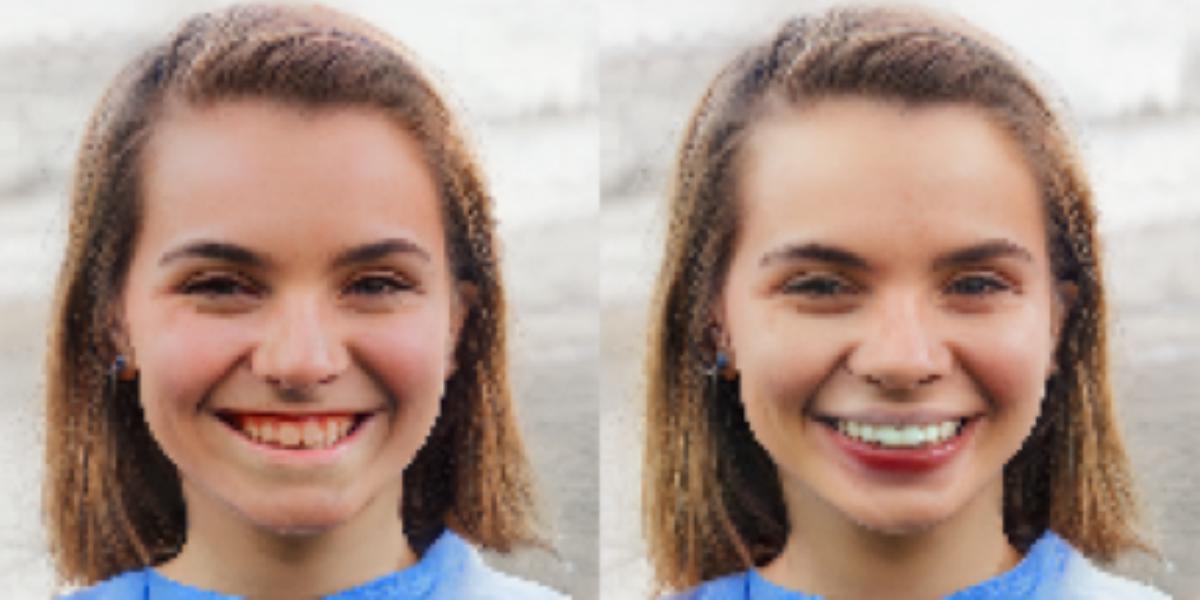} &
        \includegraphics[width=0.19\textwidth]{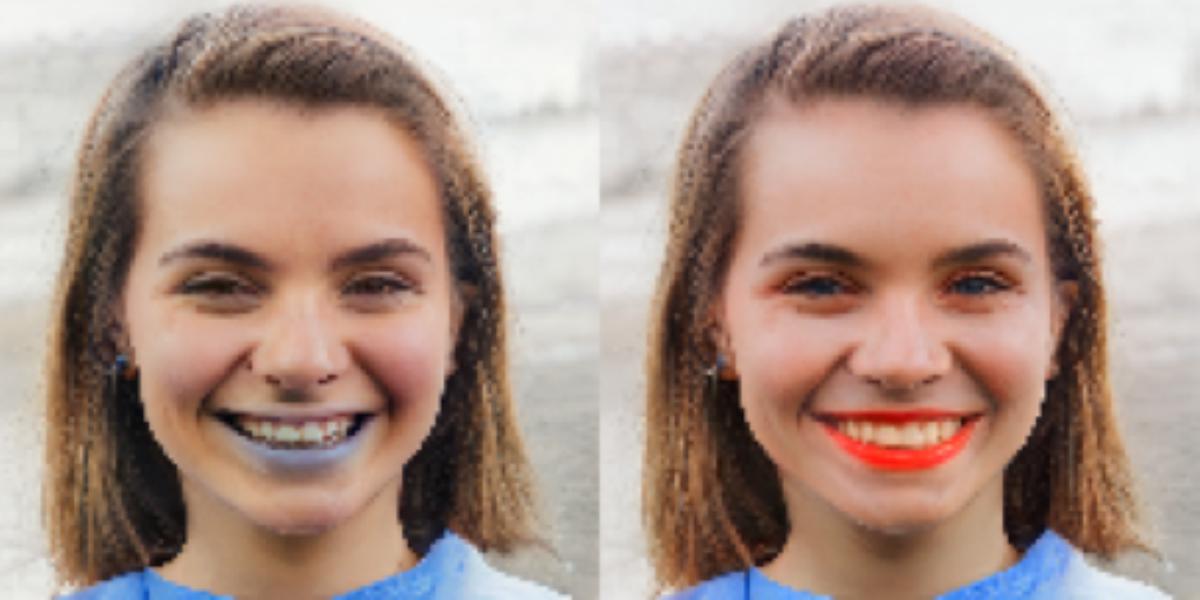} &
        \includegraphics[width=0.19\textwidth]{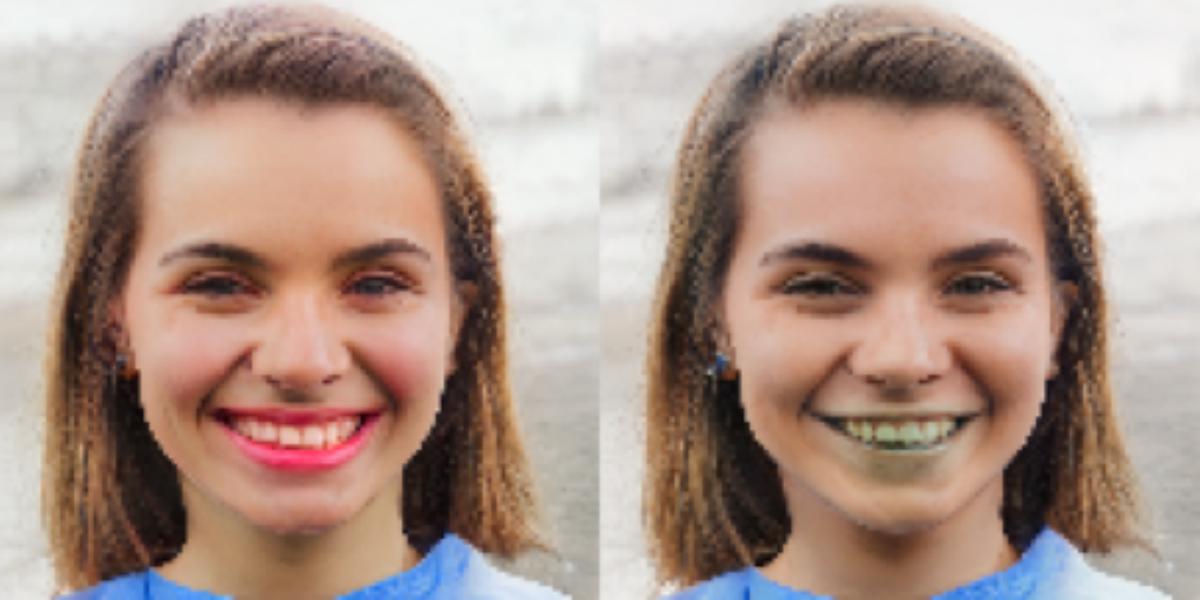} \\
        & Red/white & Red/white & Lip thickness & Blue/red & Green/magenta \\
         \rotatebox{90}{Eye photometry}& 
        \includegraphics[width=0.19\textwidth]{figs/subspace_viz/eye_photom/r0_b0_v0.jpg} &
        \includegraphics[width=0.19\textwidth]{figs/subspace_viz/eye_photom/r1_b0_v0.jpg} &
        \includegraphics[width=0.19\textwidth]{figs/subspace_viz/eye_photom/r2_b0_v0.jpg} &
        \includegraphics[width=0.19\textwidth]{figs/subspace_viz/eye_photom/r3_b0_v0.jpg} &
        \includegraphics[width=0.19\textwidth]{figs/subspace_viz/eye_photom/r4_b0_v0.jpg} \\
        & Open/close & Dark/bright & Brow ridge & Eye height and brow & Eye width and brow\\
         \rotatebox{90}{Eye shape}& 
        \includegraphics[width=0.19\textwidth]{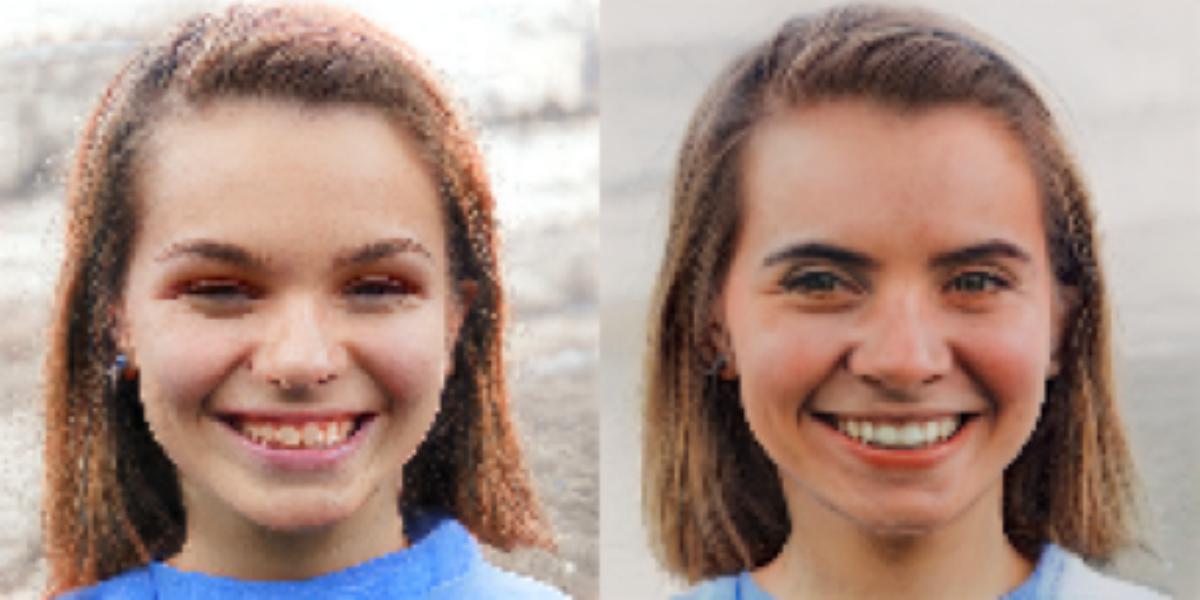} &
        \includegraphics[width=0.19\textwidth]{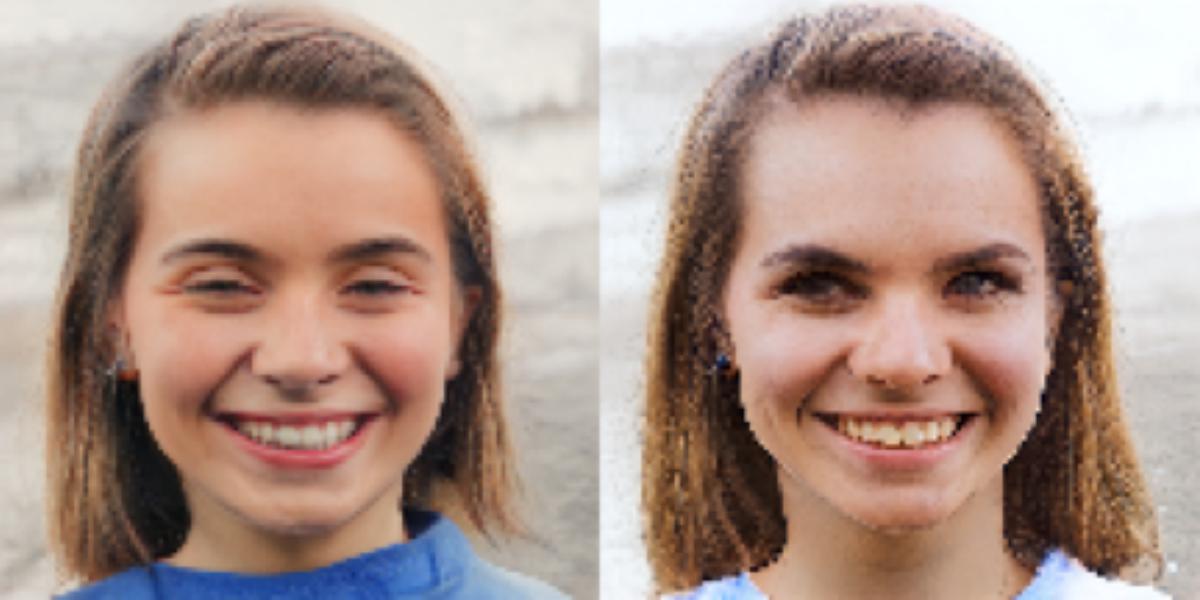} &
        \includegraphics[width=0.19\textwidth]{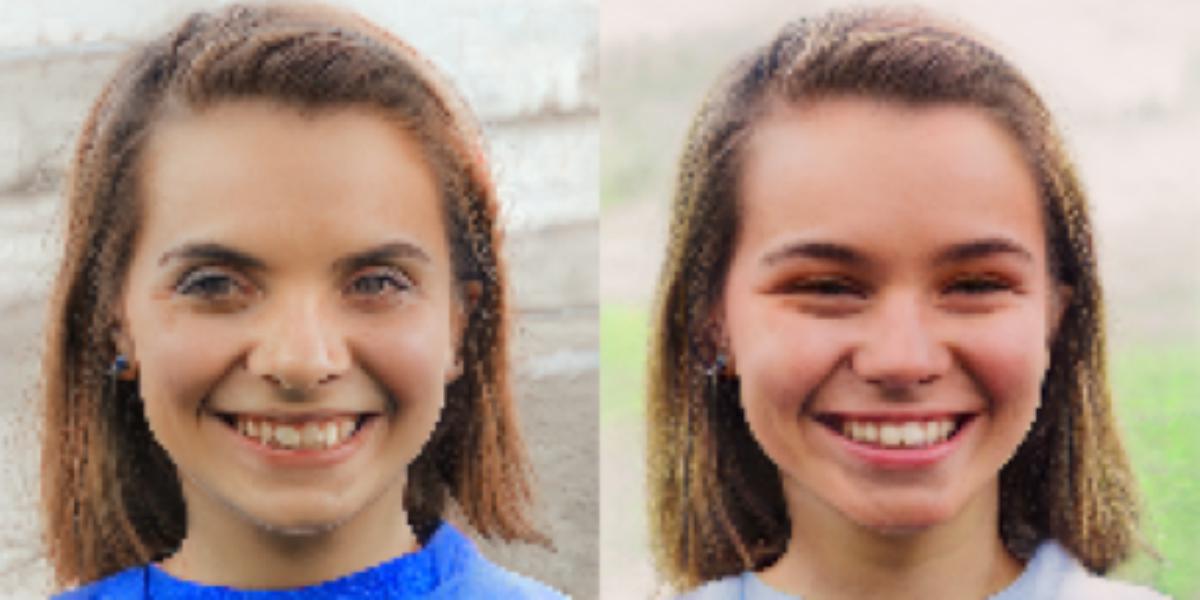} &
        \includegraphics[width=0.19\textwidth]{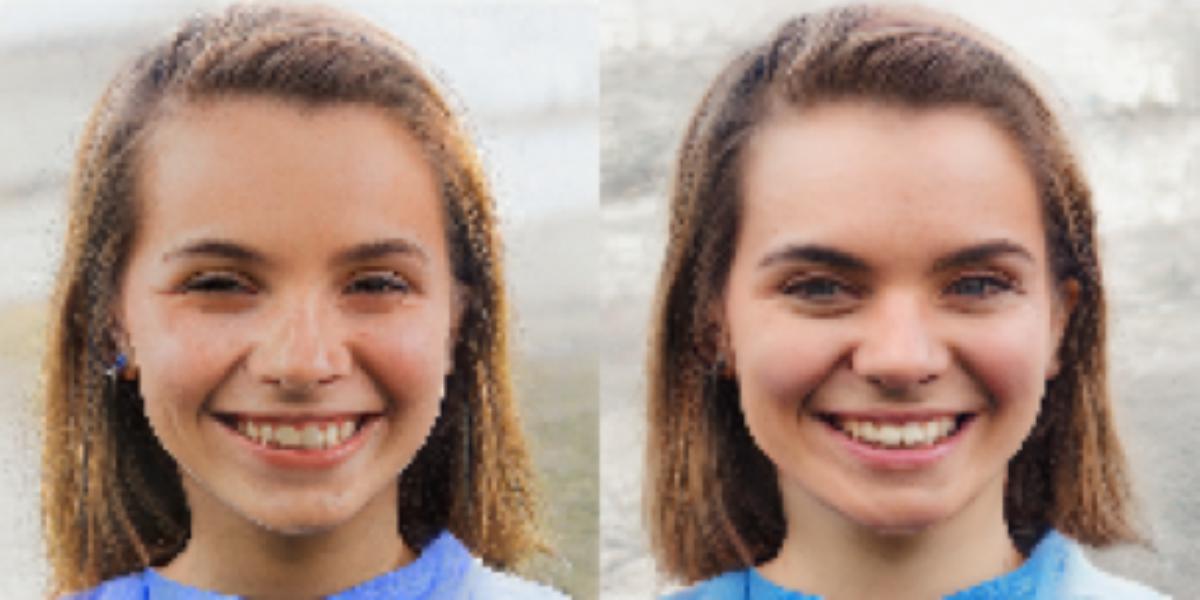} &
        \includegraphics[width=0.19\textwidth]{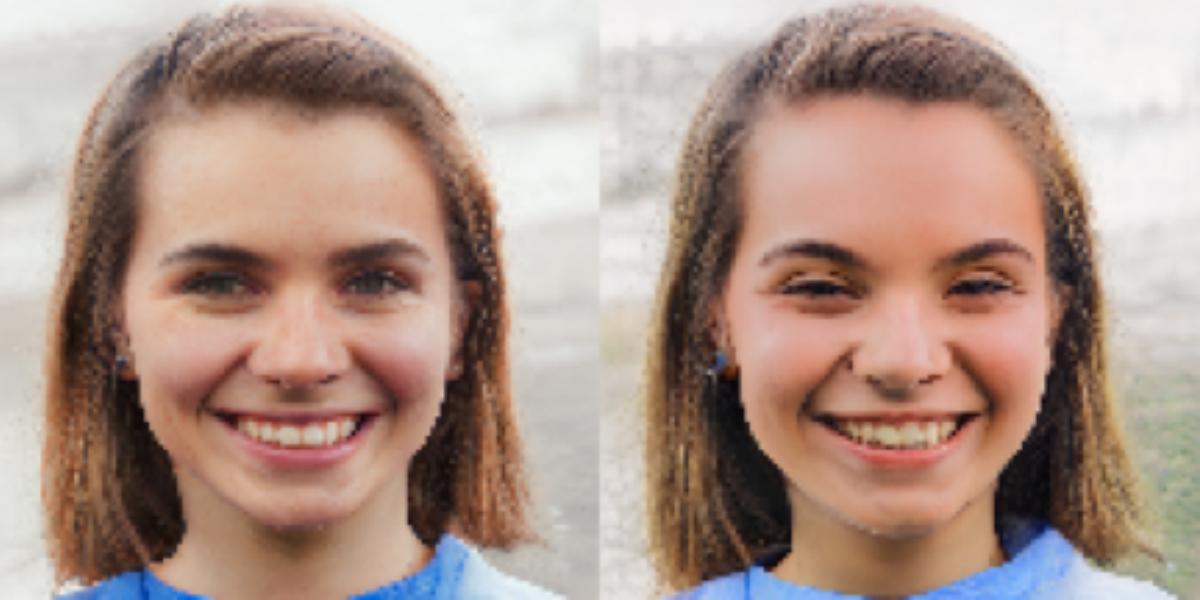} \\
        & Brow ridge and eye size & Open/close & Eye width & Narrow eyes & Narrow eyes\\
         \rotatebox{90}{Nose photometry}& 
        \includegraphics[width=0.19\textwidth]{figs/subspace_viz/nose_photom/r0_b0_v0.jpg} &
        \includegraphics[width=0.19\textwidth]{figs/subspace_viz/nose_photom/r1_b0_v0.jpg} &
        \includegraphics[width=0.19\textwidth]{figs/subspace_viz/nose_photom/r4_b0_v0.jpg} &
        \includegraphics[width=0.19\textwidth]{figs/subspace_viz/nose_photom/r7_b0_v0.jpg} &
        \includegraphics[width=0.19\textwidth]{figs/subspace_viz/nose_photom/r8_b0_v0.jpg} \\
        & Vertical length & Red/white & Nose size and nostril & Nose width & Orange/blue \\
         \rotatebox{90}{High frequency}& 
        \includegraphics[width=0.19\textwidth]{figs/subspace_viz/face_high_freq/r1_b0_v0.jpg} &
        \includegraphics[width=0.19\textwidth]{figs/subspace_viz/face_high_freq/r3_b0_v0.jpg} &
        \includegraphics[width=0.19\textwidth]{figs/subspace_viz/face_high_freq/r4_b0_v0.jpg} &
        \includegraphics[width=0.19\textwidth]{figs/subspace_viz/face_high_freq/r5_b0_v0.jpg} &
        \includegraphics[width=0.19\textwidth]{figs/subspace_viz/face_high_freq/r8_b0_v0.jpg} \\
         \rotatebox{90}{Hair photometry}& 
        \includegraphics[width=0.19\textwidth]{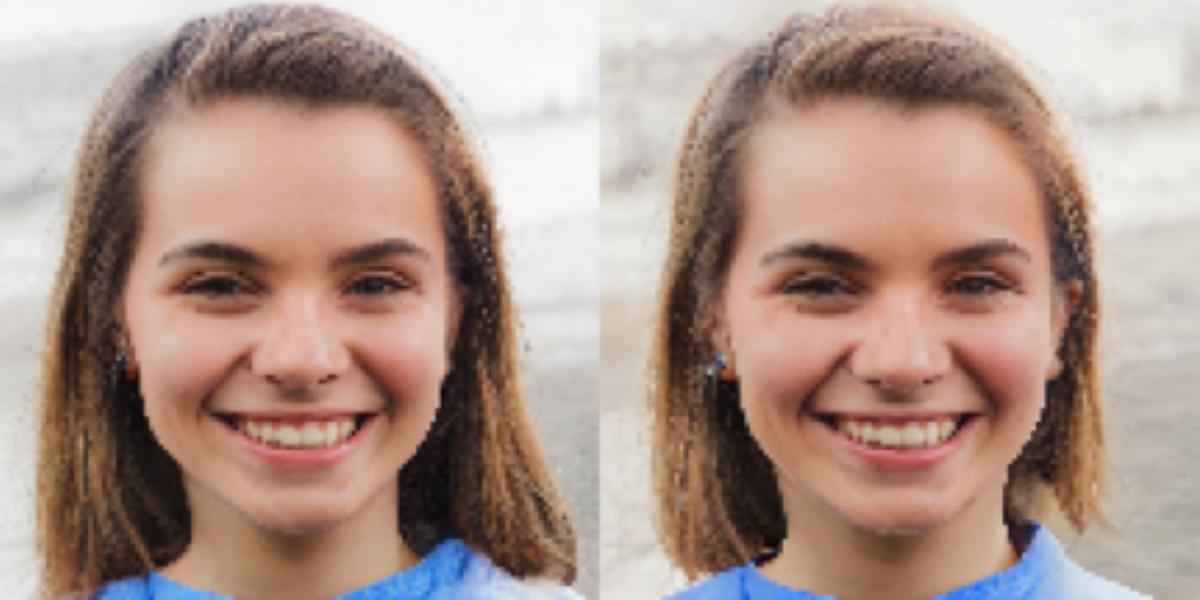} &
        \includegraphics[width=0.19\textwidth]{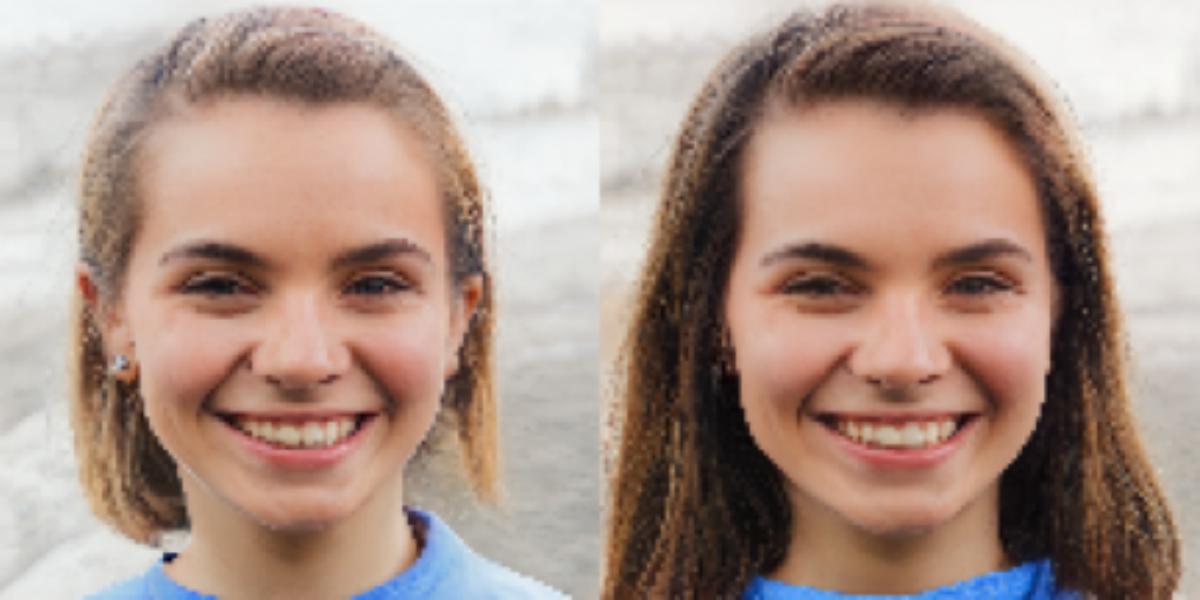} &
        \includegraphics[width=0.19\textwidth]{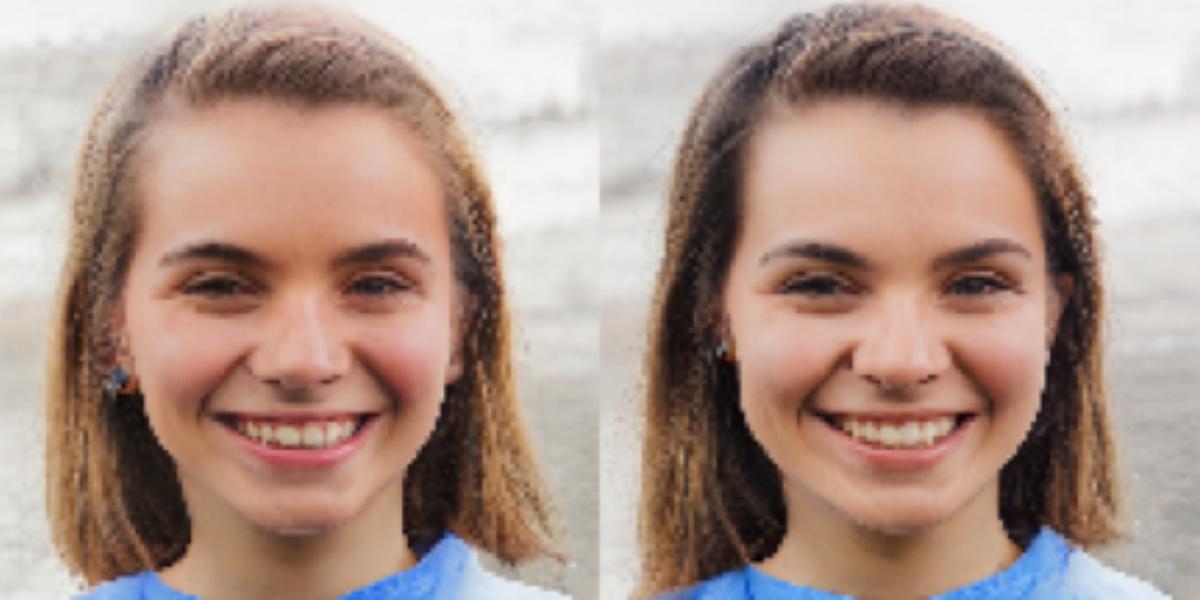} &
        \includegraphics[width=0.19\textwidth]{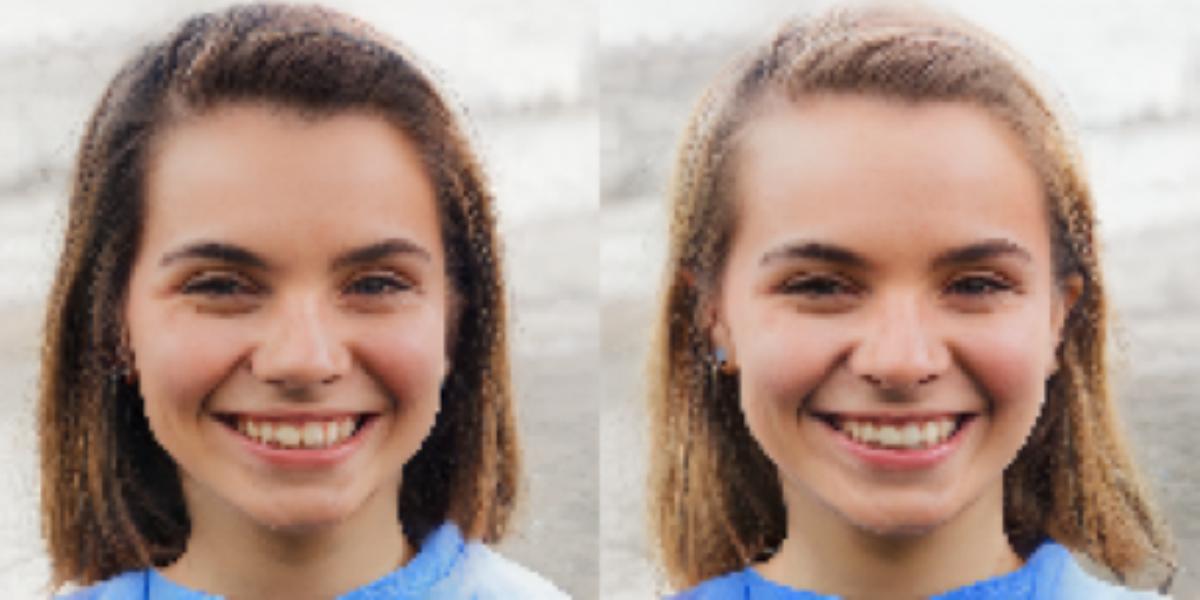} &
        \includegraphics[width=0.19\textwidth]{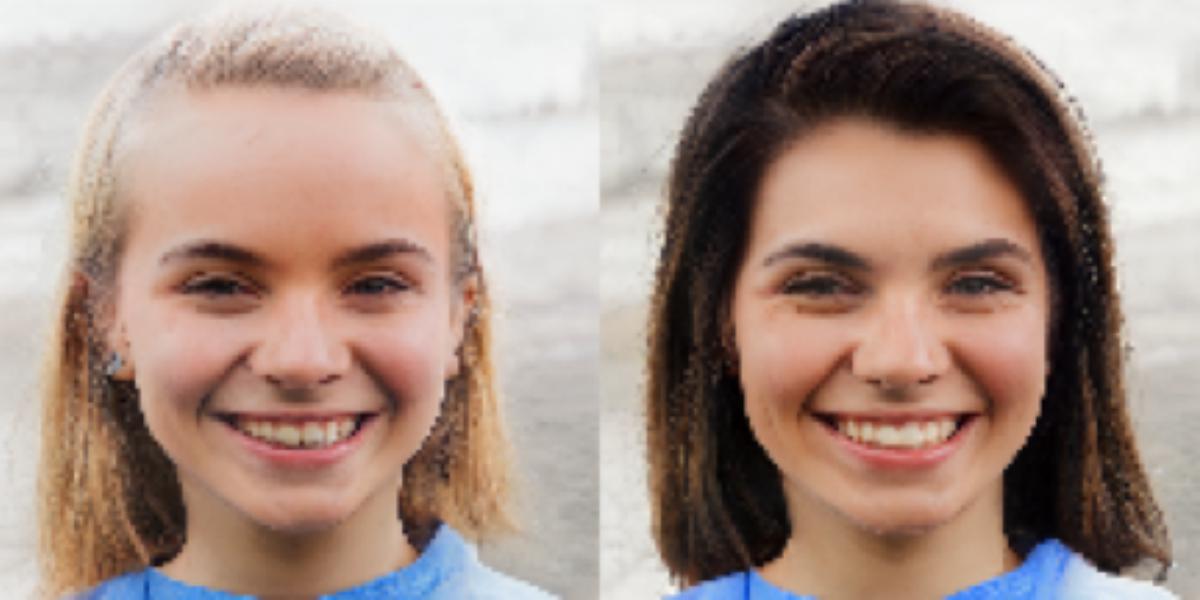} \\
         \rotatebox{90}{Background}& 
        \includegraphics[width=0.19\textwidth]{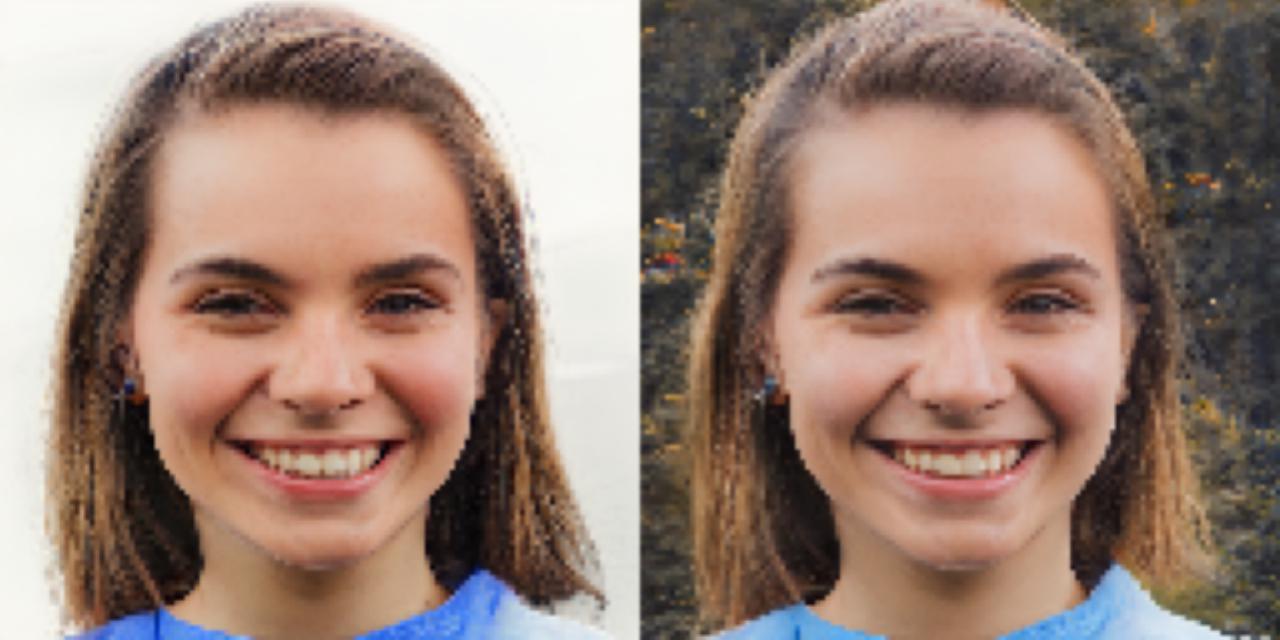} &
        \includegraphics[width=0.19\textwidth]{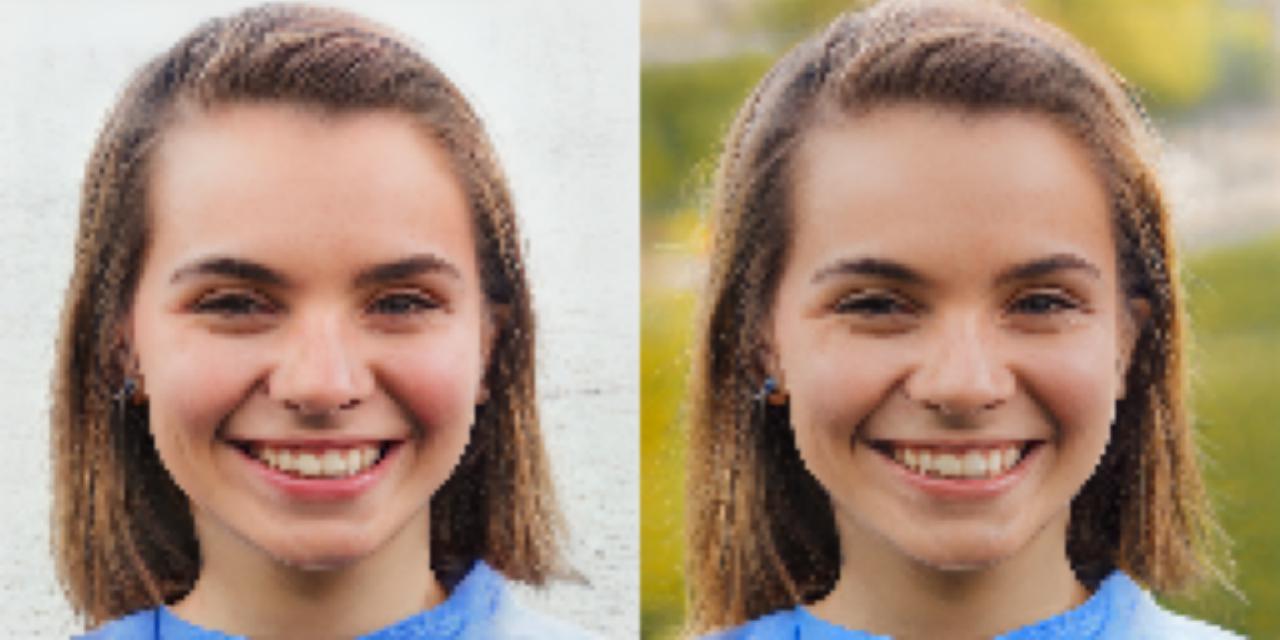} &
        \includegraphics[width=0.19\textwidth]{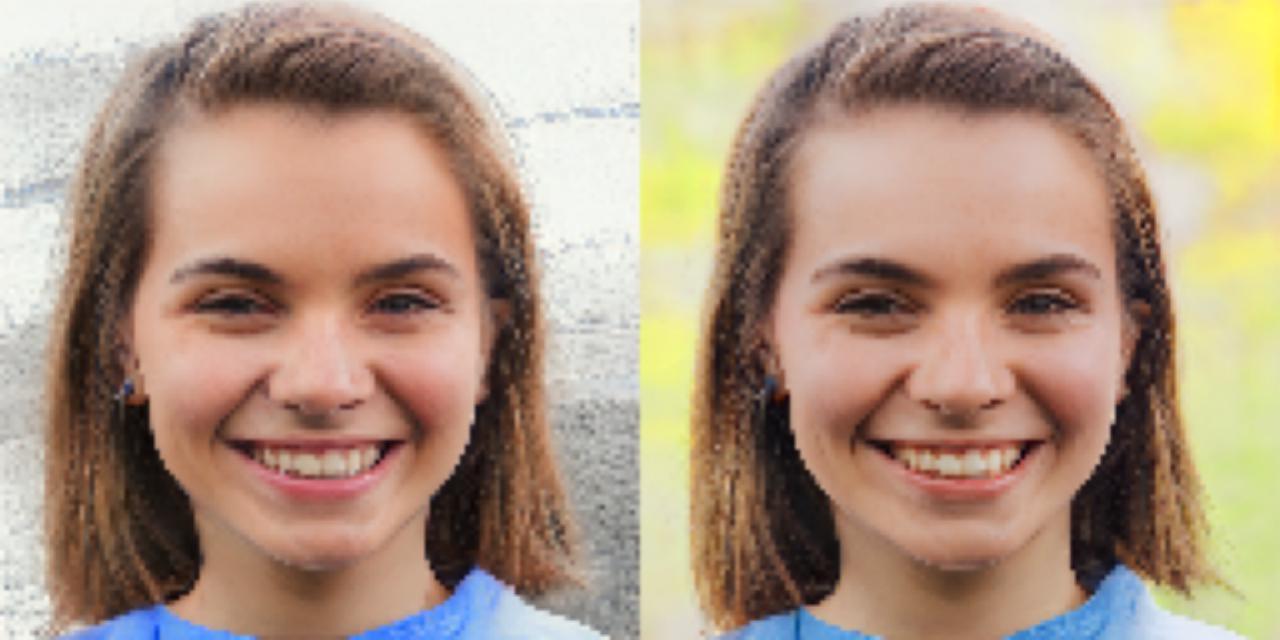} &
        \includegraphics[width=0.19\textwidth]{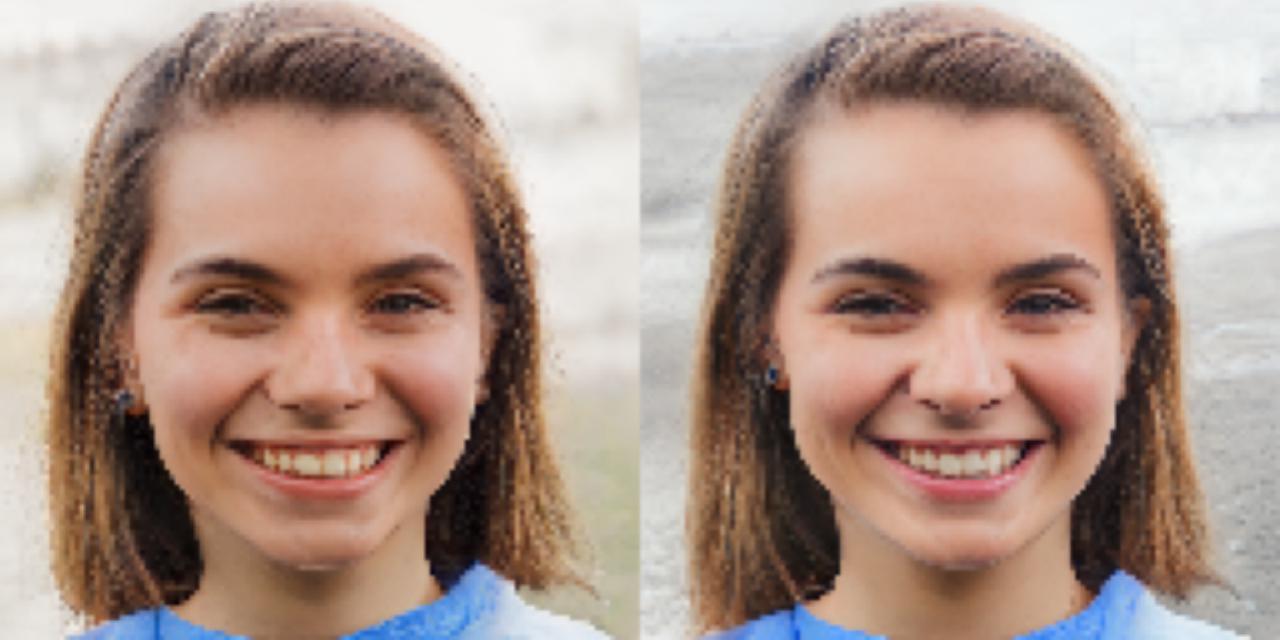} &
        \includegraphics[width=0.19\textwidth]{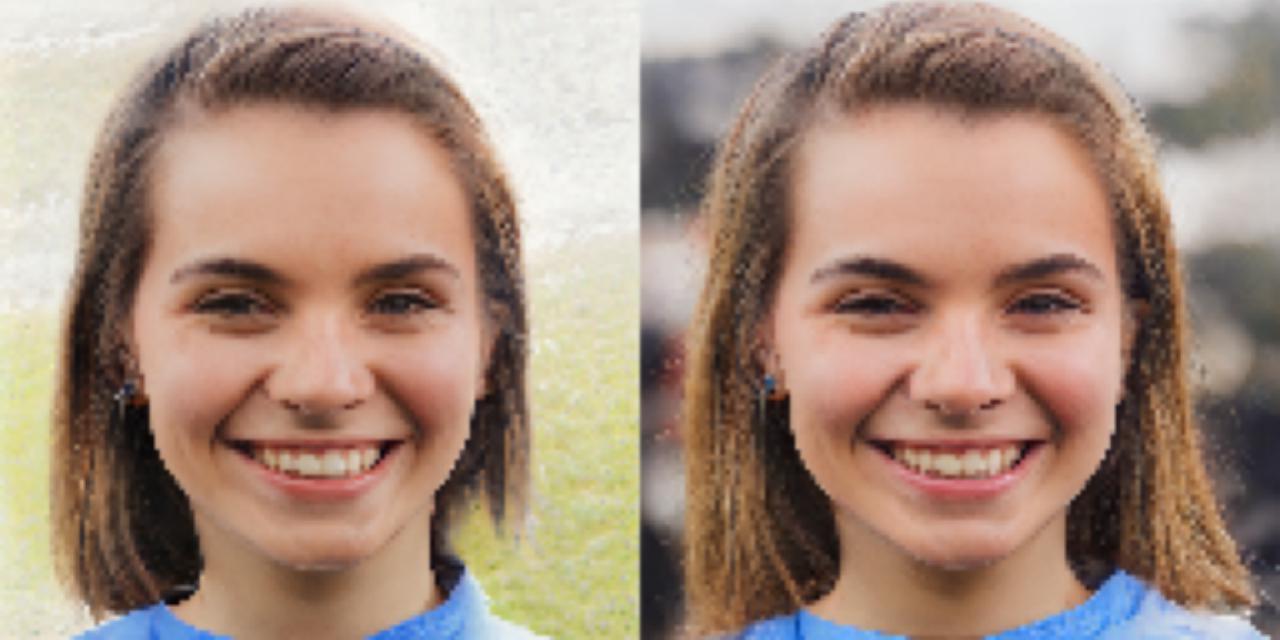} \\
    \end{tabular}
    \caption{Visualization of the discovered interpretable latent subspaces in $\Ss$. We select and visualize the manipulation effects of 5 representative principle components in each subspace. This is the long version of Figure~\ref{fig:exp-subspace}. See Section~\ref{sec:exp-setup} for details.}
    \label{fig:exp-subspace-full}
\end{figure*}

To illustrate the flexibility of our proposed subspace discovery approach, we construct 10+ criterions and solve their corresponding principle component vectors. The formulations of the subspaces are listed in Table~\ref{tab:exp-formulations}.
We visualize 5 selected component vectors in each subspaces by manipulate positvely and negatively and visualize the modified images in Figure~\ref{fig:exp-subspace-full}.

We notice that the face boundary control modifies the face color significantly, while the high frequency control change face boundary without modifying face color. We suspect that the problem with face boundary subspace might be related to the landmark detector.

\begin{table*}
    \centering
    \footnotesize
    \renewcommand{\arraystretch}{2}
    \begin{tabular}{c|c}
    \hline
        Subspace & Formulation \\\hline\hline
        Skin color & $\Span(V_\text{mac}^\text{skin}) \cap \Span(W_\text{res}^{\overline{\text{skin}}})$\\\hline
        Face boundary & $\Span(V_\text{fl}^\text{face}) \cap \Span(W_\text{fl}^{\overline{\text{face}}}) \cap \Span(W_\text{ap}^{{\text{face}}})$\\\hline
        Face boundary (id) & $\Span(V_\text{fl}^\text{face}) \cap \Span(W_\text{fl}^{\overline{\text{face}}}) \cap \Span(W_\text{ap}^{{\text{face}}}) \cap \Span(W_\text{id})$\\\hline
        Mouth shape & $\Span(V_\text{fl}^\text{mouth}) \cap \Span(W_\text{fl}^{\overline{\text{mouth}}}) \cap \Span(W_\text{ap}^{{\text{face}}})$\\\hline
        Mouth shape (id) & $\Span(V_\text{fl}^\text{mouth}) \cap \Span(W_\text{fl}^{\overline{\text{mouth}}}) \cap \Span(W_\text{ap}^{{\text{face}}}) \cap \Span(W_\text{id})$\\\hline
        Mouth photometry & $\Span(V_\text{mp}^\text{mouth}) \cap \Span(W_\text{mp}^{\overline{\text{mouth}}})$\\\hline
        Mouth photometry (id) & $\Span(V_\text{mp}^\text{mouth}) \cap \Span(W_\text{mp}^{\overline{\text{mouth}}}) \cap \Span(W_\text{id})$\\\hline
        Lip color & $\Span(V_\text{mac}^\text{lip}) \cap \Span(W_\text{res}^{\overline{\text{lip}}})$\\\hline
        Lip photometry & $\Span(V_\text{mp}^\text{lip}) \cap \Span(W_\text{mp}^{\overline{\text{lip}}})$\\\hline
        Eye photometry & $\Span(V_\text{mp}^\text{eye}) \cap \Span(W_\text{mp}^{\overline{\text{eye}}})$\\\hline
        Eye shape & $\Span(V_\text{fl}^\text{eye}) \cap \Span(W_\text{fl}^{\overline{\text{eye}}}) \cap \Span(W_\text{ap}^{{\text{eye}}})$\\\hline
        Nose photometry & $\Span(V_\text{mp}^\text{nose}) \cap \Span(W_\text{mp}^{\overline{\text{nose}}})$\\\hline
        High Freqeuncy & $\Span(V_\text{high}) \cap \Span(W_\text{low})$\\\hline
        Hair photometry & $\Span(V_\text{mp}^\text{hair}) \cap \Span(W_\text{mp}^{\overline{\text{hair}}})$\\\hline
        Background photometry & $\Span(V_\text{mp}^\text{background}) \cap \Span(W_\text{mp}^{\overline{\text{background}}})$\\\hline
    \end{tabular}
    \caption{Formulation for our discovered latent subspaces in Figure~\ref{fig:exp-subspace-full}. See also Section~\ref{sec:exp-subspace}. Obviously readers can freely formulate their own criterion to discover more latent subspaces.}
    \label{tab:exp-formulations}
\end{table*}

\section{Comparison with Previous Approaches}
\label{sec:comparison-sup}

We consider two subspaces discovered by our approach (referred as ICE for interpretable control exploration). The first is the subspace to manipulate mouth photometry, and the second is its intersection with $\Span(W_\text{id})$. See \Cref{tab:exp-formulations} for the detailed formulations.
We compare the top-20 principle components in our discovered subspaces with the top-20 channels detected by StyleSpace~\cite{wu2021stylespace} to control mouth manipulation. We consider three metrics for evaluation. The first is the photometry difference in the mouth region. The second is the photometry difference in the rest image region. A good manipulation should have large photometry difference in the mouth region but small in the rest region. The third metric is the identity cosine similarity between the original image and the manipulated image.

\begin{figure}
\centering
    \footnotesize
        \includegraphics[width=0.4\textwidth]{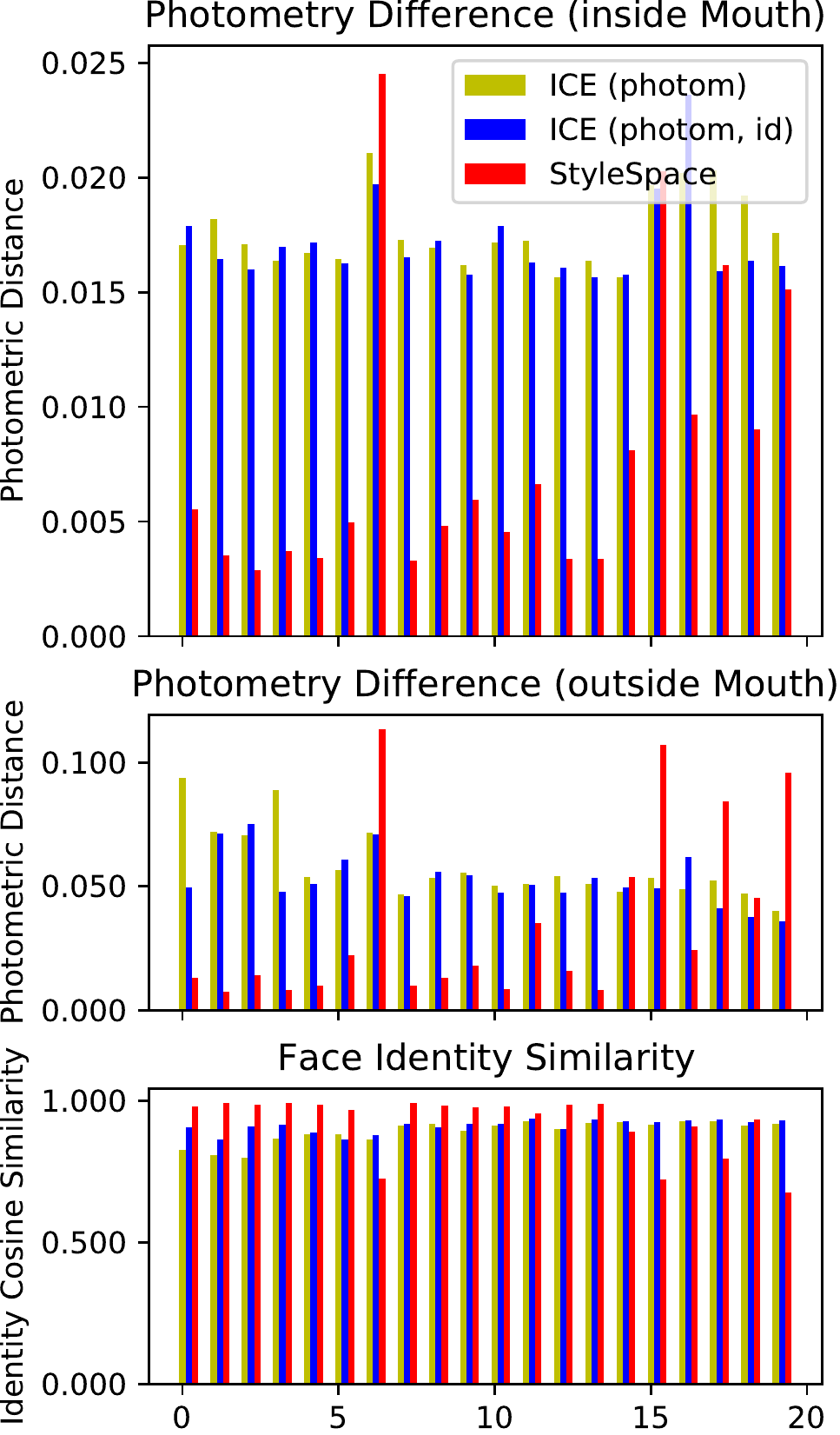}
    \caption{Quantitative comparison with StyleSpace~\cite{wu2021stylespace}}
    \label{fig:quanti-compare-basis}
\end{figure}

We randomly generate 2000 images to experiment the manipulation effect. We first modify the subspace directions at a step of 10 and plot the modification magnitudes in \Cref{fig:quanti-compare-basis}. We noticed that StyleSpace detect 4 channels (6, 15, 17, 19, referred as sig-4) with very significant modification magnitude, while the rest are not noticeable. On the other hand, our approach ICE produces 20 meaningful manipulation directions. To further compare the difference of modification, we multiply the manipulation magnitudes of those insignificant StyleSpace channels by 4 to make them noticeable and comparable with other directions. We compute the designed three metrics and list them in \Cref{tab:quanti-compare-basis}. It can be noticed that the sig-4 produces most significant modification, but might harm the background consistency and identity. ICE produces effective directions while preserve the identity.

\Cref{fig:quali-compare-basis} visualizes the representative directions of the above three approaches. We visualize the top-6 and the sig-4 directions of StyleSpace and the top-10 directions of ICE (photom) and ICE (photom, id). Note that unlike \Cref{fig:exp-subspace-full} where the visualzed components corresponds in pairs,  ``ICE~(photom)'' and ``ICE~(photom, id)'' directly visualize the top-10 directions which do not correspond in pairs. It can be noticed that ICE produces richer manipulation, and the manipulation preserves photometry outside the mouth region and identity as restricted by the criterions.

\begin{table}
    \centering
    \footnotesize
    \renewcommand{\arraystretch}{1.5}
    \setlength{\tabcolsep}{2pt}
    \begin{tabular}{lrrrr}
        \hline
        Average & Mag & Inside$\uparrow$ & Outside$\downarrow$ & Identity$\uparrow$\\\hline\hline
        StyleSpace (sig-4) & 10 & 0.100 & 0.019 & 0.730 \\\hline
        ICE (photom, top-4) & 10 & 0.081 & 0.017 & 0.826 \\\hline
        ICE (photom, id, top-4) & 10 & 0.061 & 0.017 & 0.899 \\\hline\hline

        StyleSpace (insig-16) & 40 & 0.051 & 0.018 & 0.832 \\\hline
        ICE (photom, top-20) & 10 & 0.058 & 0.018 & 0.893 \\\hline
        ICE (photom, id, top-20) & 10 & 0.052 & 0.017 & 0.912 \\\hline
    \end{tabular}
    \caption{Evaluation of our modification effects compared with StyleSpace. We manipulate insig-16 with magnitudes 4 times larger to make it noticeable.}
    \label{tab:quanti-compare-basis}
\end{table}

\begin{figure*}
    \scriptsize
    \setlength{\tabcolsep}{1pt}
    \centering
    \begin{tabular}{cccccc}
        \multirow{2}{*}[25pt]{ \rotatebox{90}{ICE (photom), top-10} } &
        \includegraphics[width=0.19\textwidth]{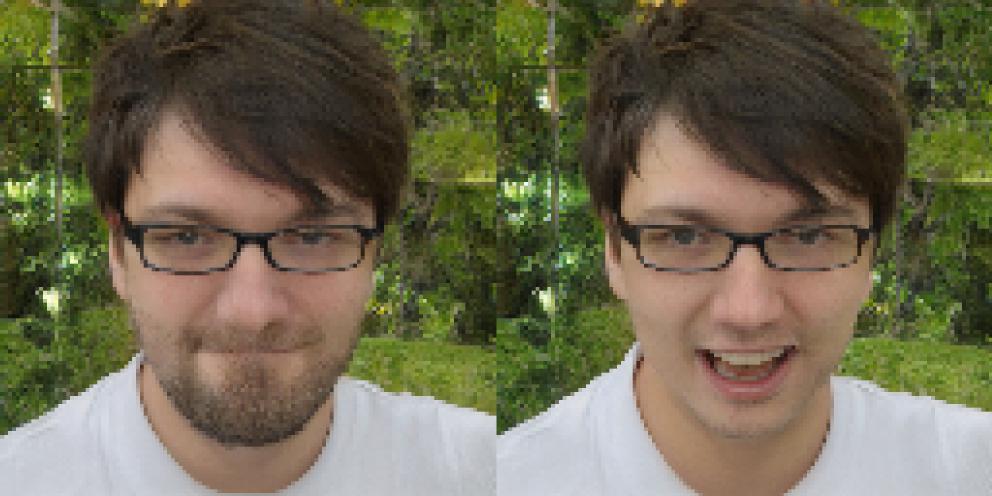} &
        \includegraphics[width=0.19\textwidth]{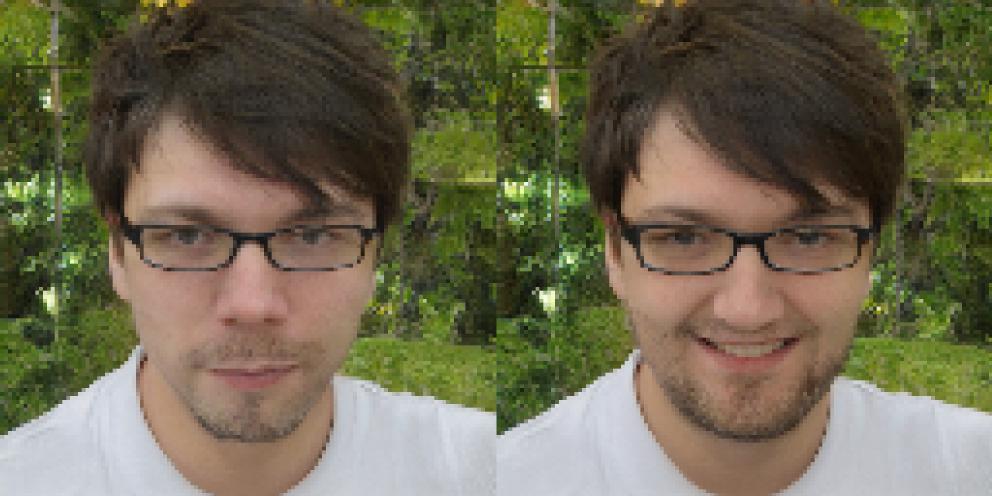} &
        \includegraphics[width=0.19\textwidth]{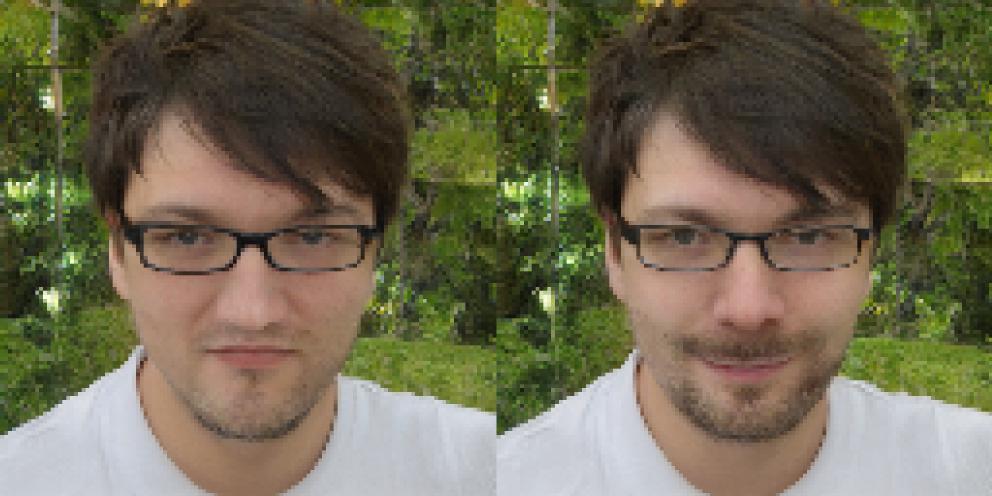} &
        \includegraphics[width=0.19\textwidth]{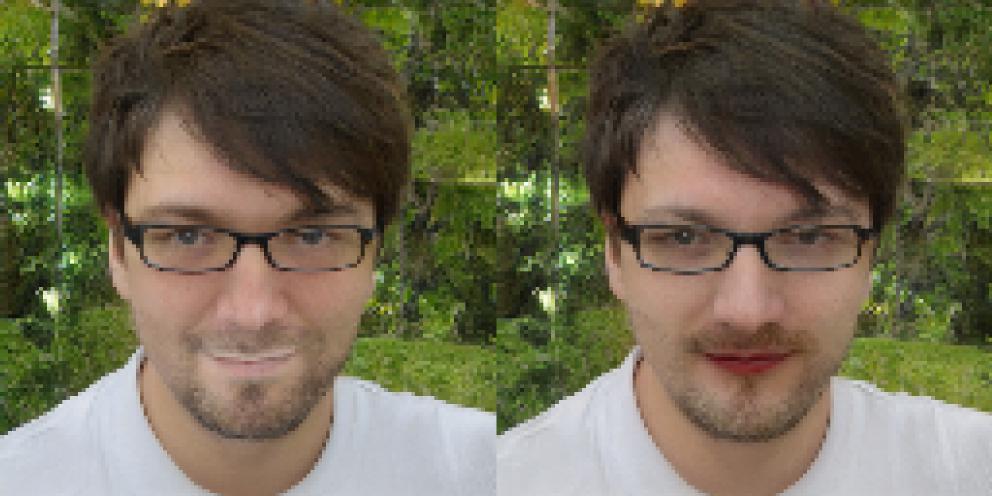} &
        \includegraphics[width=0.19\textwidth]{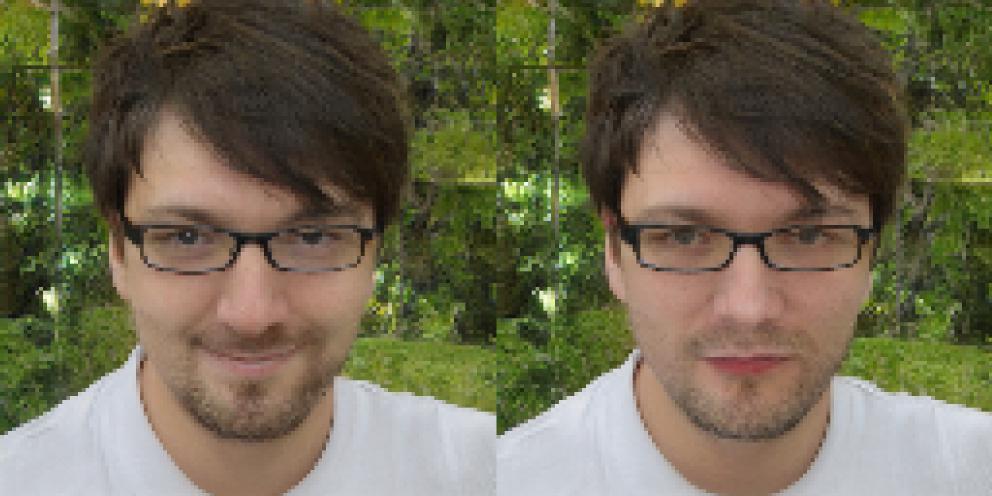} \\
        & \includegraphics[width=0.19\textwidth]{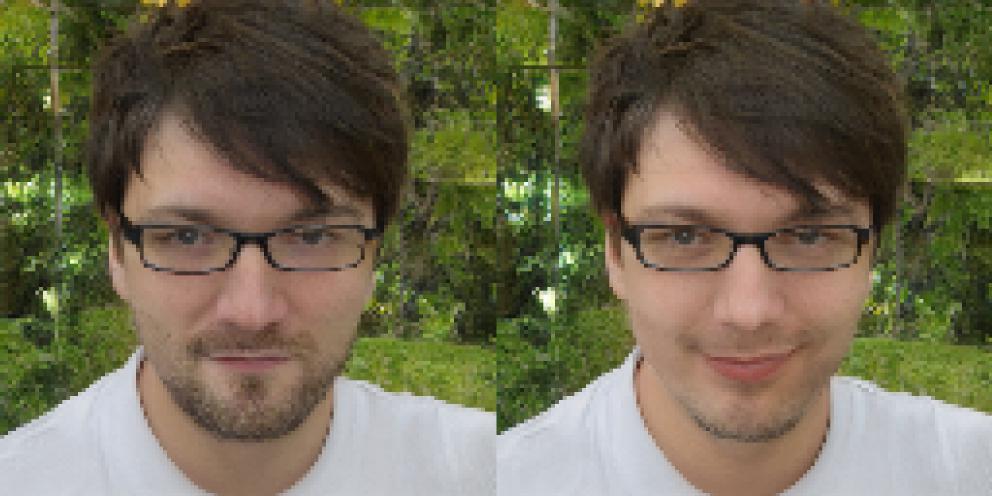} &
        \includegraphics[width=0.19\textwidth]{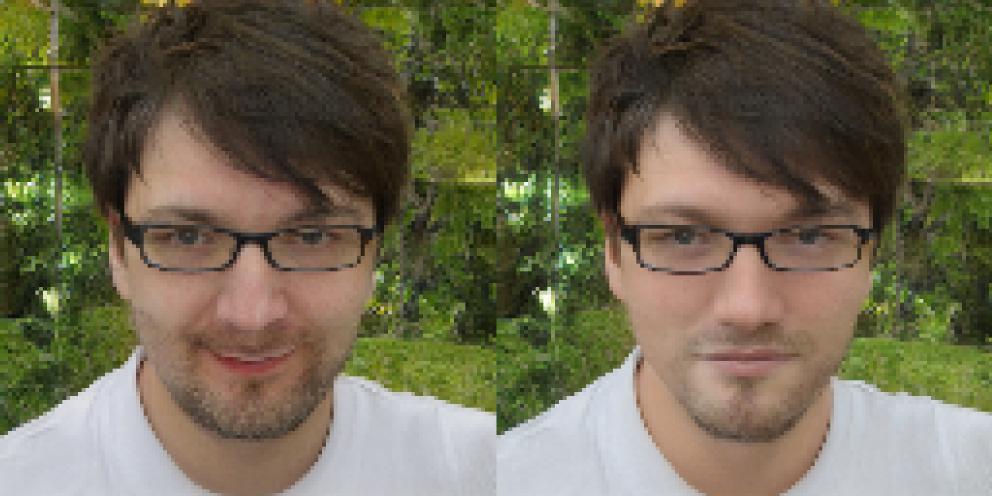} &
        \includegraphics[width=0.19\textwidth]{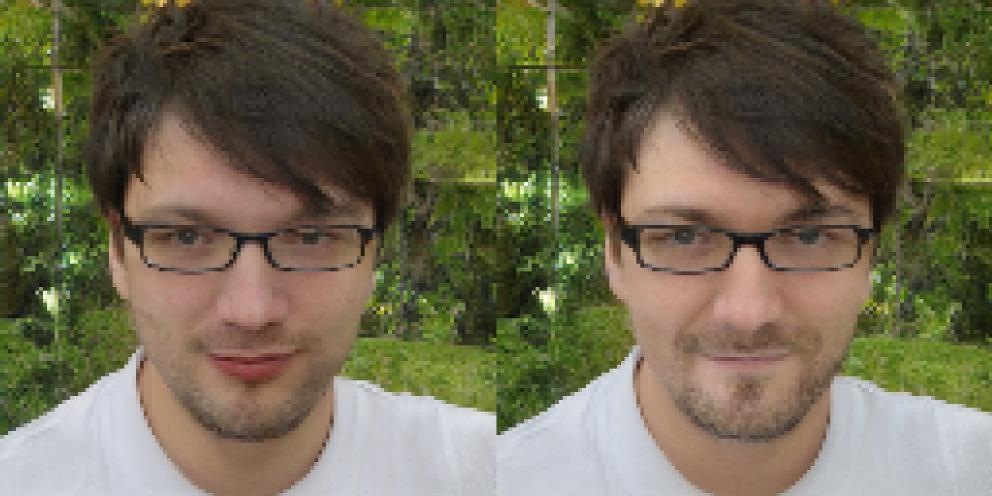} &
        \includegraphics[width=0.19\textwidth]{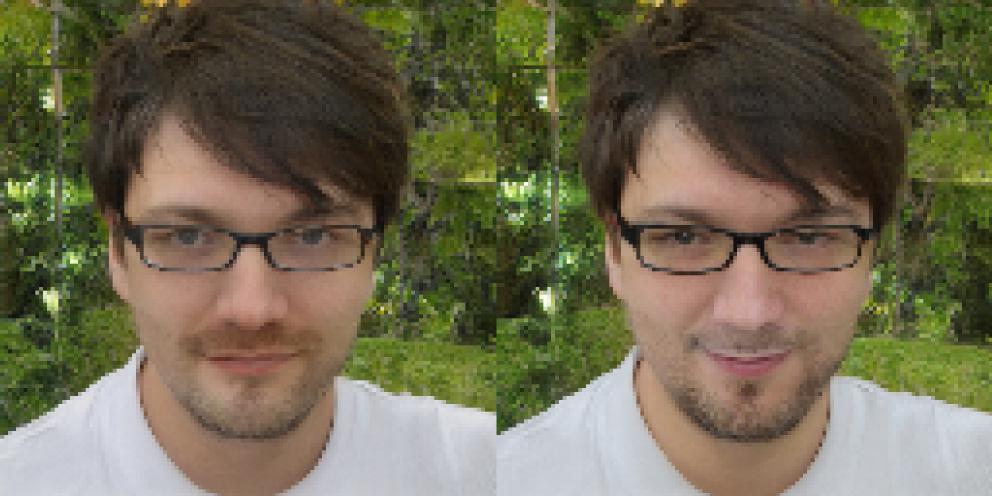} &
        \includegraphics[width=0.19\textwidth]{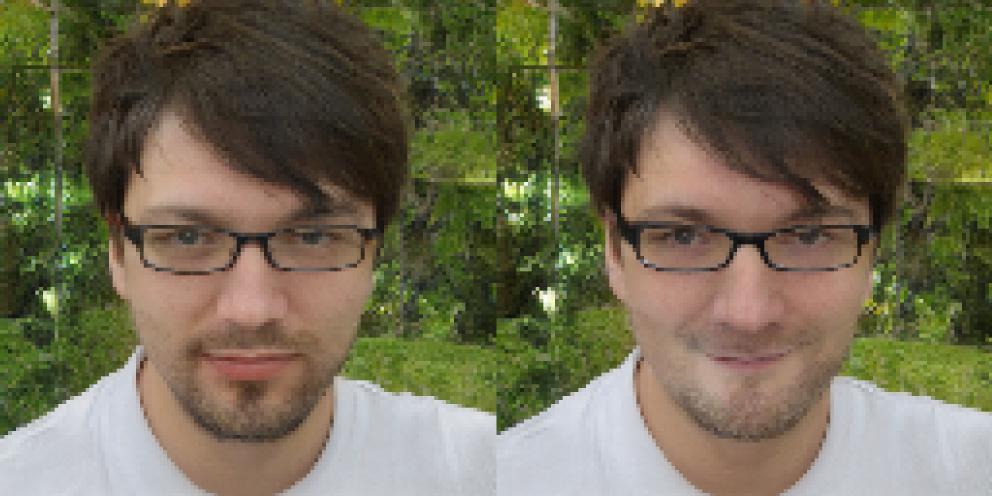} \\\hline
        \rule{0pt}{0.1\textwidth} \multirow{2}{*}[30pt]{ \rotatebox{90}{ICE (photom, id), top-10} } &
        \includegraphics[width=0.19\textwidth]{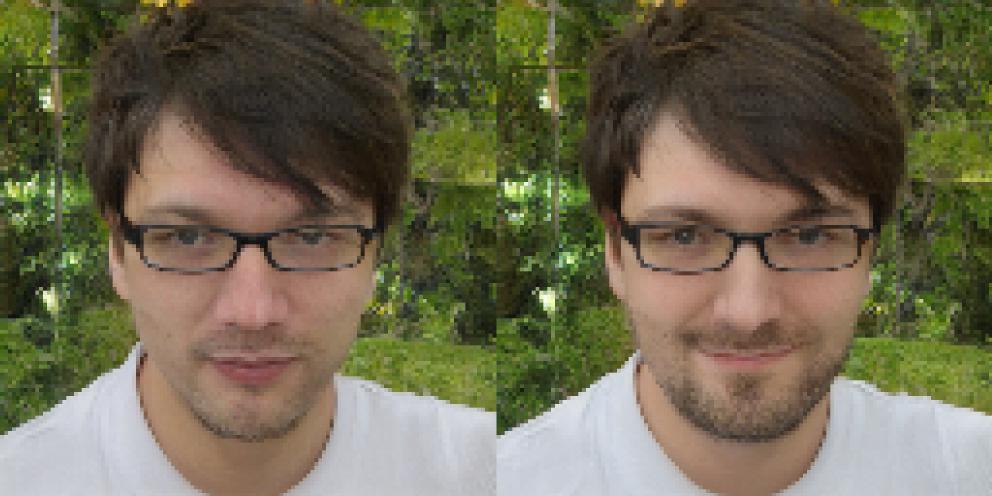} &
        \includegraphics[width=0.19\textwidth]{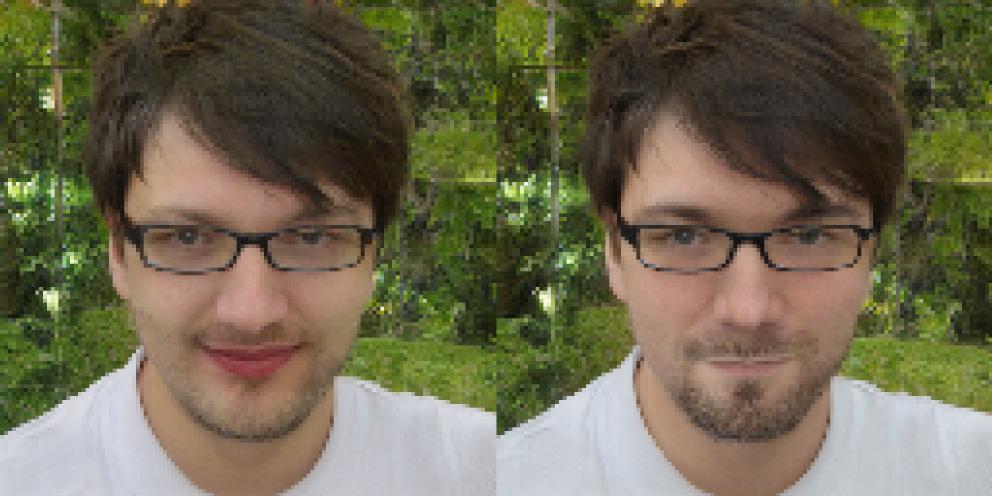} &
        \includegraphics[width=0.19\textwidth]{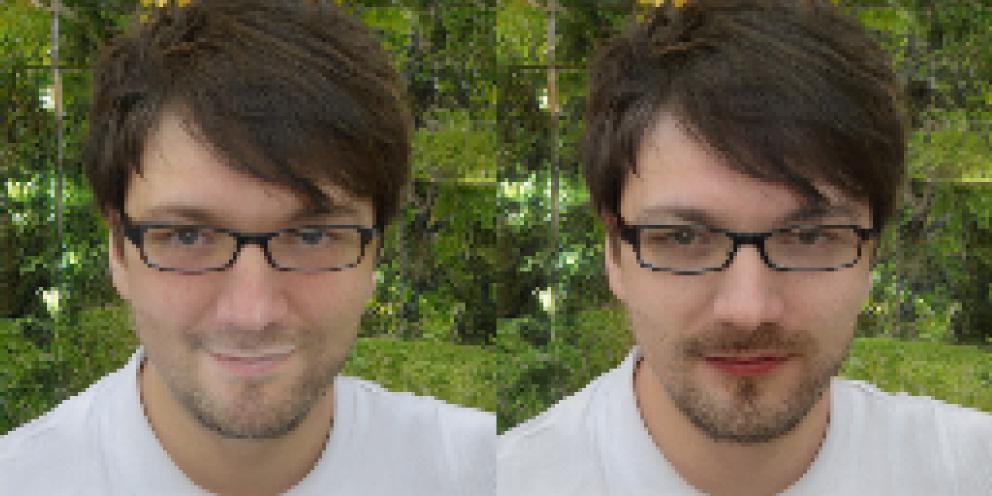} &
        \includegraphics[width=0.19\textwidth]{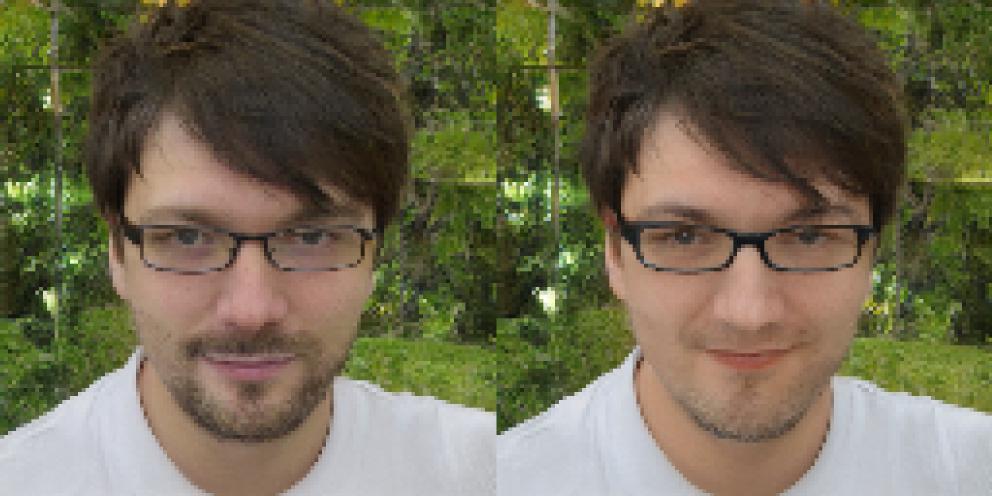} &
        \includegraphics[width=0.19\textwidth]{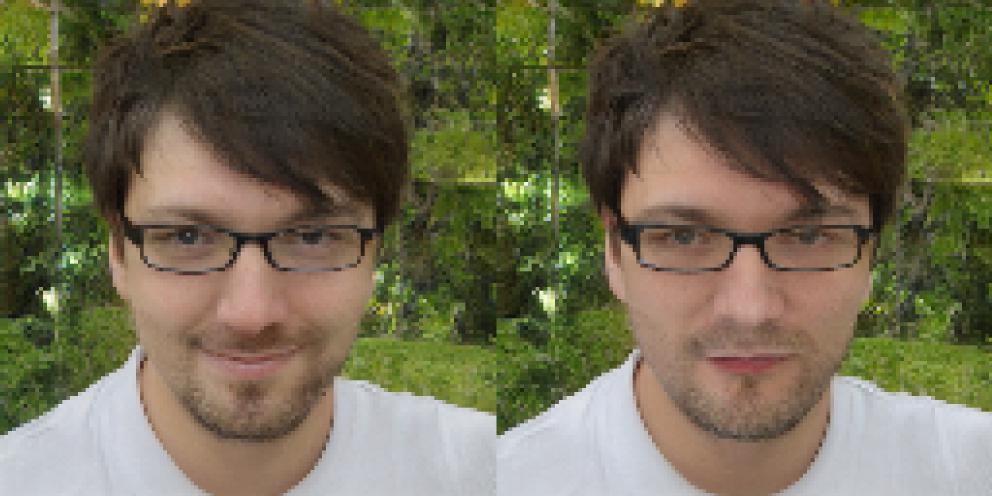} \\
        & \includegraphics[width=0.19\textwidth]{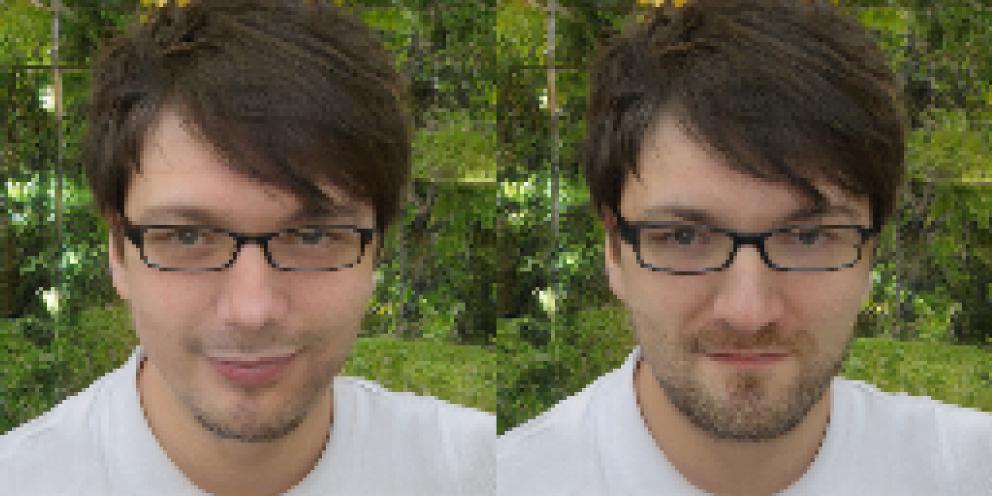} &
        \includegraphics[width=0.19\textwidth]{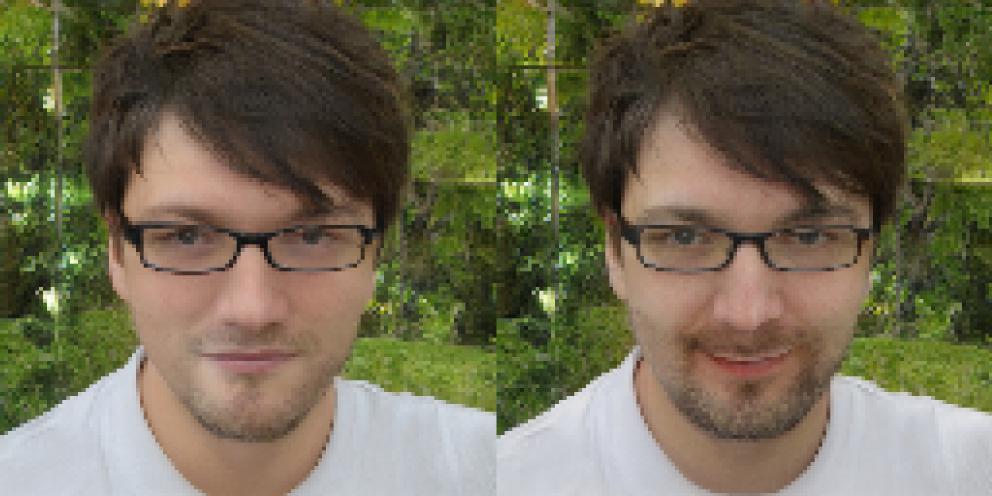} &
        \includegraphics[width=0.19\textwidth]{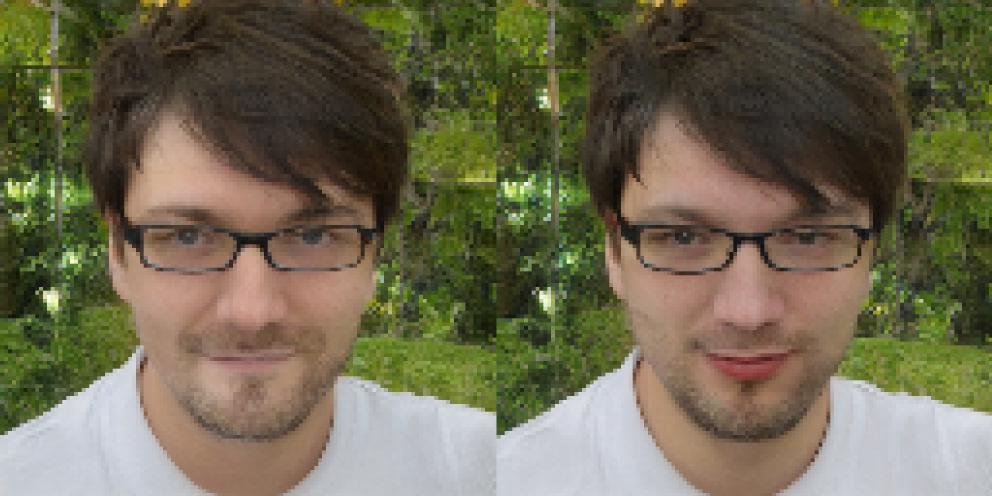} &
        \includegraphics[width=0.19\textwidth]{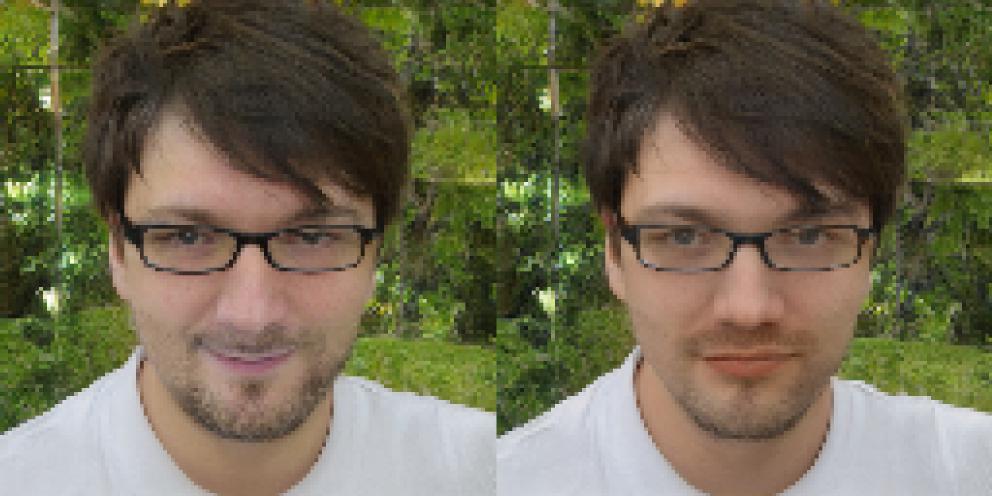} &
        \includegraphics[width=0.19\textwidth]{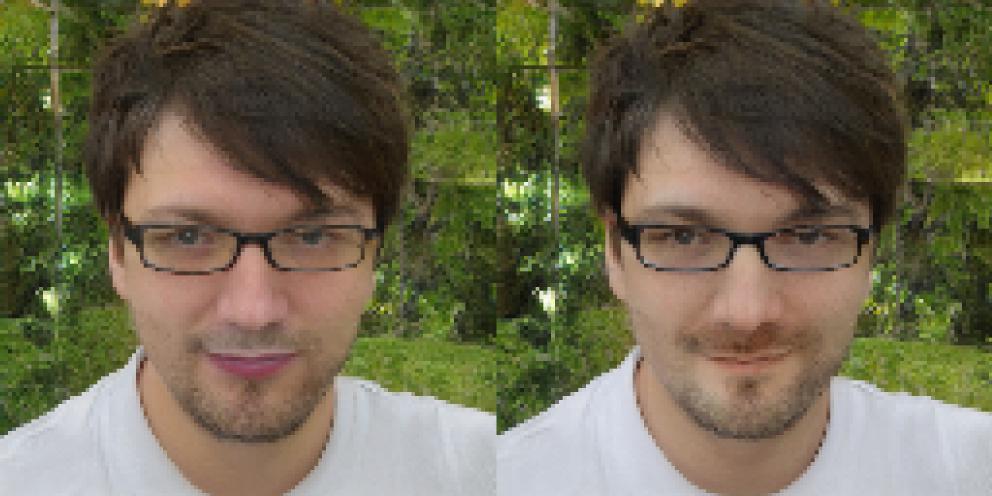} \\\hline
        \rule{0pt}{0.1\textwidth} \multirow{2}{*}[25pt]{ \rotatebox{90}{StyleSpace, top-6 + sig-4} } &
        \includegraphics[width=0.19\textwidth]{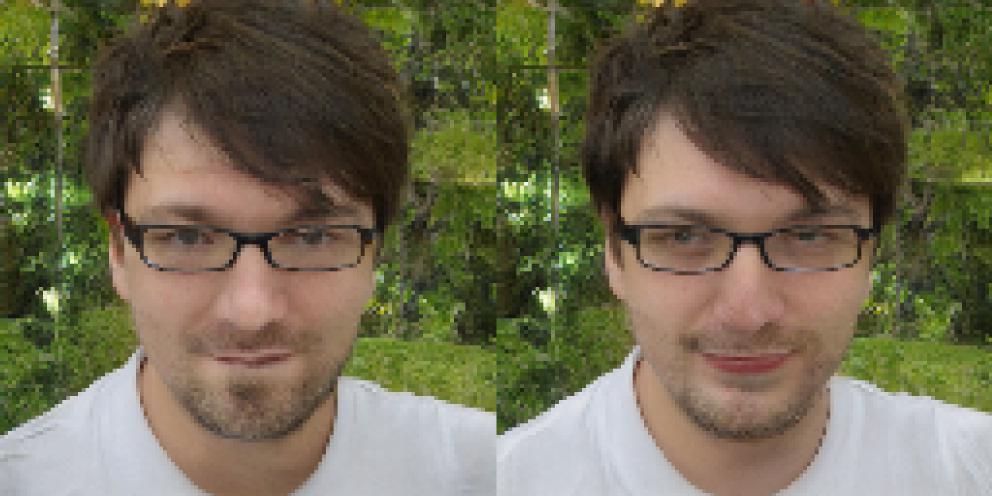} &
        \includegraphics[width=0.19\textwidth]{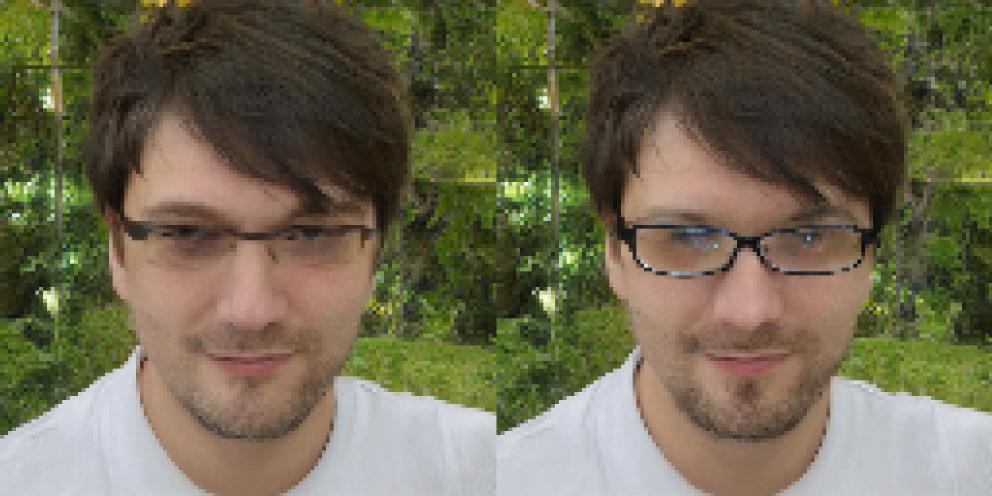} &
        \includegraphics[width=0.19\textwidth]{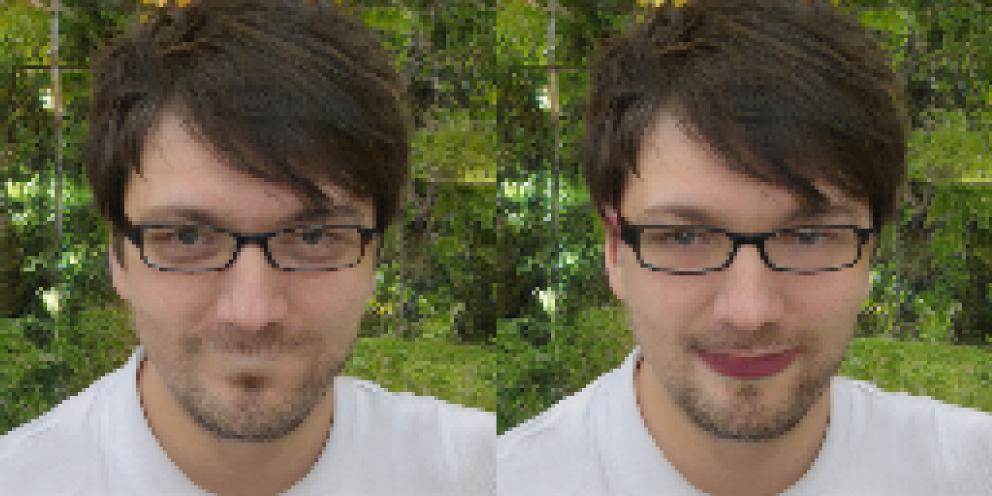} &
        \includegraphics[width=0.19\textwidth]{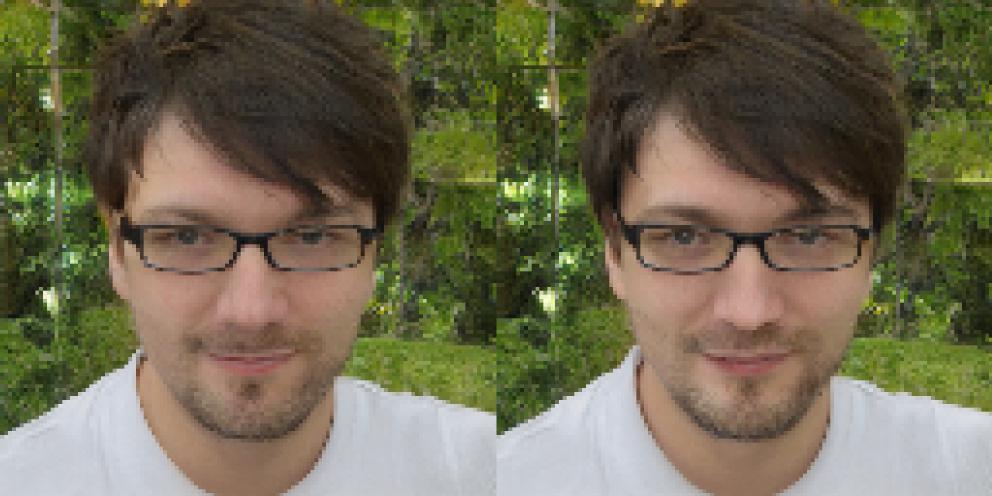} &
        \includegraphics[width=0.19\textwidth]{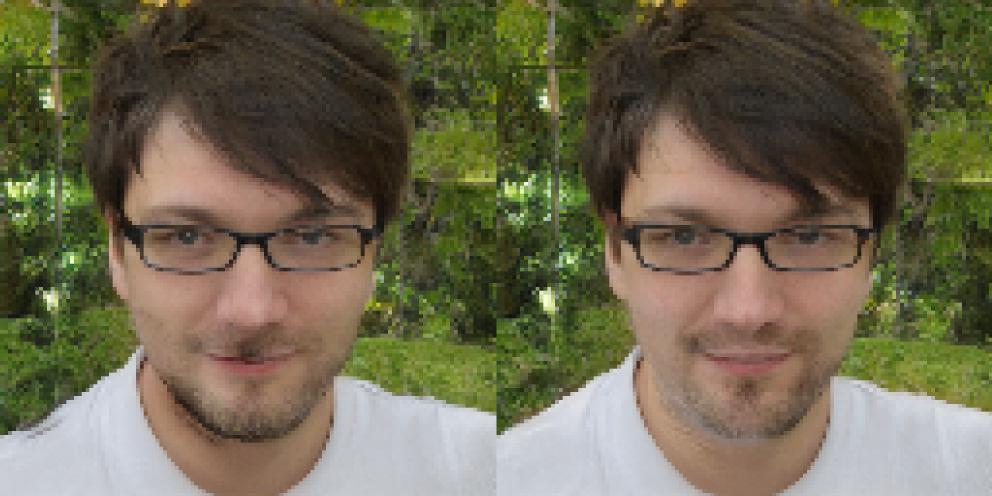} \\
        & \includegraphics[width=0.19\textwidth]{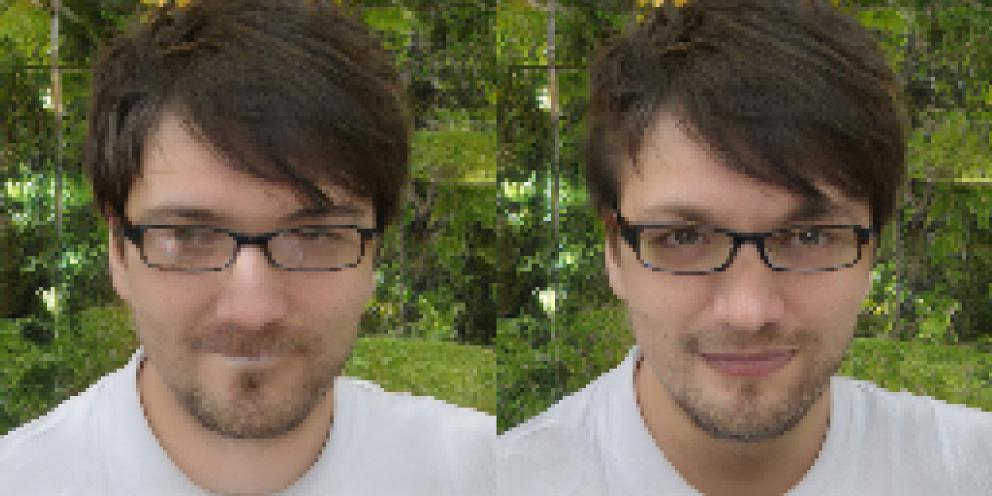} &
        \includegraphics[width=0.19\textwidth]{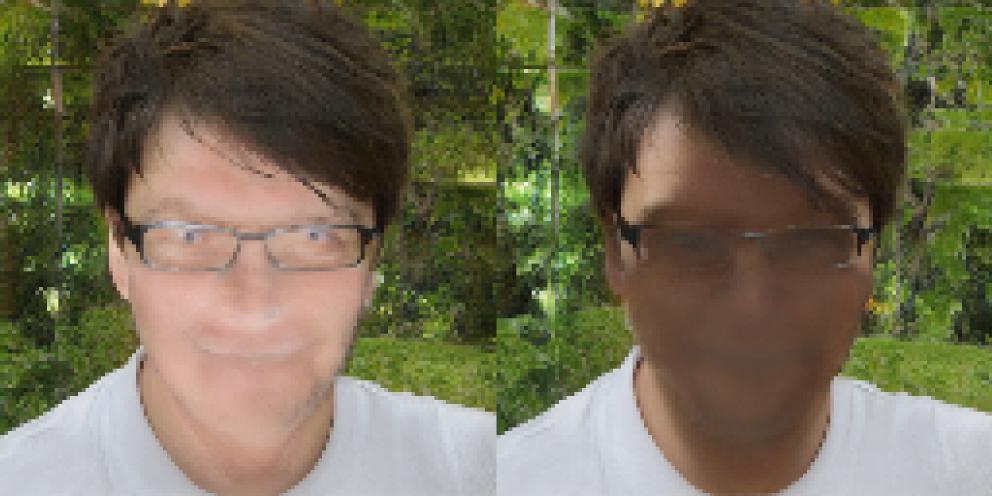} &
        \includegraphics[width=0.19\textwidth]{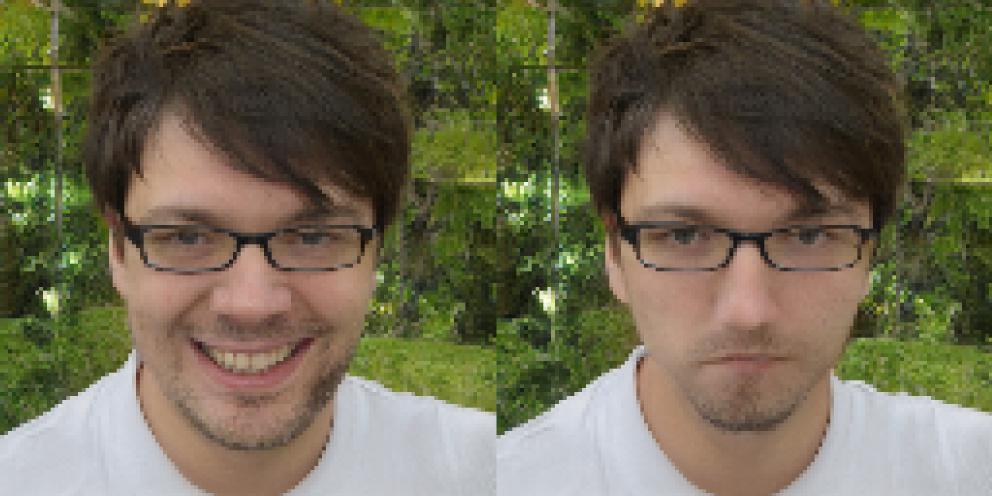} &
        \includegraphics[width=0.19\textwidth]{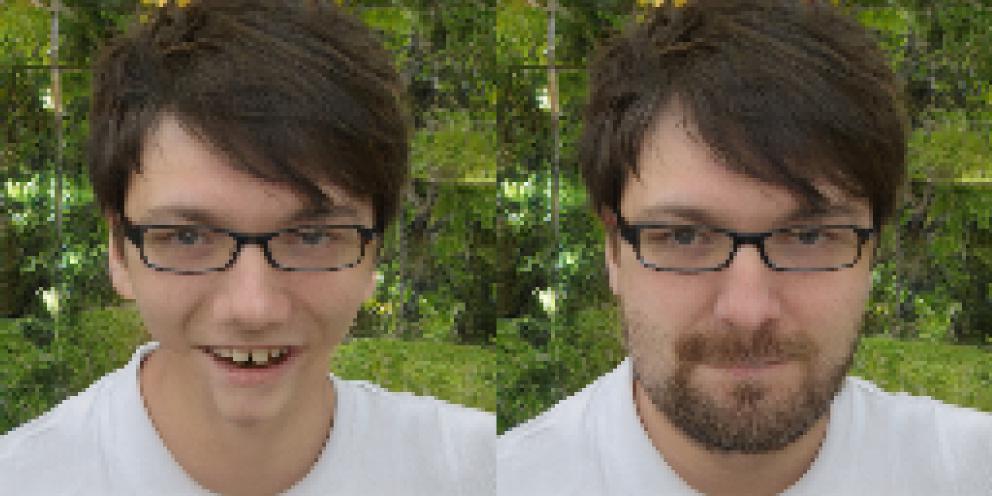} &
        \includegraphics[width=0.19\textwidth]{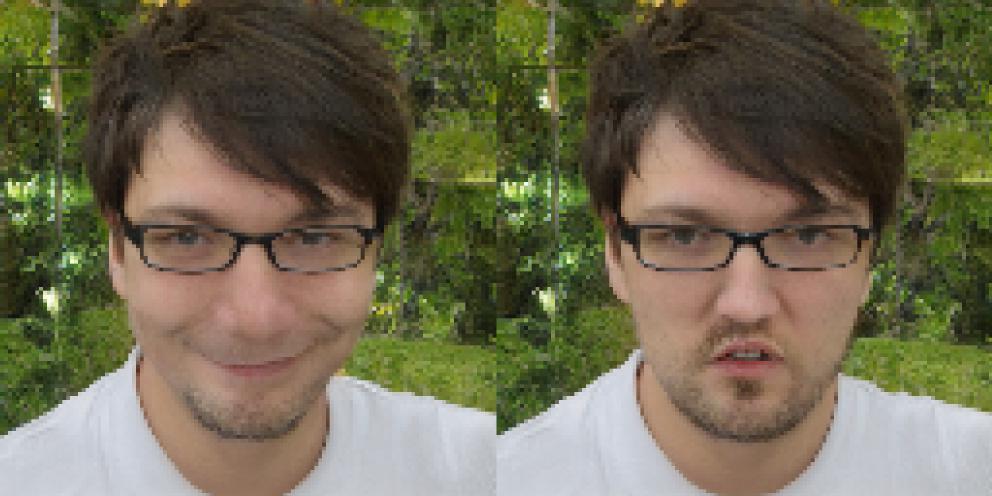}
    \end{tabular}
    \caption{Qualitative comparision with Stylespace~\cite{wu2021stylespace}. The visualized images are selected from the generated samples in \Cref{tab:quanti-compare-basis}.}
    \label{fig:quali-compare-basis}
\end{figure*}

\section{Transferability, Generality and Individuality}
\label{sec:transferability-sup}

\begin{figure*}
    \centering
    \scriptsize
    \setlength{\tabcolsep}{1pt}
    \begin{tabular}{ccccccccccc}
         \rotatebox{90}{Train}& 
        \includegraphics[width=0.19\textwidth]{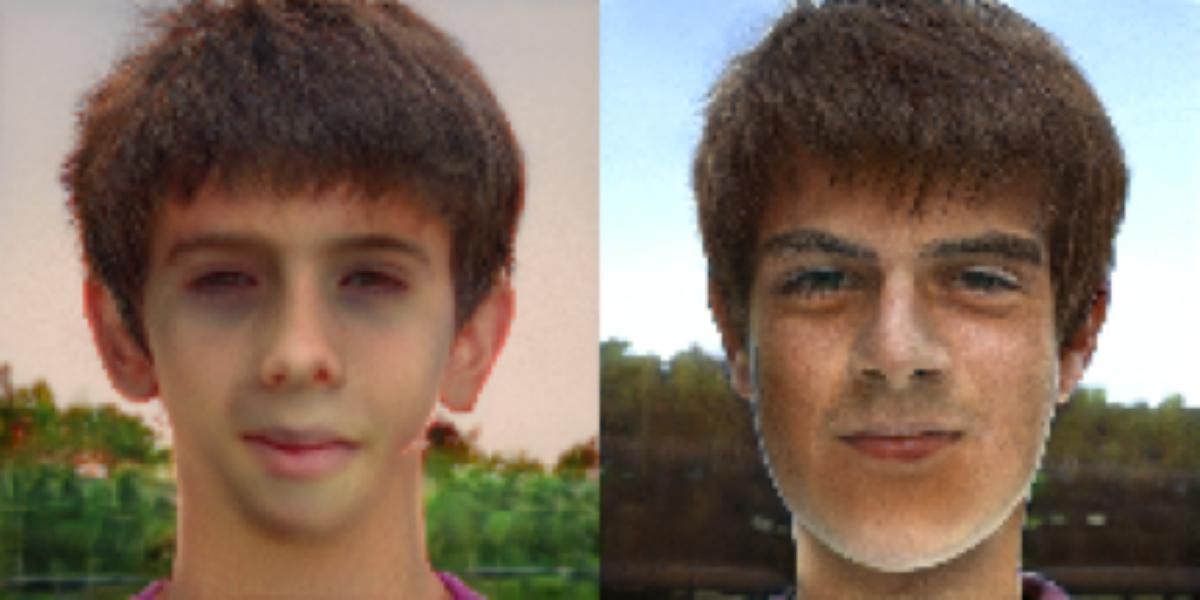} &
        \includegraphics[width=0.19\textwidth]{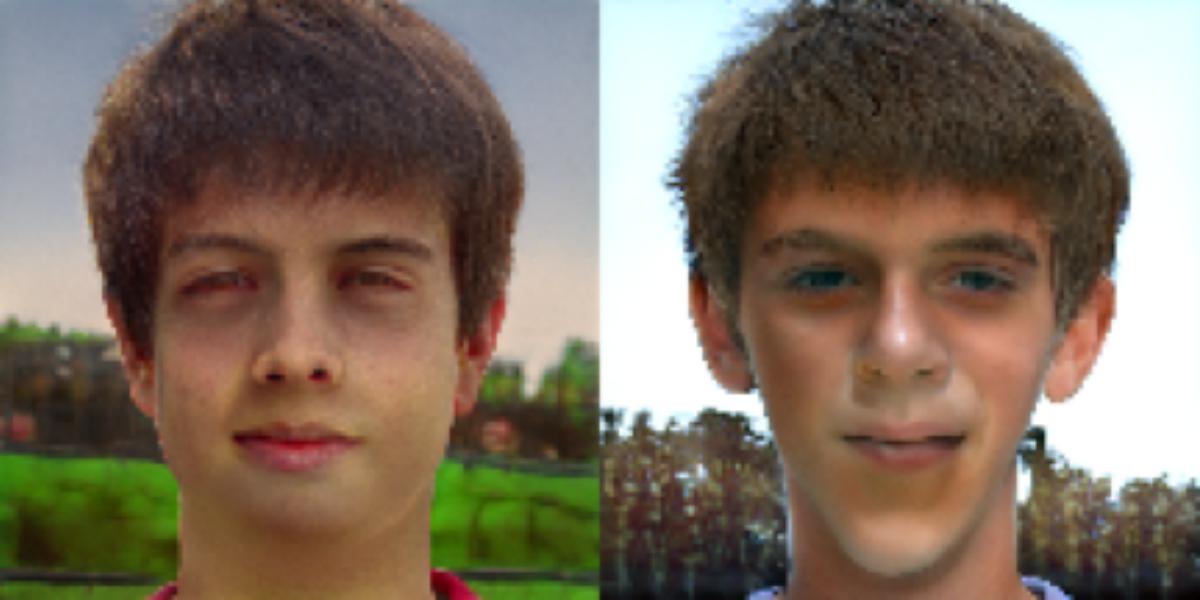} &
        \includegraphics[width=0.19\textwidth]{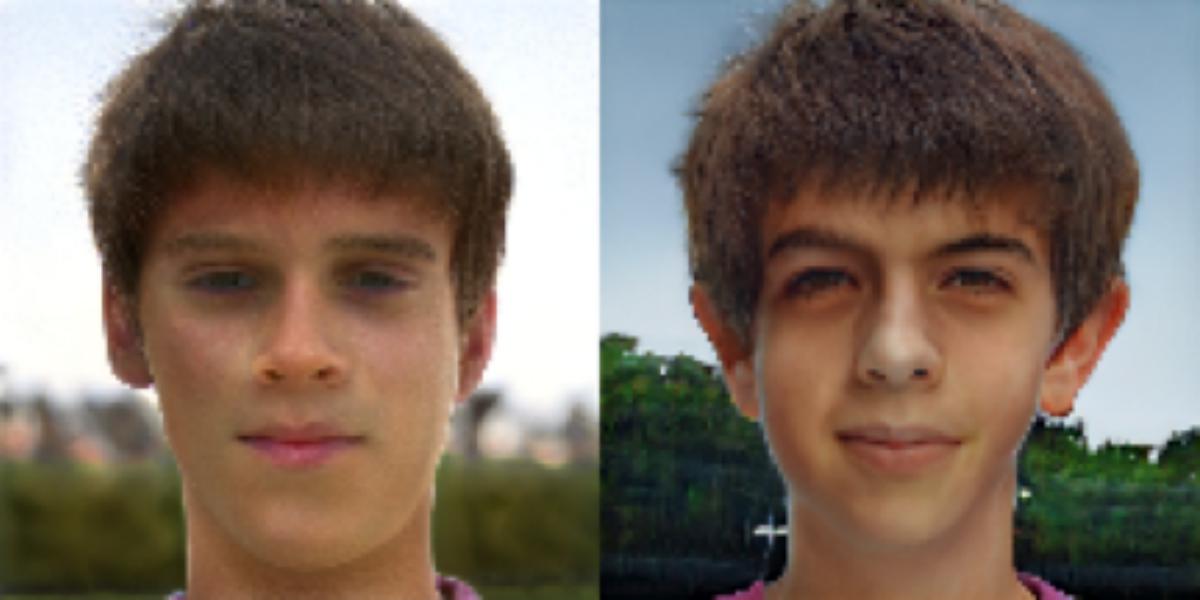} &
        \includegraphics[width=0.19\textwidth]{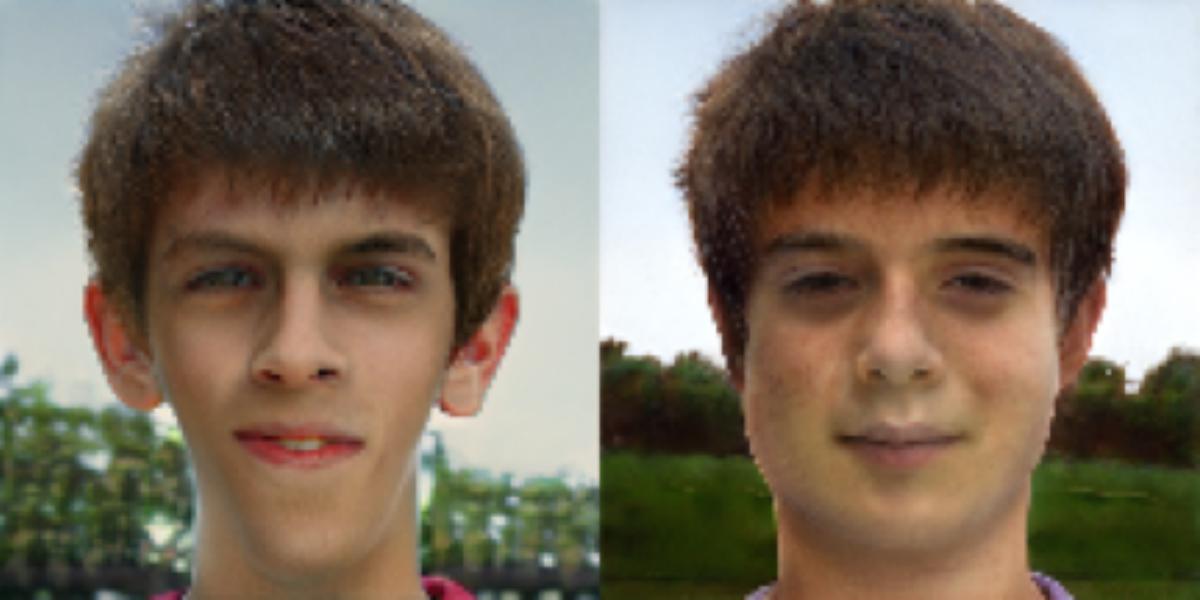} &
        \includegraphics[width=0.19\textwidth]{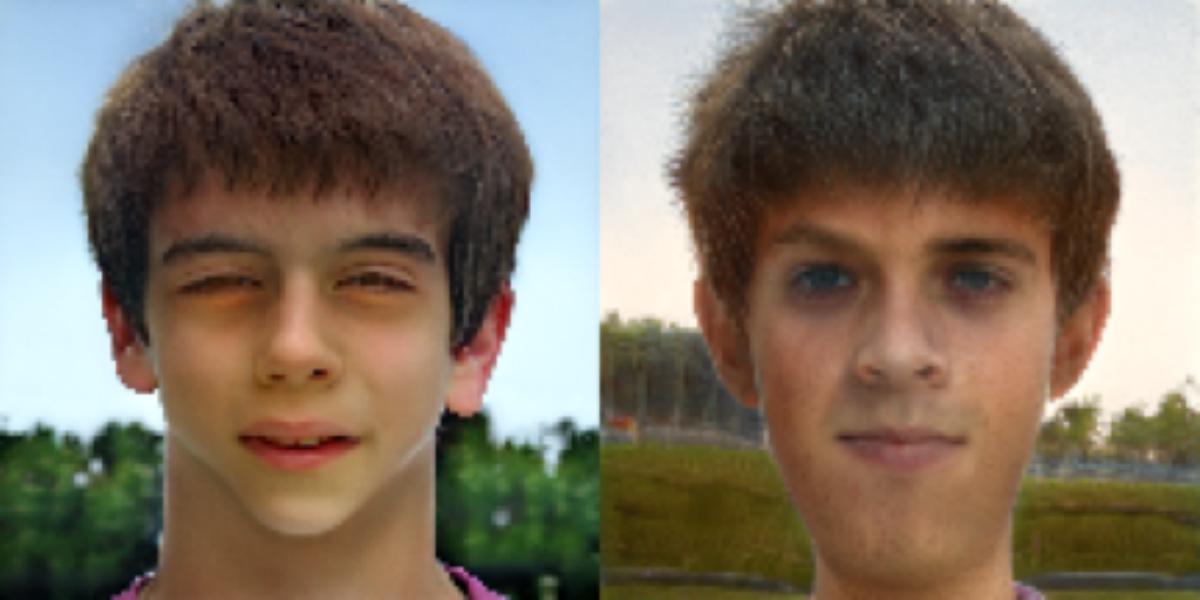} \\
         \rotatebox{90}{Test}& 
        \includegraphics[width=0.19\textwidth]{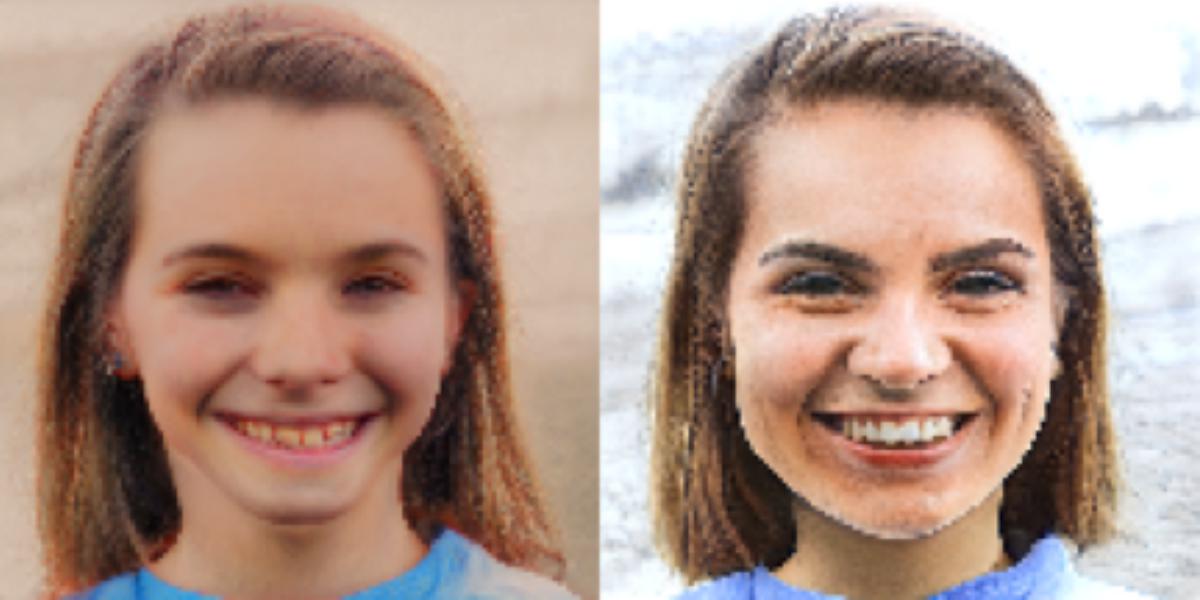} &
        \includegraphics[width=0.19\textwidth]{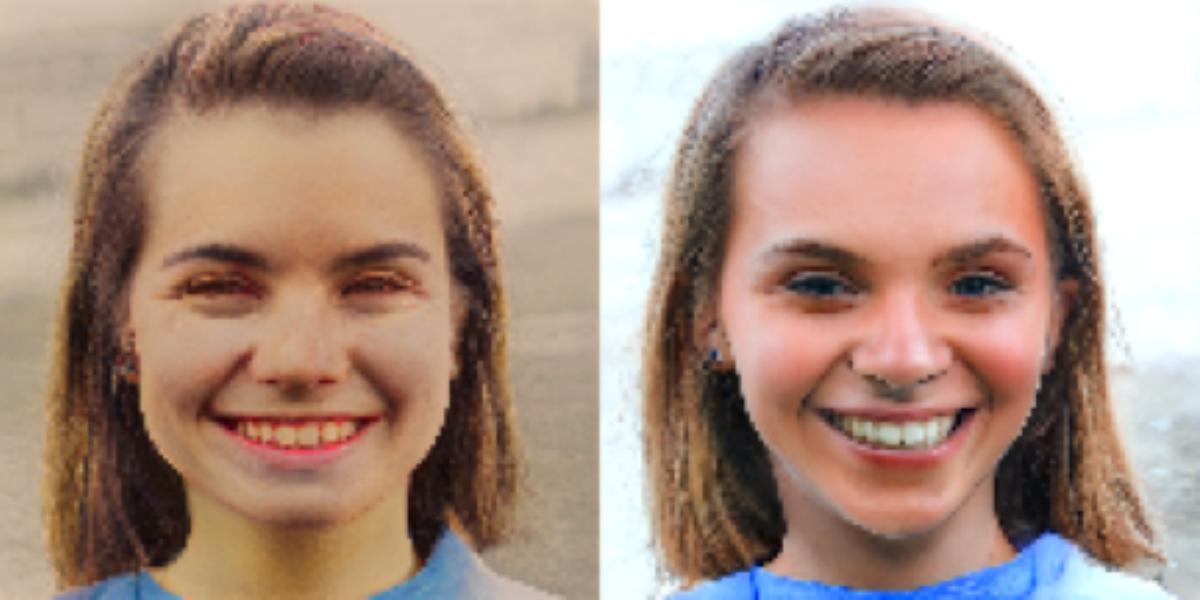} &
        \includegraphics[width=0.19\textwidth]{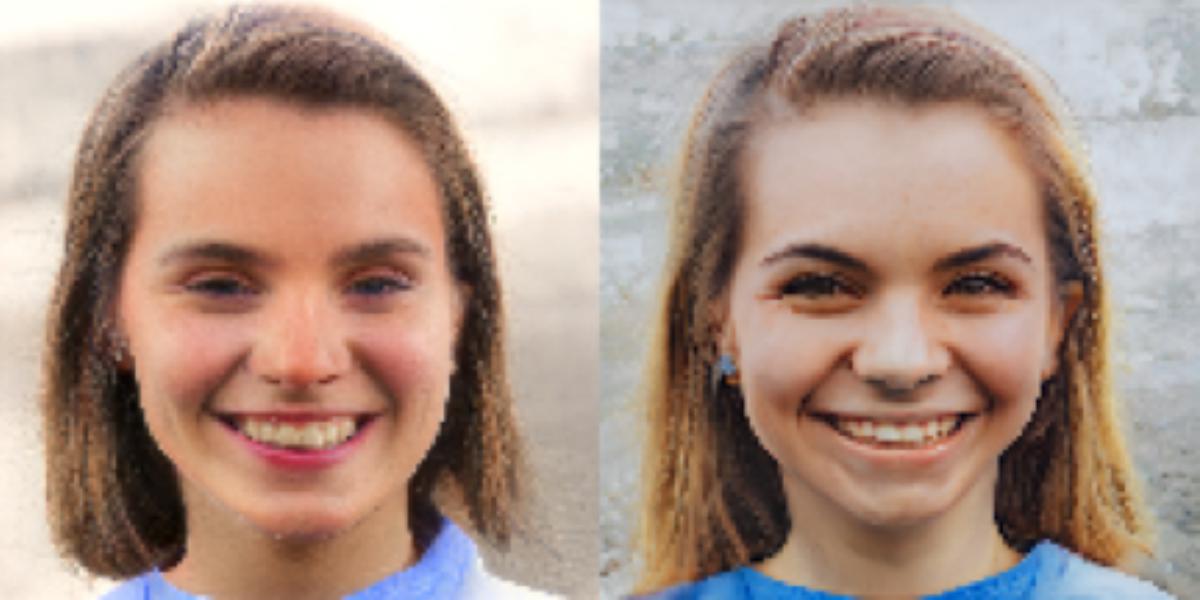} &
        \includegraphics[width=0.19\textwidth]{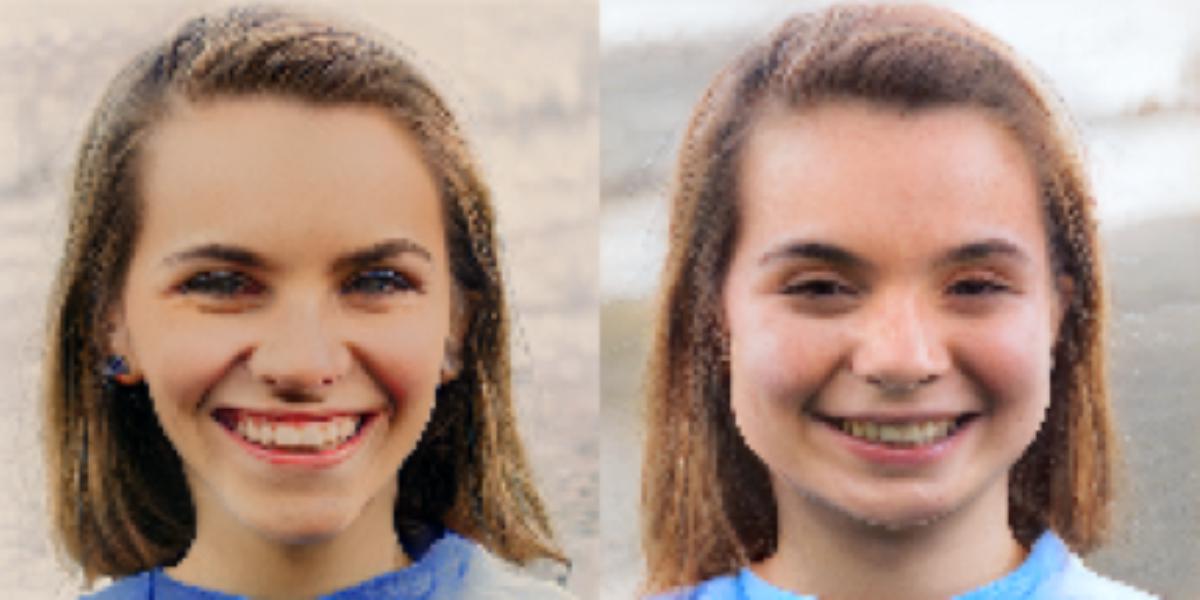} &
        \includegraphics[width=0.19\textwidth]{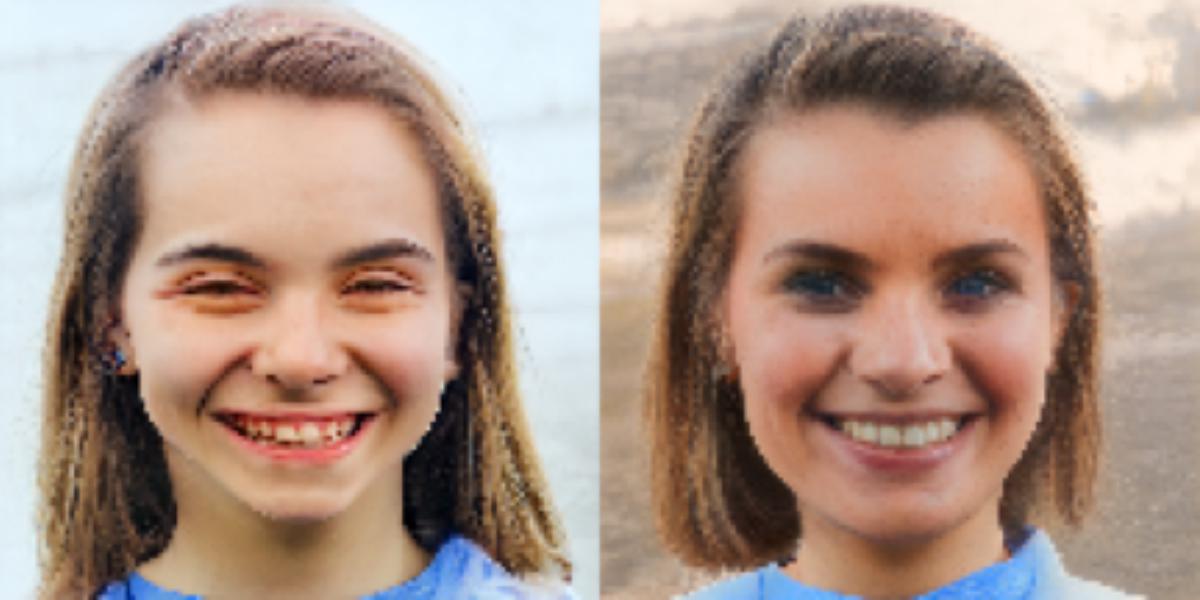} \\
         \rotatebox{90}{Test}& 
        \includegraphics[width=0.19\textwidth]{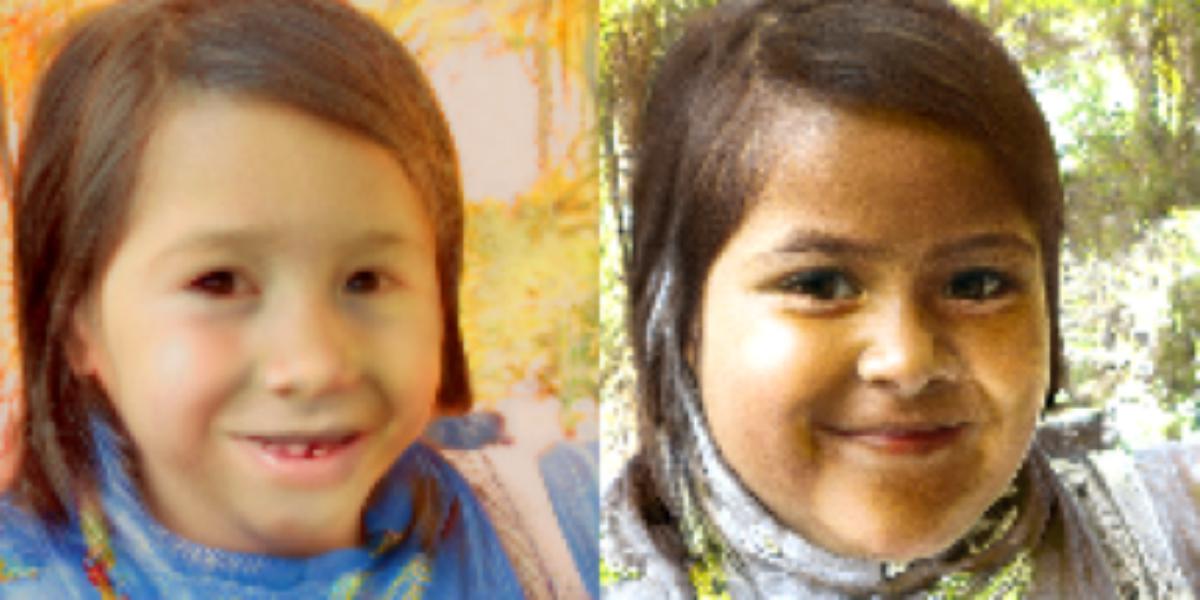} &
        \includegraphics[width=0.19\textwidth]{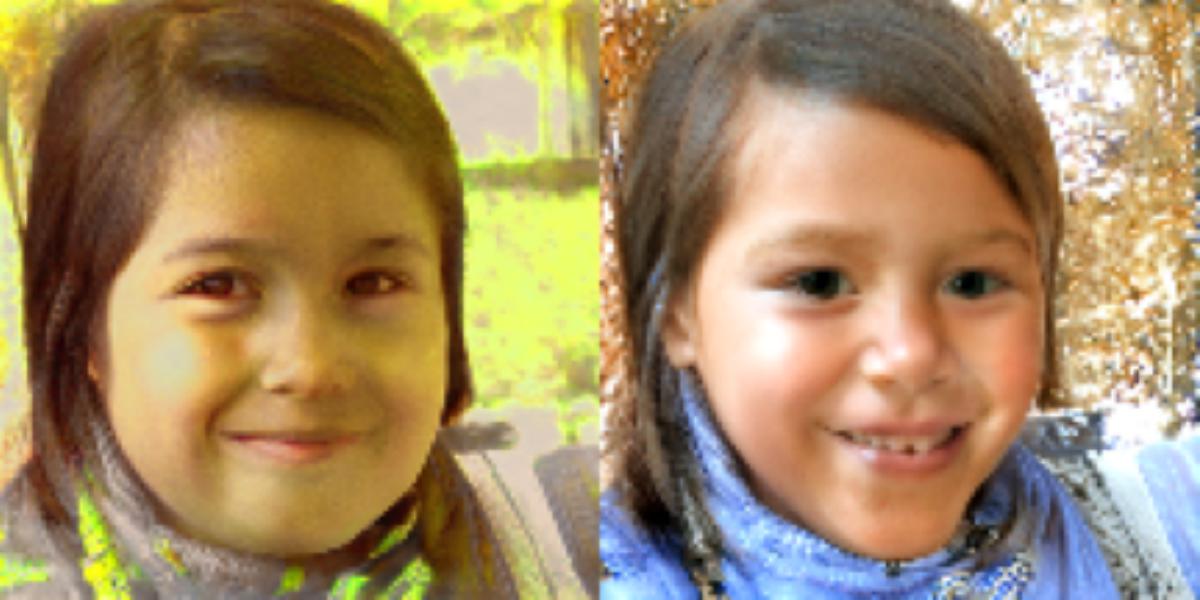} &
        \includegraphics[width=0.19\textwidth]{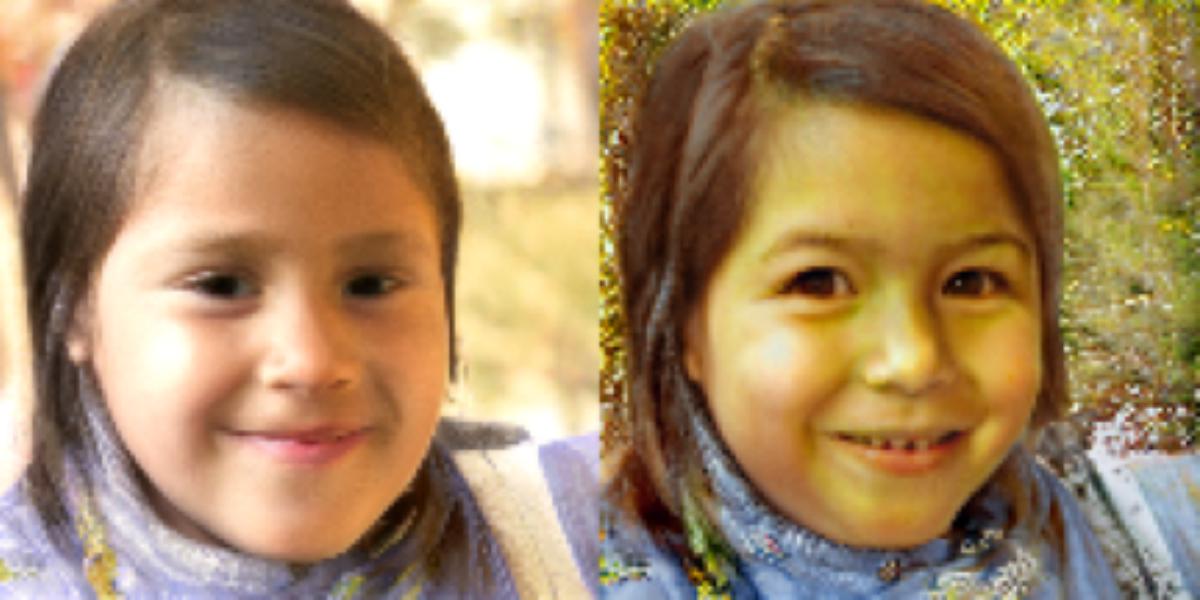} &
        \includegraphics[width=0.19\textwidth]{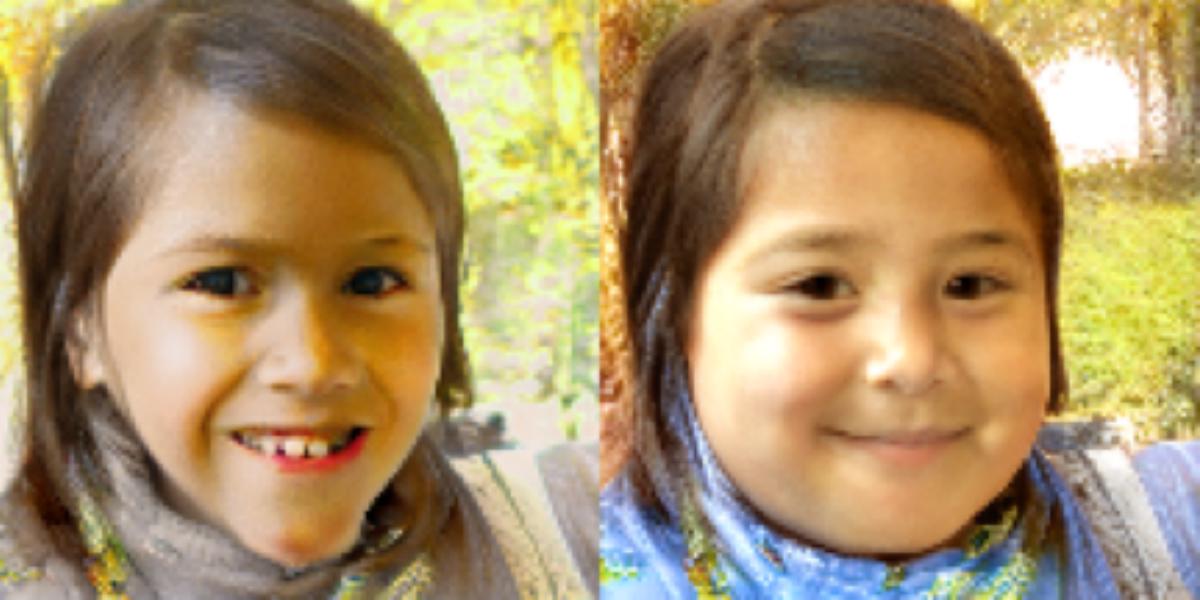} &
        \includegraphics[width=0.19\textwidth]{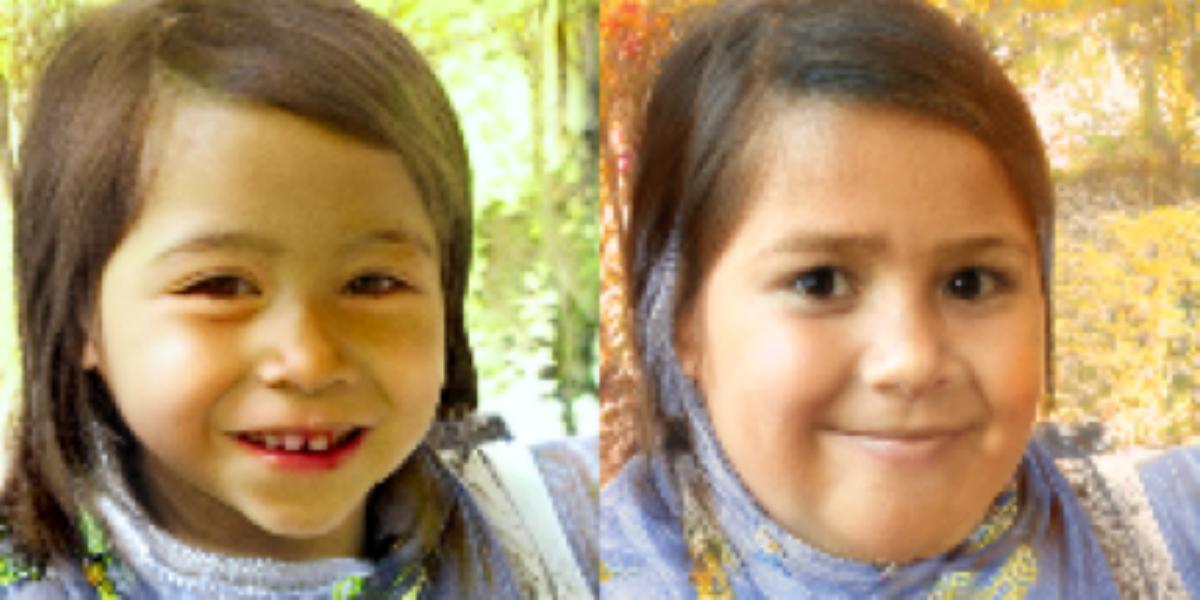} \\\hline
        \rule{0pt}{0.1\textwidth} \rotatebox{90}{Train}& 
        \includegraphics[width=0.19\textwidth]{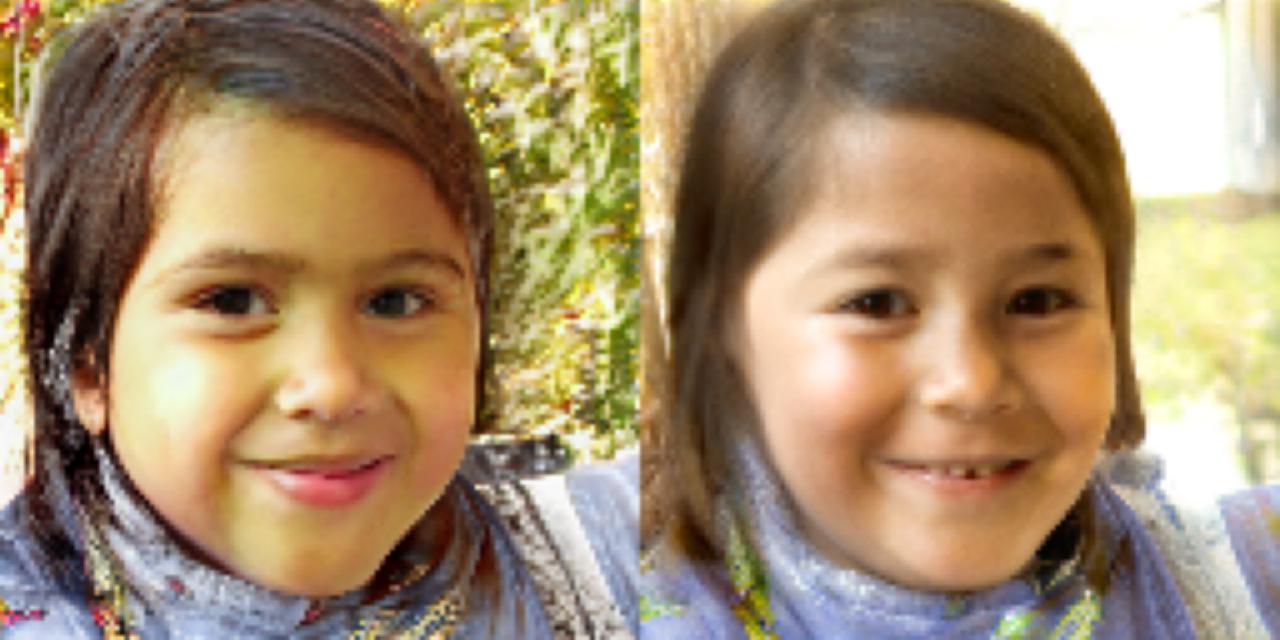} &
        \includegraphics[width=0.19\textwidth]{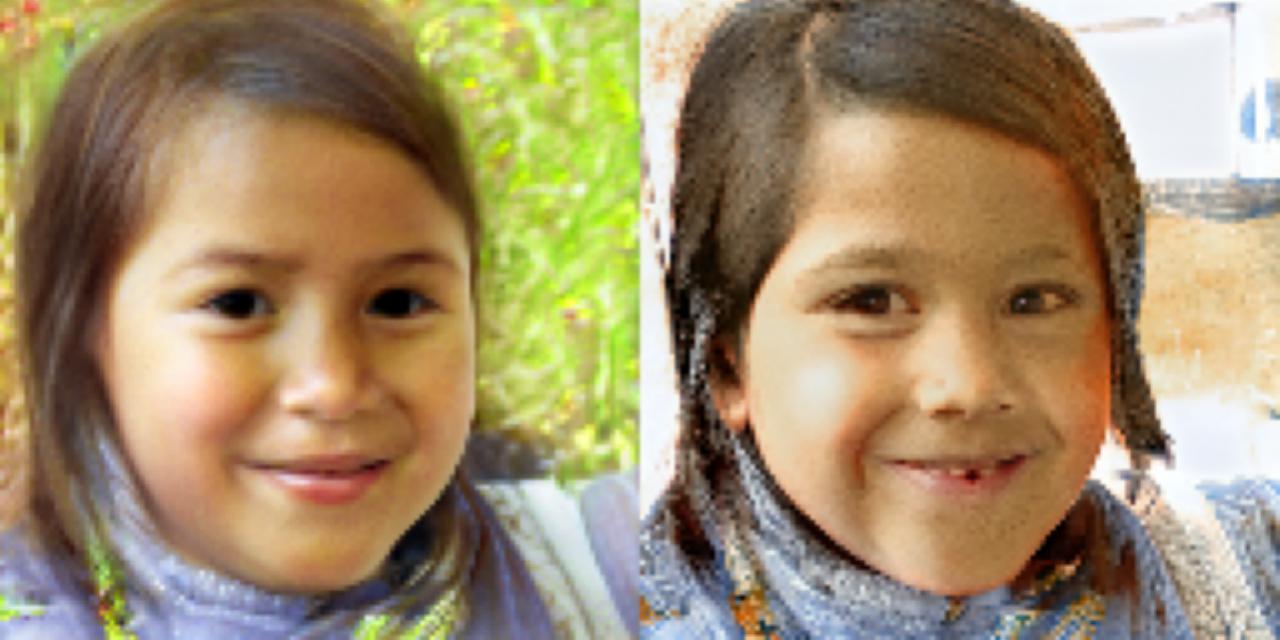} &
        \includegraphics[width=0.19\textwidth]{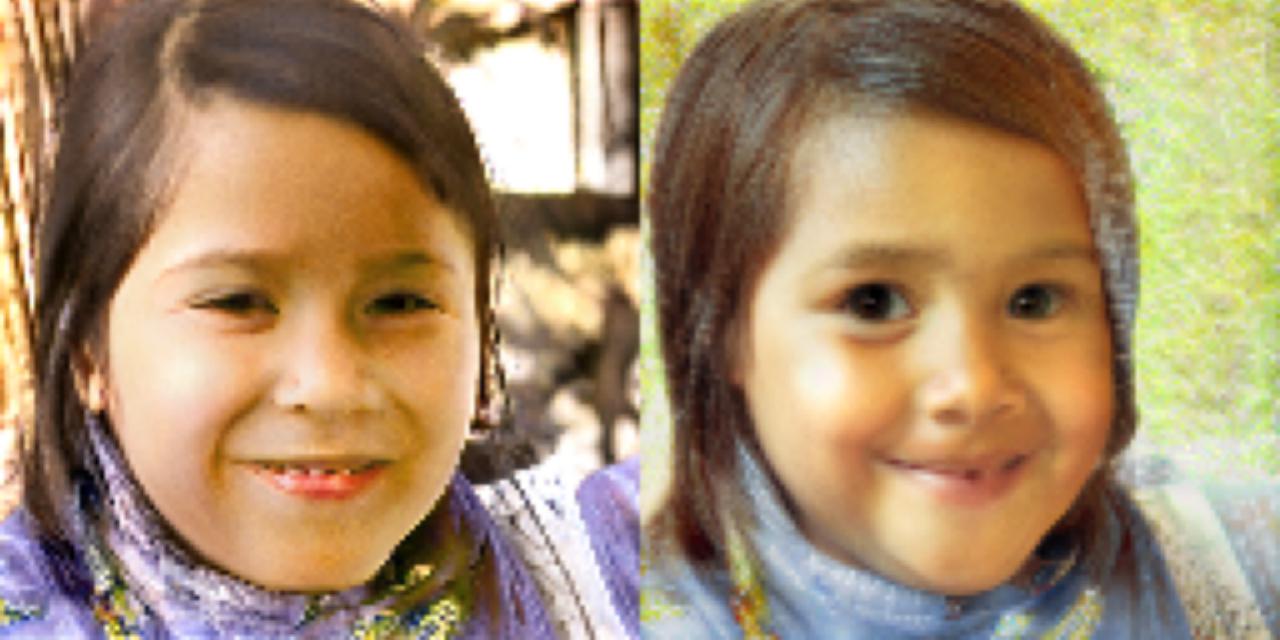} &
        \includegraphics[width=0.19\textwidth]{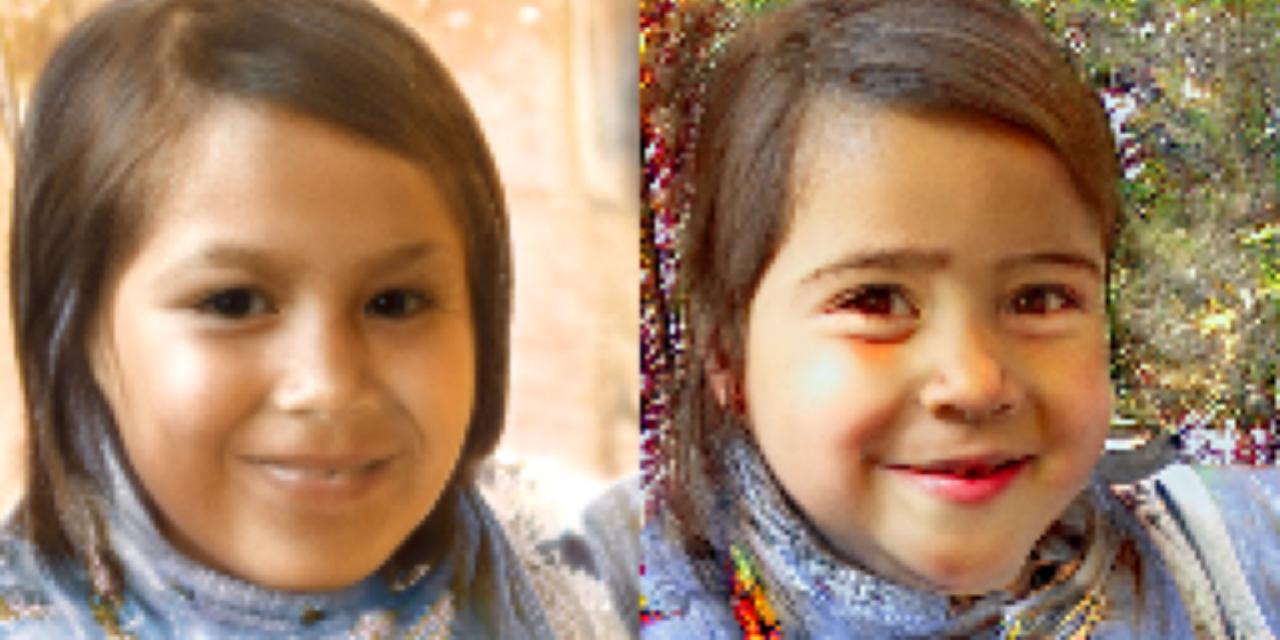} &
        \includegraphics[width=0.19\textwidth]{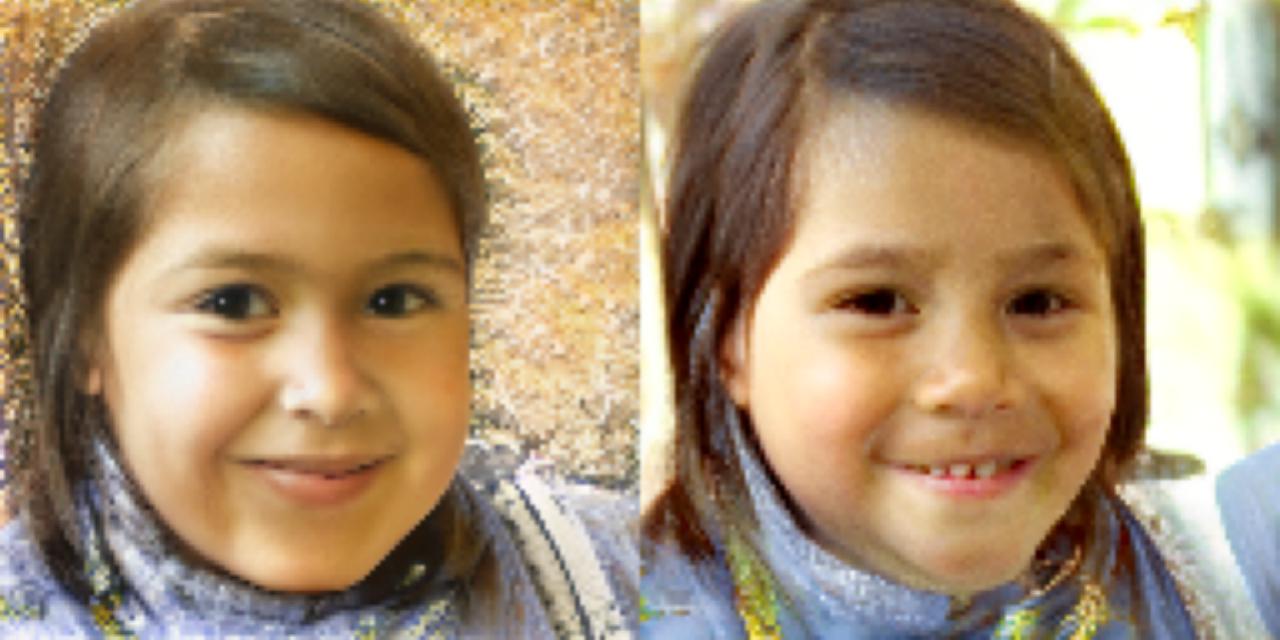} \\
         \rotatebox{90}{Test}& 
        \includegraphics[width=0.19\textwidth]{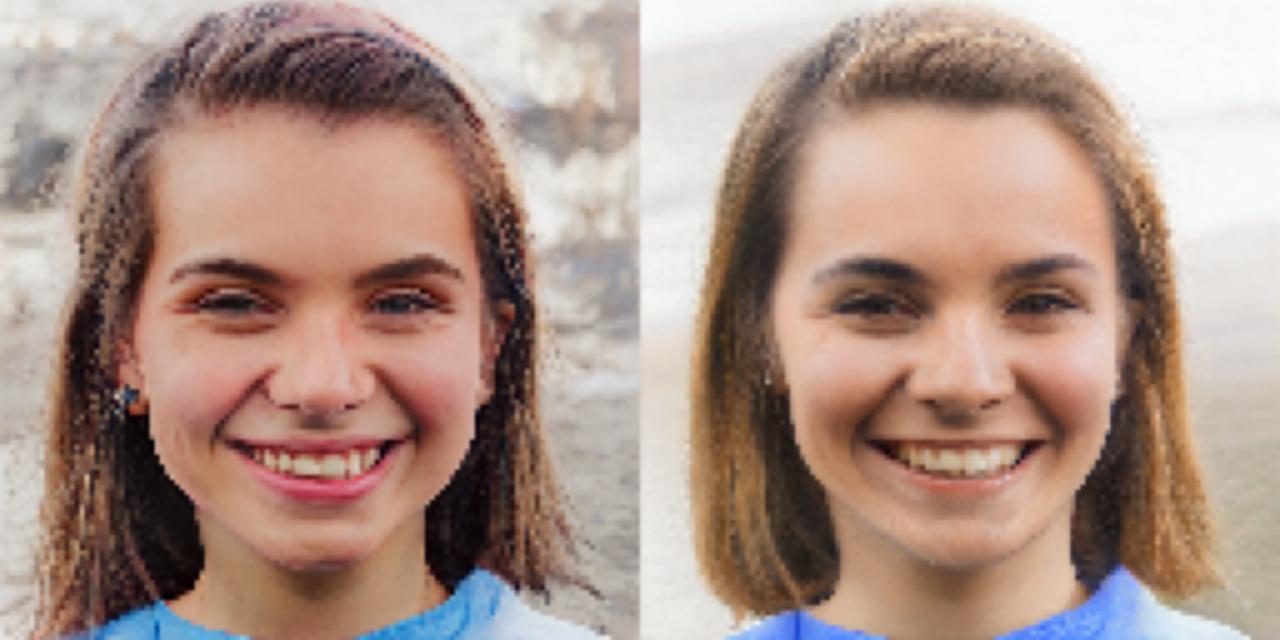} &
        \includegraphics[width=0.19\textwidth]{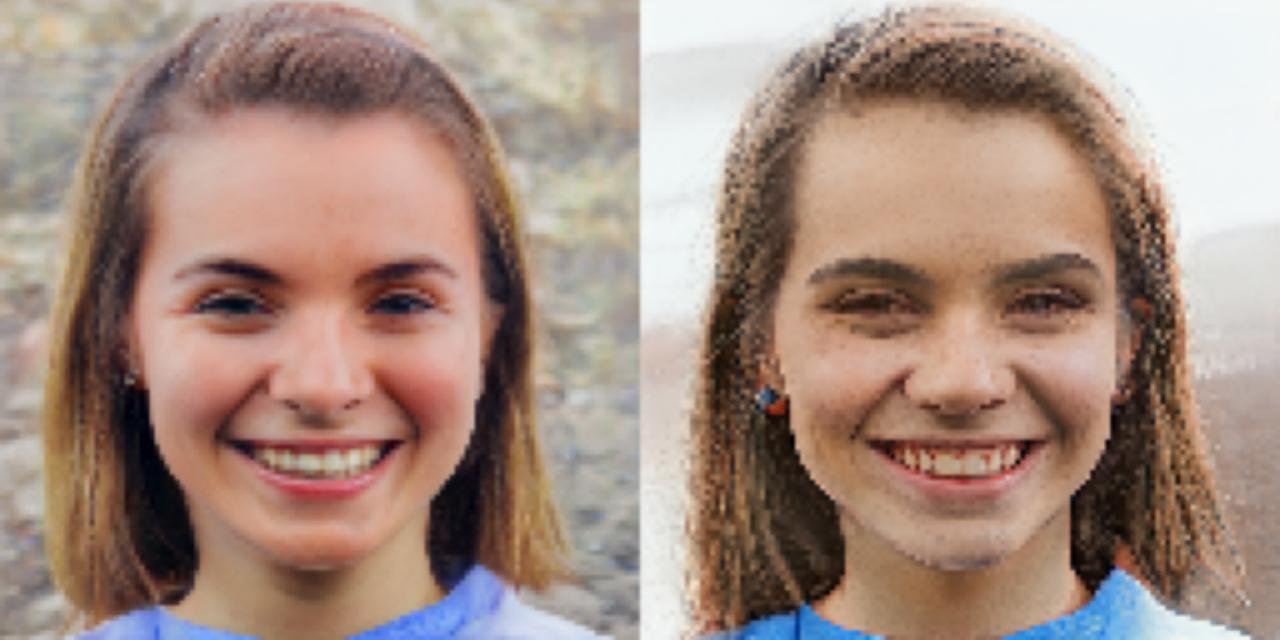} &
        \includegraphics[width=0.19\textwidth]{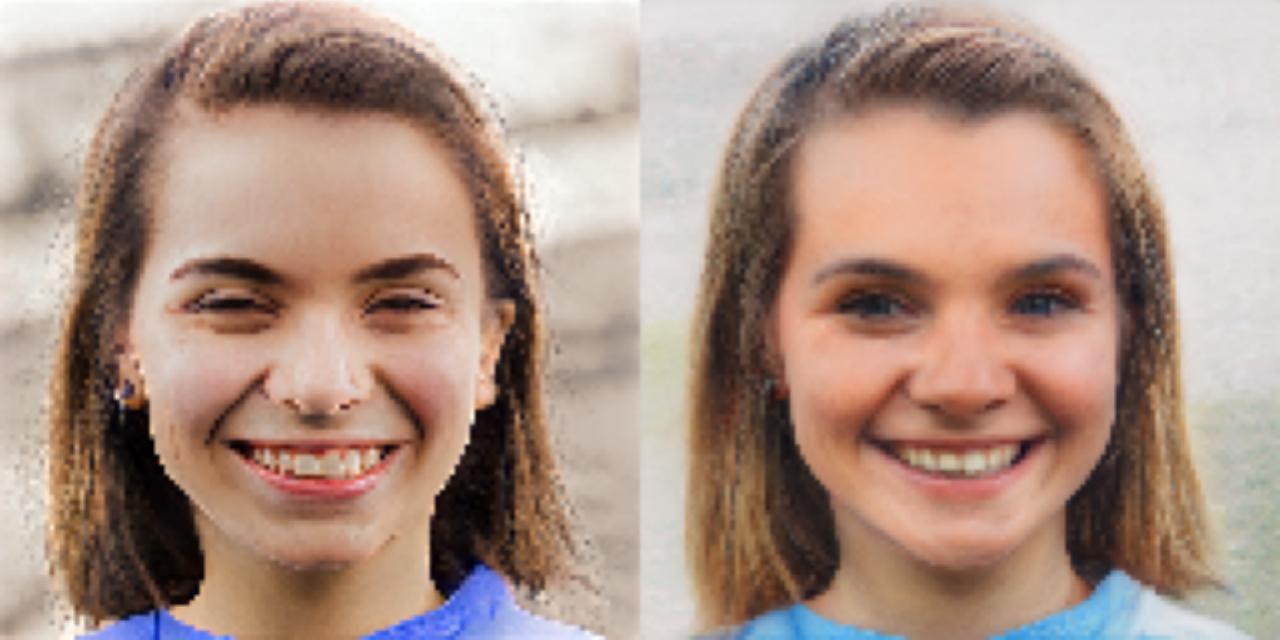} &
        \includegraphics[width=0.19\textwidth]{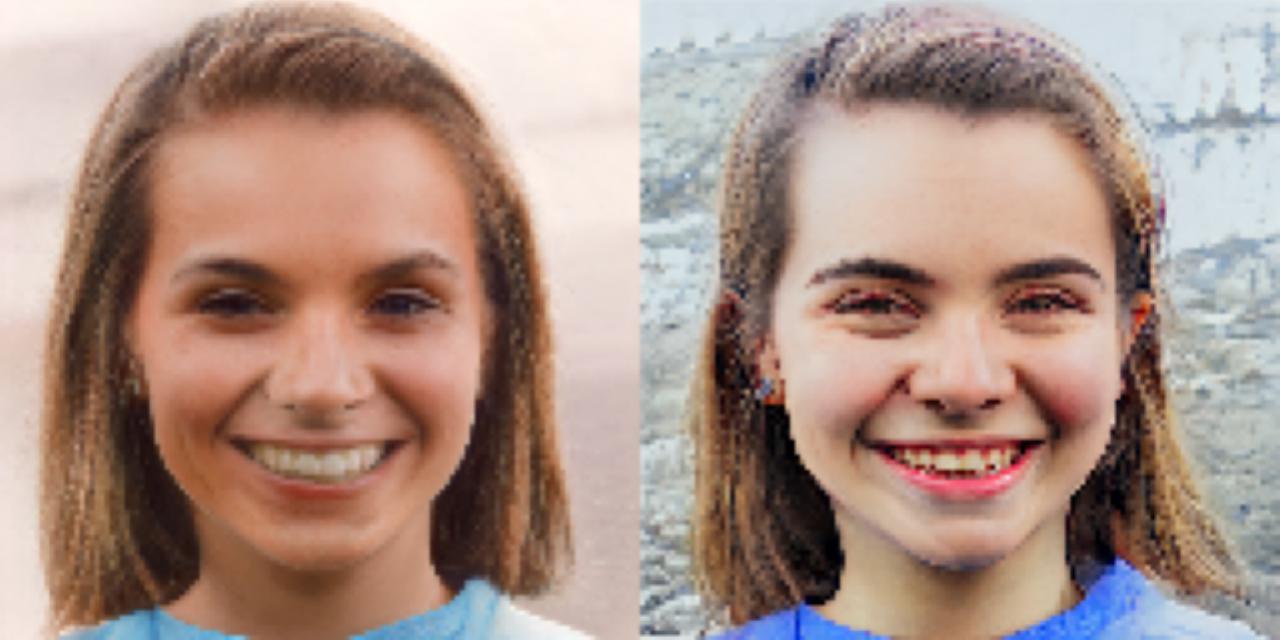} &
        \includegraphics[width=0.19\textwidth]{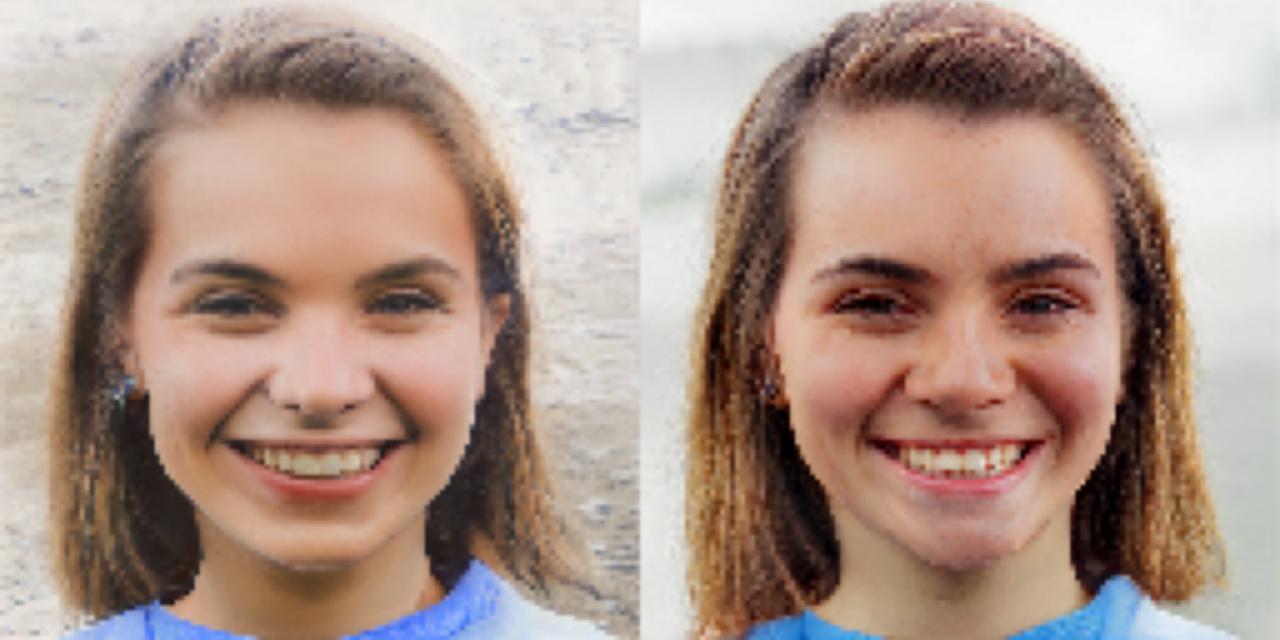} \\\hline
        \rule{0pt}{0.1\textwidth} \rotatebox{90}{Train}& 
        \includegraphics[width=0.19\textwidth]{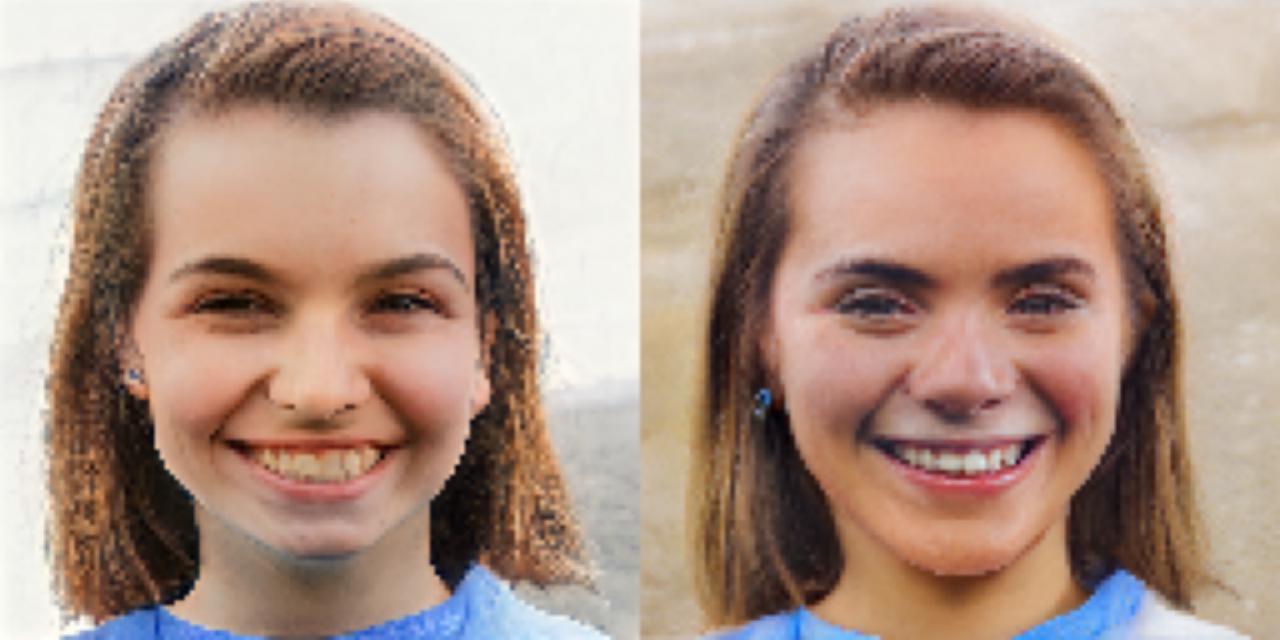} &
        \includegraphics[width=0.19\textwidth]{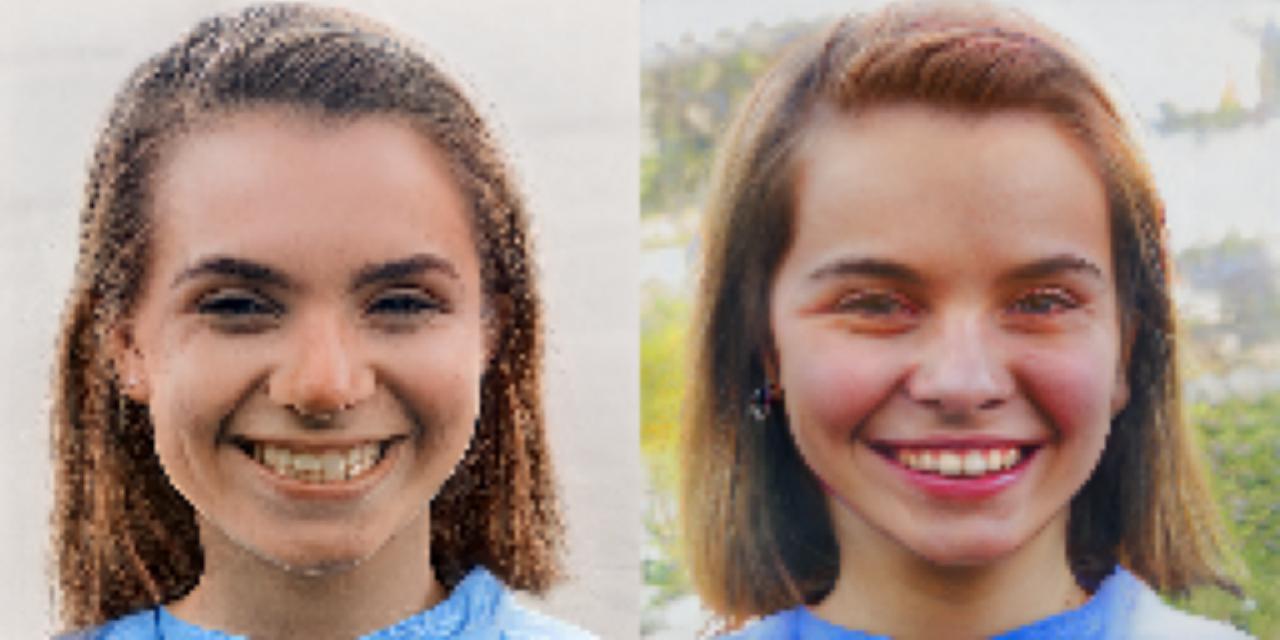} &
        \includegraphics[width=0.19\textwidth]{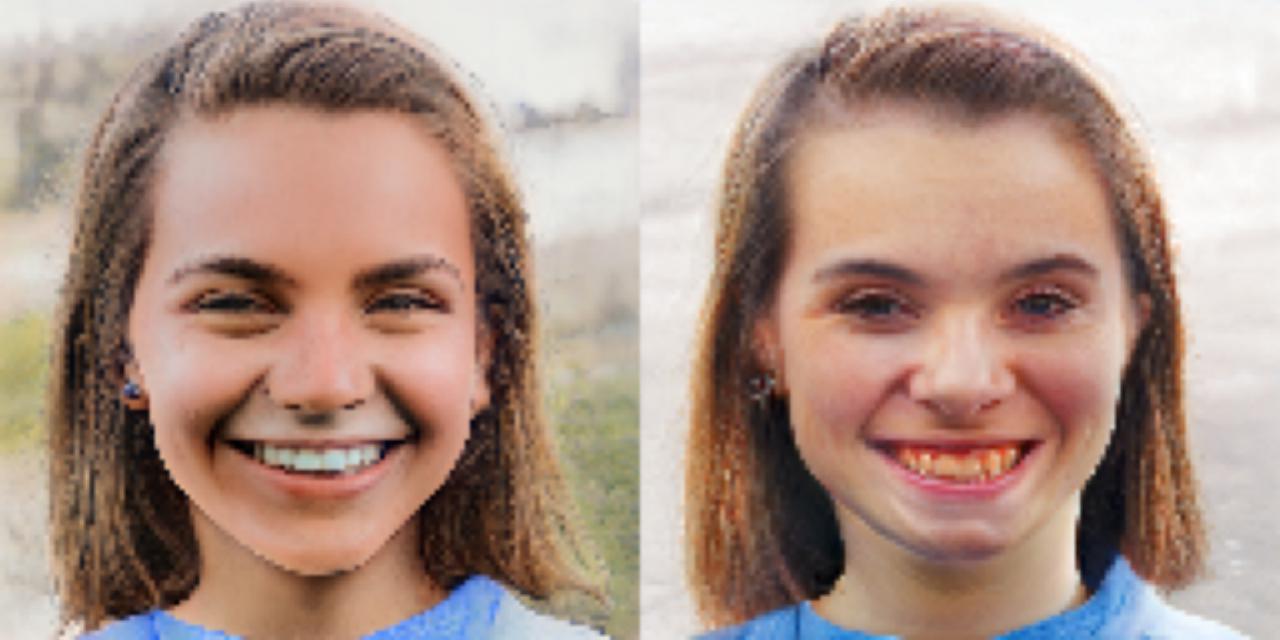} &
        \includegraphics[width=0.19\textwidth]{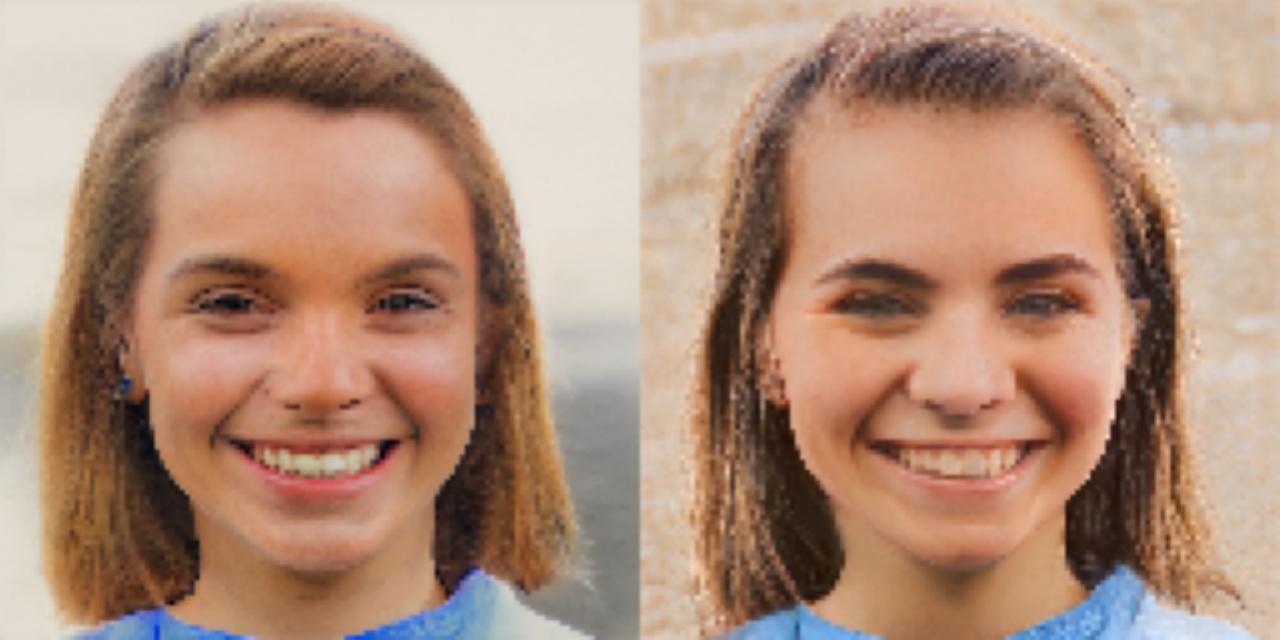} &
        \includegraphics[width=0.19\textwidth]{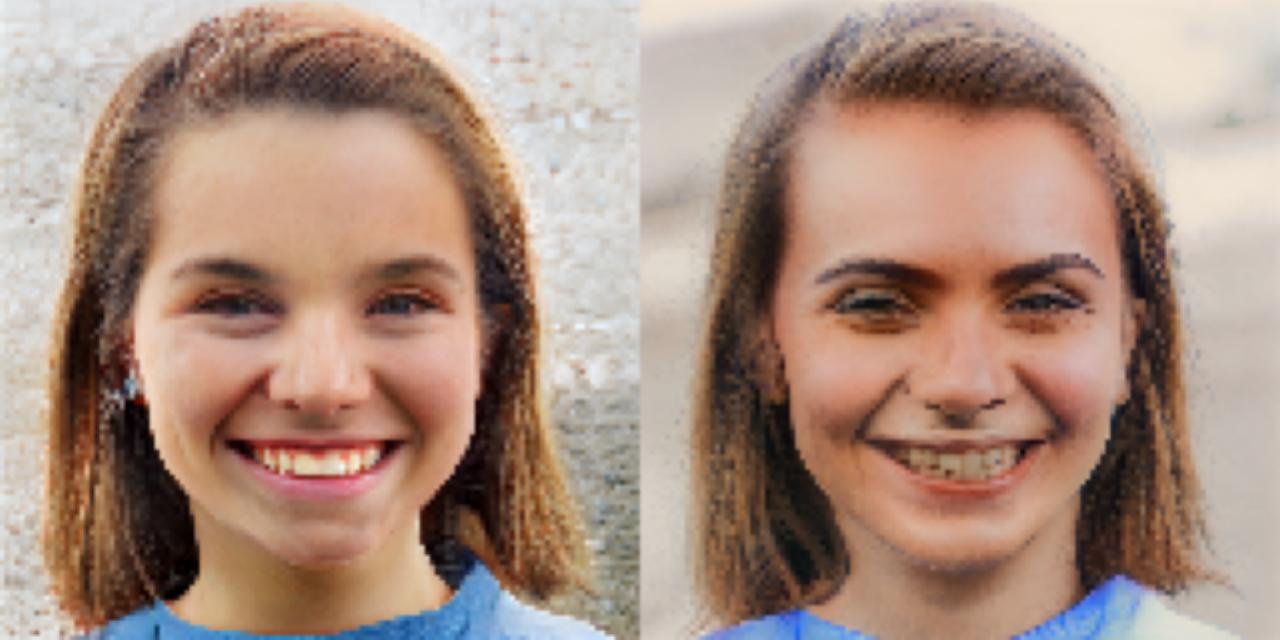} \\
         \rotatebox{90}{Test}& 
        \includegraphics[width=0.19\textwidth]{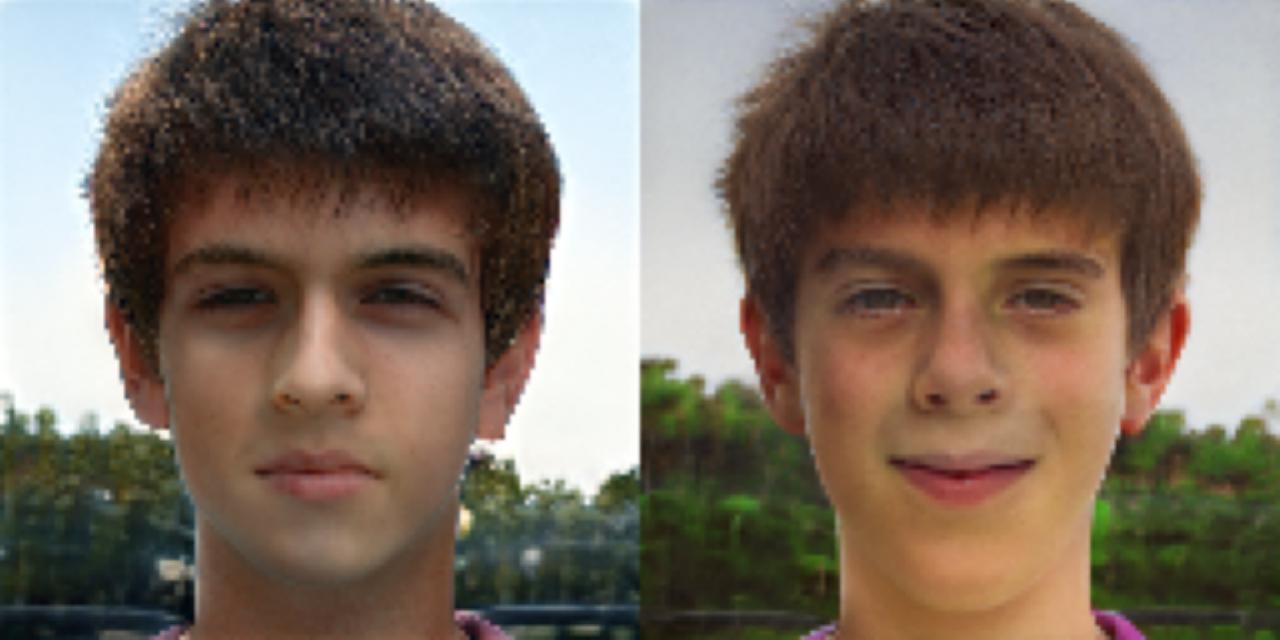} &
        \includegraphics[width=0.19\textwidth]{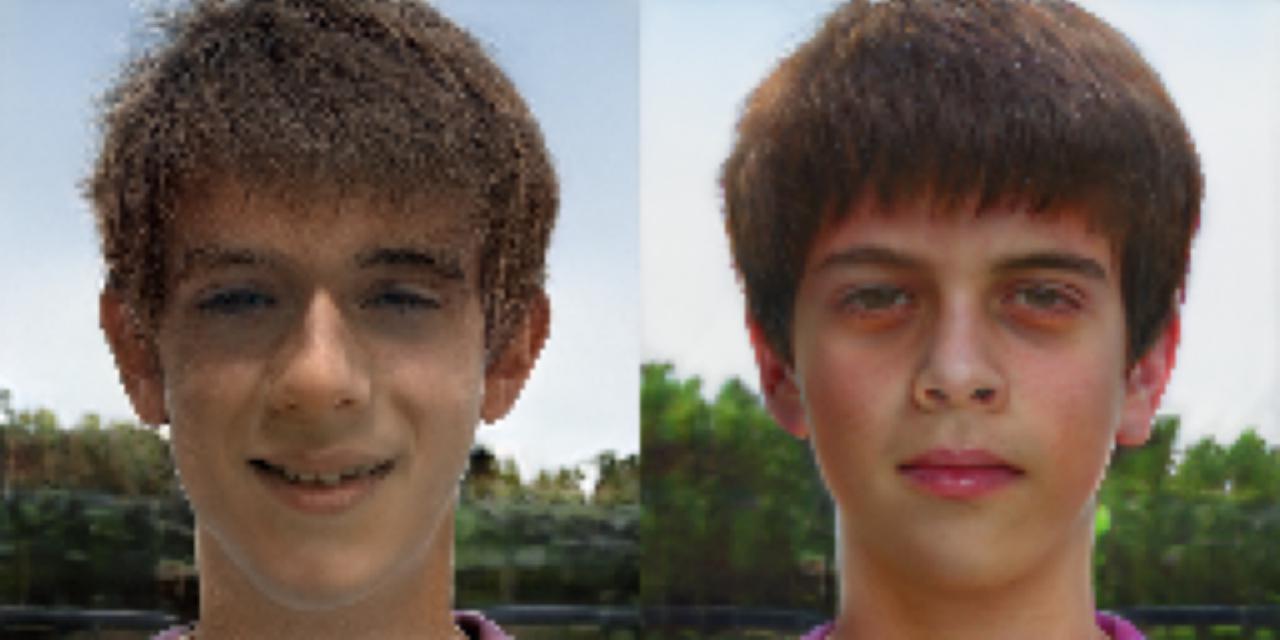} &
        \includegraphics[width=0.19\textwidth]{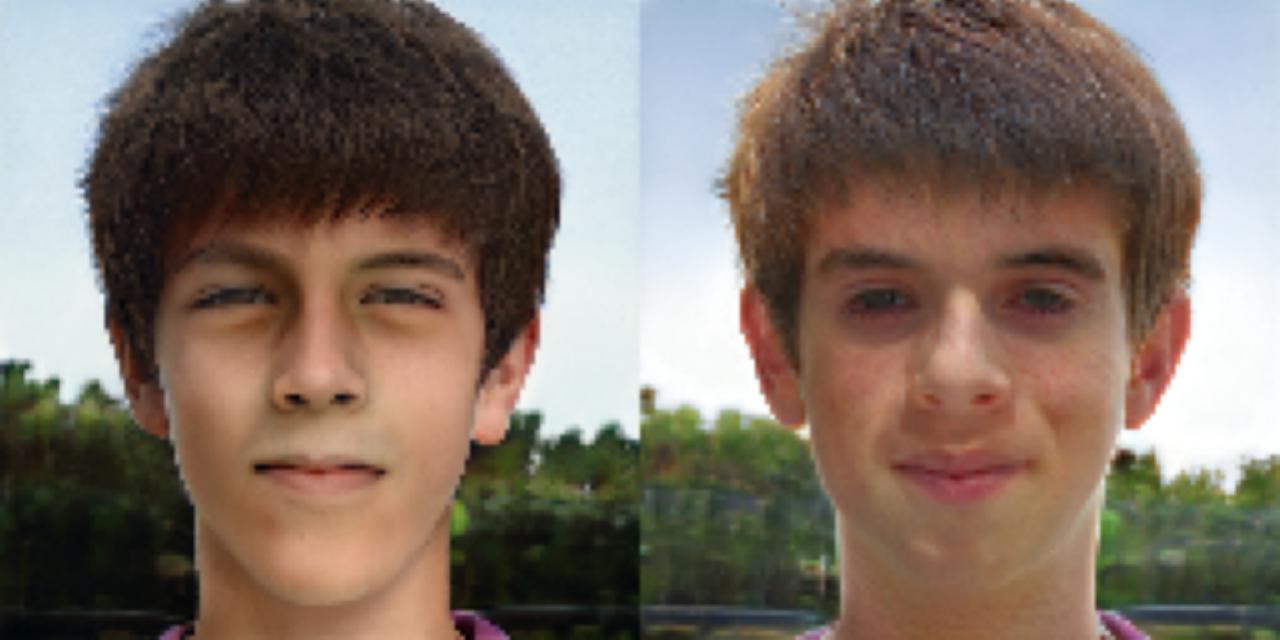} &
        \includegraphics[width=0.19\textwidth]{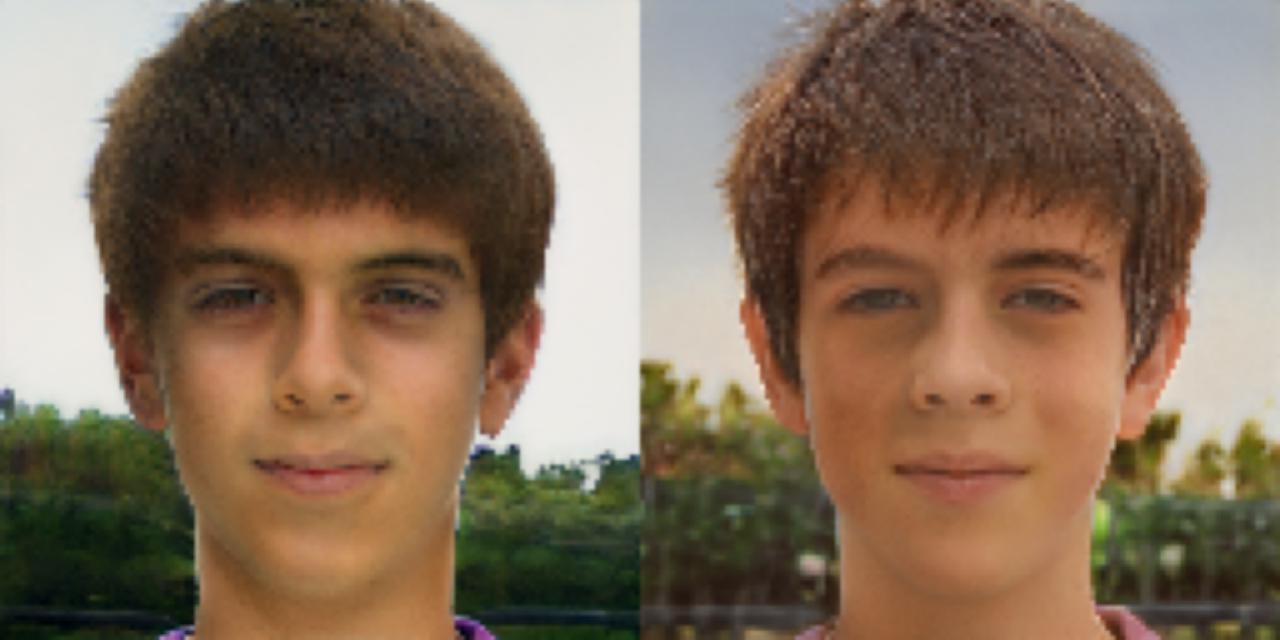} &
        \includegraphics[width=0.19\textwidth]{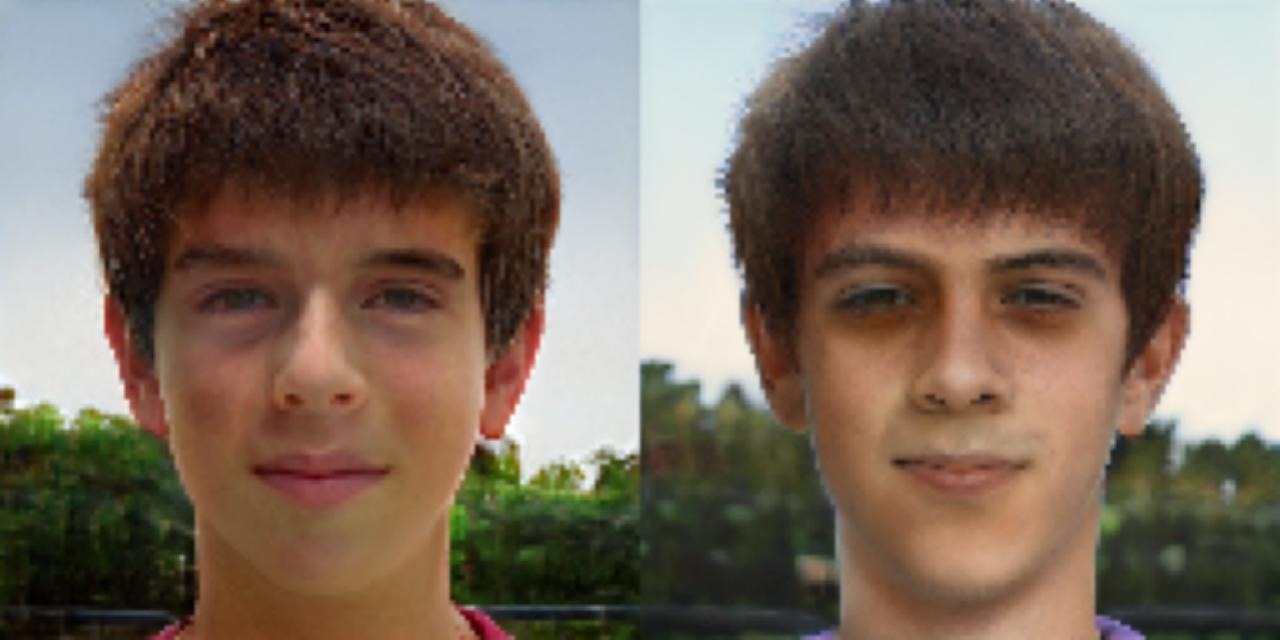} \\
    \end{tabular}
    \caption{Illustration of a failure case with poor transferability. ``Train'' indicates that the susbpaces are solved from the corresponding original images and the manipulation of training data is visualized. ``Test'' indicates the manipulation on test data. The upper and lower parts swap training and testing data. See \Cref{sec:transferability-sup}. }
    \label{fig:transferability}
\end{figure*}

Though the most of our discovered interpretable subspaces is shows to have good transferability, we do observe some cases where the effectiveness of discovered subspaces is closely relevant to the traning samples. We show an example of manipulating face boundary in \Cref{fig:transferability}. The corresponding criterion can be found in \Cref{tab:exp-formulations}. We use one training sample to solve the latent subspace and show the manipulation result on both training and tesing images. As can be noticed, we discover richest modification of face shape when trained with the male teenage in the first row. The modification transfers well on testing images. However, when changing training images as in the second row, the resulted controls fail to produce significant change on both training and testing images. We further use the testing image as training data in the third row, and only obtain very slight changes on the length of chin.

Among the subspaces formulated in \Cref{tab:exp-formulations}, only the face boundary case is found to have the above problem. We suspect that the landmark detector exploits inconsistent semantic features from different images, which conflicts with the semantic disentanglement in the generator. 

The formulation of \eqref{eq:l2-diff} naturally supports batch operation by summing over multiple $u$. In our experiments, we find 1-2 training images are sufficient to solve most latent subspaces. However, the above face boundary case is an exception. As is shown in \Cref{fig:transferability}, the latent subspace principle components are inconsistent on different samples. We observe that batching these samples tends to cancel out meaningful manipulation components.

\section{Counterfactual Explanation Experiments}
\label{sec:exp-counterfactual-materials}

\begin{figure*}
    \centering
    \footnotesize
    \setlength{\tabcolsep}{1pt}
    \begin{tabular}{cccccccc}
        \includegraphics[width=0.14\textwidth]{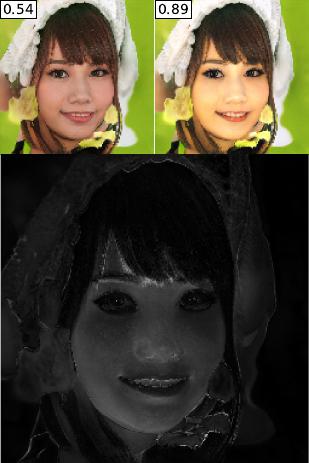} &
        \includegraphics[width=0.14\textwidth]{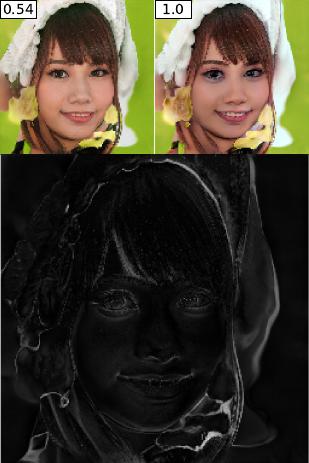} &
        \includegraphics[width=0.14\textwidth]{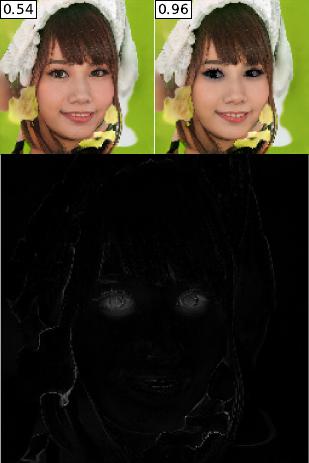} &
        \includegraphics[width=0.14\textwidth]{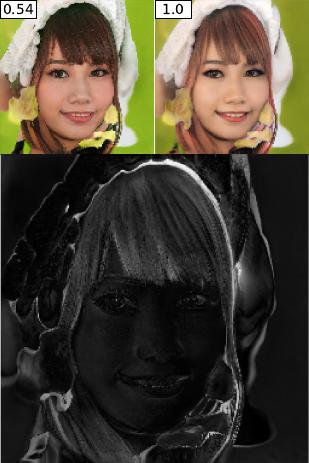} &
        \includegraphics[width=0.14\textwidth]{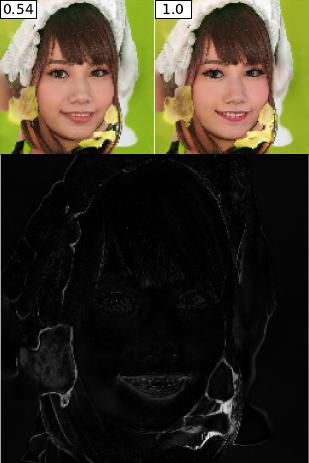} &
        \includegraphics[width=0.14\textwidth]{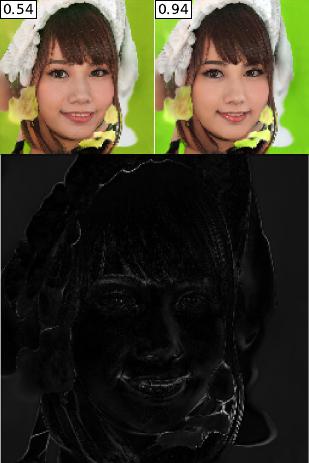} &
        \includegraphics[width=0.14\textwidth]{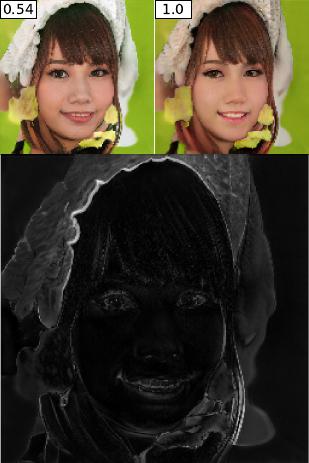} &
        \\
        \includegraphics[width=0.14\textwidth]{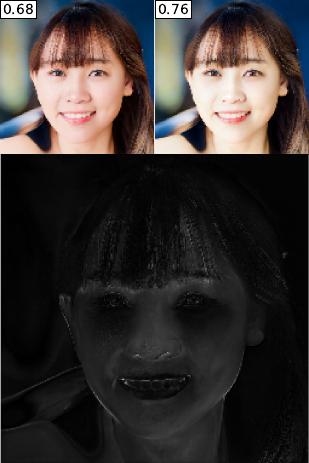} &
        \includegraphics[width=0.14\textwidth]{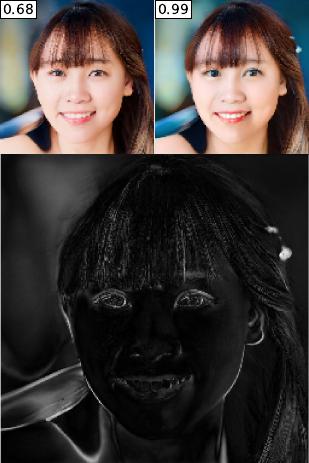} &
        \includegraphics[width=0.14\textwidth]{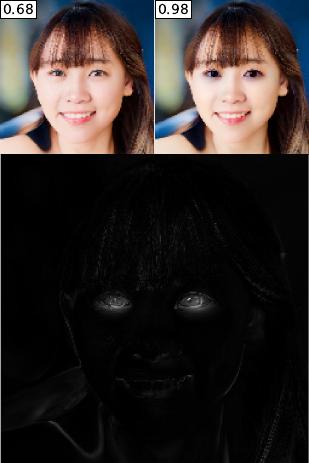} &
        \includegraphics[width=0.14\textwidth]{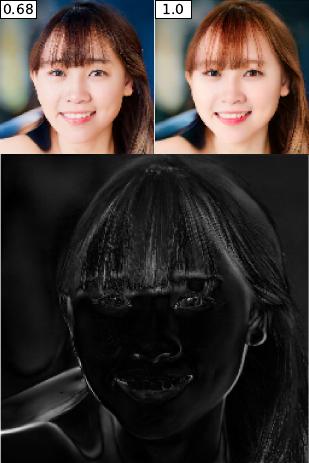} &
        \includegraphics[width=0.14\textwidth]{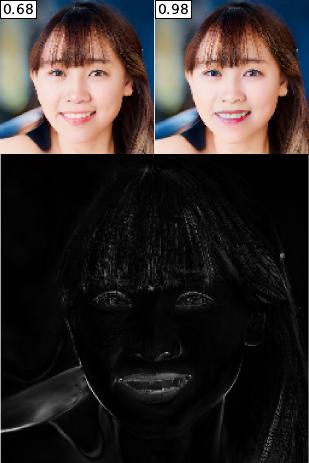} &
        \includegraphics[width=0.14\textwidth]{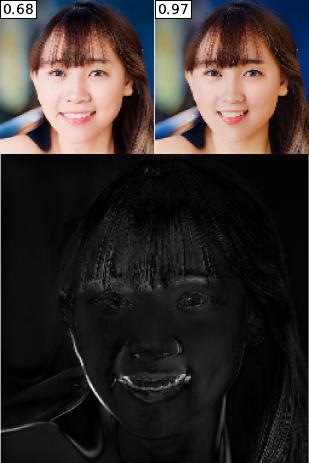} &
        \includegraphics[width=0.14\textwidth]{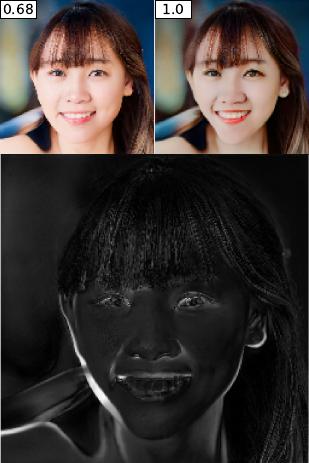} &
        \\
        \includegraphics[width=0.14\textwidth]{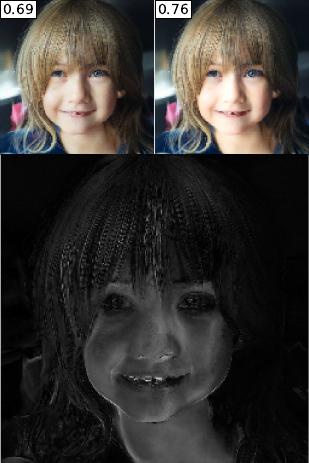} &
        \includegraphics[width=0.14\textwidth]{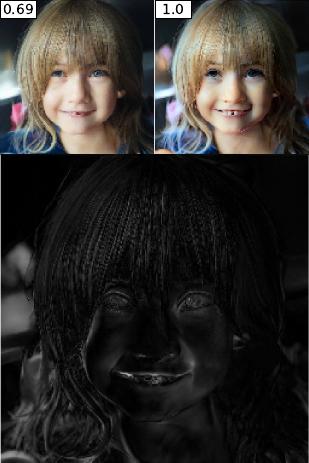} &
        \includegraphics[width=0.14\textwidth]{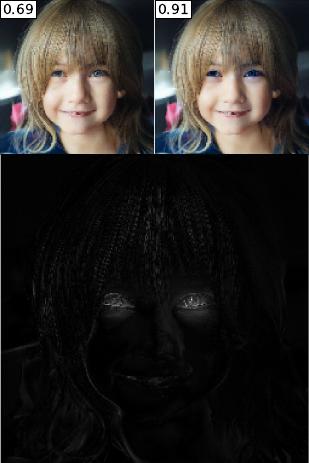} &
        \includegraphics[width=0.14\textwidth]{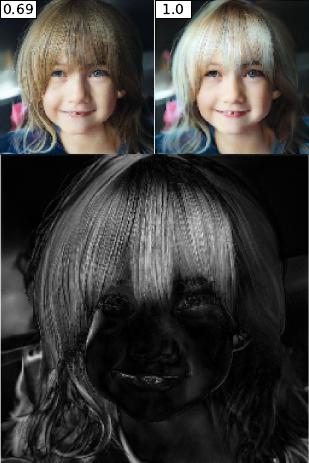} &
        \includegraphics[width=0.14\textwidth]{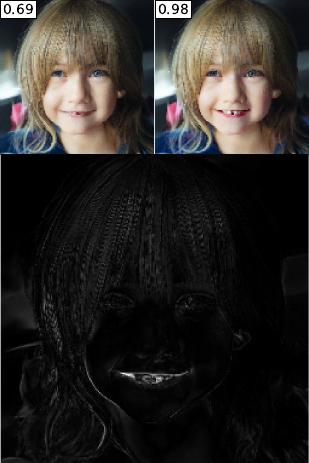} &
        \includegraphics[width=0.14\textwidth]{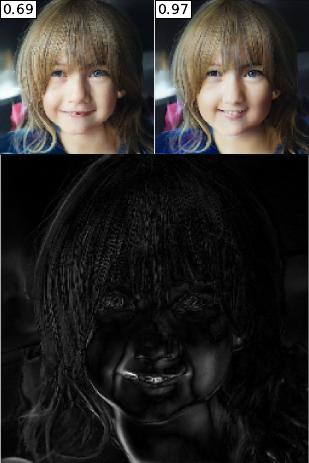} &
        \includegraphics[width=0.14\textwidth]{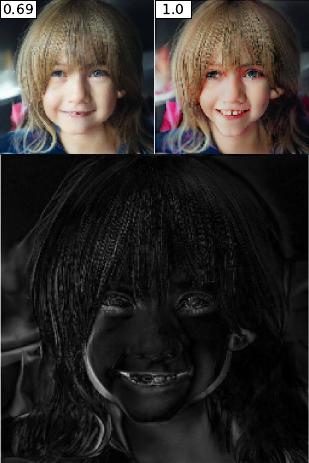} &
        \\
        \includegraphics[width=0.14\textwidth]{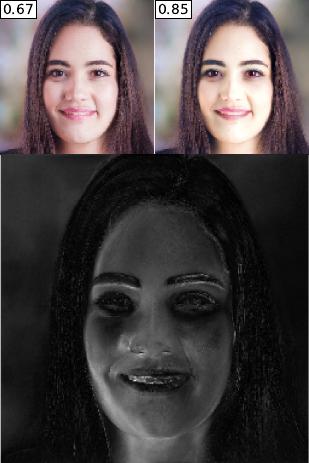} &
        \includegraphics[width=0.14\textwidth]{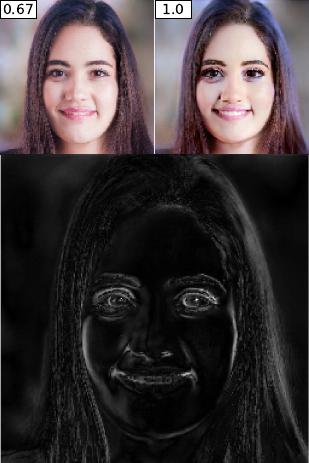} &
        \includegraphics[width=0.14\textwidth]{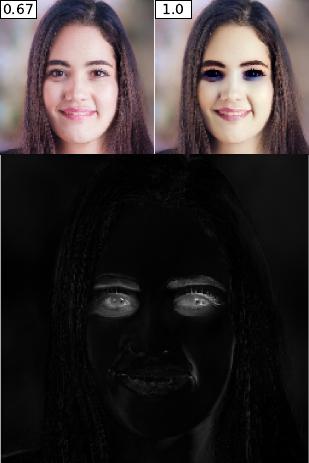} &
        \includegraphics[width=0.14\textwidth]{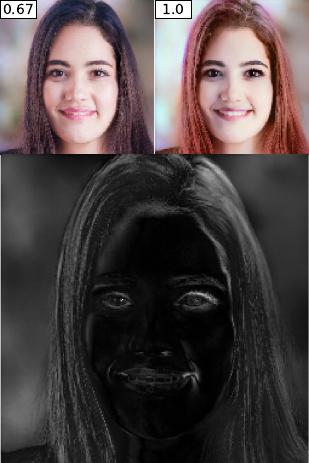} &
        \includegraphics[width=0.14\textwidth]{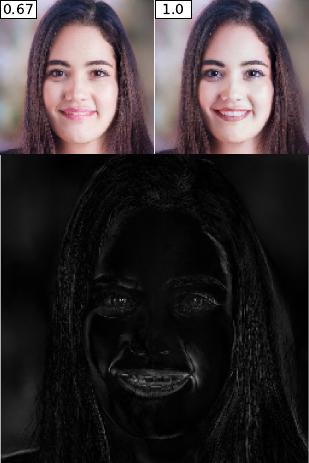} &
        \includegraphics[width=0.14\textwidth]{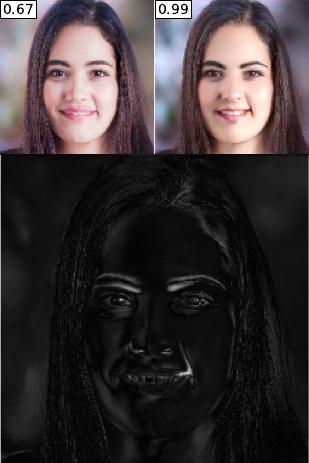} &
        \includegraphics[width=0.14\textwidth]{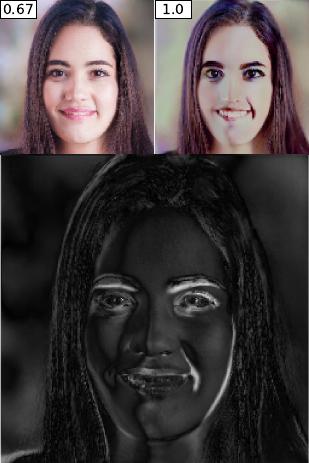} &
        \\
        Face color & Eye shape & Eye photometry & Hair photometry & Lip photometry & Mouth shape & Face boundary
    \end{tabular}
    \caption{Difference maps of counterfacts generated for the attractiveness classifier in Figure~\ref{fig:attractive}. See also Section~\ref{sec:exp-counterfactual}.}
    \label{fig:attractive-diff}
\end{figure*}

In Figure~\ref{fig:attractive-diff}, we visualize the difference maps corresponding to Figure~\ref{fig:attractive} to highlight the modification region. As mentioned in Section~\ref{sec:exp-counterfactual}, counterfacts in the subspaces of eye photometry, skin color and lip color express well-disentangled and localized manipulation. Note that we selected female samples whose original perceived attractiveness scores are near $0.5$. We find that counterfactuals for male samples or samples with very low attractiveness scores might not result in enhancement of human-perceived attractiveness. This is not surprising since the attractiveness is highly subjective measurement and difficult to represent by a simple classifier trained on very limited data.

\begin{figure*}
    \centering
    \footnotesize
    \setlength{\tabcolsep}{1pt}
    \begin{tabular}{cccccc}
        \includegraphics[width=0.19\textwidth]{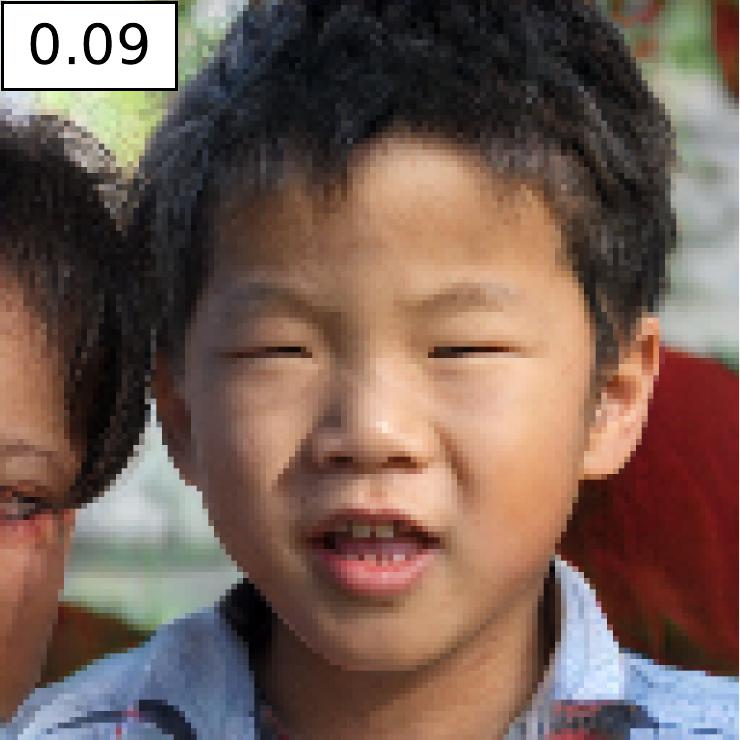} &
        \includegraphics[width=0.19\textwidth]{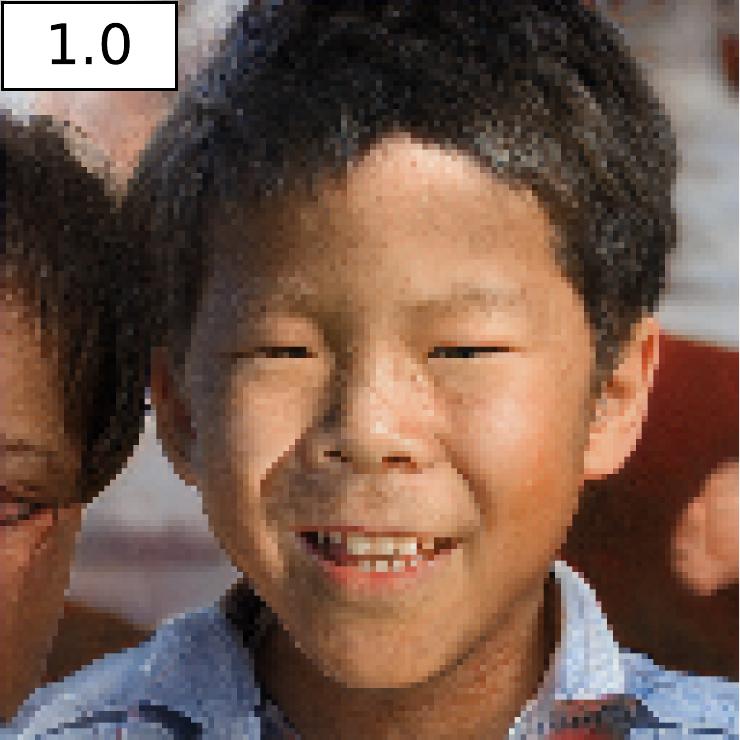} &
        \includegraphics[width=0.19\textwidth]{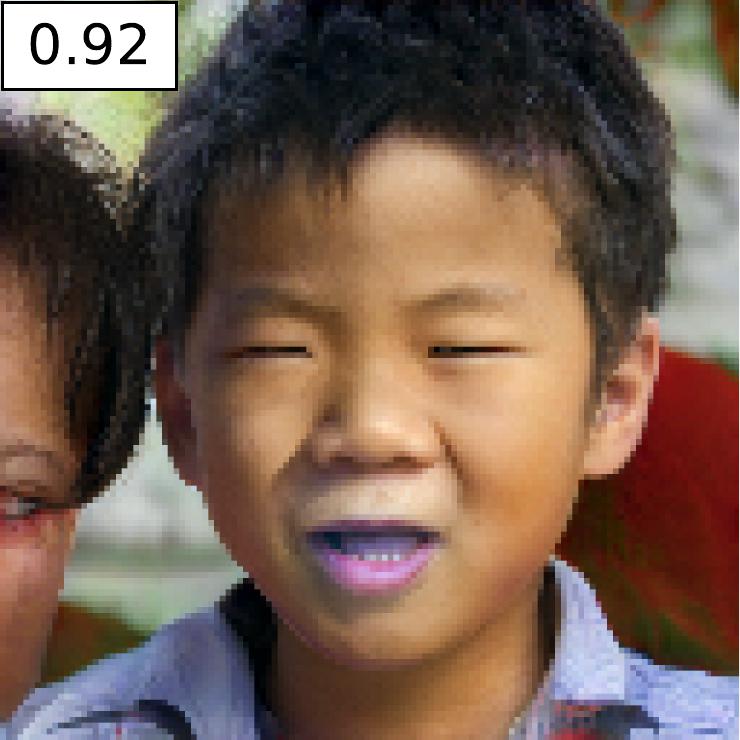} &
        \includegraphics[width=0.19\textwidth]{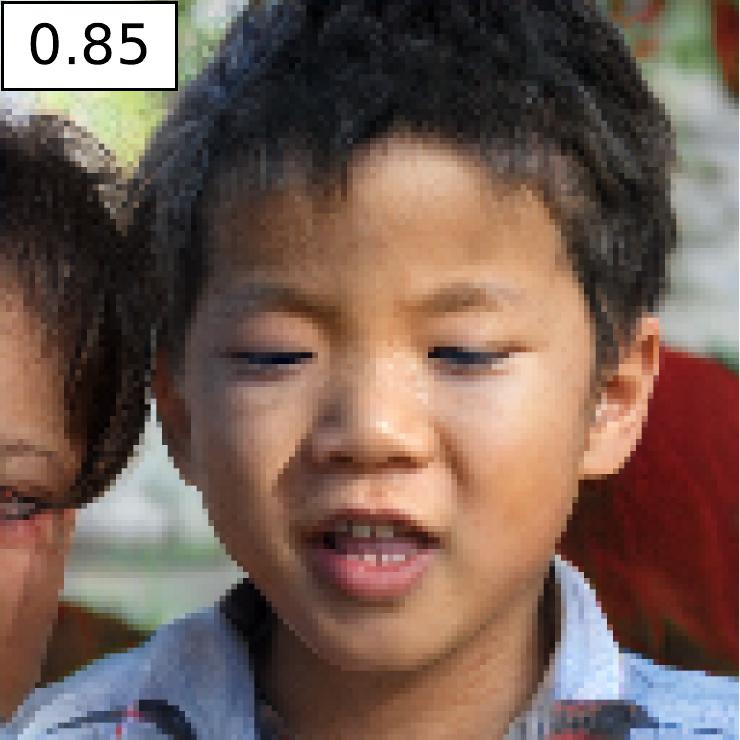} &
        \includegraphics[width=0.19\textwidth]{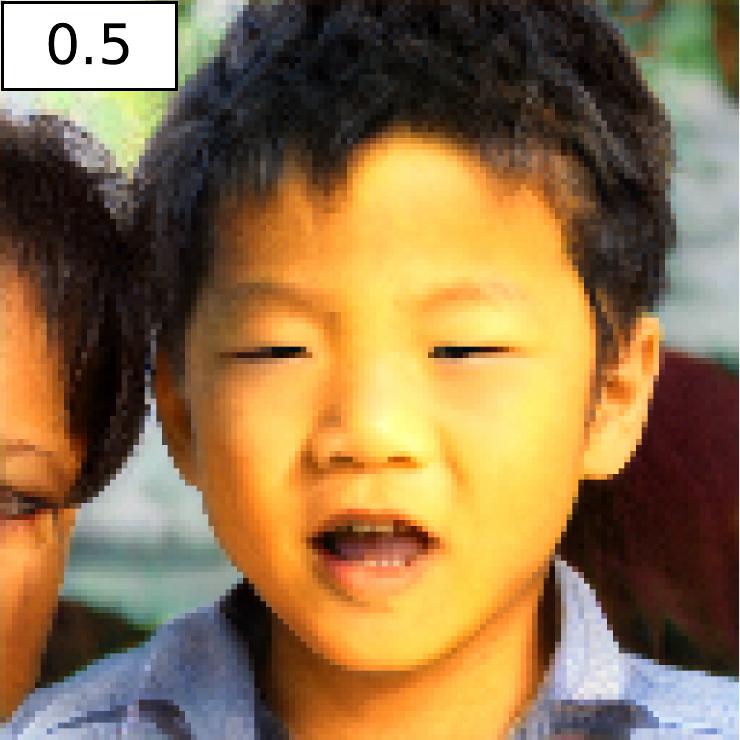} &
        \\
        \includegraphics[width=0.19\textwidth]{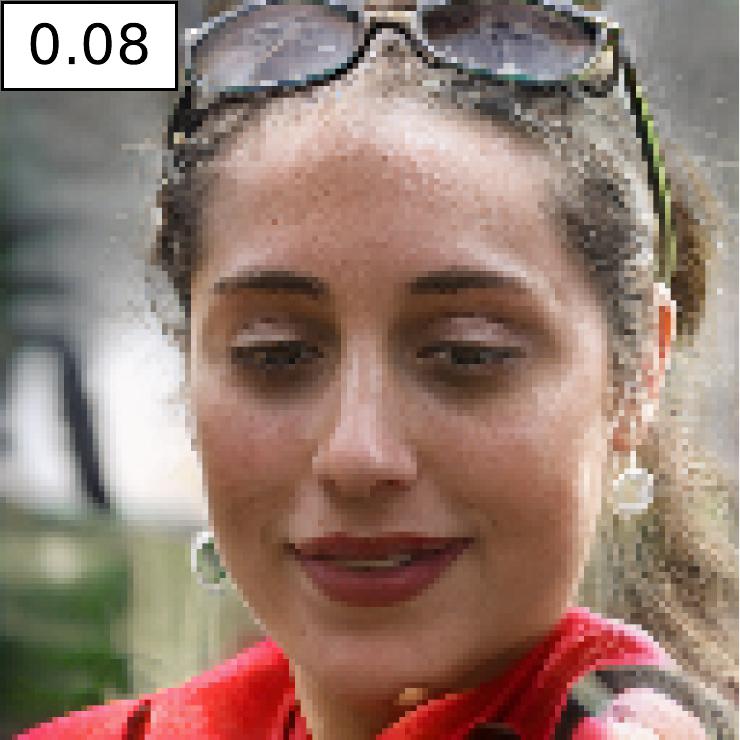} &
        \includegraphics[width=0.19\textwidth]{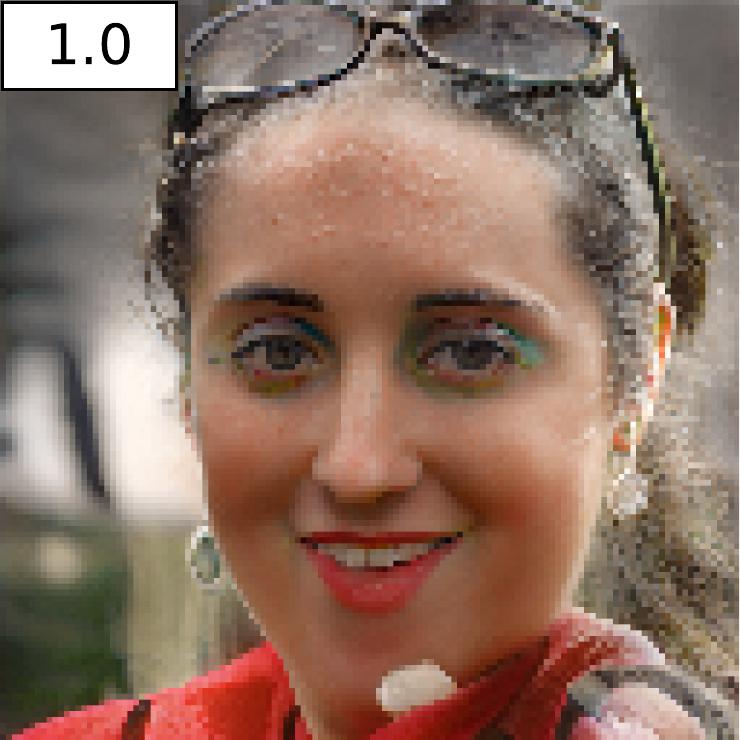} &
        \includegraphics[width=0.19\textwidth]{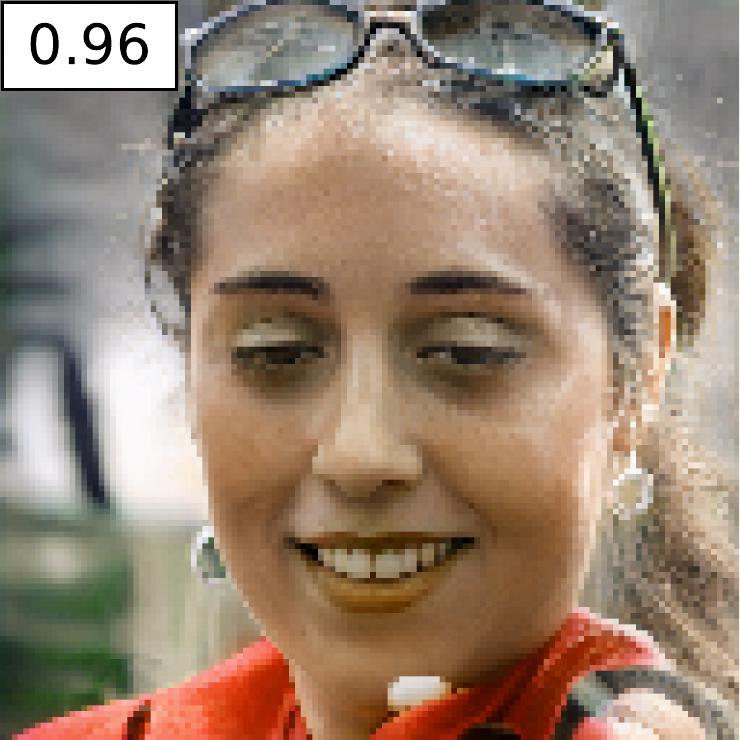} &
        \includegraphics[width=0.19\textwidth]{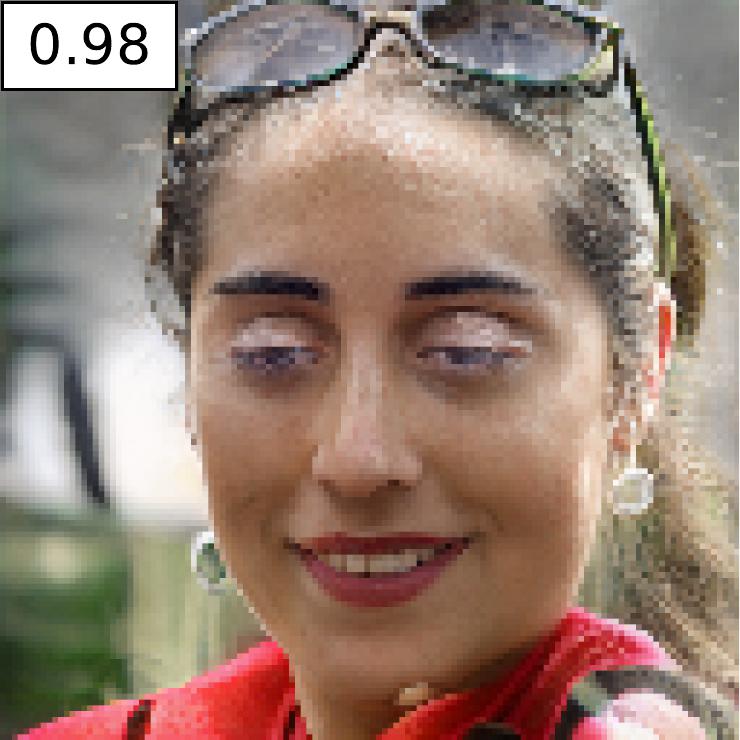} &
        \includegraphics[width=0.19\textwidth]{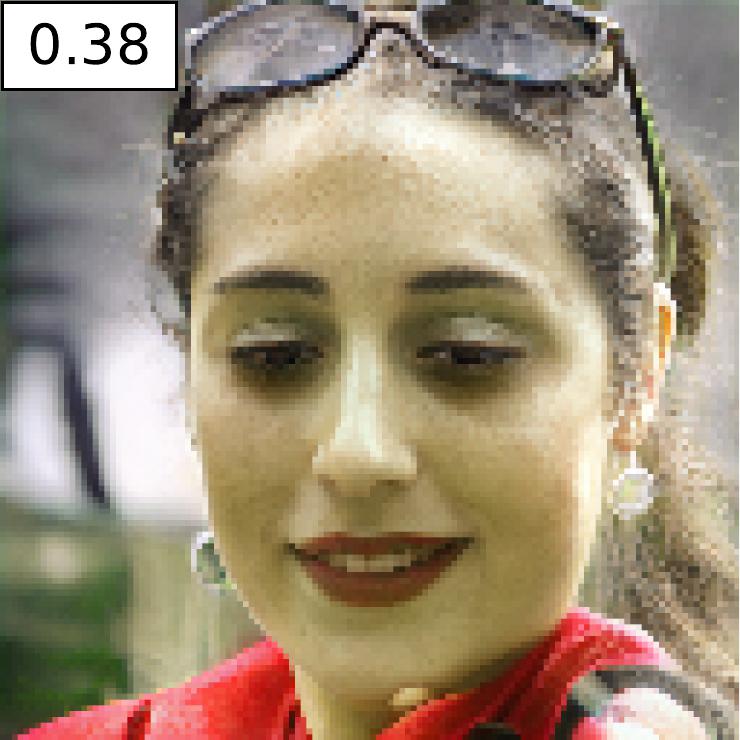} &
        \\
        \includegraphics[width=0.19\textwidth]{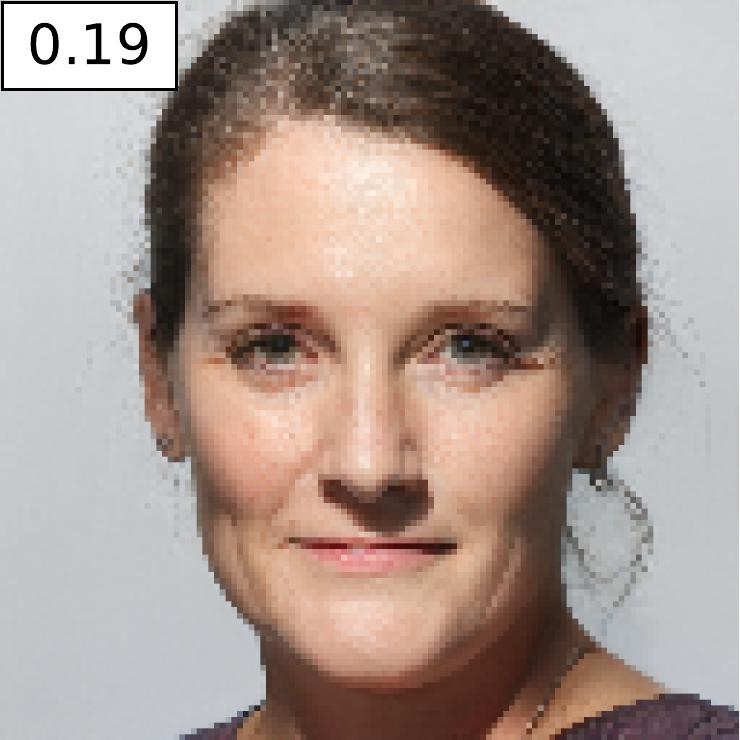} &
        \includegraphics[width=0.19\textwidth]{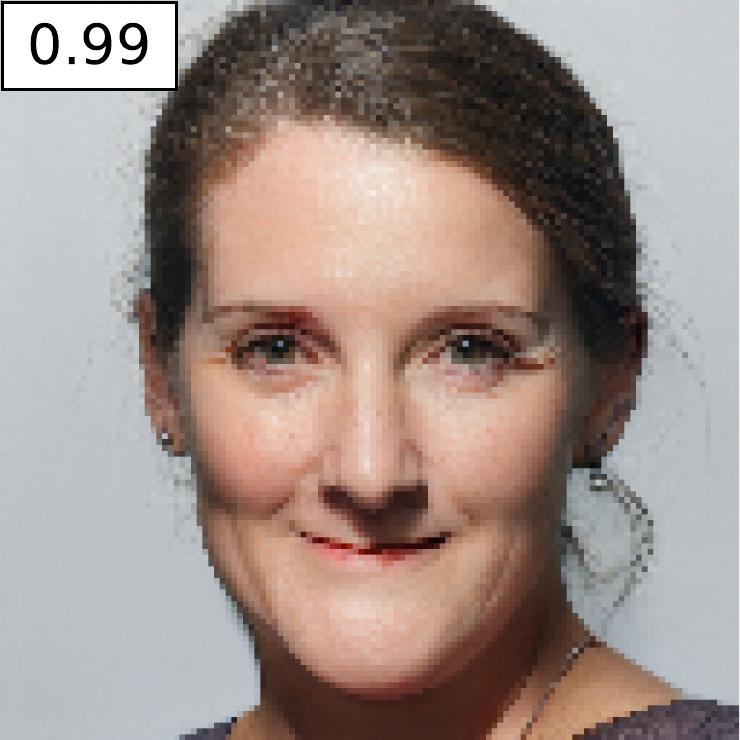} &
        \includegraphics[width=0.19\textwidth]{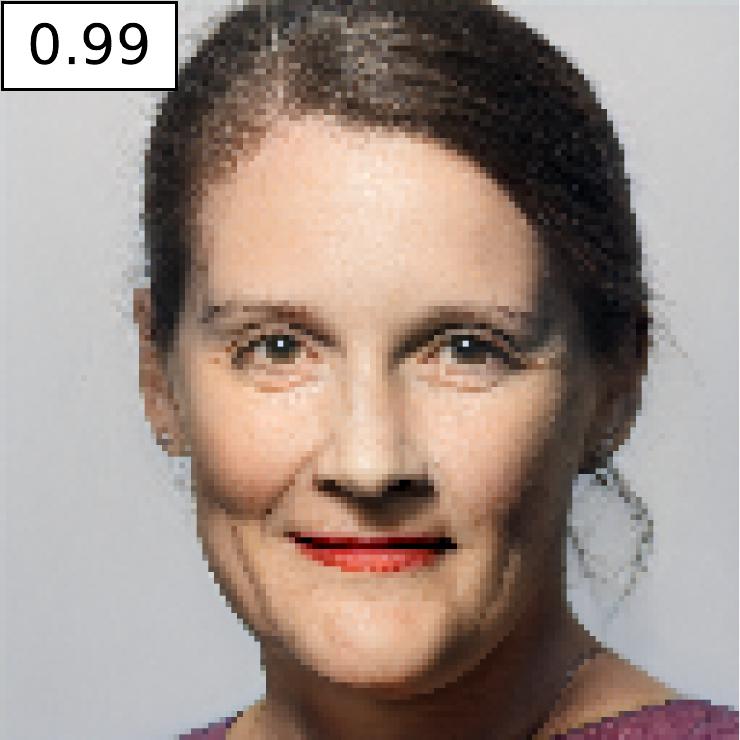} &
        \includegraphics[width=0.19\textwidth]{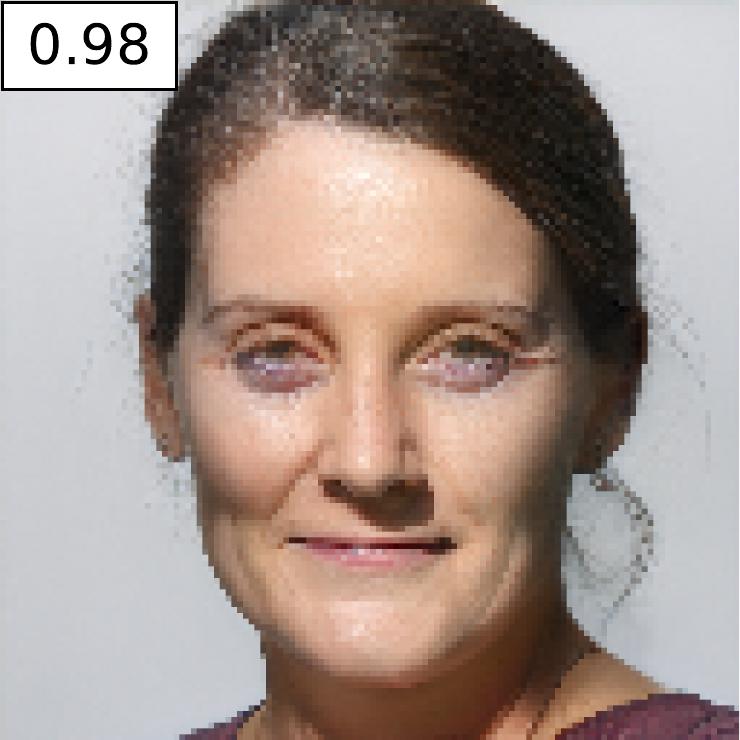} &
        \includegraphics[width=0.19\textwidth]{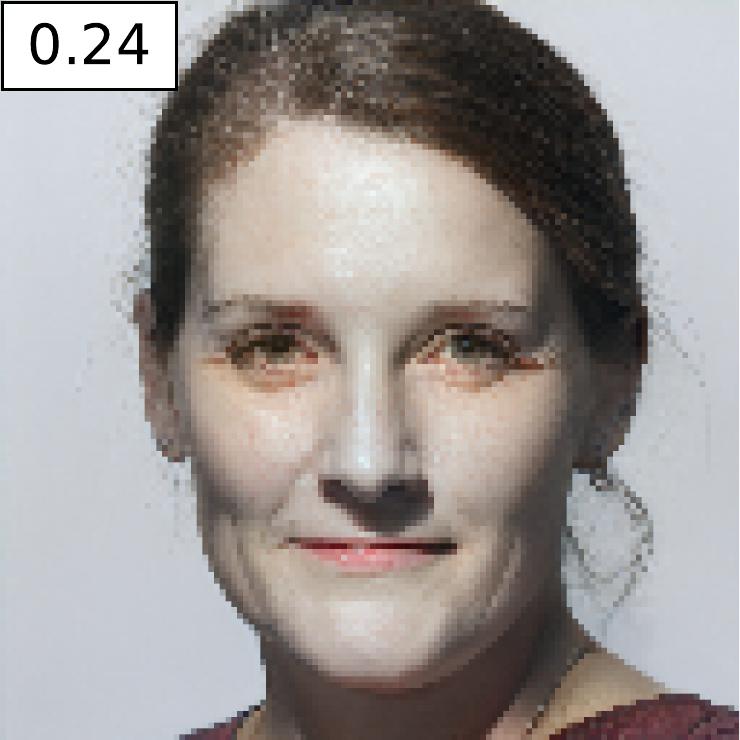} &
        \\
        \includegraphics[width=0.19\textwidth]{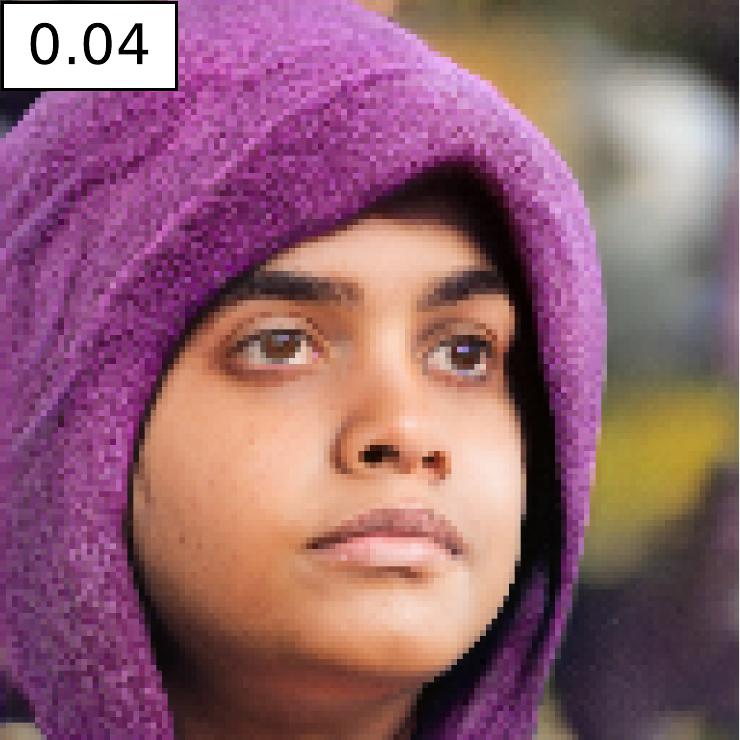} &
        \includegraphics[width=0.19\textwidth]{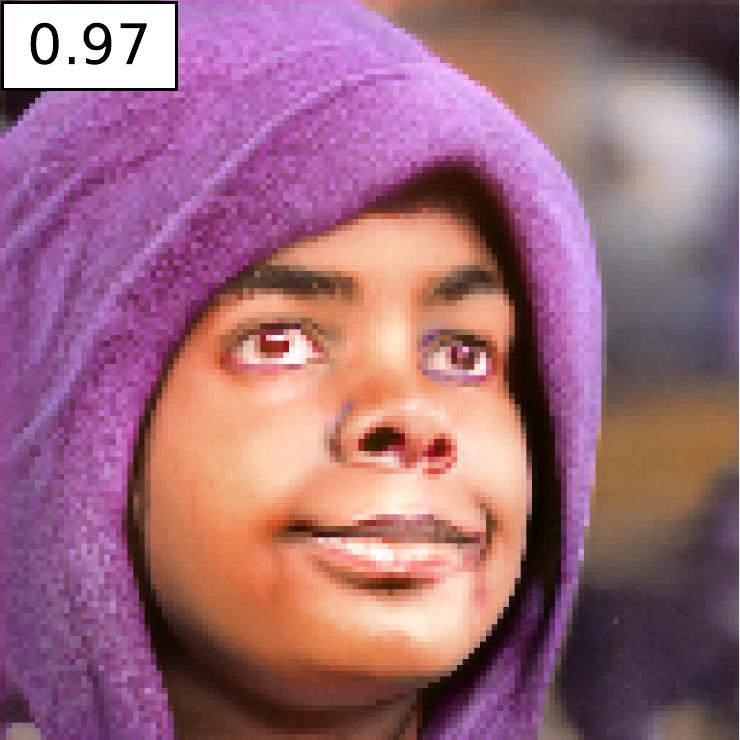} &
        \includegraphics[width=0.19\textwidth]{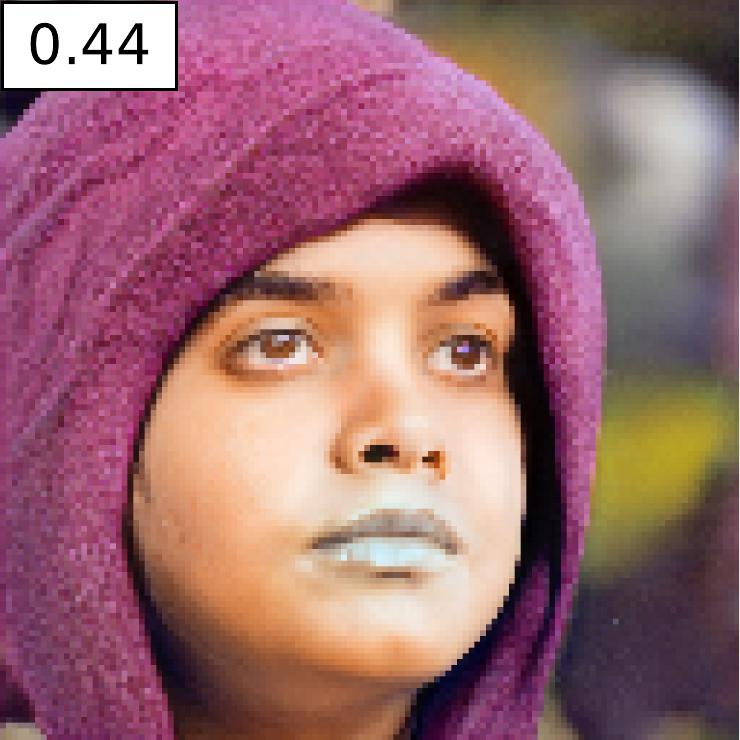} &
        \includegraphics[width=0.19\textwidth]{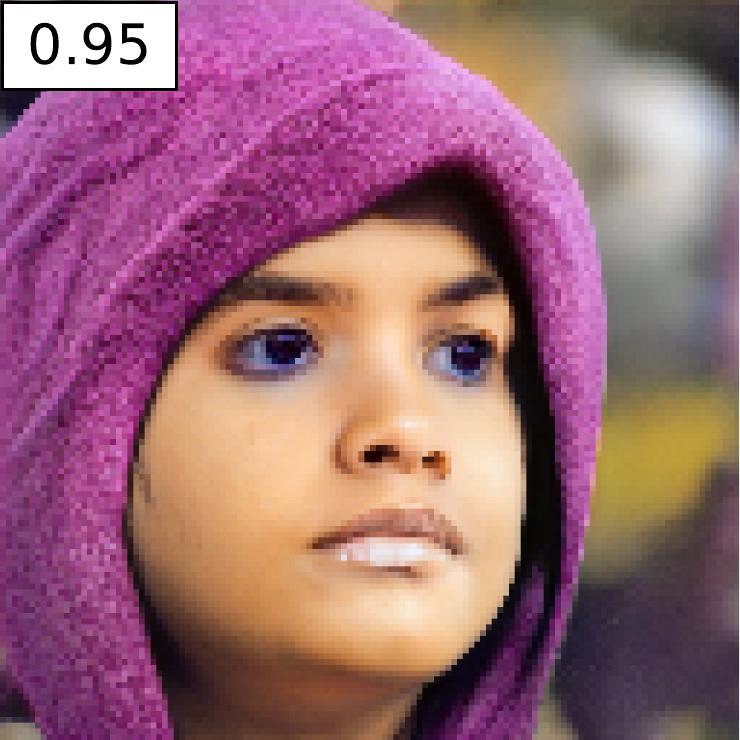} &
        \includegraphics[width=0.19\textwidth]{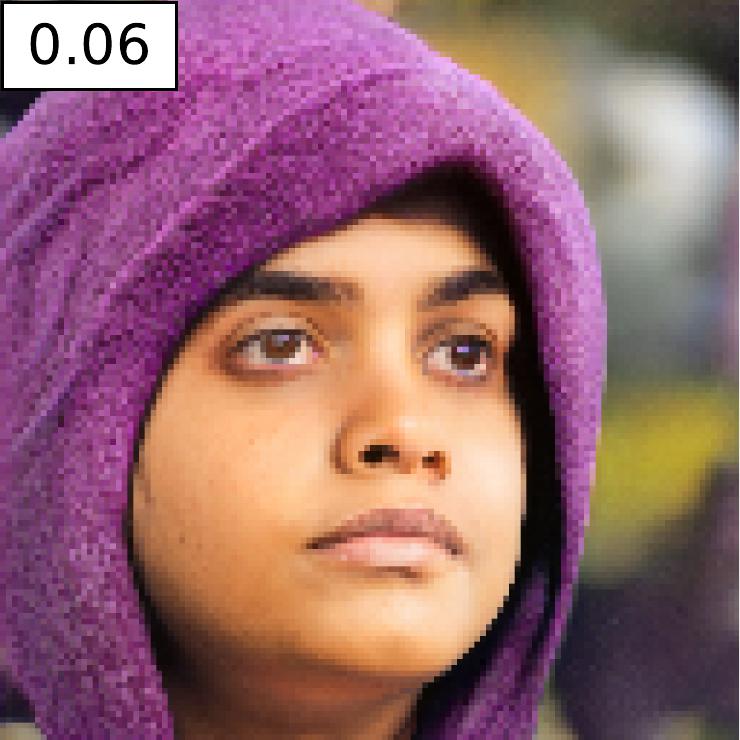} &
        \\
        Original image & Mouth shape & Lip photometry & Eye photometry & Face color
    \end{tabular}
    \caption{Counterfacts generated in interpretable subspaces to increase predicted score of the smile classifier. See \Cref{sec:exp-counterfactual-materials} for details. Corresponding difference maps can be found in \Cref{fig:smiling-diff} in the appendix.}
    \label{fig:smiling}
\end{figure*}

In Figure~\ref{fig:smiling}, we visualize the counterfacts generated to increase the smiling score of a given classifier. We select the subspaces of mouth shape, lip color, eye photometry, and face color. As can be noticed, mouth shape is closely correlated with the smiling score and the generated counterfacts align with human cognition. The eye photometry subspace is also shown to be correlated with prediction score. However, only the 3rd row shows noticeable laughter lines on eyes. The rest counterfacts do not provide explainable modification. We suspect this is related to the fact that the classifier does not exploit the eye clue well in prediction. Skin color subspace does not produce valid couterfacts, which indicates the prediction is uncorrelated with this subspace.

\begin{figure*}
    \centering
    \footnotesize
    \setlength{\tabcolsep}{1pt}
    \begin{tabular}{cccccc}
        \includegraphics[width=0.19\textwidth]{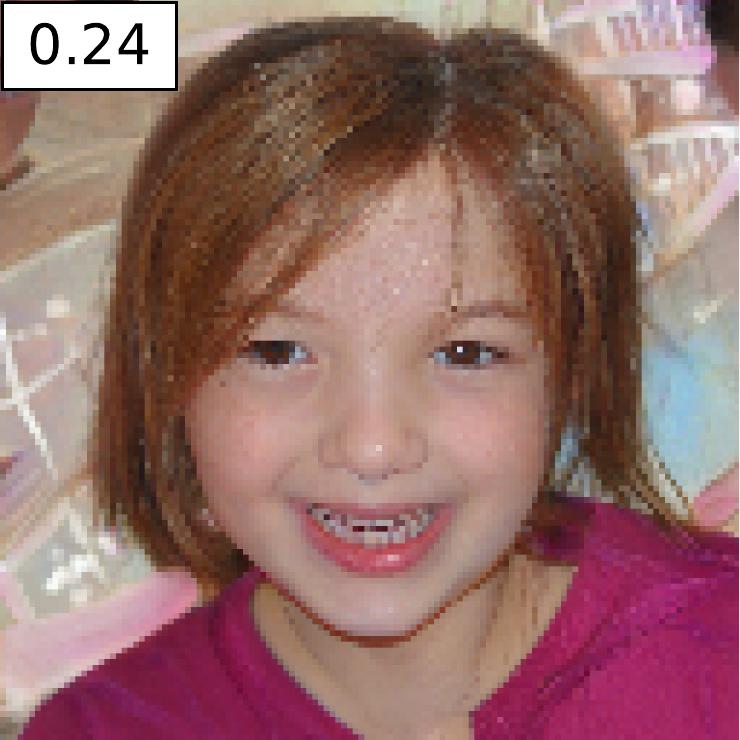} &
        \includegraphics[width=0.19\textwidth]{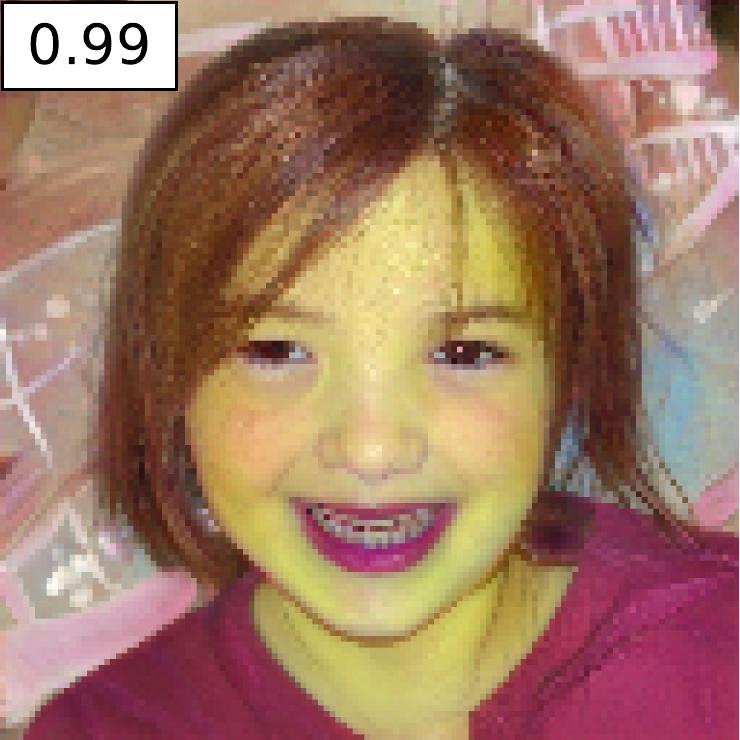} &
        \includegraphics[width=0.19\textwidth]{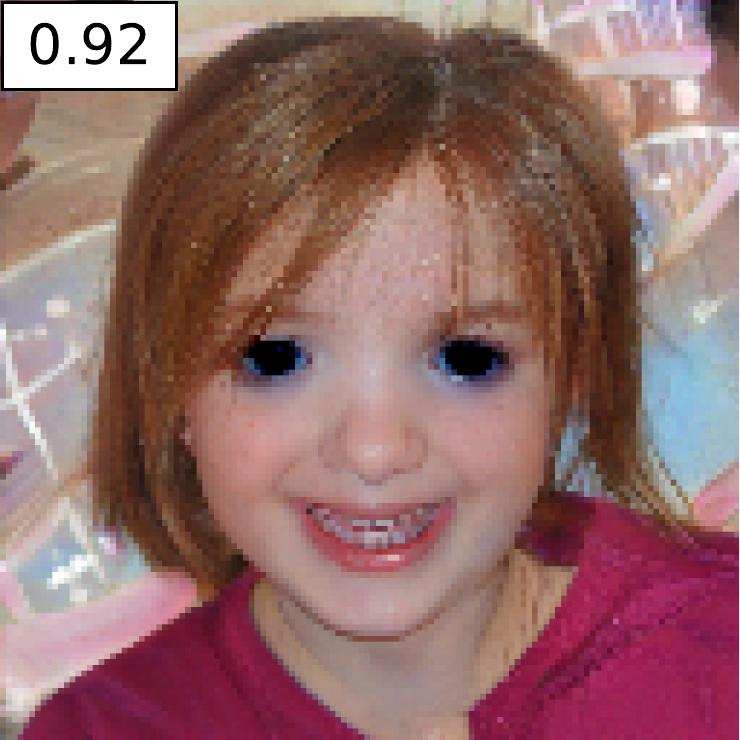} &
        \includegraphics[width=0.19\textwidth]{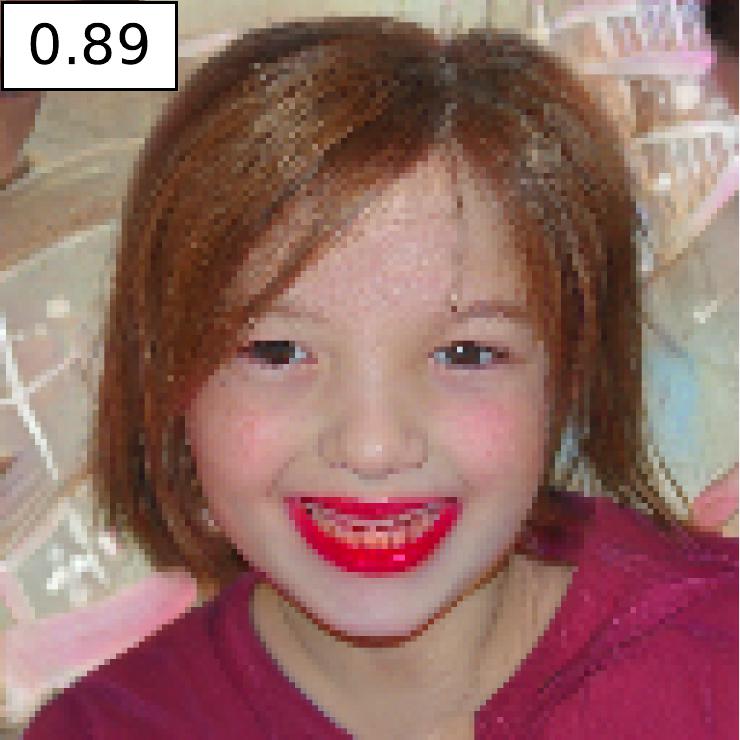} &
        \includegraphics[width=0.19\textwidth]{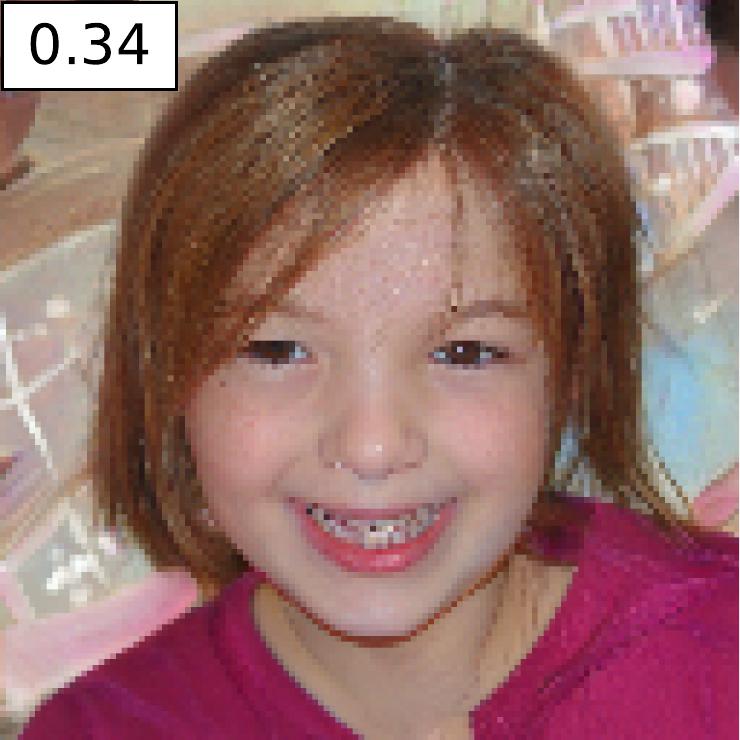} &
        \\
        \includegraphics[width=0.19\textwidth]{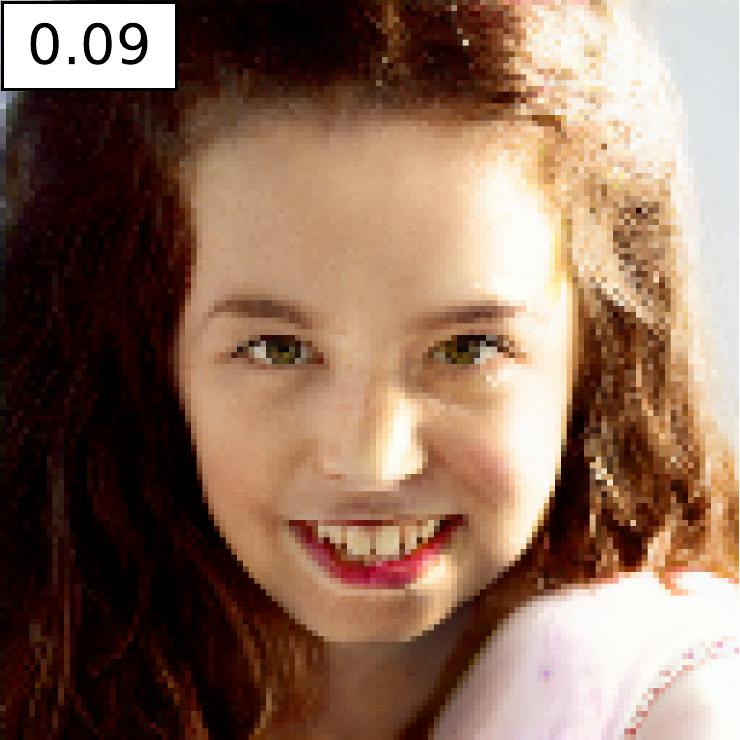} &
        \includegraphics[width=0.19\textwidth]{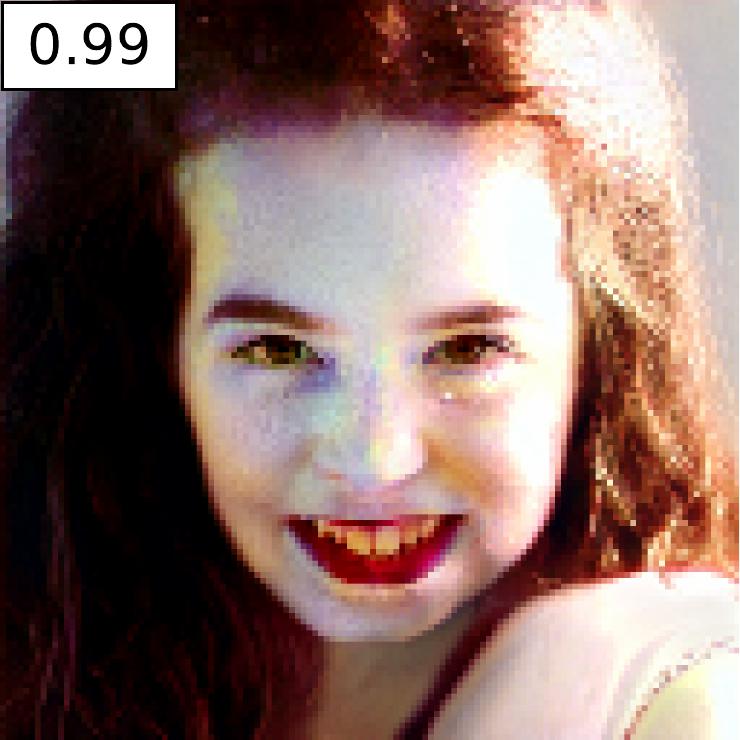} &
        \includegraphics[width=0.19\textwidth]{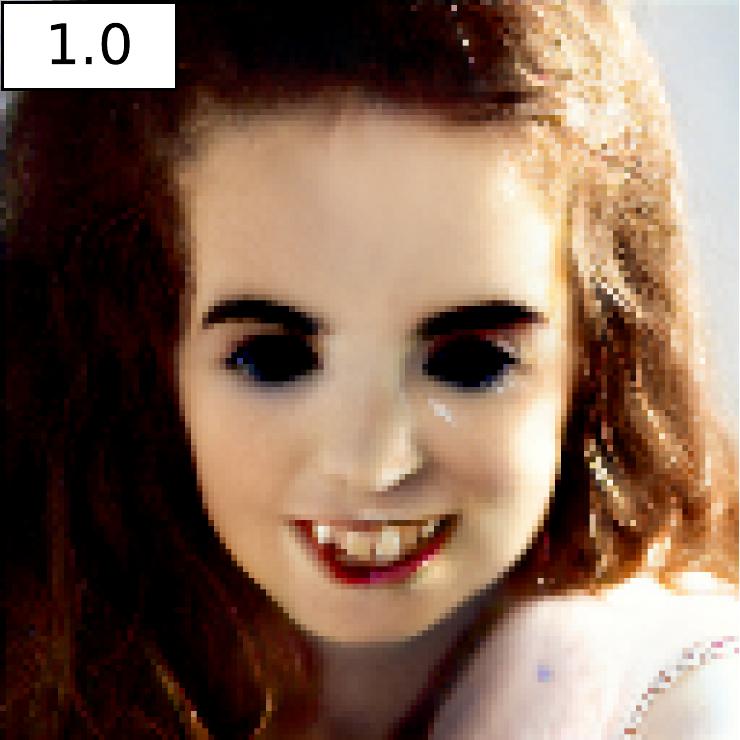} &
        \includegraphics[width=0.19\textwidth]{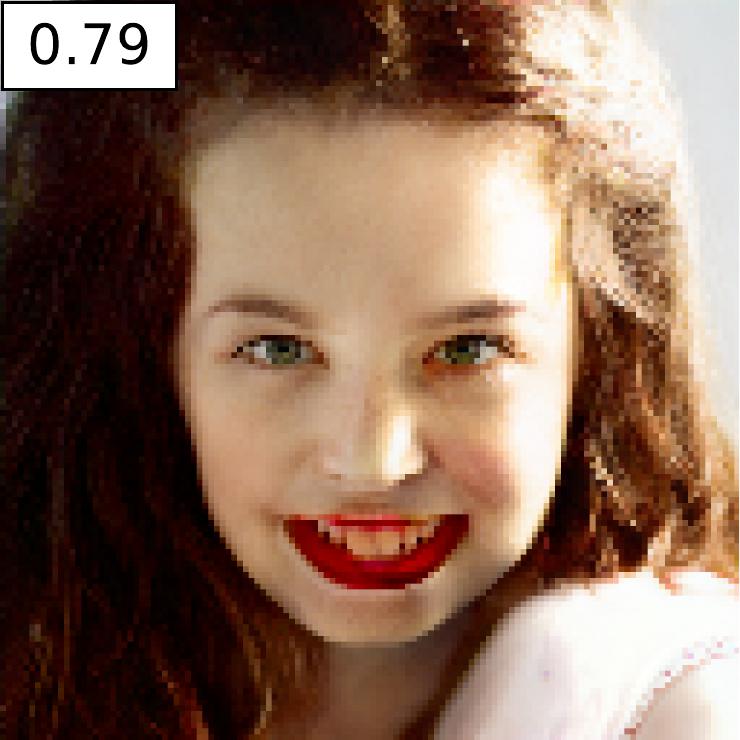} &
        \includegraphics[width=0.19\textwidth]{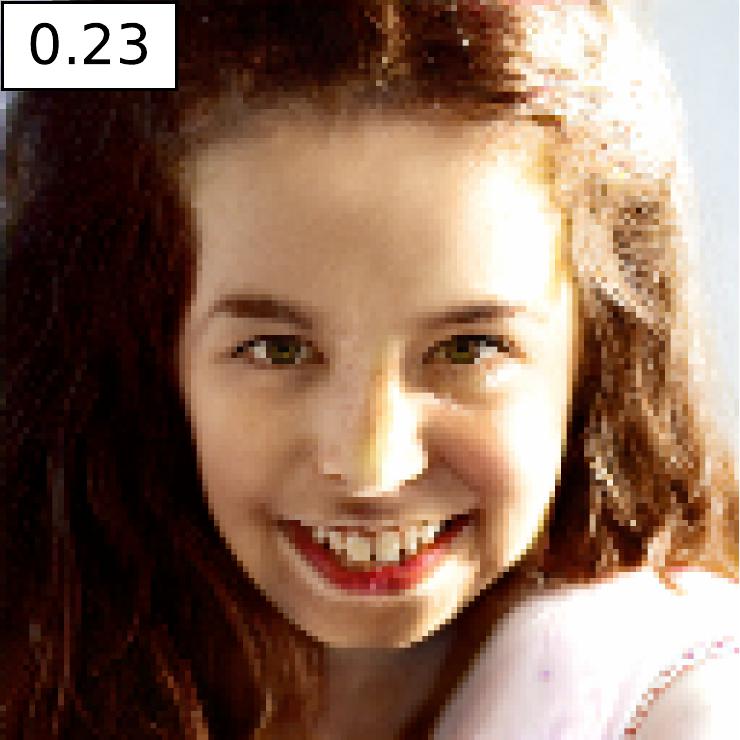} &
        \\
        \includegraphics[width=0.19\textwidth]{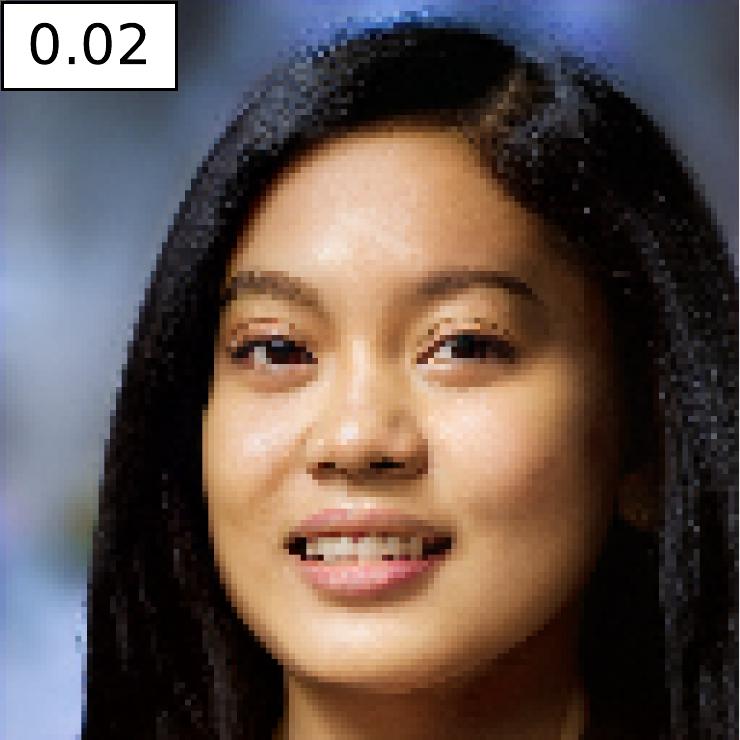} &
        \includegraphics[width=0.19\textwidth]{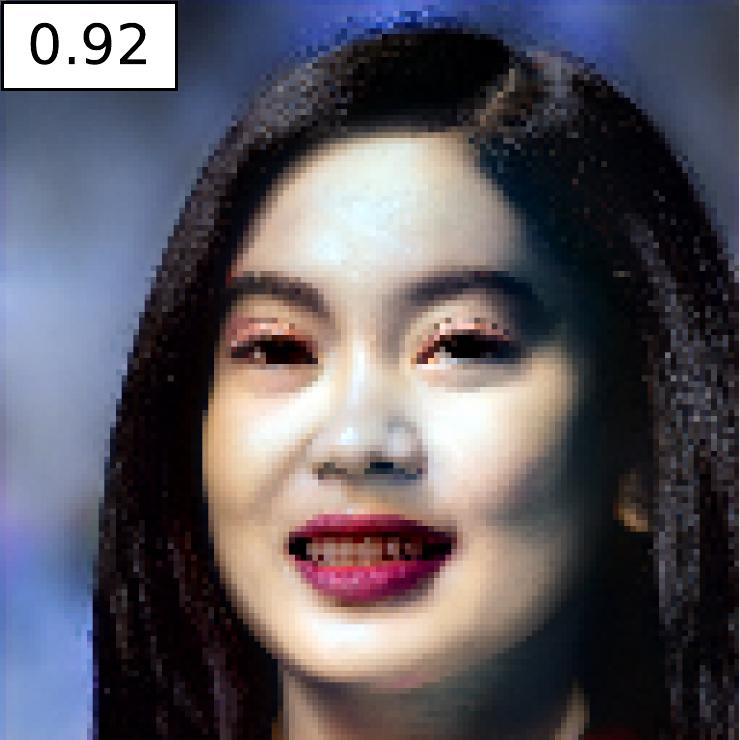} &
        \includegraphics[width=0.19\textwidth]{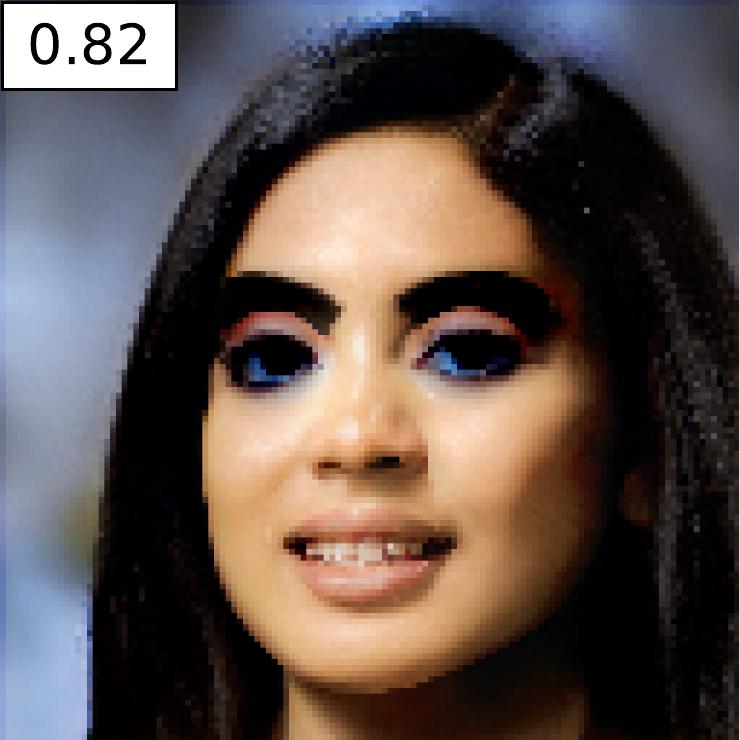} &
        \includegraphics[width=0.19\textwidth]{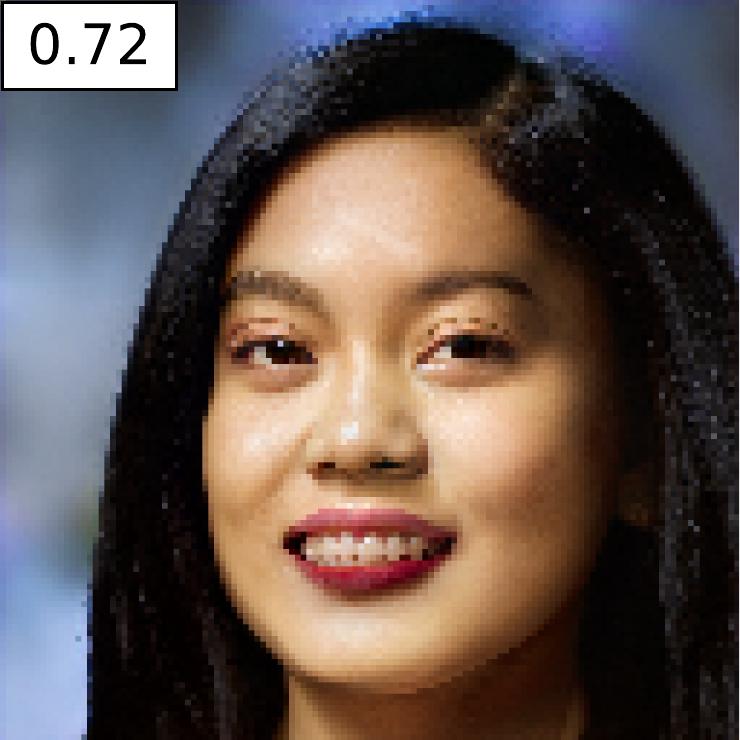} &
        \includegraphics[width=0.19\textwidth]{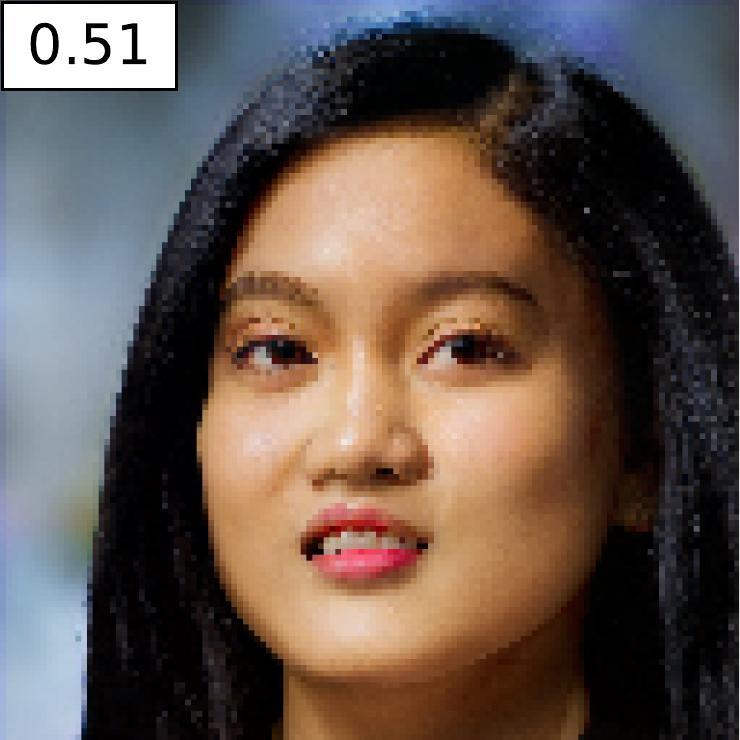} &
        \\
        \includegraphics[width=0.19\textwidth]{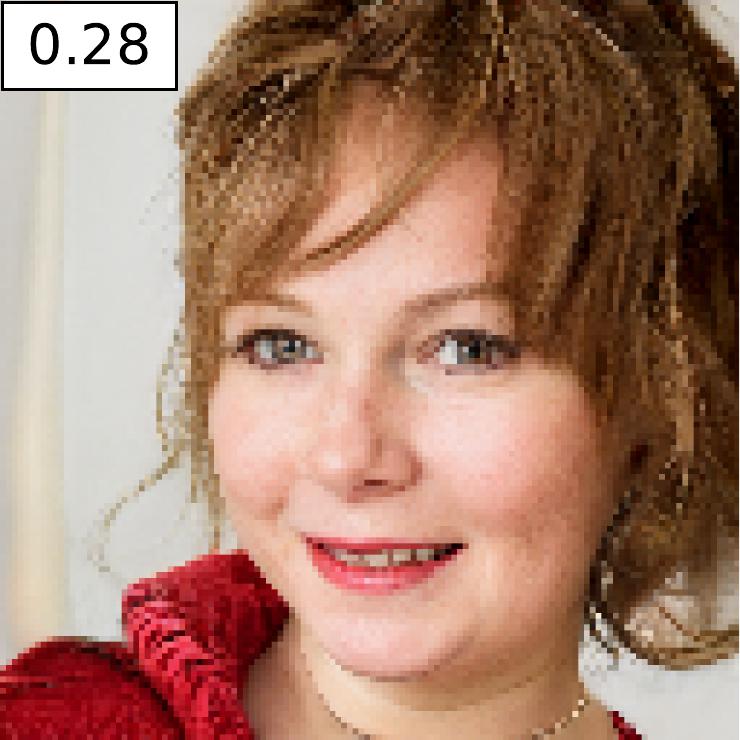} &
        \includegraphics[width=0.19\textwidth]{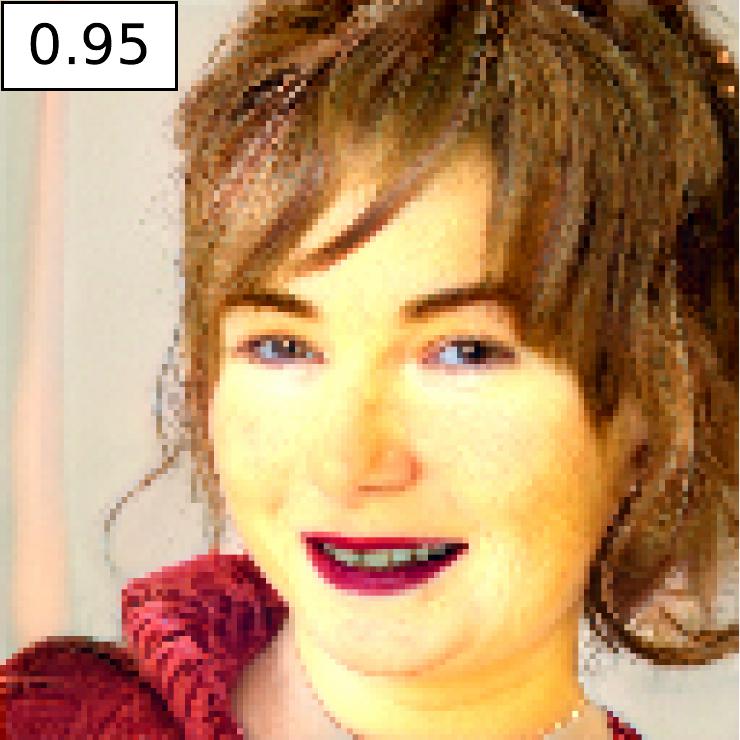} &
        \includegraphics[width=0.19\textwidth]{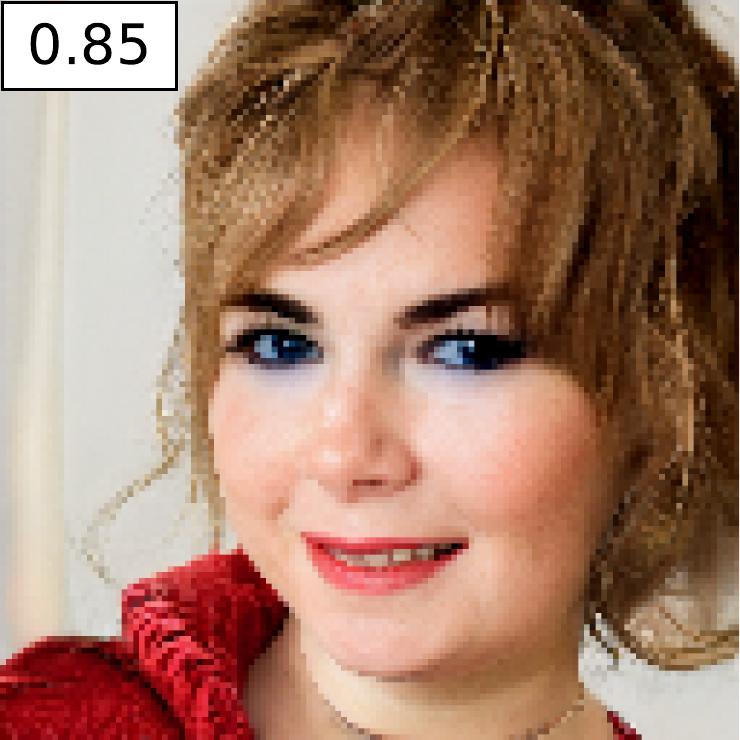} &
        \includegraphics[width=0.19\textwidth]{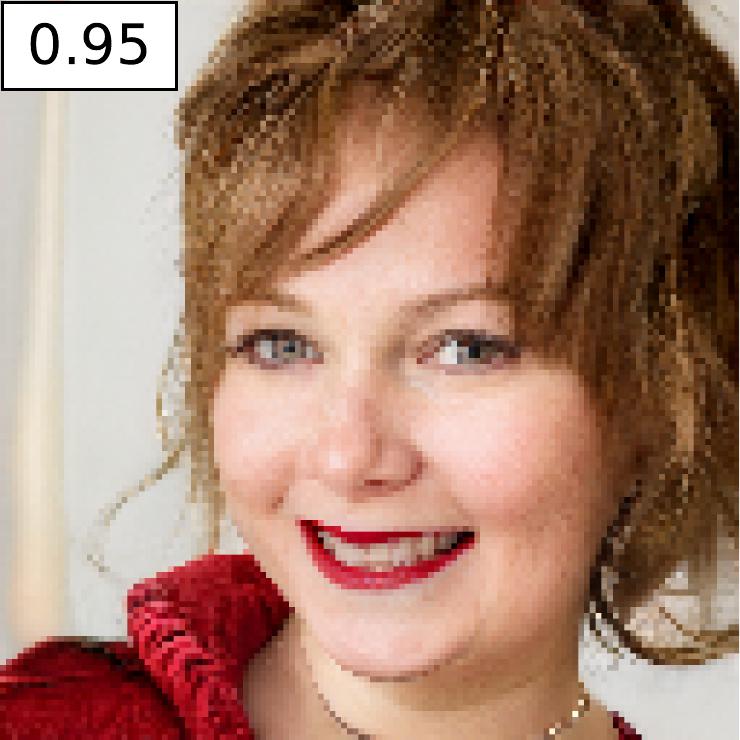} &
        \includegraphics[width=0.19\textwidth]{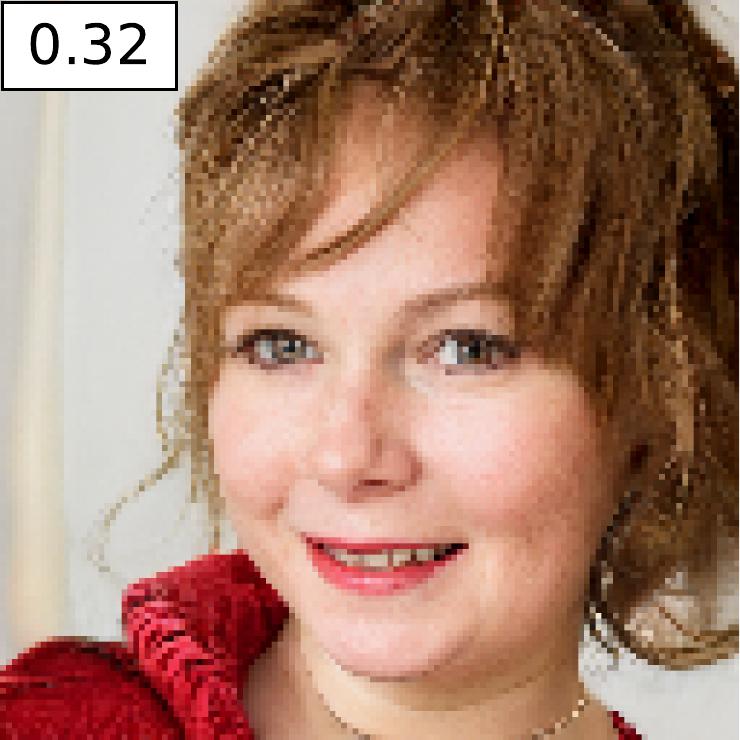} &
        \\
        Original image & Face color & Eye photometry & Lip photometry & Mouth shape
    \end{tabular}
    \caption{Counterfacts generated in interpretable subspaces to increase predicted score of the heavy make-upclassifier. See \Cref{sec:exp-counterfactual-materials} for details. Corresponding difference maps can be found in \Cref{fig:heavy_makeup-diff} in the appendix.}
    \label{fig:heavy_makeup}
\end{figure*}

In Figure~\ref{fig:heavy_makeup}, we show the counterfacts generated for the classifier of heavy make-up. Counterfacts with modification in eye photometry, lip color and mouth photometry implies that the classifier capture explainable semantics in the corresponding subspaces. However, the counterfacts in skin color subspace is puzzling, which indicates that the concept learned by this classifier is not well aligned with human perception.

\begin{figure*}
    \centering
    \footnotesize
    \setlength{\tabcolsep}{1pt}
    \begin{tabular}{cccccc}
        \includegraphics[width=0.19\textwidth]{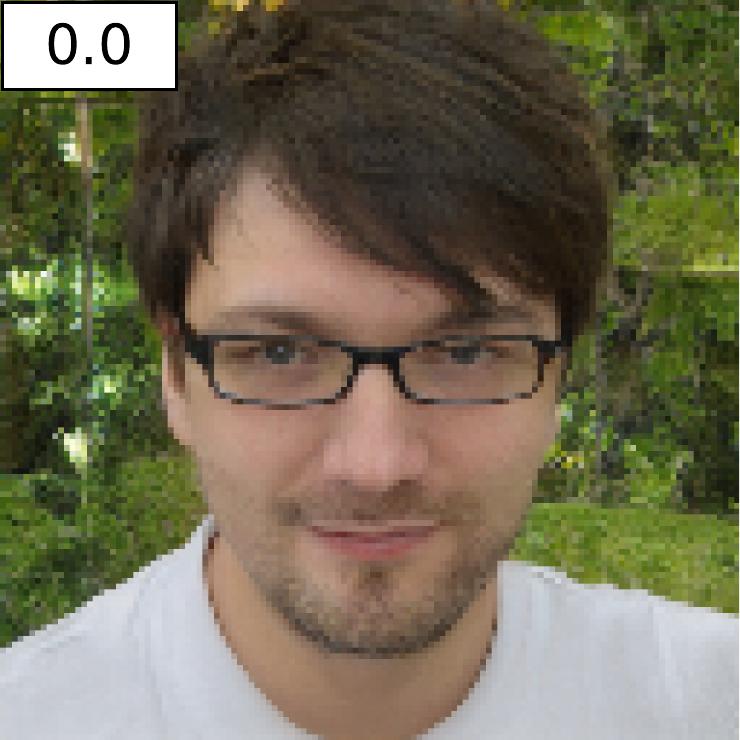} &
        \includegraphics[width=0.19\textwidth]{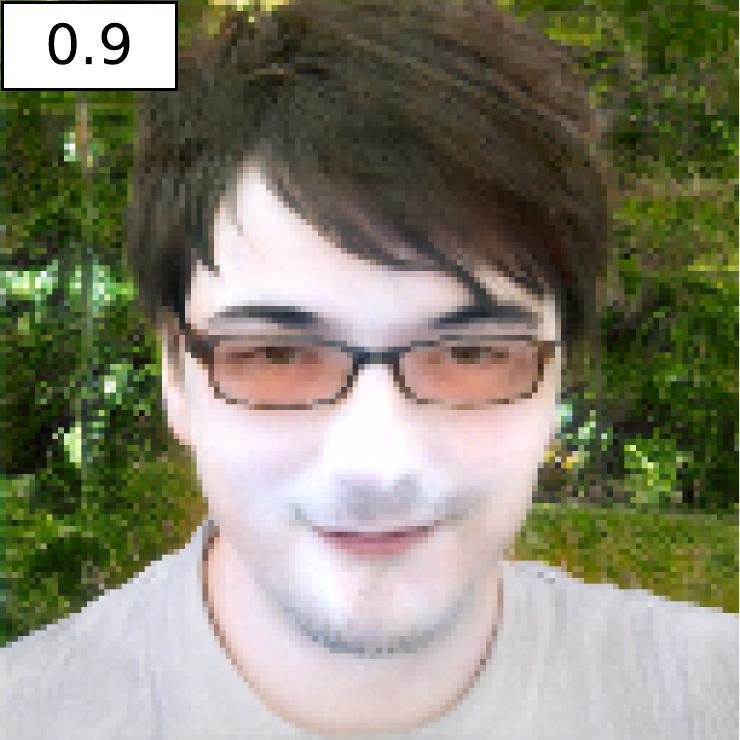} &
        \includegraphics[width=0.19\textwidth]{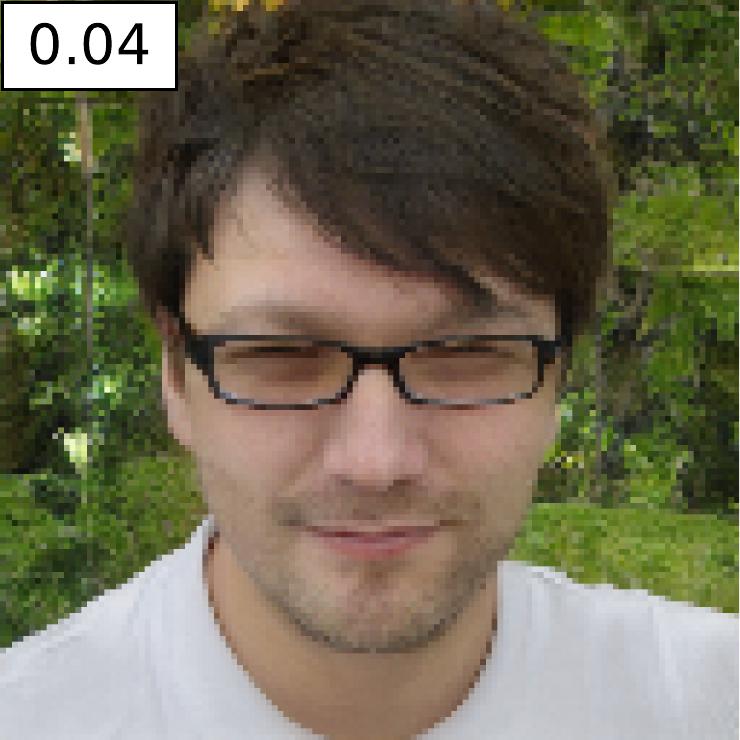} &
        \includegraphics[width=0.19\textwidth]{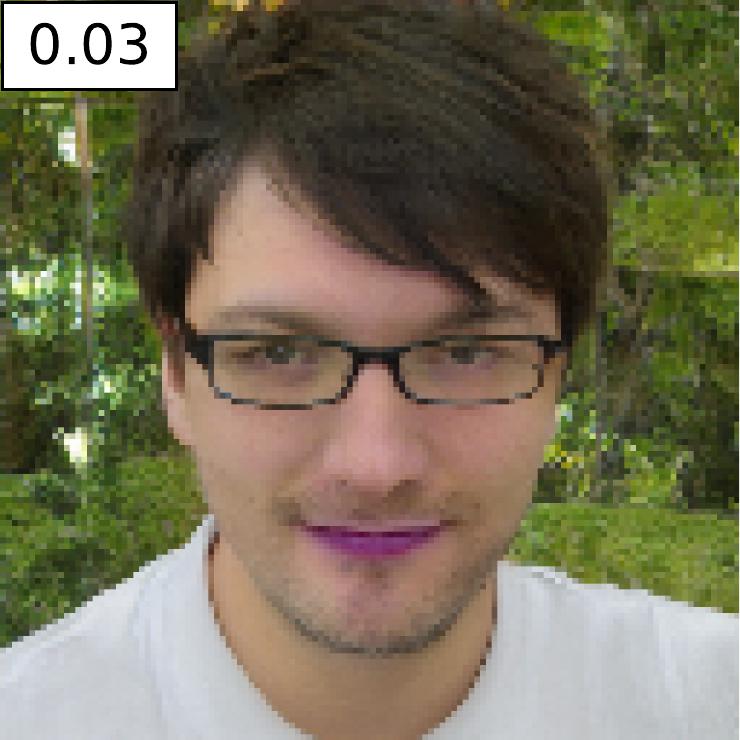} &
        \includegraphics[width=0.19\textwidth]{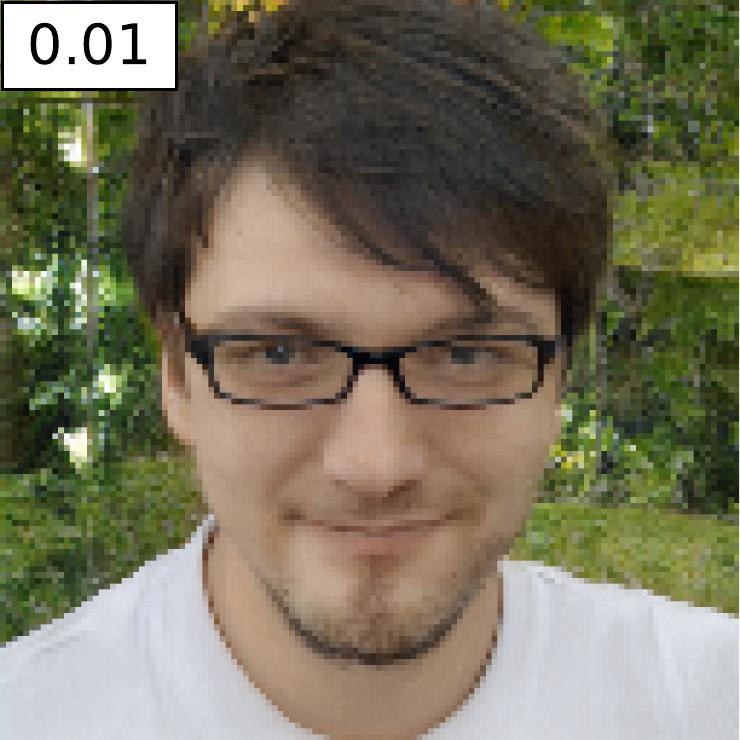} &
        \\
        \includegraphics[width=0.19\textwidth]{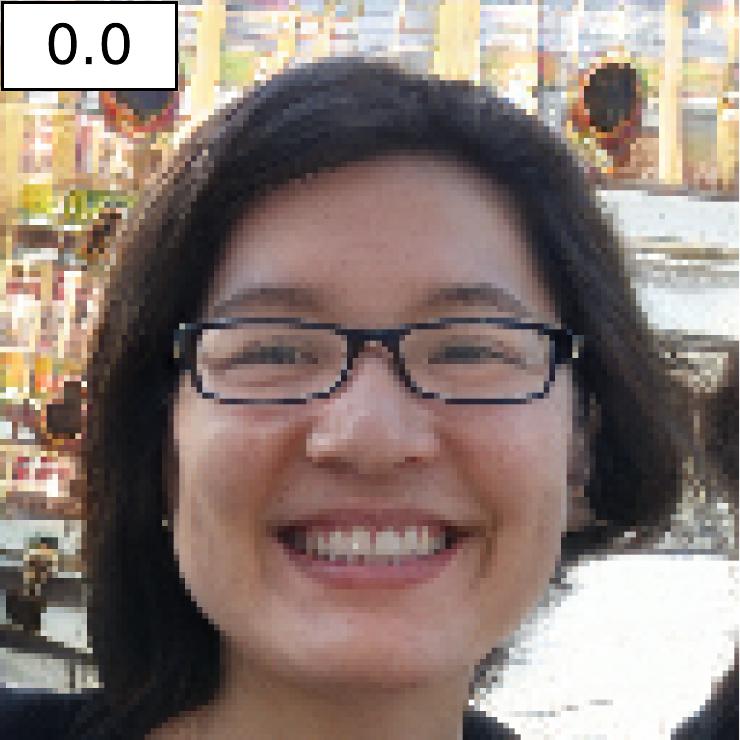} &
        \includegraphics[width=0.19\textwidth]{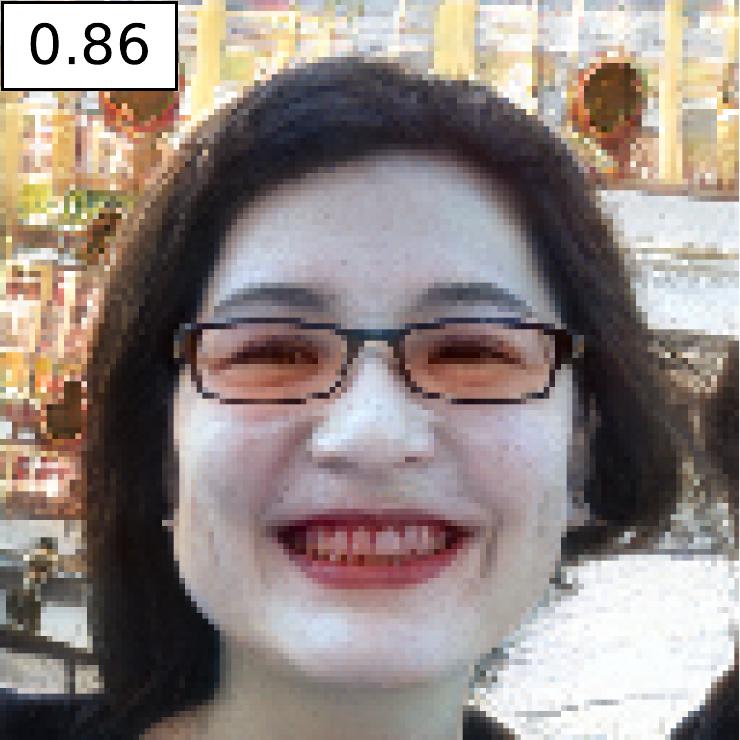} &
        \includegraphics[width=0.19\textwidth]{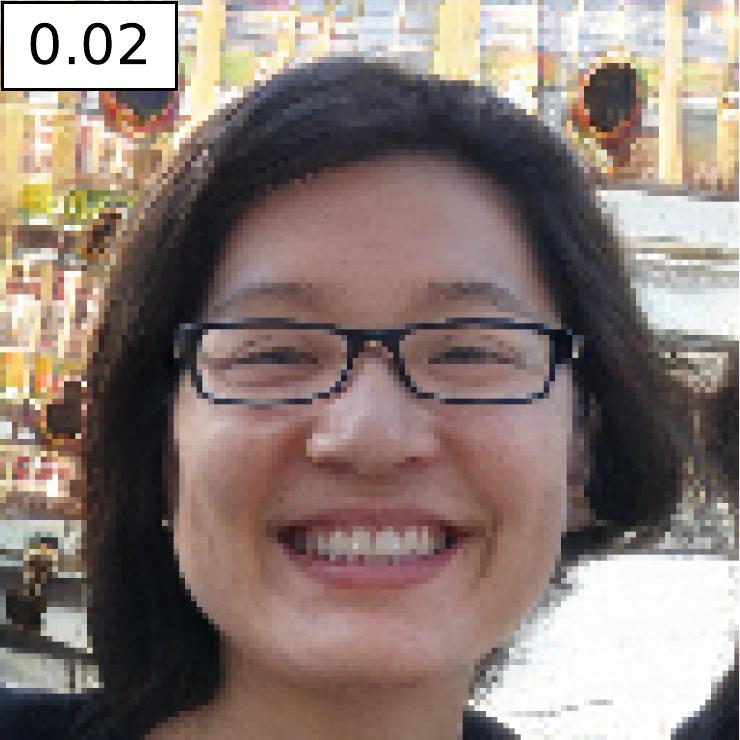} &
        \includegraphics[width=0.19\textwidth]{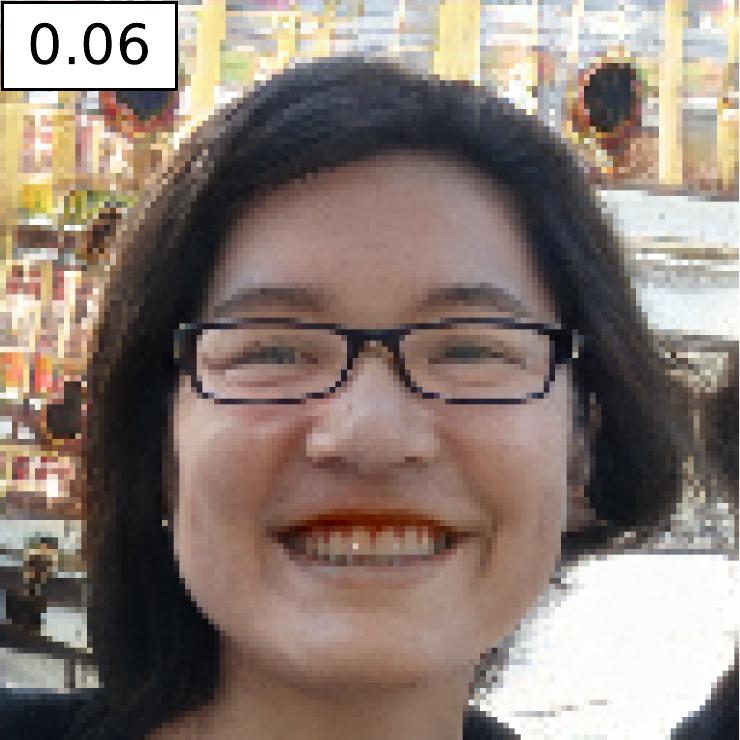} &
        \includegraphics[width=0.19\textwidth]{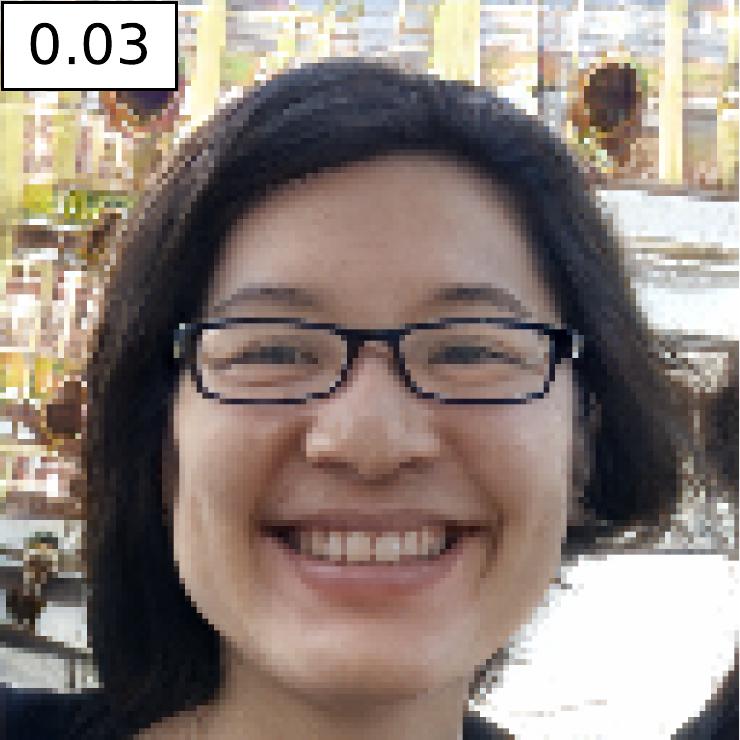} &
        \\
        \includegraphics[width=0.19\textwidth]{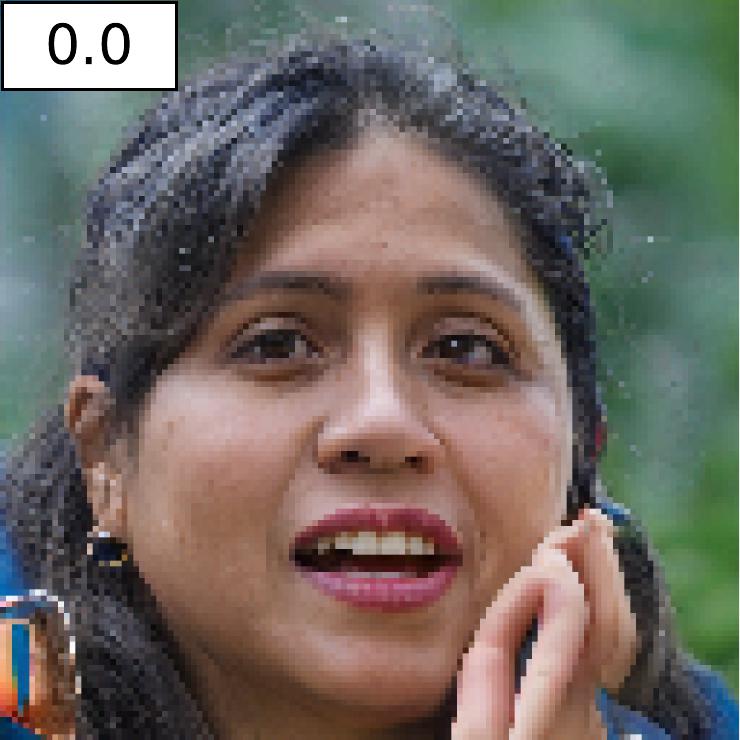} &
        \includegraphics[width=0.19\textwidth]{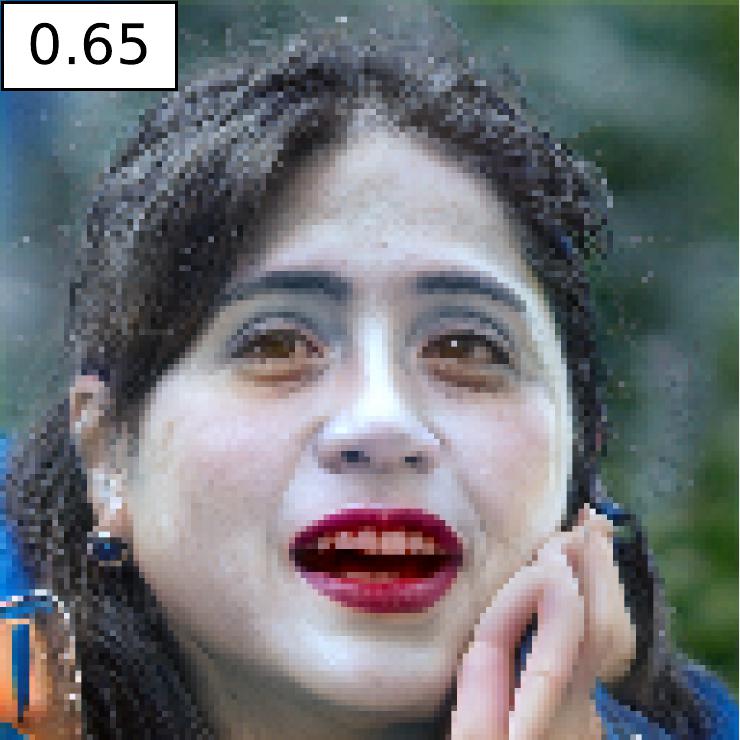} &
        \includegraphics[width=0.19\textwidth]{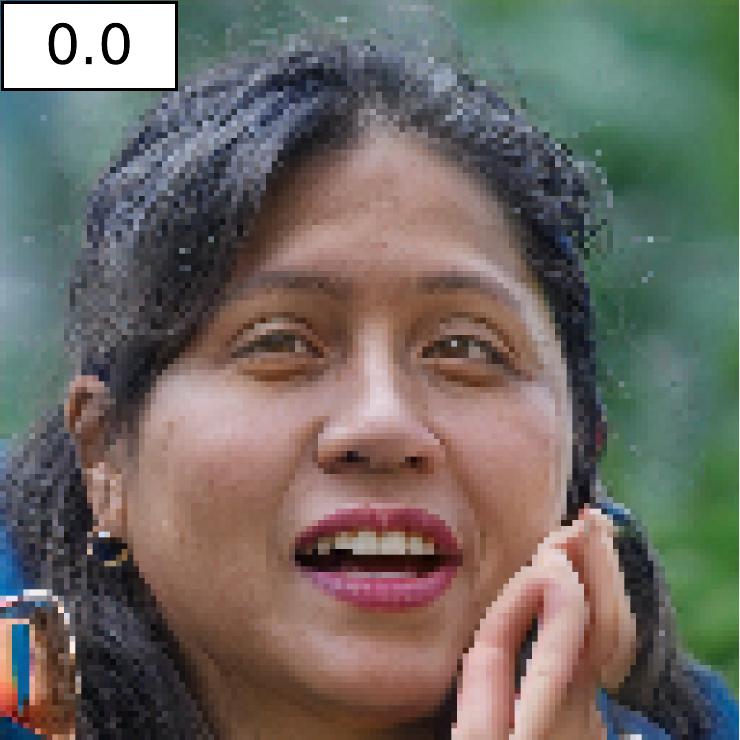} &
        \includegraphics[width=0.19\textwidth]{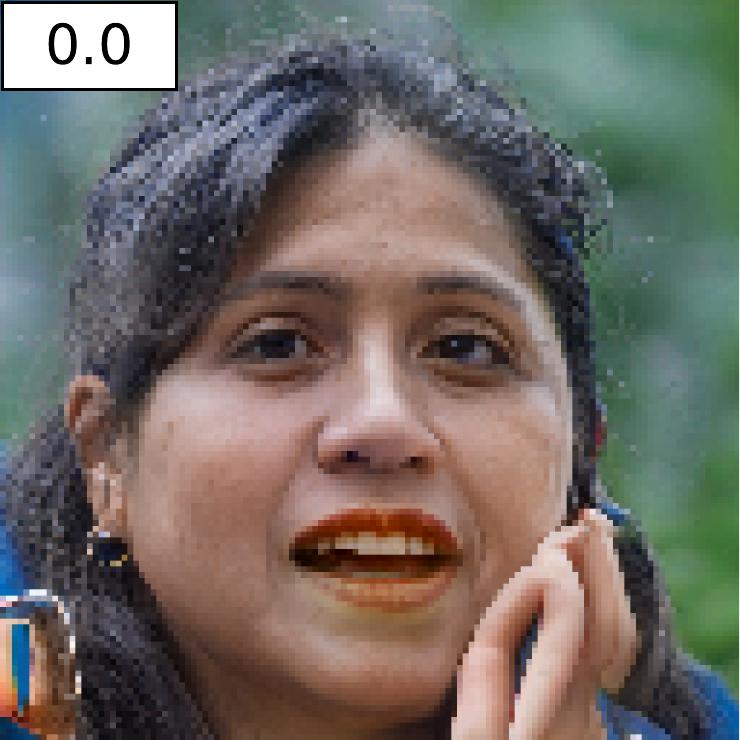} &
        \includegraphics[width=0.19\textwidth]{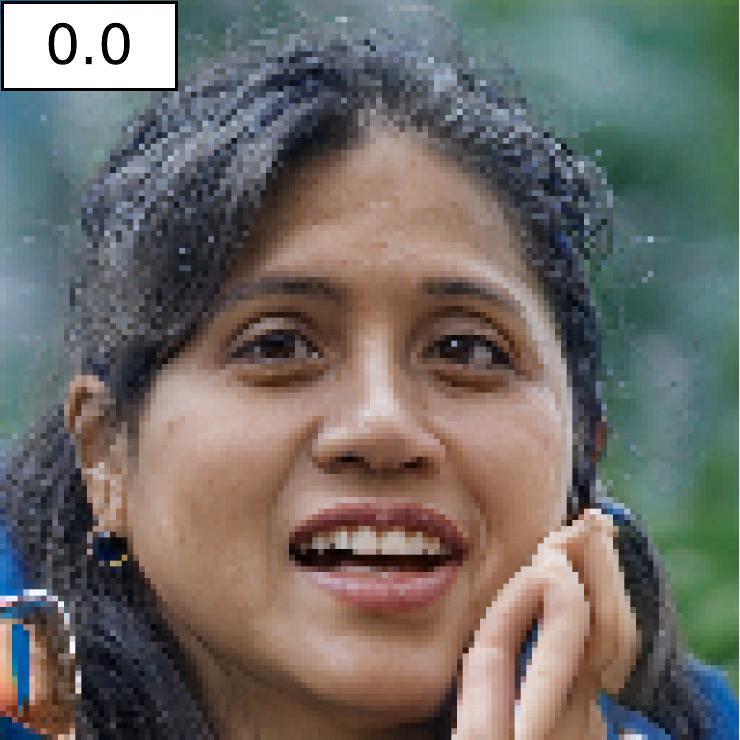} &
        \\
        \includegraphics[width=0.19\textwidth]{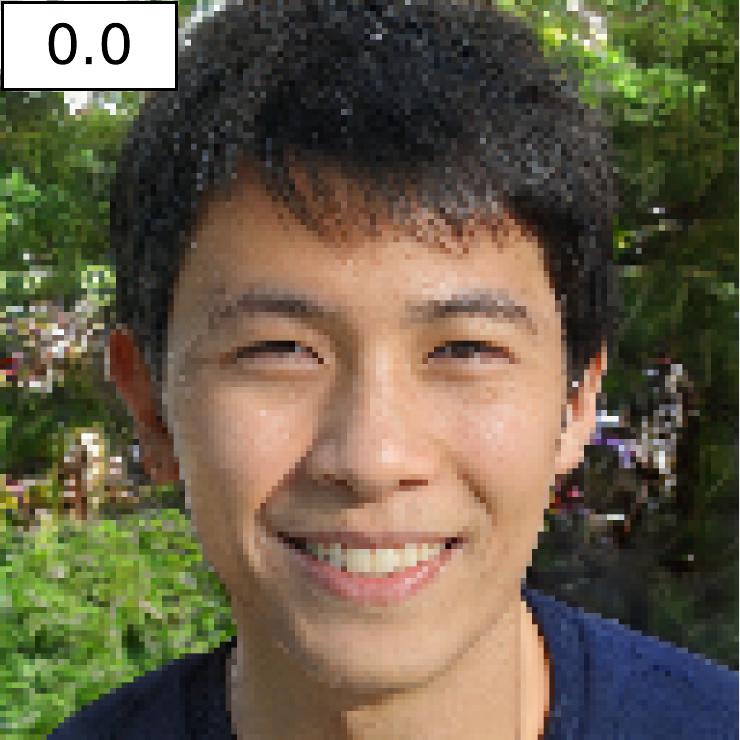} &
        \includegraphics[width=0.19\textwidth]{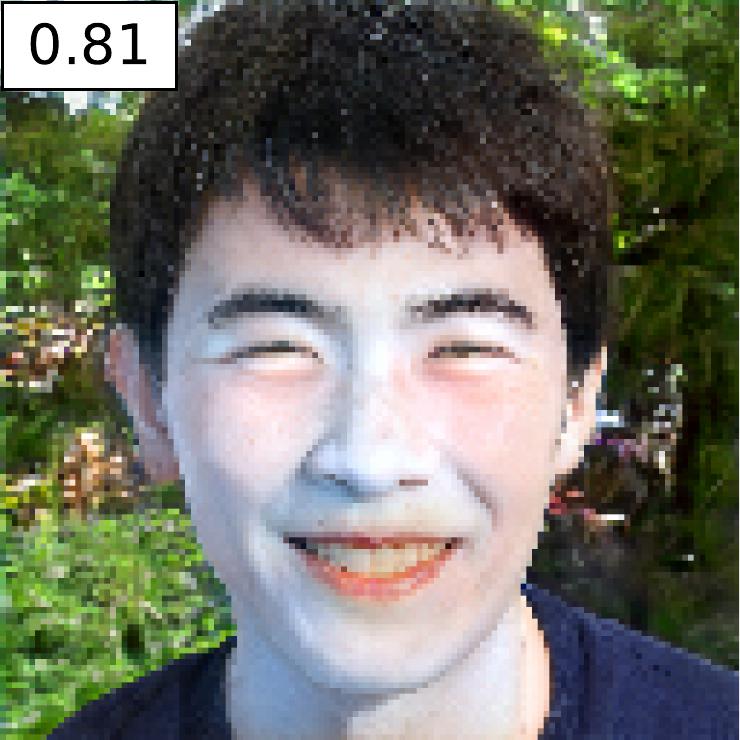} &
        \includegraphics[width=0.19\textwidth]{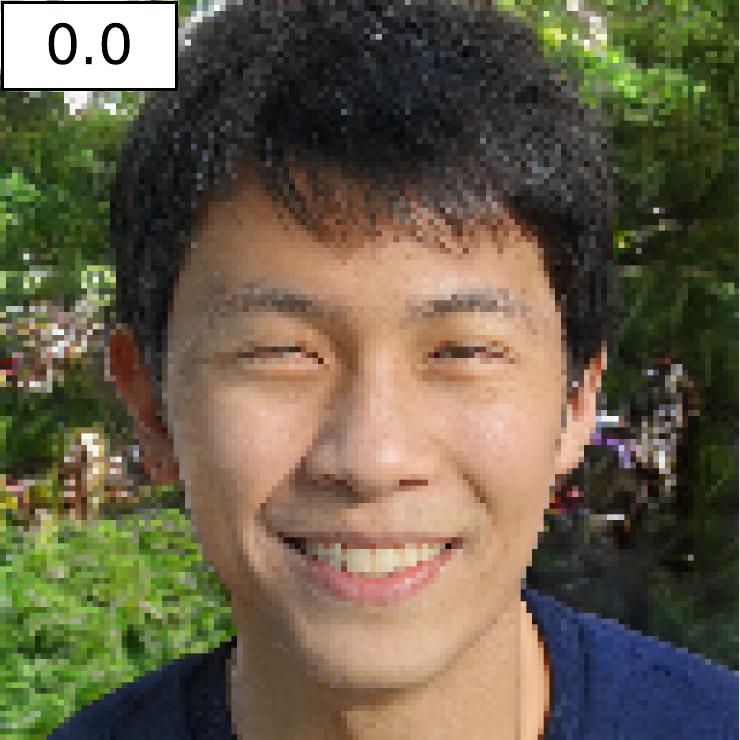} &
        \includegraphics[width=0.19\textwidth]{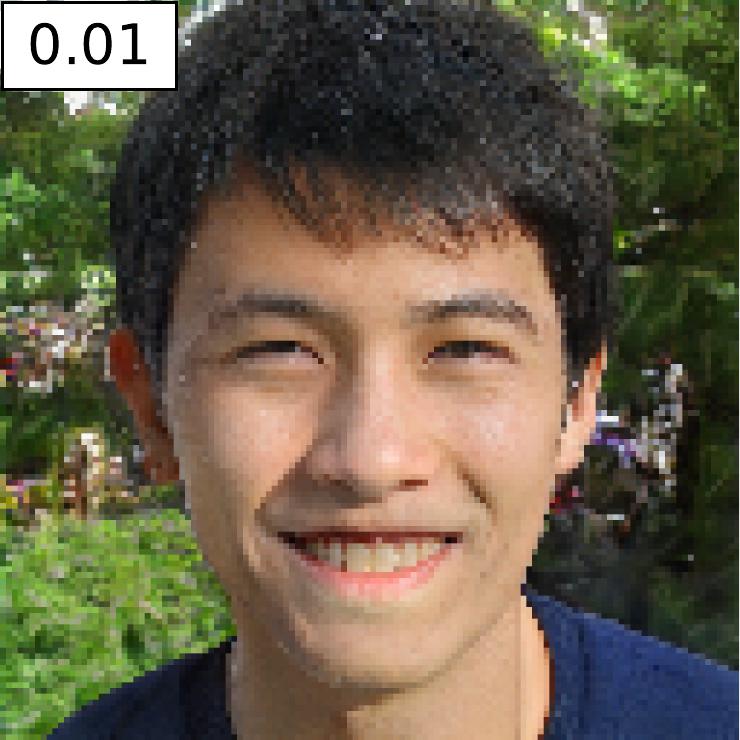} &
        \includegraphics[width=0.19\textwidth]{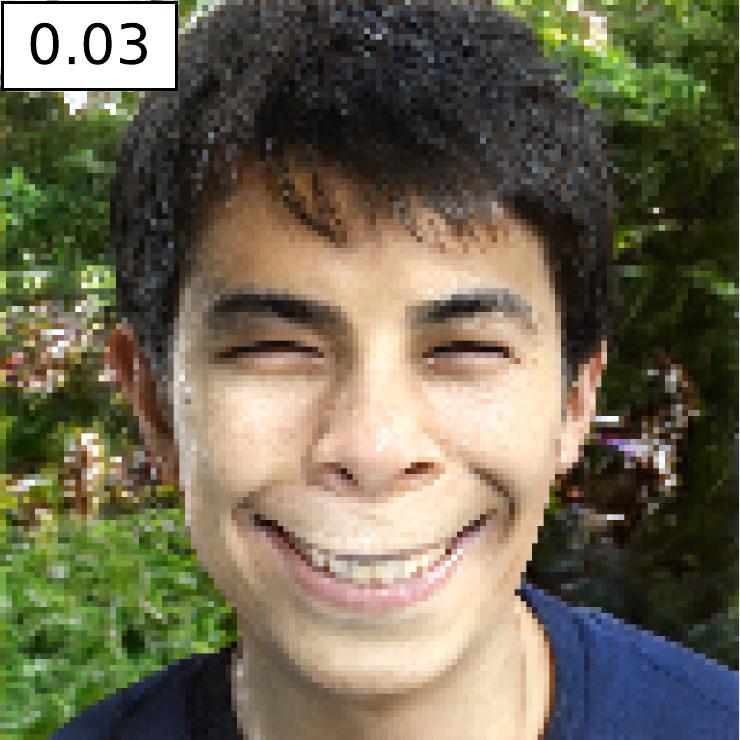} &
        \\
        Original image & Face color & Eye photometry & Lip photometry & Mouth shape
    \end{tabular}
    \caption{Counterfacts generated in interpretable subspaces to increase predicted score of the pale-skin classifier. See \Cref{sec:exp-counterfactual-materials} for details. Corresponding difference maps can be found in \Cref{fig:pale_skin-diff} in the appendix.}
    \label{fig:pale_skin}
\end{figure*}

In Figure~\ref{fig:pale_skin}, we visualize the counterfacts generated for the classifier of pain skin. Counterfacts generated on face color subspace is shown to be closely correlated with the prediction, while counterfacts in other subspaces show little influence on the prediction.

\begin{figure*}
    \centering
    \footnotesize
    \setlength{\tabcolsep}{1pt}
    \begin{tabular}{ccccc}
        \includegraphics[width=0.19\textwidth]{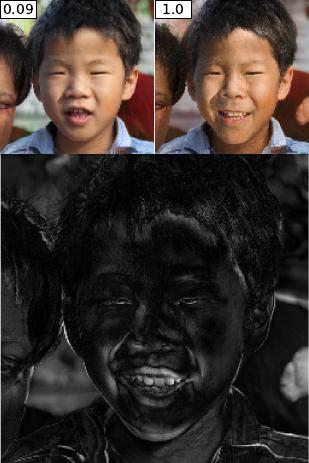} &
        \includegraphics[width=0.19\textwidth]{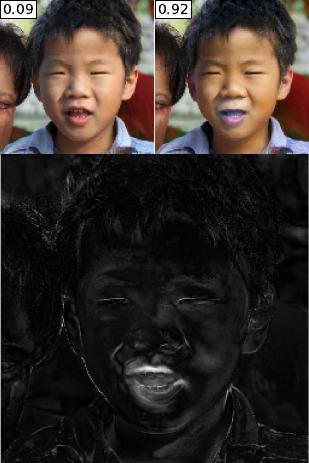} &
        \includegraphics[width=0.19\textwidth]{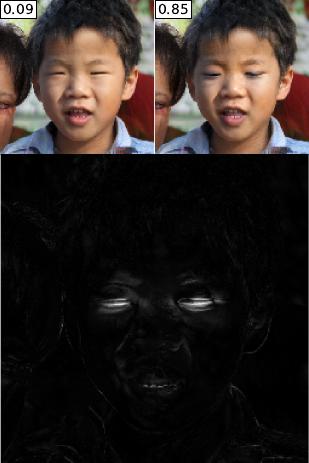} &
        \includegraphics[width=0.19\textwidth]{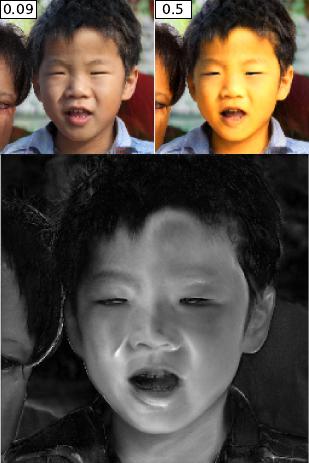} &
        \\
        \includegraphics[width=0.19\textwidth]{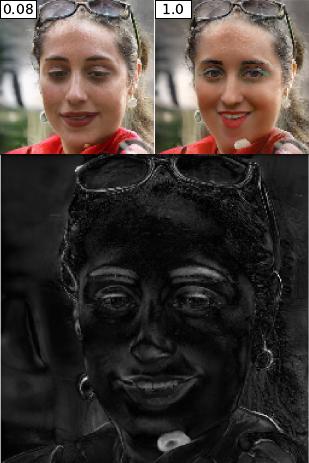} &
        \includegraphics[width=0.19\textwidth]{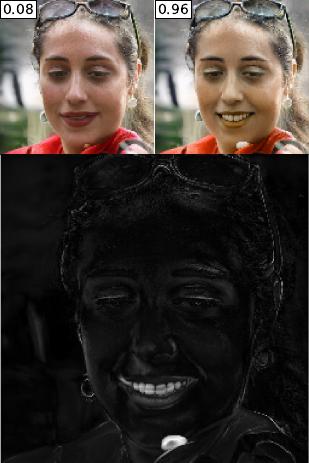} &
        \includegraphics[width=0.19\textwidth]{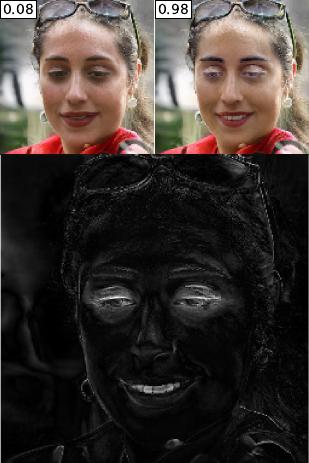} &
        \includegraphics[width=0.19\textwidth]{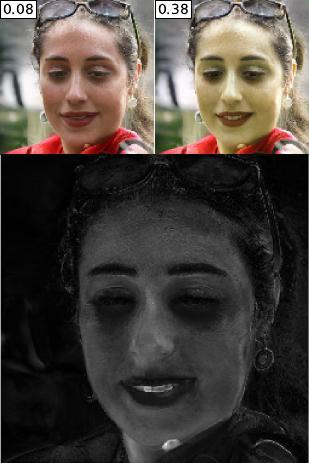} &
        \\
        \includegraphics[width=0.19\textwidth]{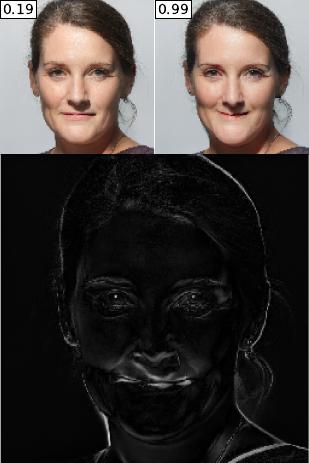} &
        \includegraphics[width=0.19\textwidth]{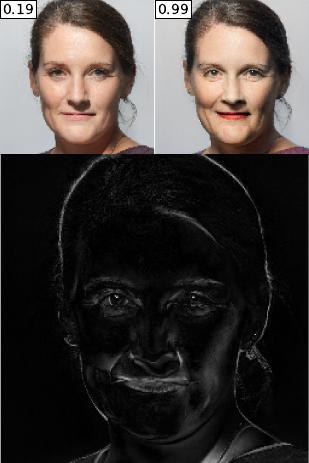} &
        \includegraphics[width=0.19\textwidth]{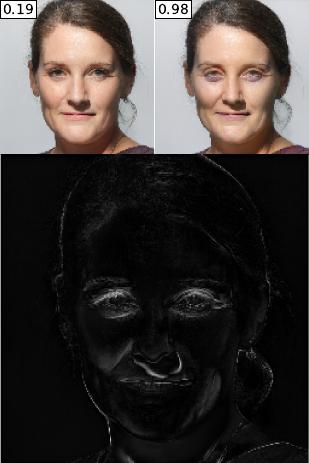} &
        \includegraphics[width=0.19\textwidth]{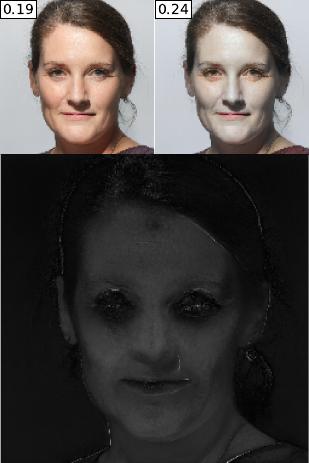} &
        \\
        \includegraphics[width=0.19\textwidth]{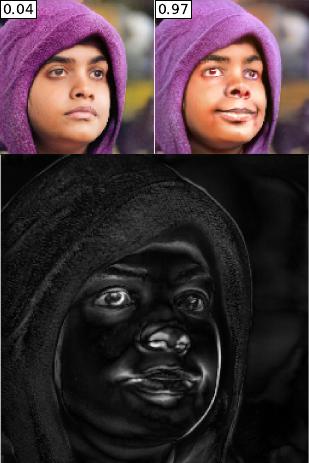} &
        \includegraphics[width=0.19\textwidth]{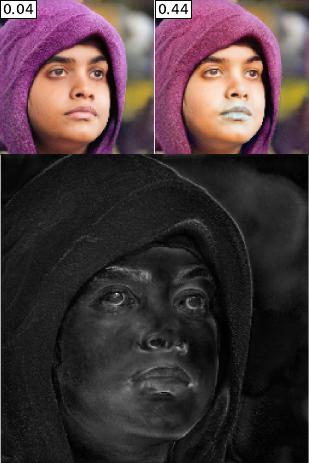} &
        \includegraphics[width=0.19\textwidth]{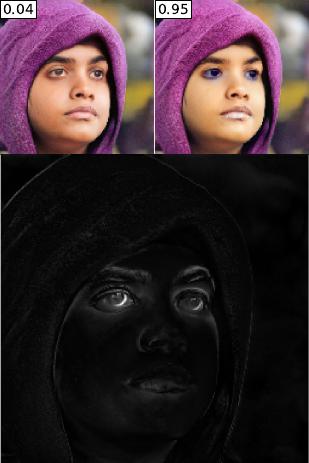} &
        \includegraphics[width=0.19\textwidth]{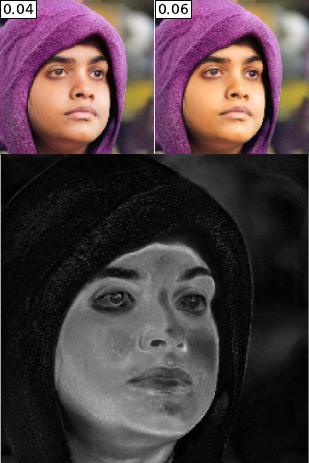} &
        \\
        Mouth shape & Lip photometry & Eye photometry & Face color
    \end{tabular}
    \caption{Difference maps of counterfacts generated for the smiling classifier in Figure~\ref{fig:smiling}. See also Section~\ref{sec:exp-counterfactual}.}
    \label{fig:smiling-diff}
\end{figure*}

\begin{figure*}
    \centering
    \footnotesize
    \setlength{\tabcolsep}{1pt}
    \begin{tabular}{ccccc}
        \includegraphics[width=0.19\textwidth]{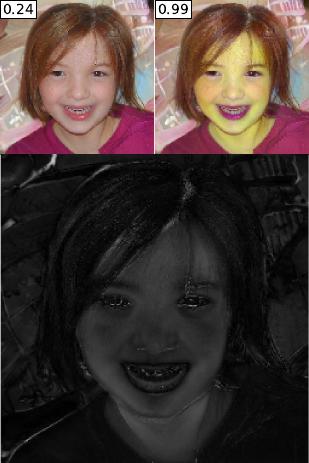} &
        \includegraphics[width=0.19\textwidth]{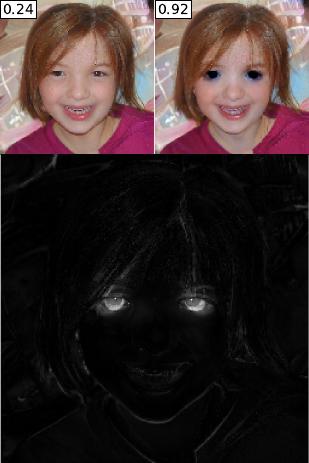} &
        \includegraphics[width=0.19\textwidth]{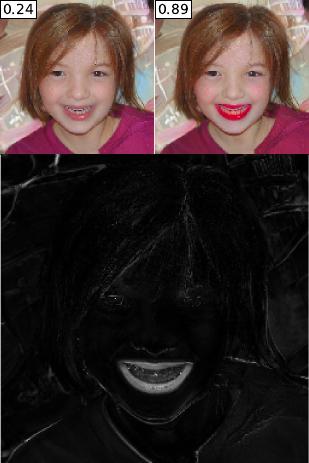} &
        \includegraphics[width=0.19\textwidth]{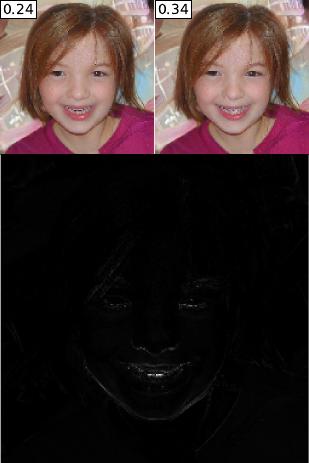} &
        \\
        \includegraphics[width=0.19\textwidth]{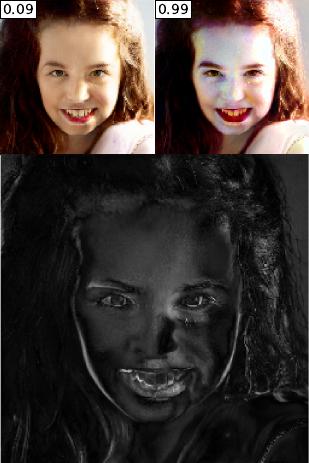} &
        \includegraphics[width=0.19\textwidth]{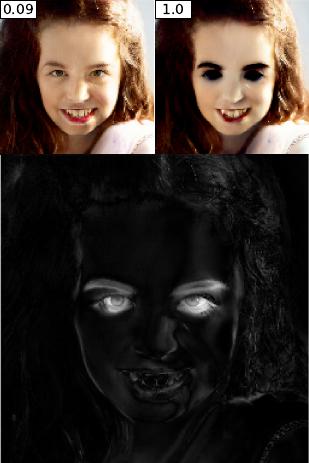} &
        \includegraphics[width=0.19\textwidth]{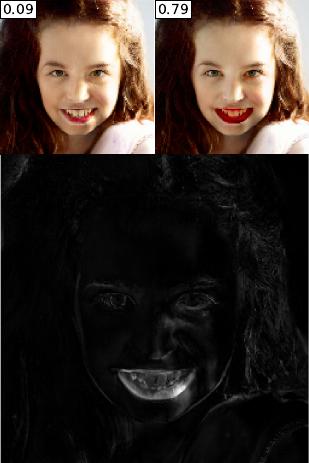} &
        \includegraphics[width=0.19\textwidth]{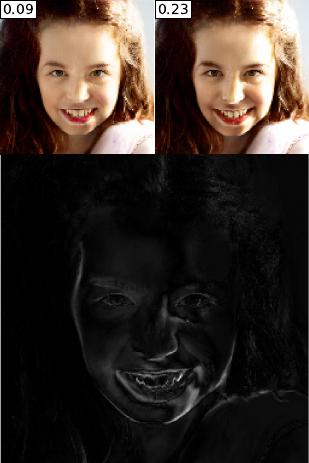} &
        \\
        \includegraphics[width=0.19\textwidth]{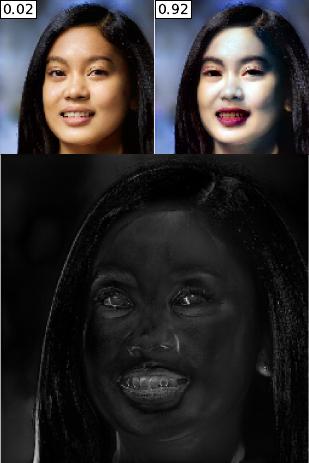} &
        \includegraphics[width=0.19\textwidth]{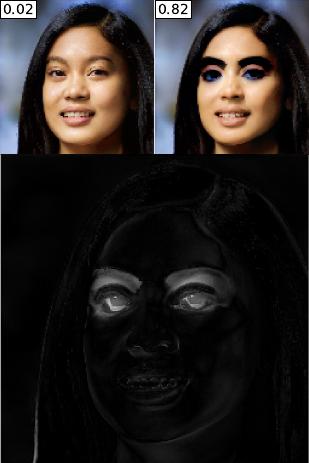} &
        \includegraphics[width=0.19\textwidth]{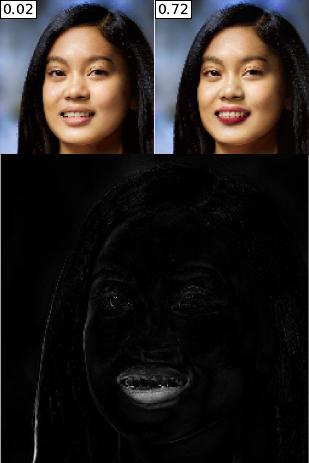} &
        \includegraphics[width=0.19\textwidth]{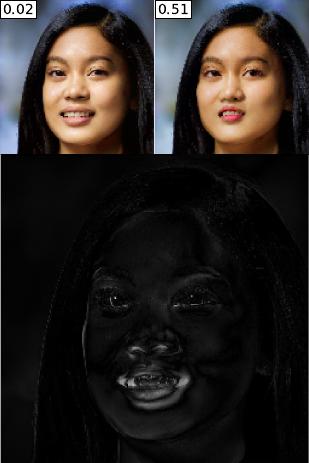} &
        \\
        \includegraphics[width=0.19\textwidth]{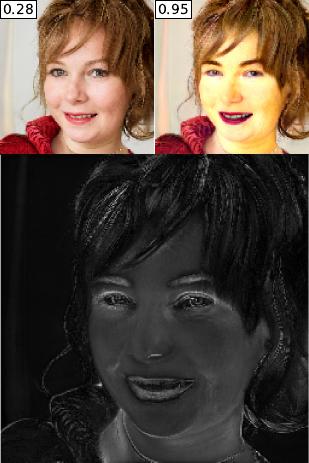} &
        \includegraphics[width=0.19\textwidth]{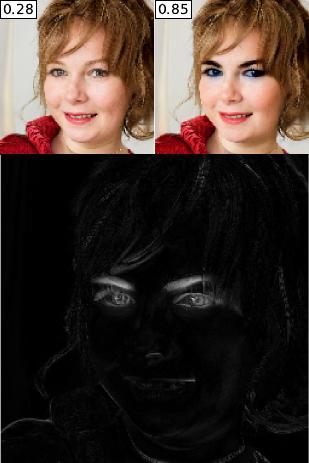} &
        \includegraphics[width=0.19\textwidth]{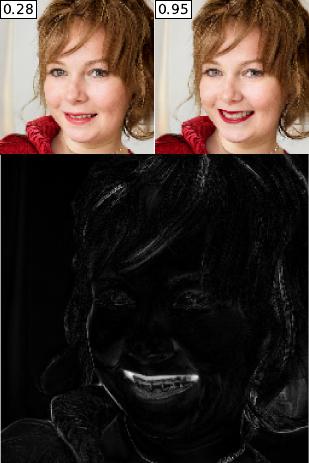} &
        \includegraphics[width=0.19\textwidth]{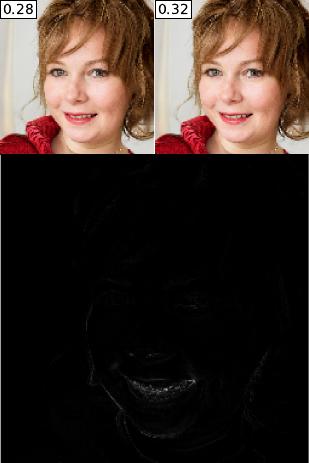} &
        \\
        Face color & Eye photometry & Lip photometry & Mouth shape
    \end{tabular}
    \caption{Difference maps of counterfacts generated for the heavy-makeup classifier in Figure~\ref{fig:heavy_makeup}. See also Section~\ref{sec:exp-counterfactual}.}
    \label{fig:heavy_makeup-diff}
\end{figure*}

\begin{figure*}
    \centering
    \footnotesize
    \setlength{\tabcolsep}{1pt}
    \begin{tabular}{ccccc}
        \includegraphics[width=0.19\textwidth]{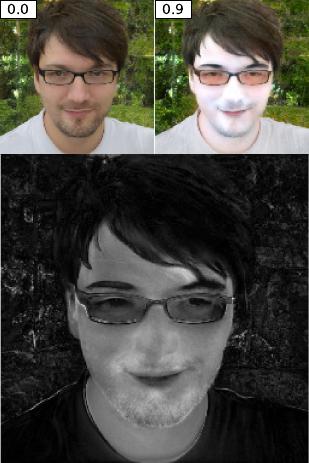} &
        \includegraphics[width=0.19\textwidth]{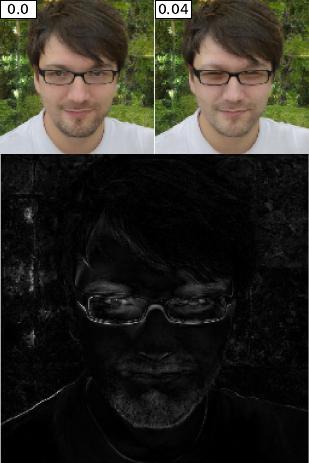} &
        \includegraphics[width=0.19\textwidth]{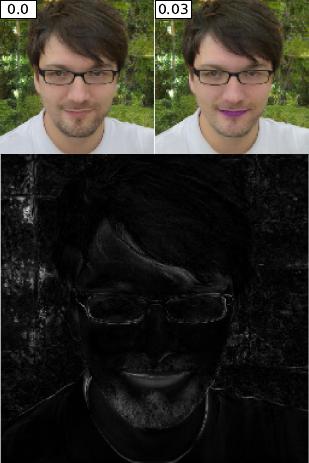} &
        \includegraphics[width=0.19\textwidth]{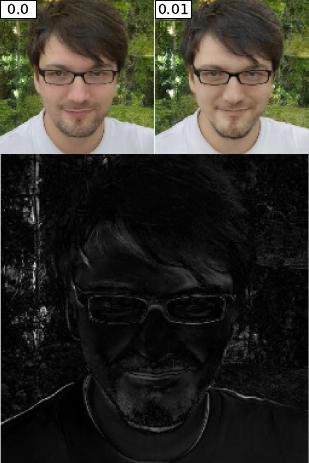} &
        \\
        \includegraphics[width=0.19\textwidth]{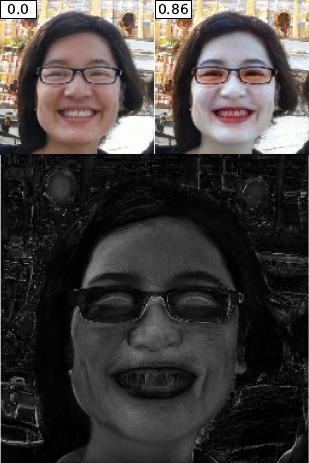} &
        \includegraphics[width=0.19\textwidth]{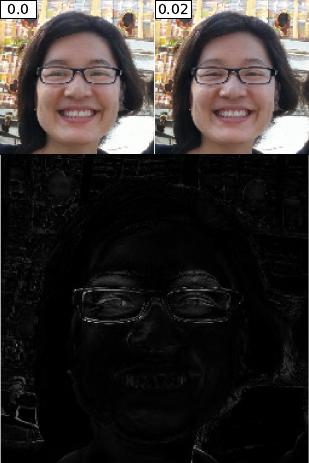} &
        \includegraphics[width=0.19\textwidth]{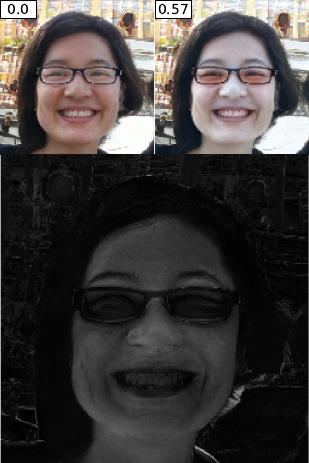} &
        \includegraphics[width=0.19\textwidth]{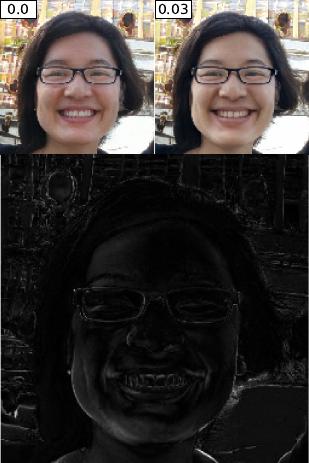} &
        \\
        \includegraphics[width=0.19\textwidth]{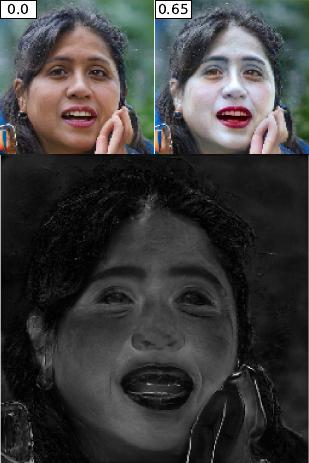} &
        \includegraphics[width=0.19\textwidth]{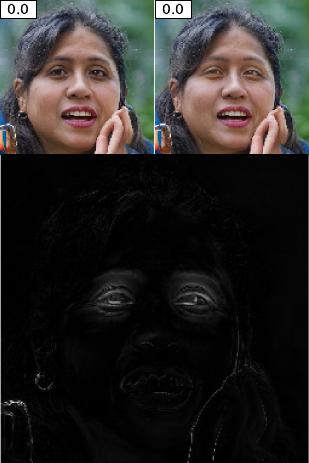} &
        \includegraphics[width=0.19\textwidth]{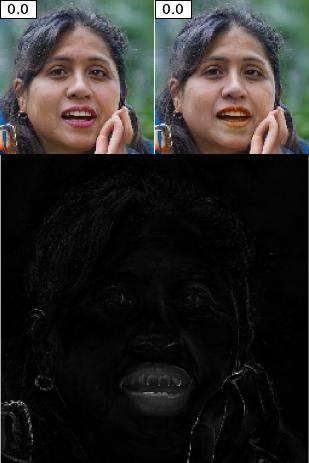} &
        \includegraphics[width=0.19\textwidth]{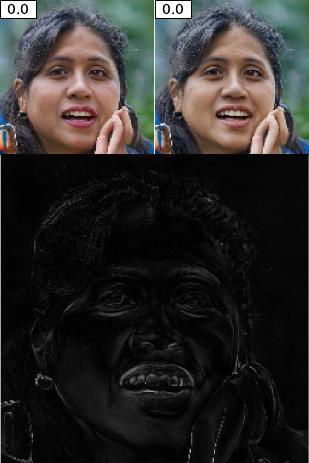} &
        \\
        \includegraphics[width=0.19\textwidth]{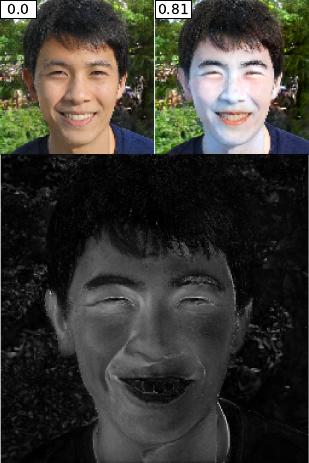} &
        \includegraphics[width=0.19\textwidth]{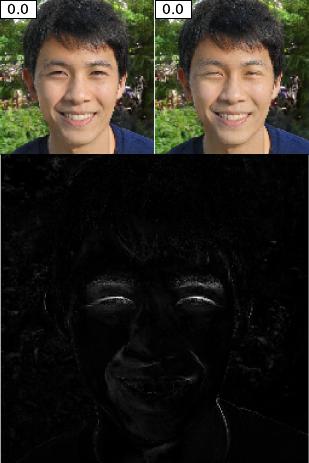} &
        \includegraphics[width=0.19\textwidth]{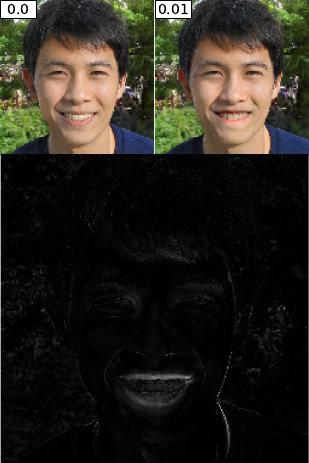} &
        \includegraphics[width=0.19\textwidth]{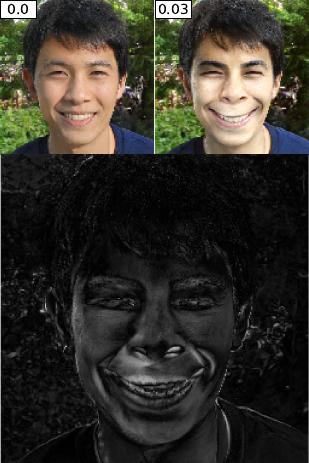} &
        \\
        Face color & Eye photometry & Lip photometry & Mouth shape
    \end{tabular}
    \caption{Difference maps of counterfacts generated for the pale-skin classifier in Figure~\ref{fig:pale_skin}. See also Section~\ref{sec:exp-counterfactual}.}
    \label{fig:pale_skin-diff}
\end{figure*}


\makeatletter\@input{PaperForReviewaux.tex}\makeatother
\makeatletter\@input{supaux.tex}\makeatother